\documentclass[10pt,twocolumn,letterpaper]{article}
\usepackage[pagenumbers]{cvpr}

\usepackage{amsmath}
\usepackage{amssymb}

\usepackage{algorithm}
\usepackage{float}
\usepackage{algpseudocode}

\usepackage{booktabs}
\usepackage{multirow}
\usepackage{graphicx}
\usepackage{xcolor}
\usepackage{colortbl}

\usepackage{cuted}
\usepackage{caption}
\usepackage{url}
\usepackage{placeins}

\renewcommand{\dblfloatpagefraction}{0.8}


\definecolor{HandRSIHeader}{HTML}{DDD4EB}
\definecolor{HandRSIInit}{HTML}{F9F7FC}
\definecolor{HandRSILite}{HTML}{F3EEF8}
\definecolor{HandRSIMedium}{HTML}{EDE5F5}
\definecolor{HandRSIHeavy}{HTML}{E5D9F0}

\usepackage{tikz}
\usepackage{pgfplots}
\pgfplotsset{compat=1.17}
\usepgfplotslibrary{groupplots,fillbetween}
\usetikzlibrary{calc,arrows.meta,shapes.geometric,backgrounds}
\pgfplotsset{
    hr/.style={
        scale only axis,
        axis lines=left, axis line style={-, black!70, line width=0.4pt},
        tick style={black!70, line width=0.4pt}, major tick length=2pt, tick align=outside,
        grid=major, grid style={black!8, line width=0.3pt},
        tick label style={font=\scriptsize}, label style={font=\footnotesize}, title style={font=\footnotesize},
        xlabel style={yshift=2pt}, ylabel style={yshift=-2pt},
        legend style={font=\scriptsize, draw=none, fill=none, inner sep=1pt},
        legend cell align=left,
        /pgf/number format/1000 sep={},
    },
}
\definecolor{HandRSIBlue}{HTML}{2A78D6}
\definecolor{CAce}{HTML}{D6646E}
\definecolor{CGray}{HTML}{8A96A3}
\definecolor{CLite}{HTML}{9F8AD3}
\definecolor{CMedium}{HTML}{6F52B5}
\definecolor{CHeavy}{HTML}{432A83}
\definecolor{CHawor}{HTML}{5E9C80}
\definecolor{CWilor}{HTML}{9AA3AE}
\definecolor{CEgo}{HTML}{8E7CC3}
\definecolor{QTag}{HTML}{5F6B75}
\definecolor{QMiss}{HTML}{D6646E}
\definecolor{LoopA}{HTML}{2A8C82}
\definecolor{LoopB}{HTML}{D9822B}
\definecolor{LoopC}{HTML}{8E7CC3}
\definecolor{Gate}{HTML}{C8553D}
\definecolor{KImg}{HTML}{6B7785}
\definecolor{KDet}{HTML}{C8553D}
\definecolor{KTemp}{HTML}{2A78D6}
\definecolor{KShape}{HTML}{E0A03A}
\definecolor{KArtic}{HTML}{8E7CC3}
\definecolor{KPlace}{HTML}{4A967E}
\newlength{\qw}
\newcommand{\qrow}[1]{\tikz[baseline=(l.center)]\node[rotate=90,inner sep=0pt,outer sep=0pt,minimum height=1em](l){\small #1};}
\newcommand{\qtile}[3]{\tikz[baseline=(q.center)]{%
    \node[inner sep=0pt,outer sep=0pt](q){\includegraphics[width=\qw]{#1}};%
    \if\relax\detokenize{#2}\relax\else
    \node[anchor=south west,fill=#3,fill opacity=0.92,text opacity=1,text=white,inner sep=1.5pt,rounded corners=1pt,
          font=\fontsize{6.5}{7.5}\selectfont] at ([xshift=2.2pt,yshift=2.2pt]q.south west){#2};%
    \fi}}
\DeclareCaptionFormat{HandRSIPurpleCaption}{%
    \setlength{\fboxsep}{5pt}%
    \colorbox{HandRSIHeader}{%
        \parbox{\dimexpr\linewidth-2\fboxsep\relax}{%
            \raggedright
            \strut #1#2#3\strut\par
        }%
    }\par
}

\definecolor{cvprblue}{rgb}{0.21,0.49,0.74}
\usepackage[
    pagebackref,
    breaklinks,
    colorlinks,
    allcolors=cvprblue
]{hyperref}

\hypersetup{
    pdftitle={HandRSI: Recursive Self-Improvement for Hand Pose Estimation},
    pdfauthor={Zhekai Wang, Luqing Luo, Wenhao Yao, Junjie Lu, Huanan Zhu, Feifei Gao, Yuhua Jiang},
}

\def\confName{}
\def\confYear{}

\newcommand{\hotAceMPJPE}{12.97}
\newcommand{\hotOursMPJPE}{8.37}
\newcommand{\arcAceMPJPE}{15.58}
\newcommand{\arcOursMPJPE}{13.85}

\newcommand{\hotRedMPJPE}{35}
\newcommand{\arcRedMPJPE}{11}
\newcommand{\hotRedJitter}{52}
\newcommand{\arcRedJitter}{44}

\newcommand{\hotCOneMPJPE}{9.91}

\newcommand{\hotRedCOneMPJPE}{16}
\newcommand{\hotRedCOneEPE}{27}
\newcommand{\hotRedCOneJitter}{51}
\newcommand{\arcRedCOneMPJPE}{11}
\newcommand{\arcRedCOneEPE}{18}
\newcommand{\arcRedCOneJitter}{44}
\newcommand{\arcFtShareMPJPE}{4}
\newcommand{\ftShareJitterMax}{3}
\newcommand{\hoiAceCT}{0.058}
\newcommand{\hoiCOneCT}{0.062}
\newcommand{\hoiCTwoFiveCT}{0.058}
\newcommand{\hoiCTwoFiveMPJPE}{22.24}

\newcommand{\liteFPS}{41.1}

\newcommand{\heavyFPS}{20.7}

\newcommand{\hvRatio}{98.5}
\newcommand{\nSearched}{124}

\makeatletter
\def\abstract{\iftoggle{cvprpagenumbers}{}{\thispagestyle{empty}}\centerline{\large\bf Abstract}\vspace*{8pt}\noindent\it\ignorespaces}

\makeatother

\newcommand{\codeurl}{\url{https://github.com/GinKenneth/Hand_RSI}}
\newcommand{\weightsurl}{\url{https://huggingface.co/HandRSI/HandRSI}}

\title{HandRSI: Recursive Self-Improvement for Hand Pose Estimation\\[4pt]
{\large\normalfont Technical Report}\vspace*{-6pt}}

\author{
Zhekai Wang$^{1,2}$\hspace{0.45em}
Luqing Luo$^{1}$\hspace{0.45em}
Wenhao Yao$^{1}$\hspace{0.45em}
Junjie Lu$^{1,3}$\\
Huanan Zhu$^{1,4}$\hspace{0.45em}
Feifei Gao$^{1,5}$\hspace{0.45em}
Yuhua Jiang$^{1,5\dagger}$\\[1pt]
{\normalsize $^{1}$LivSyn Robotics\quad
$^{2}$South China University of Technology\quad
$^{3}$University of Technology Sydney}\\
{\normalsize $^{4}$University of Science and Technology Beijing\quad
$^{5}$Tsinghua University}\\[1pt]
{\small Code: \href{https://github.com/GinKenneth/Hand_RSI}{\nolinkurl{github.com/GinKenneth/Hand_RSI}}\qquad
 Weights: \href{https://huggingface.co/HandRSI/HandRSI}{\nolinkurl{huggingface.co/HandRSI/HandRSI}}}
}

\begin{document}

\AddToShipoutPicture*{\AtTextUpperLeft{\put(0,\LenToUnit{5mm}){%
    \includegraphics[width=4.5cm,trim=28bp 28bp 34bp 18bp,clip]{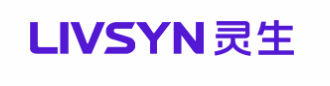}}}}

\maketitle
{\renewcommand{\thefootnote}{\fnsymbol{footnote}}\footnotetext[2]{Corresponding author, jiangyuhua@lingsheng.ai}}

\begin{strip}
    \vspace*{-1.2em}
    \centering
    \includegraphics[width=\linewidth]{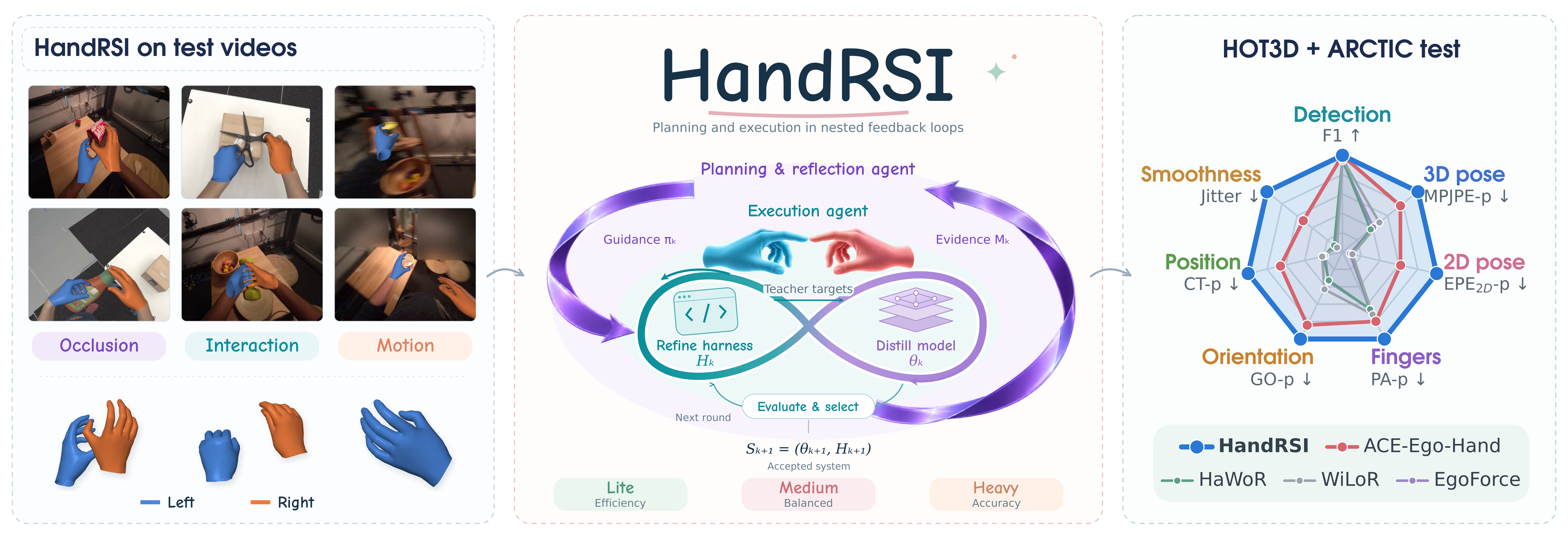}
    \par\vspace{2.3pt}
    \captionsetup{hypcap=false}
    \captionof{figure}{
\textbf{HandRSI at a glance.}
\textbf{Left:} HandRSI's predictions on frames of the sealed HOT3D and ARCTIC tests under occlusion, interaction
and fast motion; below, the MANO hands of each top frame from a novel view.
\textbf{Center:} coding agents refine the inference harness and distill it into the estimator in nested
feedback loops; Lite, Medium and Heavy systems are selected after the loop.
\textbf{Right:} HandRSI against ACE-Ego-Hand~\cite{ace_ego_hand} and the strongest re-run baselines on the
sealed HOT3D and ARCTIC tests; each axis scores a method relative to the best one (100\%).
    }
    \label{fig:teaser}
    \captionsetup{hypcap=true}
\end{strip}

\begin{abstract}
Egocentric 3D hand pose estimators still miss hands at the image border,
misplace occluded hands and jitter.
Such failures are usually fixed by hand-written inference refinements and by
fine-tuning, two interdependent steps that are tuned separately.
We present HandRSI, a recursive self-improvement framework for a hand pose
\emph{system} that consists of a video estimator and its executable inference
harness.
In each round, coding agents refine the harness, and a change is accepted only
if it improves a composite of all ten benchmark metrics on held-out data
without significantly regressing any of them.
The accepted harness then guides the estimator through selective distillation,
which supervises hand presence and temporal consistency on frames without
ground truth, while ground truth supervises pose.
Since the updated estimator may change which refinements remain useful, the
harness is re-validated on it.
Starting from the released \mbox{ACE-Ego-Hand} model, five rounds reduce
\mbox{MPJPE-p} by \hotRedMPJPE\% on HOT3D and \arcRedMPJPE\% on ARCTIC and roughly
halve jitter.
On HOT3D, HandRSI's \mbox{MPJPE-p} is \hotRedCOneMPJPE\% lower than that of
\mbox{ACE-Ego-Hand} fine-tuned on the same ground truth.
Under a unified protocol, it outperforms nine re-run published methods on all
ten metrics of both benchmarks.
Hypervolume selection on the accuracy--speed front yields Lite,
Medium and Heavy systems running at \heavyFPS{} to \liteFPS{} FPS.
This report gives the full method, every stage of every round and the
complete protocol. Code, evaluator, harnesses and weights are public.
\end{abstract}

\section{Introduction}
\label{sec:intro}

Recovering 3D hand motion from egocentric video~\cite{ego4d,epic100} is a
key step toward learning manipulation from human
demonstrations~\cite{dexmv,videodex,mimicplay}.
Video models now reconstruct hands under heavy occlusion and through short
out-of-view gaps~\cite{vidihand,ace_ego_hand}, yet a deployed estimator still
misses hands at the image border, misplaces occluded hands in depth, and
jitters at the fingertips.
Existing approaches fix such failures in two ways: with an
\emph{inference harness}, code that fuses auxiliary
detectors~\cite{shan2020hands}, enforces shape and temporal
priors~\cite{hmp} and refines hand localization using image evidence, and by
retraining or adapting the network~\cite{s2dhand,vhpot}.
Both are designed by hand and coupled, since a correction that helps one network may
become redundant or even harmful once the network is retrained.

In language-model systems, self-improving coding agents and harness
optimizers already revise executable scaffolds from execution
feedback~\cite{dgm,meta_harness,rrsi,harnesspro,vharness_srsi},
and joint harness--weight methods feed harness behavior back into model
parameters~\cite{harness_zero,co_harness,whale}.
Transferring the idea to 3D perception raises three problems.
First, quality is multi-dimensional: detection, 3D pose, 2D alignment,
placement and smoothness trade off, and a gate on one of them admits
changes that damage the others.
Second, a harness is a teacher of uneven quality: where 3D ground truth
(GT) exists its pose corrections are worse than the GT, but it knows
things the network does not, such as when a hand is visible and how it
moves.
Third, every network update changes which harness modules are still
needed.

We present \textbf{HandRSI}, a benchmark-grounded framework in which
coding agents recursively improve an egocentric hand pose system that
consists of an estimator and its harness (Figs.~\ref{fig:teaser}
and~\ref{fig:pipeline}).
In each round, eight agents write harness modules on a development split.
A gate accepts a module only if it improves a composite of all ten
benchmark metrics on held-out validation data without significantly
regressing any of them.
The accepted harness then serves as a teacher, and distillation separates
the information sources: the estimator learns pose from 3D GT, and hand
presence and temporal behavior from the teacher.
The harness is then re-validated on the updated estimator, and a
reflection agent plans the next round.
Agents never see val GT or test data.

Starting from the released ACE-Ego-Hand model~\cite{ace_ego_hand} and an
empty harness, five rounds reduce MPJPE-p from \hotAceMPJPE{} to
\hotOursMPJPE\,mm on HOT3D and from \arcAceMPJPE{} to \arcOursMPJPE\,mm on
ARCTIC's held-out subject.
Jitter drops by \hotRedJitter\% and \arcRedJitter\%.
Against nine published methods that we re-run under ACE-Ego-Hand's
protocol on the same segments, HandRSI is best in all ten metrics on both
benchmarks (Tab.~\ref{tab:main}).
Even over ACE-Ego-Hand fine-tuned on the same GT, HandRSI lowers
MPJPE-p by \hotRedCOneMPJPE\%, 2D error by \hotRedCOneEPE\% and jitter by
\hotRedCOneJitter\% on HOT3D.
Per-round controls show that the agent-written harness alone improves the
untouched base network, and on an unseen camera it still improves 3D pose
and jitter.
Distillation mainly transfers temporal stability and saturates after
round~2.
The lightest of the three tiers selected on the measured accuracy--speed
front runs within 4\% of the network's speed and keeps most of the gain.

Our contributions are:
\begin{itemize}[itemsep=2pt,topsep=4pt]
    \item A recursive self-improvement framework for a 3D perception system,
    with a ten-metric no-regression gate, distillation by information source
    and re-validation.
    \item Per-round controls that separate what the agent-written harness
    contributes from what transfers into the network.
    \item State-of-the-art results on ARCTIC and HOT3D against nine re-run
    methods, accuracy--speed tiers, and a release that reproduces every
    number.\footnote{\raggedright Code, evaluator and result files: \codeurl;\\
    weights and harnesses: \weightsurl.}
\end{itemize}
Section~\ref{sec:method} describes the method in full, Sec.~\ref{sec:setup} the
data, evaluator and baselines, and Sec.~\ref{sec:results} the results, including
every stage of every round with its controls; Appendix~\ref{sec:app_gallery}
gives extended qualitative comparisons on all three test sets.

\section{Related Work}
\label{sec:related}

\paragraph{Egocentric 3D hand pose.}
Early RGB methods lift 2D keypoints to 3D~\cite{zimmermann2017learning} or
regress MANO~\cite{mano} hand meshes~\cite{boukhayma2019,ge2019}; current
single-image methods reconstruct MANO hands with large transformers trained
at scale~\cite{hamer,wilor,hamba,handos}, including models for egocentric
images~\cite{wildhands}, interacting hands~\cite{interwild} and real-time
use~\cite{fah}.
3D labels come from instrumented capture (multi-view, motion capture or
depth) and model
fitting~\cite{freihand,interhand,ho3d,dexycb,h2o,assemblyhands,arctic,hot3d,hoi4d,reinterhand},
now also outdoors with a worn multi-camera rig~\cite{show3d}, or come from
millions of synthetic renderings~\cite{anyhand}, while large egocentric video collections~\cite{ego4d,epic100} and
hand-contact annotations~\cite{shan2020hands} remain largely without 3D
labels.
A benchmark study of egocentric hand--object pose~\cite{fan2024benchmarks}
finds fast hand motion and close contact between hands and objects still
intractable for current methods.
Video methods enforce consistency over frames~\cite{tcmr,hmp,omnihands},
also by fusing neighboring frames against occlusion~\cite{deformer},
recover 4D hand motion, also in world space under a moving
camera~\cite{haptic,dyn_hamr,hawor,stablehand,mint}, or use the forearm to
place the hand in camera space~\cite{egoforce}.
Video diffusion features now give the strongest egocentric
results~\cite{vidihand,ace_ego_hand}; ACE-Ego-Hand evaluates on
ARCTIC~\cite{arctic,fan2024benchmarks}, HOT3D~\cite{hot3d} and HOI4D~\cite{hoi4d} with penalties
for missed hands.
Concurrent work adapts a video foundation model into a tracker of hand pose,
contact and force~\cite{rlhnd}, and estimates world-space hand motion and the
camera trajectory in one streaming model~\cite{infinihand}.
Accurate egocentric hands matter downstream, for example to teach robots
dexterous skills from human video~\cite{dexmv,videodex,mimicplay}.
We keep ACE-Ego-Hand's model and protocol and improve the system around the
network.

\paragraph{Agents that write and improve code.}
Coding agents edit repositories through tool
interfaces~\cite{sweagent,openhands} and are benchmarked on machine-learning
engineering~\cite{mlebench}; research agents run the loop from idea to
experiment~\cite{aiscientist}, improve published state-of-the-art
models~\cite{autosota} or curate fine-tuning data~\cite{autosft}.
Language models that evolve programs against an automatic evaluator have
discovered new algorithms~\cite{funsearch,alphaevolve}, and verbal feedback
on earlier attempts improves later ones~\cite{reflexion,selfrefine}.
Self-improving agents rewrite their own code~\cite{aide2}, also from an
archive of earlier variants~\cite{dgm,mgm}, and recursive self-improvement
has been applied to such systems themselves~\cite{metarsi}.

\paragraph{Harness optimization.}
Harness optimizers search over the code around a frozen model with an
archive of earlier attempts~\cite{meta_harness}: as separable
modules~\cite{modular_rsi}, recursively for unseen open-world
tasks~\cite{harness_squared}, regularized against memorizing the training
tasks~\cite{rrsi}, with evidence governance for a video-understanding
harness under reused, noisy feedback~\cite{vharness_srsi}, or on a
performance--cost front~\cite{harnesspro}.
Harness-Zero, Co-Harness and WHALE optimize harness and weights together
and feed harness behavior back into the model's weights, WHALE by
alternating weight and harness updates~\cite{harness_zero,co_harness,whale}.
Like HarnessPro, we select on a performance--cost front, here deployment
tiers chosen by the hypervolume indicator~\cite{zitzler1999multiobjective}
with $\varepsilon$-dominance~\cite{laumanns2002combining}.
These systems, like visual programs that compose fixed vision
models~\cite{visprog}, produce text, answers or programs; HandRSI applies
the idea to metric 3D perception, where the gate must guard ten metrics at
once, and distills the harness's gains into the network.

\paragraph{Refinement as a teacher.}
Distillation trains a student on a teacher's outputs~\cite{kd},
self-training with a noisy student can surpass the teacher by using
unlabeled data~\cite{noisy_student}, and hand keypoint detectors have been
bootstrapped from their own triangulated multi-view
predictions~\cite{simon2017hand}; egocentric estimators adapt to new
cameras without labels through dual-view consensus~\cite{s2dhand} or
test-time optimization~\cite{vhpot}.
SPIN~\cite{spin} puts model fitting in the training loop, where fits
initialized by the regressor supervise it.
HandRSI's teacher is likewise the network plus a refinement, but the
refinement is agent-written code, and the student is distilled by
information source: it keeps 3D pose from GT where it exists and takes
presence, temporal behavior and, on unlabeled ARCTIC video, the hand's
articulation from its teacher (Sec.~\ref{sec:distillation}), adapting the
released network with low-rank updates~\cite{lora,loraplus}.

\paragraph{Evaluation under adaptive reuse.}
A holdout that is queried adaptively loses its statistical
validity~\cite{dwork2015reusable}, leaderboards need mechanisms that limit
what each submission reveals~\cite{blum2015ladder}, and agent evaluations
that ignore cost or lack a held-out split reward
overfitting~\cite{kapoor2024agents}.
HandRSI caps the agents' val queries, seals test from them and reports
token use.

\section{Method}
\label{sec:method}

\begin{figure*}[t]
    \centering
    \includegraphics[width=\linewidth]{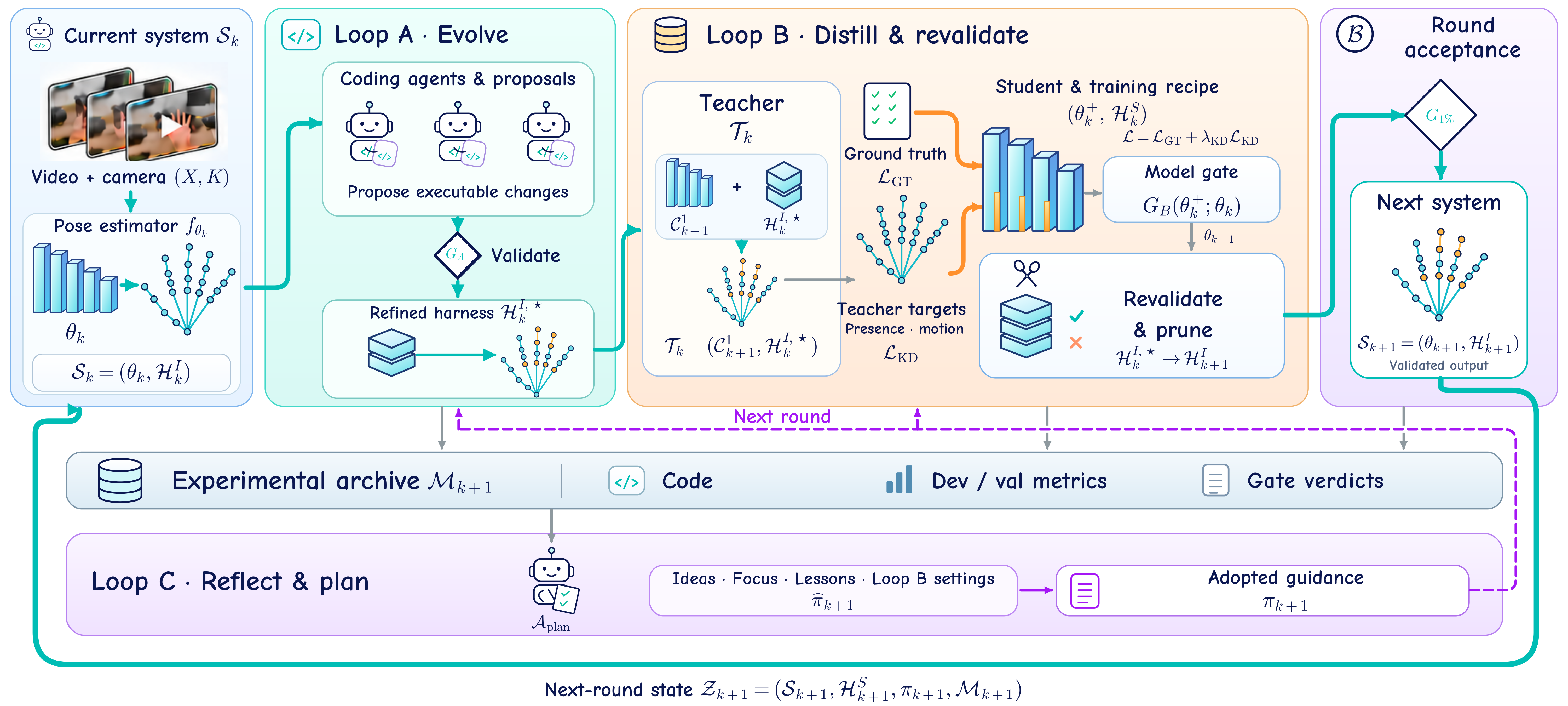}
    \caption{
        \textbf{Overview of one HandRSI round.}
        Loop~A evolves the inference harness under the gate $G_A$; Loop~B
        distills the refined harness into the estimator and re-validates it on
        the new network; the round gate $G_{1\%}$ accepts the new system on val;
        Loop~C turns the experimental archive into guidance for the next round.
    }
    \label{fig:pipeline}
\end{figure*}

\begin{algorithm}[t]
    \small
    \setlength{\fboxsep}{5pt}
    \captionsetup{
        format=HandRSIPurpleCaption,
        font=small,
        labelfont=bf,
        textfont=normalfont,
        labelsep=space,
        singlelinecheck=false,
        margin=0pt
    }
    \caption{One HandRSI round}
    \label{alg:round}
    \noindent
    \colorbox{HandRSIInit}{%
    \begin{minipage}{\dimexpr\linewidth-2\fboxsep\relax}
    \begin{algorithmic}[1]
        \Require round state $\mathcal Z_k=(\mathcal S_k,\mathcal H^S_k,\pi_k,\mathcal M_k)$ with $\mathcal S_k=(\theta_k,\mathcal H^I_k)$
        \State Capture raw outputs of $\theta_k$ on all splits
        \State \textbf{Loop A:} 8 agents edit $\mathcal H^I_k$ on dev under $\pi_k$; submissions gated by $G_A=G_{0.3\%}$ on val
        \State Merge and integrate accepted code; one val look $\Rightarrow\mathcal H_k^{I,\star}$
        \State $\mathcal C^1_{k+1}\gets$ fine-tune $\theta_k$ on GT (from round~3: used only if ahead of $\theta_k$ on dev)
        \State $\mathcal T_k\gets\bigl(\mathcal C^1_{k+1},\ \mathcal H_k^{I,\star}$ pruned on $\mathcal C^1_{k+1}$ (dev)$\bigr)$; label train and unlabeled video with $\mathcal T_k$
        \State \textbf{Loop B:} train checkpoints $\Theta_k$ with $\mathcal L_{\mathrm{GT}}+\lambda_{\mathrm{KD}}\mathcal L_{\mathrm{KD}}$ under $\mathcal H^S_k$
        \State $\theta_{k+1}\gets\theta_k^+$ if $G_B$ (some checkpoint with $L_\Delta>0$ vs.\ $\theta_k$ on val, networks alone), else $\theta_k$
        \State $\mathcal H^I_{k+1}\gets$ re-validate $\mathcal H_k^{I,\star}$ on $\theta_{k+1}$ (dev, leave-one-out, joint removal; skipped if $\theta_{k+1}=\theta_k$)
        \If{$G_{1\%}\bigl((\theta_{k+1},\mathcal H^I_{k+1});\mathcal S_k\bigr)$ on val}
            \State $\mathcal S_{k+1}\gets(\theta_{k+1},\mathcal H^I_{k+1})$
        \ElsIf{$G_{A}\bigl((\theta_k,\mathcal H_k^{I,\star});\mathcal S_k\bigr)$ on val}
            \State $\mathcal S_{k+1}\gets(\theta_k,\mathcal H_k^{I,\star})$ \Comment{harness only}
        \Else
            \State $\mathcal S_{k+1}\gets\mathcal S_k$
        \EndIf
        \State Score $\mathcal S_{k+1}$, $(\theta_{k+1},\varnothing)$, $\mathcal C^1_{k+1}$, $\mathcal C^2_{k+1}$ on sealed test (reporting only)
        \State \textbf{Loop C:} append the round to $\mathcal M_{k+1}$; $\widehat\pi_{k+1}\gets\mathcal A_{\mathrm{plan}}(\pi_k,\mathcal M_{k+1})$; adopt $\pi_{k+1}$
    \end{algorithmic}
    \end{minipage}%
    }\par
\end{algorithm}

HandRSI improves a hand pose system in rounds of harness evolution (Loop~A),
distillation (Loop~B) and reflection (Loop~C), all judged against a fixed
benchmark (Fig.~\ref{fig:pipeline}).
Language models are never trained; only the estimator and the code around
it change.
Loop~A changes the harness and Loop~B the network, each only through a
gate on the same benchmark, so their gains can be separated and compounded,
and a gate accepts a change only if it improves the benchmark as a whole
without significantly regressing any metric.
All loops share one archive, so the evidence of every attempt informs later
rounds.
Algorithm~\ref{alg:round} lists one round in full, including the raw-output
capture, the GT-only control $\mathcal C^1_{k+1}$ that the teacher is built on,
and the controls that are scored on test for reporting only.
Round~$k$ of the tables and figures is the step from $\mathcal Z_{k-1}$ to
$\mathcal Z_k$.

\subsection{Problem Formulation}
\label{sec:formulation}

\paragraph{System and round state.}
Given a clip $X=\{I_t\}_{t=1}^{T}$ and intrinsics $K$, the estimator $f_\theta$
predicts per frame and hand a presence score, MANO~\cite{mano} parameters,
camera-space translation and 2D keypoints.
An inference harness $\mathcal H^I$ is a chain of executable modules that
refines these predictions:
\begin{equation}
\begin{split}
    \mathcal S_k&=(\theta_k,\mathcal H^I_k),\\
    \mathcal S_k(X;K)&=\mathcal H^I_k\bigl(f_{\theta_k}(X;K),X,K\bigr).
\end{split}
    \label{eq:system}
\end{equation}
Modules see one segment's predictions, frames and $K$, but never ground
truth.
Together with the training recipe $\mathcal H^S$ of Loop~B, the agents'
guidance $\pi$ and the archive $\mathcal M$ of earlier attempts, a round
maps the state
\begin{equation}
    \mathcal Z_k=(\mathcal S_k,\mathcal H^S_k,\pi_k,\mathcal M_k)
    \label{eq:state}
\end{equation}
to $\mathcal Z_{k+1}$, starting from the released ACE-Ego-Hand
model~\cite{ace_ego_hand} with an empty harness, $\mathcal S_0=(\theta_0,\varnothing)$.

\paragraph{Benchmark-grounded gates.}
A specification $\mathcal B$ sets the data splits, the evaluator and
the gates outside the agents' reach: agents see only the dev split and val
only through a gate service, since an adaptively queried holdout loses its
validity~\cite{dwork2015reusable}, and test is sealed from the agents.
For a candidate $\mathcal S'$ and an incumbent $\mathcal S$, the relative
improvement of an error metric $m\in\mathcal Q_{\downarrow}$ (seven columns)
and of a detection metric $m\in\mathcal Q_{\uparrow}$ (FAcc, Recall, F1) is
\begin{equation}
    r_m=
    \begin{cases}
        \dfrac{e_m(\mathcal S)-e_m(\mathcal S')}{e_m(\mathcal S)}, & m\in\mathcal Q_{\downarrow},\\[8pt]
        \dfrac{a_m(\mathcal S')-a_m(\mathcal S)}{\max(1-a_m(\mathcal S),\,0.05)}, & m\in\mathcal Q_{\uparrow},
    \end{cases}
    \label{eq:rel}
\end{equation}
so detection is normalized by the remaining error, floored so that a few
false positives near saturation cannot dominate.
Over the ten metrics $\mathcal Q=\mathcal Q_{\downarrow}\cup\mathcal Q_{\uparrow}$,
the composite gain is
\begin{equation}
    \Delta(\mathcal S';\mathcal S)=\frac{1}{|\mathcal Q|}\sum_{m\in\mathcal Q}
    \operatorname{clip}(r_m,-0.05,0.05).
    \label{eq:composite}
\end{equation}
With $[L_\Delta,U_\Delta]$ its 90\% paired-bootstrap
interval~\cite{efron1994bootstrap}, resampling sequences, a gate requires a
significant gain and no regressing metric, where $\mathcal G_{\mathrm{safe}}$
rejects any $r_m<-2\%$ whose interval lies below zero and any $r_m<-5\%$:
\begin{equation}
    G_\tau(\mathcal S';\mathcal S)=[L_\Delta>0]\land[\Delta\geq\tau]\land\mathcal G_{\mathrm{safe}}.
    \label{eq:gate}
\end{equation}
Clipping keeps a single metric from dominating $\Delta$, and because $\Delta$
covers every column of the results table, a change that trades 2D
alignment for 3D placement, or pose for detection, is rejected.
Every gate of the method is an instance of Eq.~\eqref{eq:gate}, with the
settings of Tab.~\ref{tab:gates}.
$\Delta$ and the regression rule use HOT3D; ARCTIC is scored and reported
but does not block, because $\theta_0$ was trained on all ARCTIC subjects
except the test subject, so drops there measure forgetting rather than
generalization.
Test is scored only for reporting (Alg.~\ref{alg:round}, line~16).

\begin{table}[t]
    \centering
    \caption{\textbf{Gate settings.} $\Delta$ and the regression check use HOT3D; ARCTIC
    is scored and reported.}
    \label{tab:gates}
    \footnotesize
    \setlength{\tabcolsep}{4pt}
    \setlength{\aboverulesep}{0pt}
    \setlength{\belowrulesep}{0pt}
    \begin{tabular}{@{}l>{\raggedright\arraybackslash}p{0.47\linewidth}@{}}
        \toprule
        \rowcolor{HandRSIHeader}
        Setting & Value \\
        \midrule
        Per-metric clip of $r_m$ & $\pm5\%$ \\
        Detection floor ($1-a_m$) & $0.05$ \\
        Regression & $r_m<-2\%$ with CI$_{90}<0$, or $r_m<-5\%$ \\
        Bootstrap & paired~\cite{efron1994bootstrap}, over sequences, 90\% \\
        Loop~A pre-check (dev, uncharged) & $\Delta>0$, $L_\Delta>0$ \\
        Loop~A gate $G_A$ (val) & $G_{0.3\%}$, 1{,}000 resamples \\
        Greedy merge (dev) & $L_\Delta>0$ per added module \\
        Model gate $G_B$ (val, networks) & some checkpoint with $L_\Delta>0$ \\
        Re-validation (dev) & module useful iff $L_\Delta>0$ \\
        Round gate (val) & $G_{1\%}$, 2{,}000 resamples \\
        Harness only (val) & $G_{0.3\%}$ on the previous network \\
        \bottomrule
    \end{tabular}
\end{table}

\subsection{Loop A: Harness Evolution}
\label{sec:loopA}

With $\theta_k$ fixed, eight coding agents work in parallel for four hours
per round, each from $\mathcal H^I_k$ with a focus area from $\pi_k$
(detection, root placement, orientation, fingers, image evidence, temporal
behavior, 2D keypoints, out-of-sight hands); four of them run Claude
Opus~5.5 in Claude Code and four GPT-6-sol in Codex.
Each agent runs in a sandbox that exposes only the dev views, the cached dev
outputs of $f_{\theta_k}$, its workspace and the shared archive
$\mathcal M_k$.
Its tools replay a candidate chain on the cached outputs against its parent,
with diagnostics by joint, frame and failure type; agents may also search
papers and download source code, and they read the archive of every earlier
submission with its code and verdict, so that they build on earlier
attempts instead of repeating them.
A module sees one 81-frame segment, no sequence name and no GT, and must be
deterministic.

A submission is pre-checked on dev free of charge and then gated by
$G_A=G_{0.3\%}$ on val against the agent's incumbent, with ten charged
submissions per agent and round, since a holdout that answers adaptive
queries freely stops being a valid
estimate~\cite{dwork2015reusable,blum2015ladder}.
Accepted submissions are merged greedily on dev, best val gain first, each
added module having to bring a significant gain; an integration agent
reconciles overlapping modules, and one val comparison picks the refined
harness $\mathcal H_k^{I,\star}$ among the best submission, the greedy merge
and the integrated chain.

\subsection{Loop B: Distillation and Re-validation}
\label{sec:distillation}

\paragraph{Supervision by information source.}
The refined harness on $\mathcal C^1_{k+1}$, the network fine-tuned on GT,
forms the teacher $\mathcal T_k=(\mathcal C^1_{k+1},\mathcal H_k^{I,\star})$,
with the harness re-validated on $\mathcal C^1_{k+1}$ (from round~3 on,
$\mathcal C^1_{k+1}$ is used only if it is ahead of $\theta_k$ on dev); it
labels the training sequences and unlabeled video.
Where 3D GT exists, harness pose targets only conflict with it, but on
unlabeled frames the teacher knows what GT cannot: whether a hand is present
and how it moves.
The student therefore learns each signal from its reliable source,
\begin{equation}
\begin{split}
    \mathcal L&=\mathcal L_{\mathrm{GT}}+\lambda_{\mathrm{KD}}\mathcal L_{\mathrm{KD}},\\
    \mathcal L_{\mathrm{GT}}&=\ell_{2D}+\ell_{\mathrm{rel}}+\ell_{\mathrm{root}}
    +\lambda_{R}\ell_{\mathrm{GO}}+\lambda_{e}\ell_{\mathrm{exist}},\\
    \mathcal L_{\mathrm{KD}}&=\lambda_{e}\ell_{\mathrm{exist}}^{\mathcal T}
    +\lambda_{\mathrm{temp}}\ell_{\mathrm{temp}}^{\mathcal T},
\end{split}
\label{eq:losses}
\end{equation}
where $\mathcal L_{\mathrm{GT}}$ supervises 2D keypoints, wrist-relative and
root joints, orientation and presence on annotated frames, and
$\mathcal L_{\mathrm{KD}}$ transfers the teacher's presence and temporal
behavior to frames without GT, including unlabeled video:
$\ell_{\mathrm{exist}}^{\mathcal T}$ is a cross-entropy to the teacher's
presence, $\ell_{\mathrm{temp}}^{\mathcal T}$ a Huber loss between the
student's and the teacher's joint velocities and accelerations where the
hand is fully in view (terms below).

\paragraph{Loss terms.}
For a hand-frame, let $\hat{\mathbf J}$ be the student's MANO joints in
camera space, $\hat{\mathbf u}=\Pi_K(\hat{\mathbf J})$ their projection,
$\hat{\mathbf u}^{D}$ the direct 2D head, $\hat p_{3}$ and $\hat p_{2}$
the existence and visibility probabilities, and $\rho$ the Huber
loss~\cite{huber1964robust} with knee $0.05$.
With $s$ the target hand's 2D span ($\geq$40\,px) and
$s_{3D}=0.15$\,m,
\begin{align}
\ell_{2D}&=\bigl\langle\rho(\|\hat{\mathbf u}_j-\mathbf u_j\|/s)\bigr\rangle_{j\in\mathrm{img}}\nonumber\\
 &\quad+\bigl\langle\rho(\|\hat{\mathbf u}^D_j-\mathbf u_j\|/s)\bigr\rangle_{j\in\mathrm{img}},\\
\ell_{\mathrm{rel}}&=\bigl\langle\rho\bigl(\|(\hat{\mathbf J}_j-\hat{\mathbf J}_0)
 -(\mathbf J_j-\mathbf J_0)\|/s_{3D}\bigr)\bigr\rangle_j,\\
\ell_{\mathrm{root}}&=\rho(\|\hat{\mathbf J}_0-\mathbf J_0\|/s_{3D}),\\
\ell_{\mathrm{GO}}&=\|\hat R-R\|_F^2,
\end{align}
averaged over annotated hand-frames, and
$\ell_{\mathrm{exist}}=\mathrm{BCE}(\hat p_3,1)
+\mathrm{BCE}(\hat p_2,[\exists j\in\mathrm{img}])$.
The temporal term on frames where the target hand is fully inside the
image with a 16\,px margin is
\begin{equation}
\ell_{\mathrm{temp}}=\bigl\langle\rho(\|\Delta\hat{\mathbf J}-\Delta\mathbf J\|/\delta)\bigr\rangle
+\bigl\langle\rho(\|\Delta^2\hat{\mathbf J}-\Delta^2\mathbf J\|/\delta)\bigr\rangle
\end{equation}
with $\delta=5$\,mm and knee 1.
Teacher presence on frames without GT for that hand uses
$\mathrm{BCE}(\hat p_3,\tilde p)
+\mathrm{BCE}(\hat p_2,\tilde p\,[\exists j\in\mathrm{img}])$,
with no label where the teacher's hand is visible but within 16\,px of
the border.
On ARCTIC windows the teacher also supervises wrist-relative joints in
the hand's own frame and finger rotations, weighted per hand-frame by a
quality weight fixed before round~1: the product of a term falling with
the teacher's wrist acceleration and one falling with the distance
between the teacher's 2D joints and WiLoR's~\cite{wilor} where WiLoR
detects the hand (1 where it does not).

\begin{table}[t]
    \centering
    \caption{\textbf{Loop~B settings.}}
    \label{tab:loopB}
    \footnotesize
    \setlength{\tabcolsep}{4pt}
    \setlength{\aboverulesep}{0pt}
    \setlength{\belowrulesep}{0pt}
    \begin{tabular}{@{}l>{\raggedright\arraybackslash}p{0.58\linewidth}@{}}
        \toprule
        \rowcolor{HandRSIHeader}
        Setting & Value \\
        \midrule
        Trainable & rank-32 LoRA~\cite{lora} (DiT~\cite{dit} blocks) + heads \\
        Optimizer & AdamW~\cite{adamw}, $\beta=(0.9,0.99)$, no decay \\
        Learning rate & $10^{-4}$; LoRA+~\cite{loraplus} ratio 4; heads $\times0.1$ \\
        Schedule & 50 warm-up steps, cosine~\cite{sgdr}; EMA~\cite{polyak1992averaging} $1-5/\text{steps}$ \\
        Batch & 8 windows of 85 frames \\
        GT fine-tuning $\mathcal C^1_k$ & 500 (R1) / 2{,}000 (R2--R5) steps \\
        Student steps & 500 (R1) / 2{,}000, best of 4 ckpts on val \\
        Data mix & HOT3D 0.57, ARCTIC 0.43 (+ s03) \\
        $\lambda_{\mathrm{GT}},\lambda_{\mathrm{KD}},\lambda_e,\lambda_R$ & 2, 0.25, 0.5, 0.25 (R3: $\lambda_R$ 0.5) \\
        Temporal (GT / teacher) & 0.1 / 0.1 \\
        Latent noise & 0.02 \\
        \bottomrule
    \end{tabular}
\end{table}

\paragraph{Student and model gate.}
The student $\theta_k^+$ trains LoRA~\cite{lora} adapters and the prediction
heads with the settings of Tab.~\ref{tab:loopB}.
It is saved at several checkpoints, each compared, network alone, with
$\theta_k$ on val, and replaces $\theta_k$ only if it passes the model gate
$G_B(\theta_k^+;\theta_k)$, a significant gain of the network alone;
$\theta_{k+1}$ is then the best checkpoint without a blocking regression.

\begin{figure}[t]
    \centering
    \includegraphics[width=\linewidth]{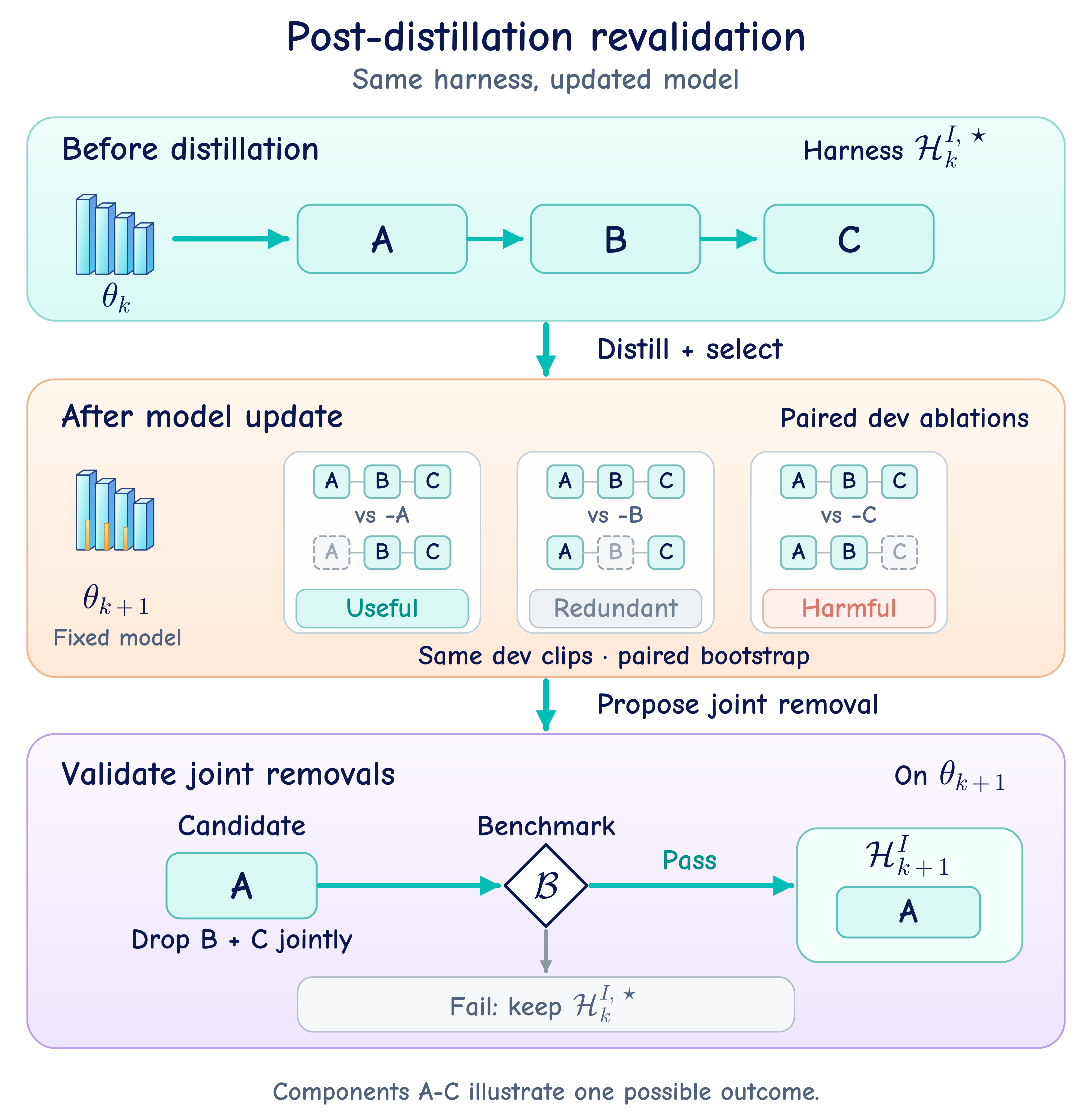}
    \caption{
        \textbf{Re-validation after a model update} (schematic).
        With $\theta_{k+1}$ fixed, paired dev ablations on the same clips
        reassess every module of $\mathcal H_k^{I,\star}$ as useful, redundant or
        harmful; the weak modules are removed jointly, and the pruned chain is
        kept only if it passes the benchmark's rule, otherwise
        $\mathcal H_k^{I,\star}$ is kept.
        Modules A--C illustrate one possible outcome.
    }
    \label{fig:reval}
\end{figure}

\paragraph{Re-validation.}
$\mathcal H_k^{I,\star}$ was written for $\theta_k$, and on the new network some of
its modules may correct errors that are gone (Fig.~\ref{fig:reval}).
On $\theta_{k+1}$'s dev captures, every module is compared with the chain
without it, using the gate's composite gain and paired bootstrap.
A module is weak if it is redundant ($L_\Delta\leq0$), harmful (a significant
regression in some column), or makes some dev recording's
EPE$_{\mathrm{2D}}$-p, CT-p or GO-p more than twice as bad; all weak modules
are removed jointly, and the full, pruned and empty chains are judged on dev
against $\mathcal S_k$ with the round gate's rule, which gives
$\mathcal H^I_{k+1}$.
The round accepts $\mathcal S_{k+1}=(\theta_{k+1},\smash[t]{\mathcal H^I_{k+1}})$ if it passes
$G_{1\%}$ against $\mathcal S_k$ on val; otherwise only the new harness is kept,
on the old network, provided it passes $G_A$.
When no student passes $G_B$, the network does not change and re-validation
is skipped.

\subsection{Loop C: Reflection}
\label{sec:loopC}

After each round a planning agent (Claude Opus~5.5) reads a read-only
dossier of the archive, \ie gate outcomes, aggregate val metrics, dev
diagnostics and every submission, but never test scores, and proposes the
next guidance
\begin{equation}
    \widehat{\pi}_{k+1}=\mathcal A_{\mathrm{plan}}(\pi_k,\mathcal M_{k+1}).
    \label{eq:guidance}
\end{equation}
The guidance consists of idea cards with dev evidence, one focus per Loop~A
agent, at most one page of lessons (a verbal memory of earlier rounds in the
spirit of Reflexion~\cite{reflexion}), papers to pre-fetch, and at most two
whitelisted Loop~B settings, which update $\mathcal H^S_{k+1}$; a validator
rejects any out-of-range item.
Changes to the pipeline itself are proposals that require an operator's
approval (Sec.~\ref{sec:data_use}), and Loop~C cannot change the evaluator or the gates, so
$\pi_{k+1}$ changes how candidates are found, never how they are judged.

Loop~C turns the archive into focused work.
After round~2, for example, it gave each agent a dev-grounded focus such as
``156 of 218 HOT3D dev misses are hands the chain places fully off-screen;
fix border placement before gating existence'', and three of its proposals
were adopted: a guard that removes any module making a single dev recording
more than twice as bad, gate scheduling with a WiLoR failure check before
scoring, and a revised brief for the integration agent.

\subsection{Deployment Tiers}
\label{sec:tiers}

After the loop we search sub-chains of the final harness, time those near
the accuracy--speed front $F$, and choose three tiers that maximize its
hypervolume~\cite{zitzler1999multiobjective} in $(\log\mathrm{FPS},\Delta)$
with respect to a reference point $\mathbf r$ below the front,
\begin{equation}
    T^\ast=\operatorname*{arg\,max}_{T\subseteq F,\,|T|=3}\mathrm{HV}(T;\mathbf r),
    \label{eq:tiers}
\end{equation}
using val and timing only.
Here $\mathrm{HV}(T;\mathbf r)$ is the area of the region of the objective plane
that some tier in $T$ dominates and that itself dominates $\mathbf r$.
Unlike a weighted score, the hypervolume needs no trade-off weight between
accuracy and speed, and it favors tiers spread along the front, from close
to the network's speed to close to the full system's accuracy.

\paragraph{Search.}
Starting from $\mathcal H^I_5$ on $\theta_2$, five cost levels restrict which
modules may stay (from modules below 25\,ms per segment up to all modules);
at each level, modules that do not help on dev are dropped, and a
continuation path then removes one module at a time, lowest dev gain per
millisecond first.
Every chain is scored on val against $\theta_2$ (Fig.~\ref{fig:supp_search}).

\paragraph{Selection.}
Configurations that are not $\varepsilon$-dominated~\cite{laumanns2002combining}
in $(\mathrm{FPS},\Delta)$, with $\varepsilon=1\%$ on FPS (our timing noise),
form the front $F$.
The reference point $\mathbf r$ is the nadir of the front moved out by half
its range, the setting recommended for selecting three solutions on a
two-objective front~\cite{ishibuchi2018reference}, and the 3-subset
maximization of Eq.~\eqref{eq:tiers} (hypervolume subset
selection~\cite{kuhn2016hypervolume,bringmann2014two}) is solved exactly by
enumeration.
The selected tiers are nested: Medium adds eight modules to Lite, and Heavy
adds the WiLoR-based module to Medium (Tab.~\ref{tab:tiers}), so a deployment
switches between them by enabling or disabling modules.
Heavy's val $\Delta$ ($+$4.93\%, $L_\Delta$ $+$4.24\%) exceeds the full chain's
($+$4.84\%, $+$4.10\%) by less than either interval's width, and the order
reverses on HOT3D test (8.41 vs.\ 8.37\,mm): the loop accepted modules one
submission at a time against the incumbent, whereas sub-chains are scored
jointly after the loop, so dropping ten modules costs nothing measurable on
val.

\paragraph{Timing.}
Each timed candidate is measured end to end (VAE encoding, network,
harness) on one 81-frame ARCTIC clip, five runs after a warm-up, on one idle
RTX PRO~5000, so the tiers report the speed that a deployment will see;
configurations timed twice agree within 0.3--1.7\%.

\section{Experimental Setup}
\label{sec:setup}

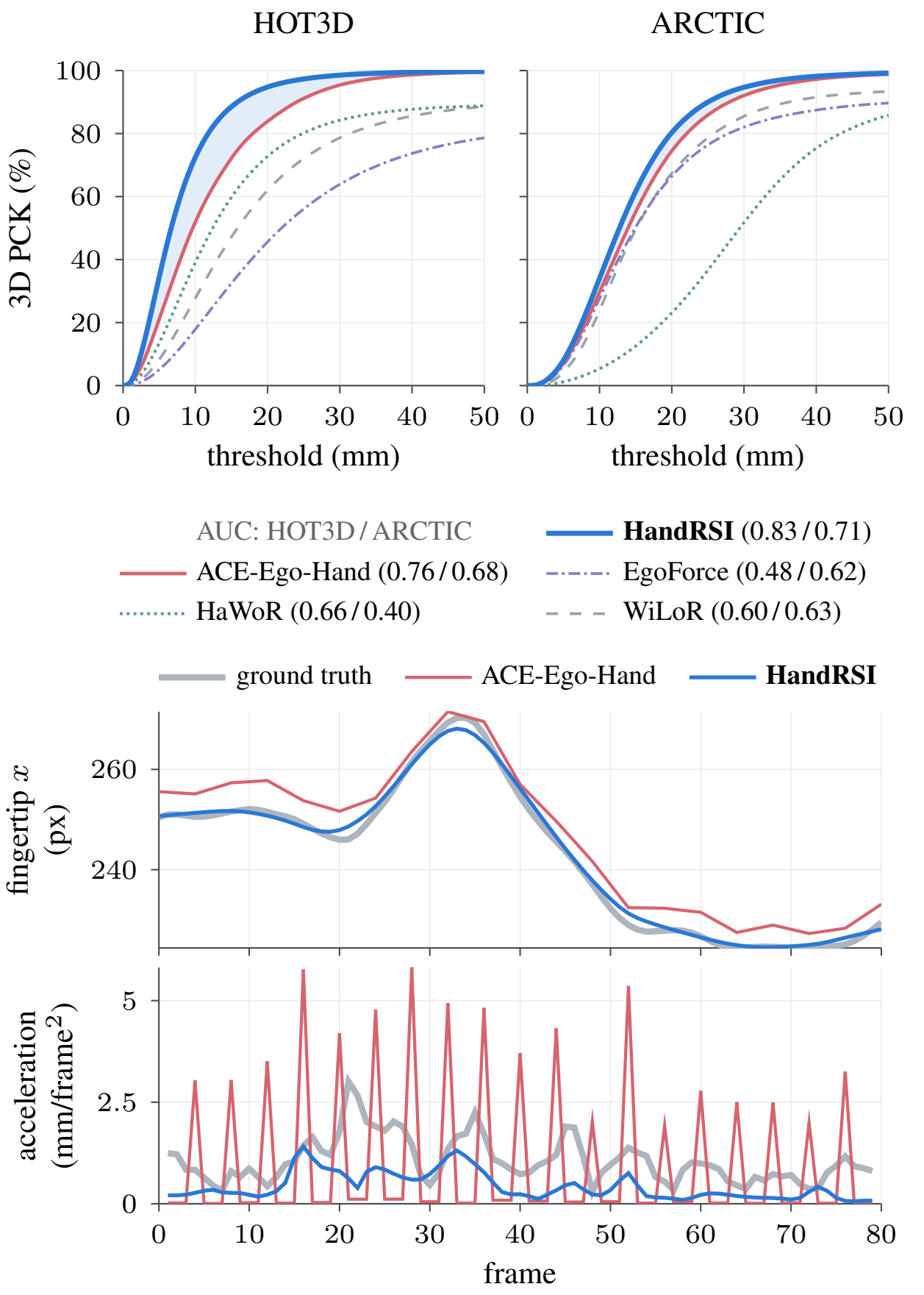
\begin{figure}[t]
    \centering
\begin{tikzpicture}
\begin{groupplot}[hr, group style={group size=2 by 1, horizontal sep=11pt, y descriptions at=edge left},
    width=0.39\linewidth, height=0.34\linewidth, xmin=0, xmax=50, ymin=0, ymax=100,
    xtick={0,10,...,50}, ytick={0,20,...,100}, xlabel={threshold (mm)},
    legend style={at={(1.04,-0.39)}, anchor=north, legend columns=2, /tikz/every even column/.append style={column sep=7pt}},
    reverse legend]
\nextgroupplot[title={HOT3D}, ylabel={3D PCK (\%)}]
\addplot[draw=none, name path=ours, forget plot] table[x=t, y=handrsi]{fig/data/pck_hot3d.dat};
\addplot[draw=none, name path=ace, forget plot] table[x=t, y=ace]{fig/data/pck_hot3d.dat};
\addplot[HandRSIBlue!13, forget plot] fill between[of=ours and ace];
\addplot[CWilor, line width=0.7pt, dashed] table[x=t, y=wilor]{fig/data/pck_hot3d.dat};
\addlegendentry{WiLoR (0.60\,/\,0.63)}
\addplot[CHawor, line width=0.75pt, densely dotted] table[x=t, y=hawor]{fig/data/pck_hot3d.dat};
\addlegendentry{HaWoR (0.66\,/\,0.40)}
\addplot[CEgo, line width=0.7pt, densely dashdotted] table[x=t, y=egoforce]{fig/data/pck_hot3d.dat};
\addlegendentry{EgoForce (0.48\,/\,0.62)}
\addplot[CAce, line width=0.9pt] table[x=t, y=ace]{fig/data/pck_hot3d.dat};
\addlegendentry{ACE-Ego-Hand (0.76\,/\,0.68)}
\addplot[HandRSIBlue, line width=1.3pt] table[x=t, y=handrsi]{fig/data/pck_hot3d.dat};
\addlegendentry{\textbf{HandRSI} (0.83\,/\,0.71)}
\addlegendimage{empty legend}\addlegendentry{\color{black!60}AUC: HOT3D\,/\,ARCTIC}
\nextgroupplot[title={ARCTIC}]
\addplot[draw=none, name path=ours, forget plot] table[x=t, y=handrsi]{fig/data/pck_arctic.dat};
\addplot[draw=none, name path=ace, forget plot] table[x=t, y=ace]{fig/data/pck_arctic.dat};
\addplot[HandRSIBlue!13, forget plot] fill between[of=ours and ace];
\addplot[CWilor, line width=0.7pt, dashed] table[x=t, y=wilor]{fig/data/pck_arctic.dat};
\addplot[CHawor, line width=0.75pt, densely dotted] table[x=t, y=hawor]{fig/data/pck_arctic.dat};
\addplot[CEgo, line width=0.7pt, densely dashdotted] table[x=t, y=egoforce]{fig/data/pck_arctic.dat};
\addplot[CAce, line width=0.9pt] table[x=t, y=ace]{fig/data/pck_arctic.dat};
\addplot[HandRSIBlue, line width=1.3pt] table[x=t, y=handrsi]{fig/data/pck_arctic.dat};
\end{groupplot}
\end{tikzpicture}\\[3pt]
\begin{tikzpicture}
\begin{groupplot}[hr, group style={group size=1 by 2, vertical sep=5pt, x descriptions at=edge bottom},
    width=0.78\linewidth, height=0.255\linewidth, xmin=0, xmax=80, xtick={0,10,...,80}, unbounded coords=jump,
    ylabel style={align=center}]
\nextgroupplot[ylabel={fingertip $x$\\(px)}, ytick={240,260},
    legend style={at={(0.5,1.04)}, anchor=south, legend columns=3, /tikz/every even column/.append style={column sep=6pt}}]
\addplot[CGray!70, line width=1.6pt] table[x=t, y=u_gt]{fig/data/temporal.dat}; \addlegendentry{ground truth}
\addplot[CAce, line width=0.8pt] table[x=t, y=u_ace]{fig/data/temporal.dat}; \addlegendentry{ACE-Ego-Hand}
\addplot[HandRSIBlue, line width=1.0pt] table[x=t, y=u_ours]{fig/data/temporal.dat}; \addlegendentry{\textbf{HandRSI}}
\nextgroupplot[ylabel={acceleration\\(mm/frame$^2$)}, xlabel={frame}, ymin=0, ytick={0,2.5,5}]
\addplot[CGray!70, line width=1.6pt] table[x=t, y=a_gt]{fig/data/temporal.dat};
\addplot[CAce, line width=0.8pt] table[x=t, y=a_ace]{fig/data/temporal.dat};
\addplot[HandRSIBlue, line width=1.0pt] table[x=t, y=a_ours]{fig/data/temporal.dat};
\end{groupplot}
\end{tikzpicture}
    \caption{
        \textbf{Top: 3D PCK} on the sealed tests (wrist-aligned; missed hands count as errors); legend: area
        under the curve up to 50\,mm, HOT3D\,/\,ARCTIC; shaded: HandRSI's margin over \mbox{ACE-Ego-Hand}.
        \textbf{Bottom: temporal stability} on one HOT3D test clip, image position of the index fingertip and
        per-frame joint acceleration: ACE-Ego-Hand's acceleration peaks every four frames, HandRSI follows the
        ground truth without these steps (clip jitter 4.83$\to$1.70\,mm/frame$^2$).
    }
    \label{fig:pck}
    \label{fig:temporal}
\end{figure}

\begin{table*}[t]
    \centering
    \caption{
        \textbf{Comparison with existing methods} on the sealed test sets, all
        re-run by us on the same segments, intrinsics and evaluator; the strongest
        baselines are shown (all ten re-run configurations in Tab.~\ref{tab:all}, none of the
        omitted ones is best in any column).
        Ours: the network alone, the three deployment tiers and the full system, with
        end-to-end FPS on one RTX PRO~5000 (Heavy and the full system also run WiLoR);
        ACE-Ego-Hand (K-given) is our starting point $\theta_0$; controls: $\theta_0$
        fine-tuned on our GT ($\mathcal C^1_1$) and the final harness on $\theta_0$ ($\mathcal C^2_5$).
        MPJPE-p, PA-p, MPJPE$^{+\mathrm{OOS}}$ in mm; EPE$_{\mathrm{2D}}$-p in px;
        GO-p in degrees; CT-p in m; Jitter in mm/frame$^2$; best in bold.
    }
    \label{tab:main}
    \footnotesize
    \setlength{\tabcolsep}{3.2pt}
    \setlength{\aboverulesep}{0pt}
    \setlength{\belowrulesep}{0pt}
    \begin{tabular}{@{}l*{10}{c}@{}}
        \toprule
        \rowcolor{HandRSIHeader}
        & \multicolumn{3}{c}{\textbf{Detection}}
        & \multicolumn{3}{c}{\textbf{3D pose}}
        & \multicolumn{3}{c}{\textbf{Placement}}
        & \textbf{Temporal} \\
        \rowcolor{HandRSIHeader}
        \textbf{Method}
        & FAcc$\uparrow$ & Recall$\uparrow$ & F1$\uparrow$
        & MPJPE-p$\downarrow$ & PA-p$\downarrow$ & MPJPE$^{+\mathrm{OOS}}{\downarrow}$
        & EPE$_{\mathrm{2D}}$-p$\downarrow$ & GO-p$\downarrow$ & CT-p$\downarrow$
        & Jitter$\downarrow$ \\
        \midrule
        \multicolumn{11}{l}{\textit{HOT3D (27 test recordings, 1{,}162 segments)}} \\
        WiLoR~\cite{wilor} & 0.861 & 0.911 & 0.949 & 26.82 & 7.44 & 41.67 & 61.48 & 22.75 & 0.082 & 16.22 \\
        \cmidrule[0.2pt](lr){1-11}
        HaWoR~\cite{hawor} & 0.835 & 0.895 & 0.942 & 24.94 & 6.92 & 40.12 & 55.05 & 22.20 & 0.072 & 11.86 \\
        EgoForce~\cite{egoforce} & 0.781 & 0.858 & 0.917 & 38.52 & 7.43 & 53.01 & 88.39 & 33.99 & 0.105 & 18.34 \\
        ACE-Ego-Hand, K-free~\cite{ace_ego_hand} & 0.922 & 0.952 & 0.961 & 17.34 & 6.38 & 13.61 & 21.20 & 12.21 & 0.045 & 2.59 \\
        ACE-Ego-Hand, K-given~\cite{ace_ego_hand} & 0.970 & 0.986 & 0.990 & 12.97 & 5.90 & 12.84 & 9.66 & 6.90 & 0.024 & 2.82 \\
        \cmidrule[0.2pt](lr){1-11}
        ACE-Ego-Hand fine-tuned on our GT ($\mathcal C^1_1$) & 0.982 & 0.994 & 0.994 & 9.91 & 5.00 & 10.76 & 5.56 & 5.71 & 0.014 & 2.77 \\
        HandRSI network ($\theta_2$), 42.5 FPS & 0.982 & 0.994 & 0.994 & 9.37 & 4.71 & 10.35 & 5.31 & 5.80 & 0.013 & 2.51 \\
        \rowcolor{HandRSILite}
        HandRSI-Lite, 41.1 FPS & 0.982 & \textbf{0.996} & 0.994 & 9.06 & 4.69 & 10.31 & 4.76 & 5.44 & \textbf{0.012} & 1.47 \\
        \rowcolor{HandRSIMedium}
        HandRSI-Medium, 34.0 FPS & 0.982 & \textbf{0.996} & 0.994 & 9.04 & 4.64 & 10.30 & 4.79 & 5.46 & 0.013 & \textbf{1.33} \\
        \rowcolor{HandRSIHeavy}
        HandRSI-Heavy, 20.7 FPS & \textbf{0.983} & \textbf{0.996} & \textbf{0.995} & 8.41 & \textbf{4.35} & 9.71 & 4.16 & 5.17 & 0.013 & 1.35 \\
        \rowcolor{HandRSIHeavy}
        \textbf{HandRSI} ($\mathcal S_5$), 16.2 FPS & 0.982 & \textbf{0.996} & 0.994 & \textbf{8.37} & 4.37 & \textbf{9.70} & \textbf{4.07} & \textbf{5.16} & 0.013 & 1.34 \\
        \midrule
        \multicolumn{11}{l}{\textit{ARCTIC (held-out subject s05, 291 segments)}} \\
        WiLoR~\cite{wilor} & 0.906 & 0.945 & 0.971 & 20.80 & 7.48 & 29.12 & 52.23 & 18.38 & 0.068 & 23.55 \\
        \cmidrule[0.2pt](lr){1-11}
        HaWoR~\cite{hawor} & 0.880 & 0.932 & 0.963 & 33.62 & 9.24 & 41.02 & 59.60 & 27.65 & 0.083 & 13.37 \\
        EgoForce~\cite{egoforce} & 0.860 & 0.920 & 0.956 & 23.14 & 8.82 & 31.57 & 66.62 & 22.19 & 0.076 & 14.87 \\
        ACE-Ego-Hand, K-free~\cite{ace_ego_hand} & 0.969 & 0.984 & 0.987 & 16.61 & 7.71 & 16.43 & 16.77 & 13.99 & 0.030 & 2.45 \\
        ACE-Ego-Hand, K-given~\cite{ace_ego_hand} & 0.986 & 0.995 & 0.996 & 15.58 & 7.43 & 16.54 & 9.36 & 12.45 & 0.024 & 2.58 \\
        \cmidrule[0.2pt](lr){1-11}
        ACE-Ego-Hand fine-tuned on our GT ($\mathcal C^1_1$) & 0.988 & 0.996 & 0.997 & 15.51 & 7.36 & 16.49 & 9.28 & 12.40 & 0.023 & 2.57 \\
        HandRSI network ($\theta_2$), 42.5 FPS & 0.987 & 0.995 & 0.997 & 15.15 & 7.28 & 16.11 & 9.12 & 12.34 & 0.021 & 2.36 \\
        \rowcolor{HandRSILite}
        HandRSI-Lite, 41.1 FPS & 0.990 & 0.999 & 0.997 & 14.53 & 7.25 & 15.84 & 8.80 & 11.76 & \textbf{0.019} & 1.58 \\
        \rowcolor{HandRSIMedium}
        HandRSI-Medium, 34.0 FPS & 0.990 & 0.999 & 0.997 & 14.47 & 7.14 & 15.78 & 8.78 & 11.76 & \textbf{0.019} & \textbf{1.45} \\
        \rowcolor{HandRSIHeavy}
        HandRSI-Heavy, 20.7 FPS & \textbf{0.991} & \textbf{1.000} & \textbf{0.998} & 14.00 & 6.44 & 15.33 & 7.69 & \textbf{11.59} & \textbf{0.019} & 1.46 \\
        \rowcolor{HandRSIHeavy}
        \textbf{HandRSI} ($\mathcal S_5$), 16.2 FPS & 0.990 & \textbf{1.000} & 0.997 & \textbf{13.85} & \textbf{6.34} & \textbf{15.18} & \textbf{7.61} & 11.61 & \textbf{0.019} & \textbf{1.45} \\
        \midrule
        \multicolumn{11}{l}{\textit{HOI4D (zero-shot, 498 segments)}} \\
        WiLoR~\cite{wilor} & 0.978 & 0.976 & 0.988 & 34.10 & 10.37 & 34.11 & 30.59 & 25.36 & 0.105 & 16.44 \\
        HaWoR~\cite{hawor} & 0.985 & 0.984 & 0.991 & 33.04 & 11.43 & 32.93 & 22.73 & 25.66 & 0.063 & 16.76 \\
        ACE-Ego-Hand, K-free~\cite{ace_ego_hand} & \textbf{0.996} & \textbf{0.996} & \textbf{0.997} & 23.14 & 11.14 & 22.82 & 17.42 & \textbf{18.33} & \textbf{0.050} & 2.22 \\
        ACE-Ego-Hand, K-given~\cite{ace_ego_hand} & 0.994 & 0.994 & 0.994 & 23.45 & 11.02 & 22.93 & \textbf{15.60} & 19.39 & 0.058 & 2.10 \\
        \cmidrule[0.2pt](lr){1-11}
        ACE-Ego-Hand fine-tuned on our GT ($\mathcal C^1_1$) & 0.994 & 0.994 & 0.994 & 23.83 & 11.01 & 23.29 & 15.71 & 19.35 & 0.062 & 2.07 \\
        Final harness on ACE-Ego-Hand ($\mathcal C^2_5$) & 0.993 & 0.993 & 0.994 & \textbf{22.24} & \textbf{10.03} & \textbf{21.59} & 16.13 & 19.22 & 0.058 & 1.05 \\
        HandRSI network ($\theta_2$), 42.5 FPS & 0.992 & 0.992 & 0.993 & 24.11 & 11.13 & 23.42 & 17.39 & 19.57 & 0.064 & 1.87 \\
        \rowcolor{HandRSILite}
        HandRSI-Lite, 41.1 FPS & 0.991 & 0.991 & 0.992 & 24.03 & 11.13 & 23.18 & 19.86 & 19.68 & 0.065 & 1.14 \\
        \rowcolor{HandRSIMedium}
        HandRSI-Medium, 34.0 FPS & 0.991 & 0.991 & 0.992 & 23.93 & 10.98 & 23.07 & 19.83 & 19.68 & 0.065 & 1.03 \\
        \rowcolor{HandRSIHeavy}
        HandRSI-Heavy, 20.7 FPS & 0.991 & 0.991 & 0.992 & 22.89 & 10.32 & 22.02 & 18.75 & 19.48 & 0.065 & 1.02 \\
        \rowcolor{HandRSIHeavy}
        \textbf{HandRSI} ($\mathcal S_5$), 16.2 FPS & 0.991 & 0.991 & 0.992 & 22.77 & 10.21 & 21.93 & 18.61 & 19.44 & 0.064 & \textbf{1.01} \\
 
        \bottomrule
    \end{tabular}
\end{table*}

\begin{figure*}[t]
    \centering
    {\small
\setlength{\qw}{\dimexpr(\linewidth-1em-10.5pt)/5\relax}%
\begin{tabular}{@{}c@{\hspace{2pt}}c@{\hspace{2pt}}c@{\hspace{2pt}}c@{\hspace{2pt}}c@{\hspace{2pt}}c@{}}
 & Input + ground truth & WiLoR~\cite{wilor} & HaWoR~\cite{hawor} & ACE-Ego-Hand~\cite{ace_ego_hand} & \textbf{\color{HandRSIBlue}HandRSI (ours)} \\[1pt]
\qrow{HOT3D} & \qtile{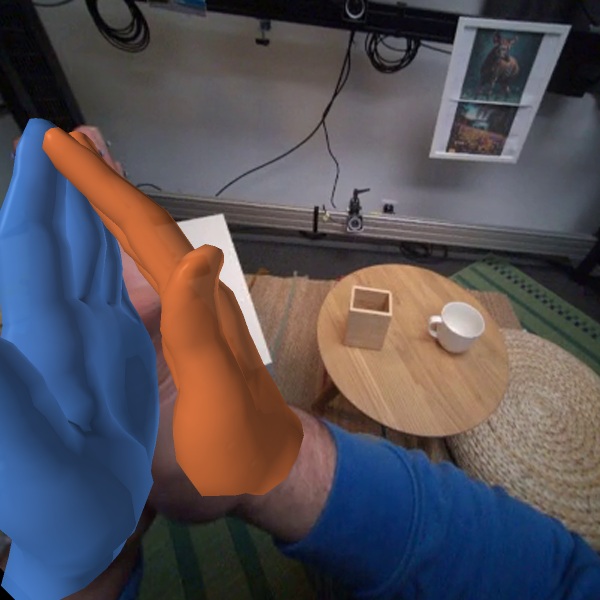}{}{QTag} & \qtile{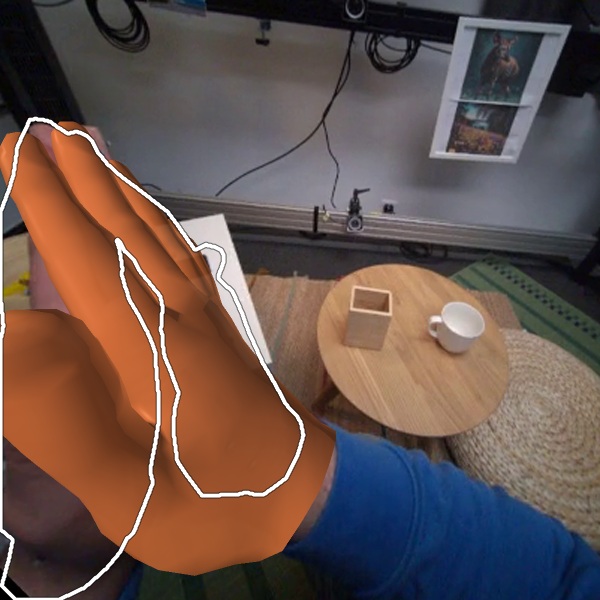}{missed}{QTag} & \qtile{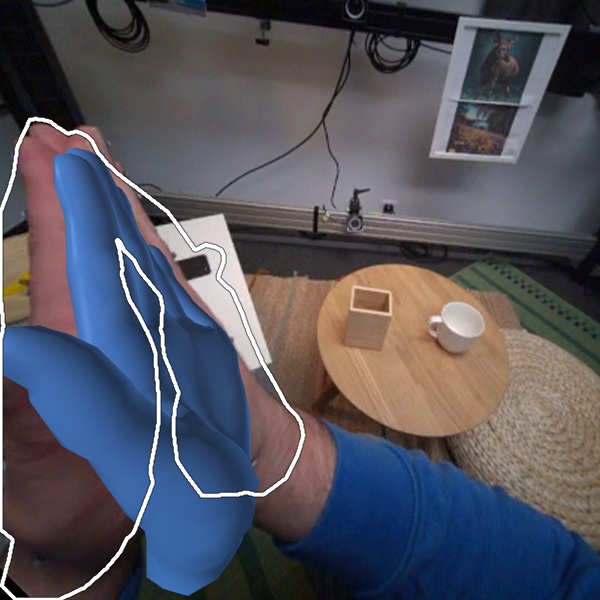}{45\,px}{QTag} & \qtile{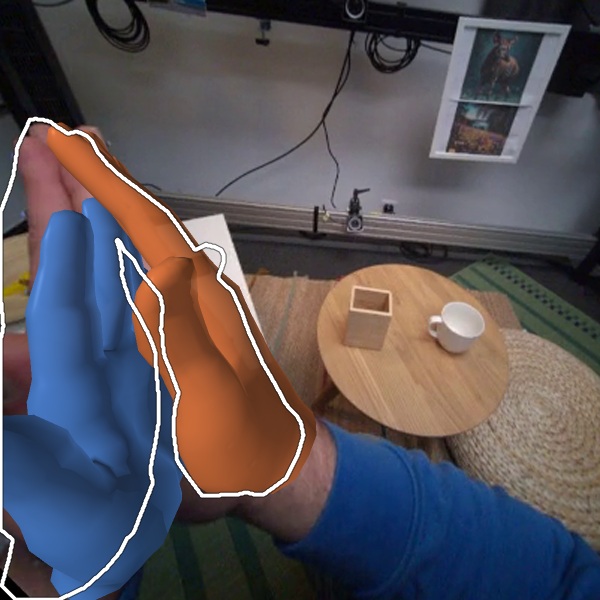}{39\,px}{QTag} & \qtile{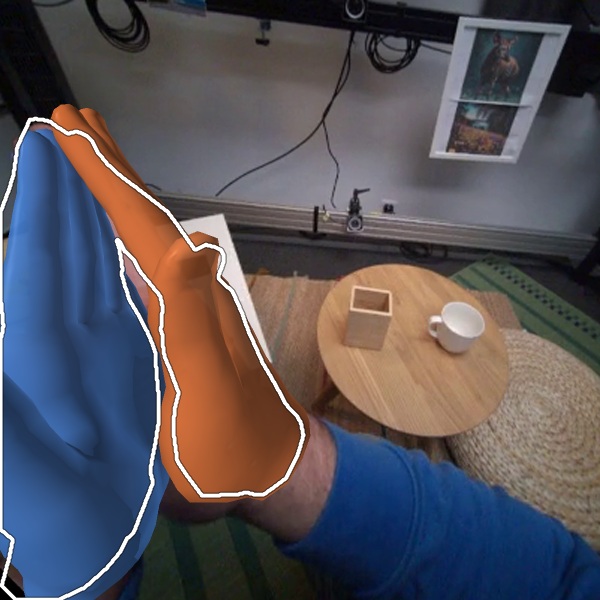}{9\,px}{HandRSIBlue} \\[2pt]
\qrow{HOT3D} & \qtile{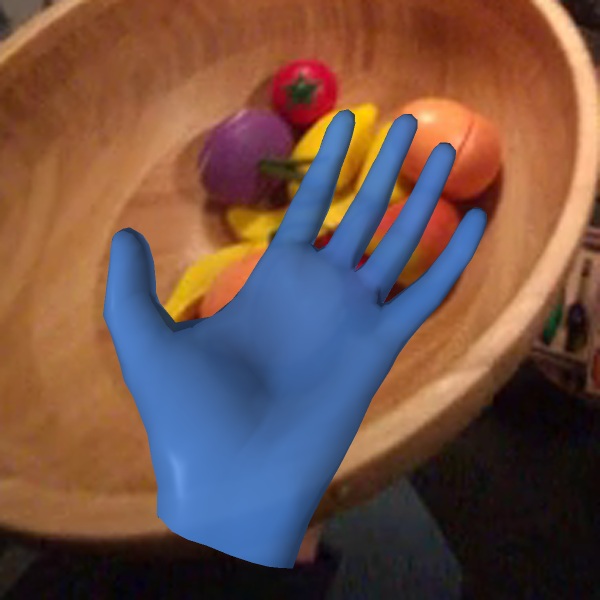}{}{QTag} & \qtile{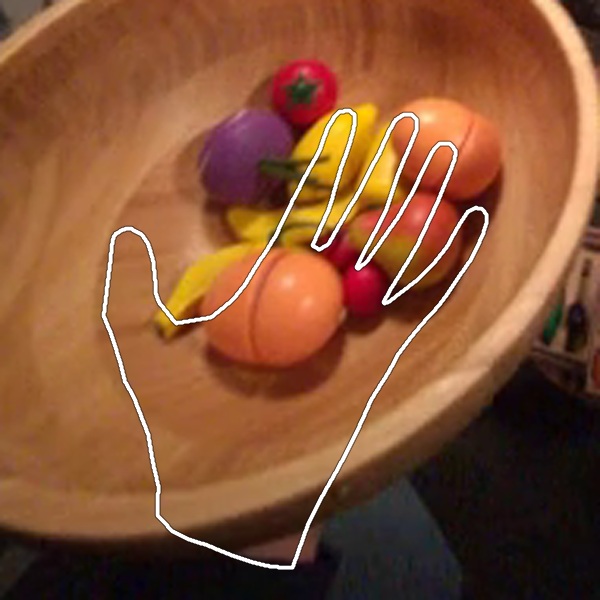}{missed}{QTag} & \qtile{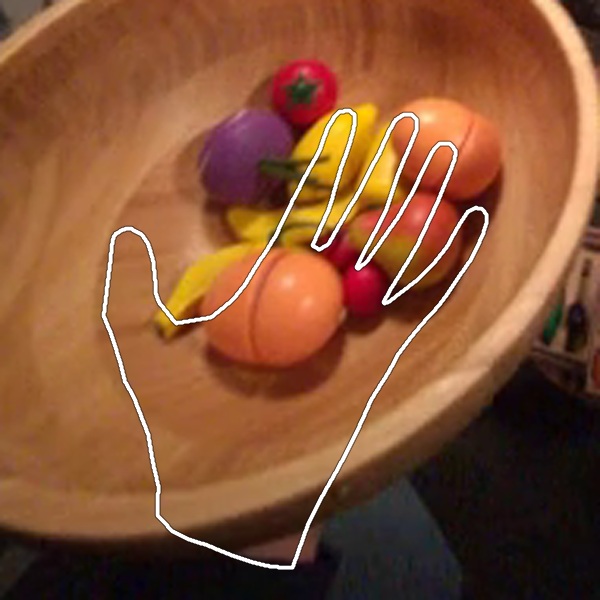}{missed}{QTag} & \qtile{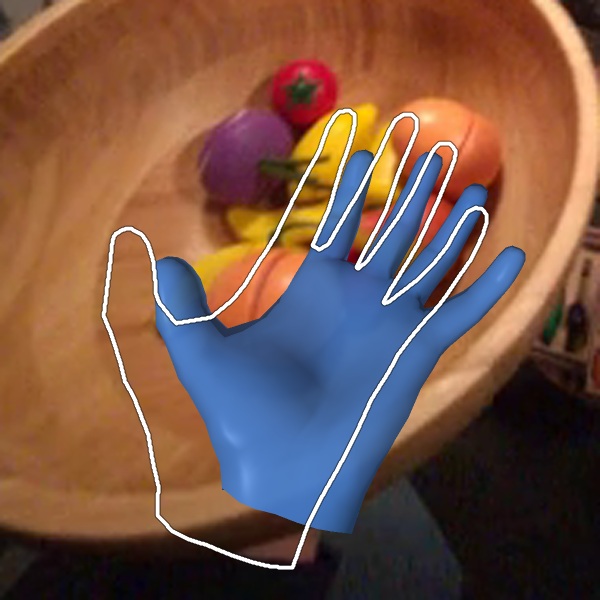}{14\,px}{QTag} & \qtile{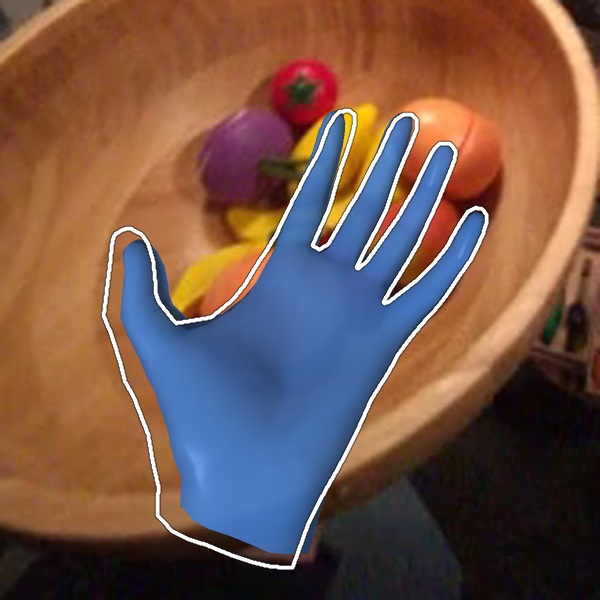}{2\,px}{HandRSIBlue} \\[2pt]
\qrow{ARCTIC} & \qtile{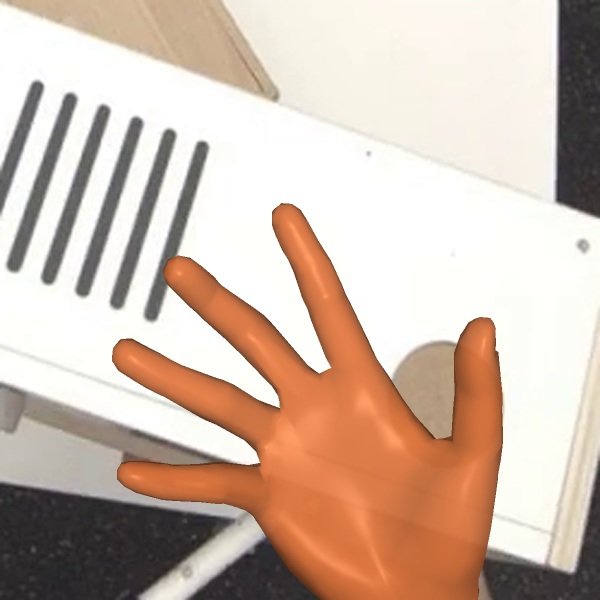}{}{QTag} & \qtile{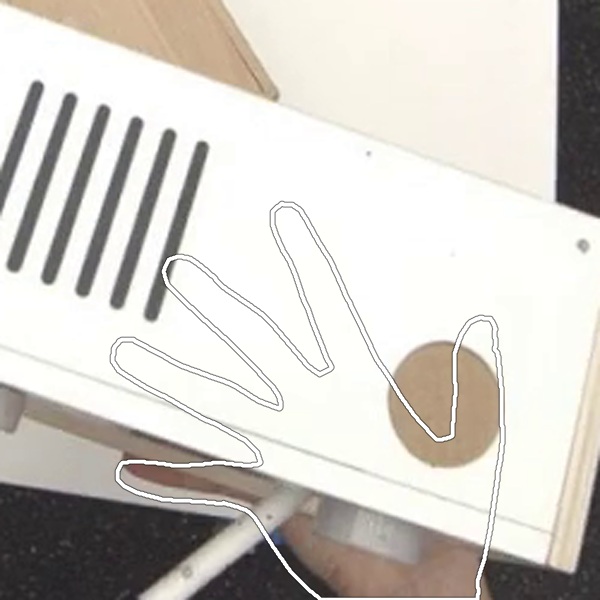}{missed}{QTag} & \qtile{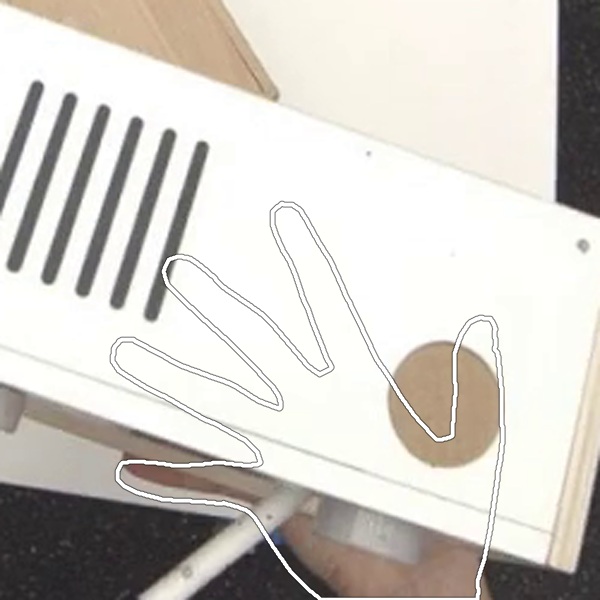}{missed}{QTag} & \qtile{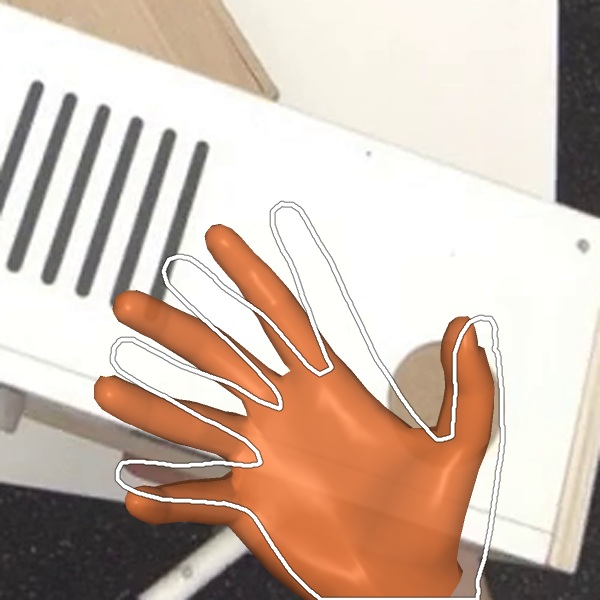}{15\,px}{QTag} & \qtile{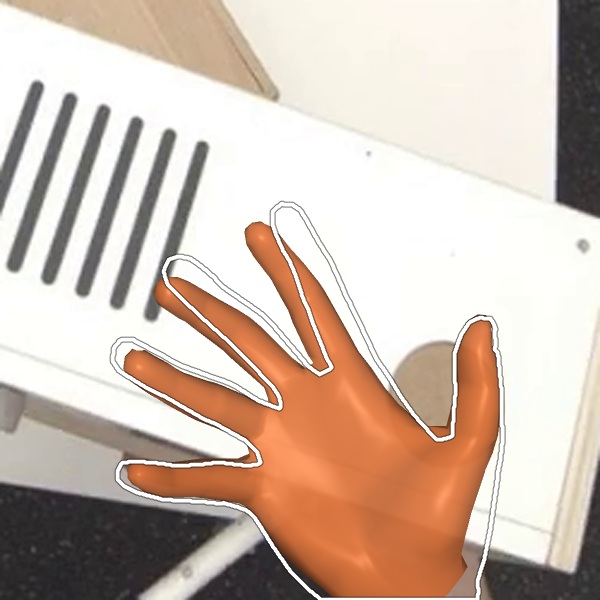}{6\,px}{HandRSIBlue} \\[2pt]
\qrow{ARCTIC} & \qtile{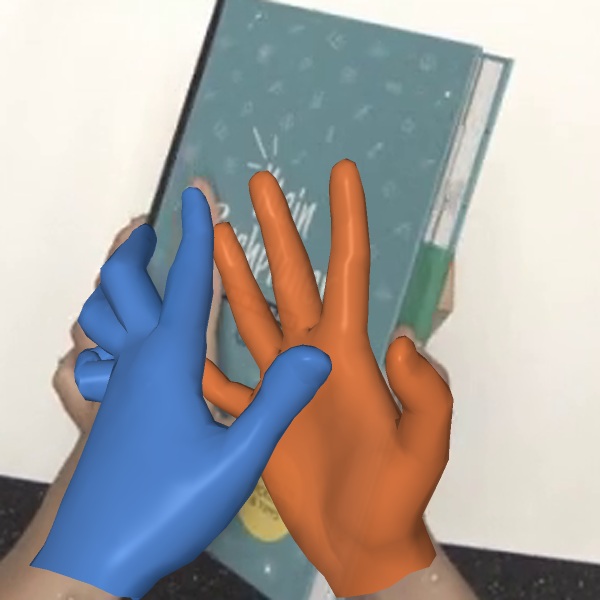}{}{QTag} & \qtile{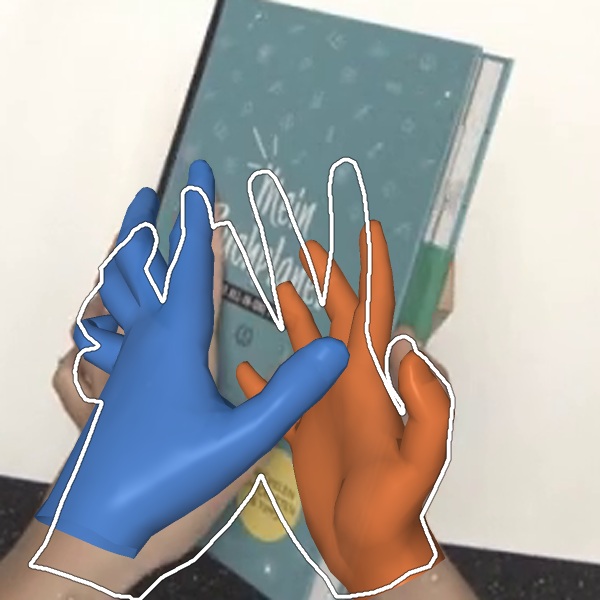}{26\,px}{QTag} & \qtile{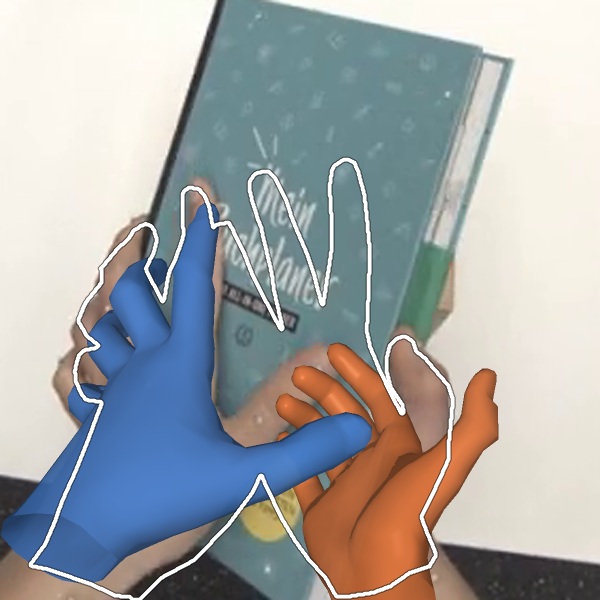}{61\,px}{QTag} & \qtile{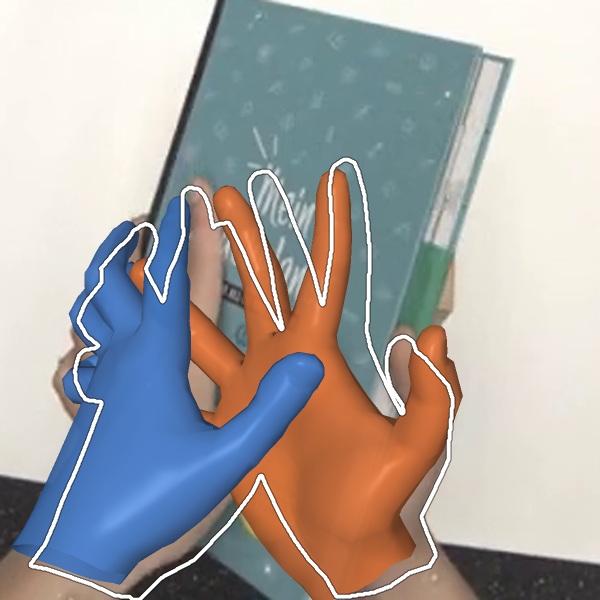}{20\,px}{QTag} & \qtile{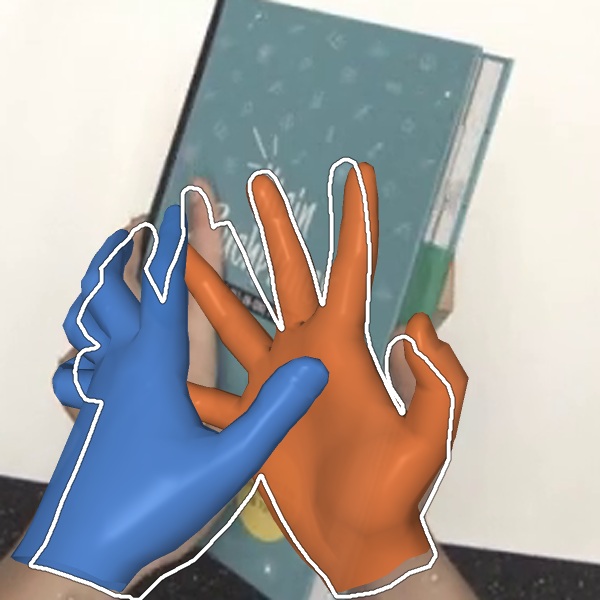}{9\,px}{HandRSIBlue} \\
\end{tabular}
}
    \caption{
        \textbf{Comparison with the strongest baselines} on the sealed tests (same crop per row; white
        outline: ground truth; tag: 2D error in that frame). The baselines miss occluded or close hands and
        misplace interacting ones; HandRSI recovers and aligns them.
    }
    \label{fig:qualitative}
\end{figure*}

\subsection{Data and Splits}
\label{sec:data}

We follow the K-given protocol of ACE-Ego-Hand~\cite{ace_ego_hand} on
81-frame segments, with missed hands penalized, and evaluate on
ARCTIC~\cite{arctic} (held-out subject s05, ACE-Ego-Hand's test set),
HOT3D~\cite{hot3d} (the HaWoR test split~\cite{hawor}, as ACE-Ego-Hand's is
not public) and, zero-shot, HOI4D~\cite{hoi4d}.
The remaining data form train, dev and val splits (Tab.~\ref{tab:splits}).
Splits are by sequence (HOT3D, HOI4D) or subject (ARCTIC) and do not
overlap; the agents' sandbox mounts only the dev views.
HOT3D recordings without public GT and ARCTIC subject s03 (official ARCTIC
test subject, GT withheld) are used only as unlabeled training video for the
teacher terms of Loop~B; HOI4D is used only for testing.

\begin{table}[t]
    \centering
    \caption{\textbf{Splits} (sequences; segments of 81 frames).}
    \label{tab:splits}
    \small
    \setlength{\tabcolsep}{3pt}
    \resizebox{\linewidth}{!}{%
    \setlength{\aboverulesep}{0pt}
    \setlength{\belowrulesep}{0pt}
    \begin{tabular}{@{}lllll@{}}
        \toprule
        \rowcolor{HandRSIHeader}
        & train & dev & val & test (sealed) \\
        \midrule
        HOT3D & 81 seq. + 62 w/o GT & 15 seq. & 13 seq. & 27 seq., 1{,}162 seg. \\
        ARCTIC & 6 subjects + s03 w/o GT & 1 subject & 1 subject & s05, 291 seg. \\
        HOI4D & -- & -- & -- & 166 rec., 498 seg. \\
        \bottomrule
    \end{tabular}}
\end{table}

\subsection{Evaluator and Metrics}
\label{sec:eval}

We report \mbox{ACE-Ego-Hand's} ten metrics: detection (FAcc, Recall, F1), 3D
pose (\mbox{MPJPE-p}, \mbox{PA-p}, MPJPE$^{+\mathrm{OOS}}$), placement
(\mbox{EPE$_{\mathrm{2D}}$-p}, \mbox{GO-p}, \mbox{CT-p}) and Jitter.

\paragraph{Evaluator.}
ACE-Ego-Hand's evaluator is not public, so we implemented its
Appendix~A.1: a hand is on-screen if any GT joint projects into the
image at depth above 1\,cm; a prediction is active if its 3D existence
score exceeds 0.5 and matches a same-side on-screen GT hand by overlap of
the projected mesh box with the GT box dilated by 10\%; every missed
on-screen hand is charged a canonical MANO~\cite{mano} hand at the camera origin;
PA-p aligns each matched hand to the GT by the least-squares similarity
transform~\cite{umeyama1991} before averaging joint errors;
EPE$_{\mathrm{2D}}$-p charges missed in-image joints the image diagonal;
Jitter is the mean joint acceleration over matched runs of $\geq$3
frames.
On ACE-Ego-Hand's own outputs, an independent
re-implementation agrees with ours within 0.6\% on nine of ten columns;
EPE$_{\mathrm{2D}}$-p differs because the paper does not state whether the
miss penalty is per hand or per joint (the matched part agrees,
6.77 vs.\ 6.76\,px).
Our re-run of ACE-Ego-Hand reproduces its published pose metrics, but not
its detection metrics: on HOT3D, 72\% of the hands it misses have
1--3 joints in the image, and the network marks them
absent.

\paragraph{Detection rule.}
All tables use the rule of the released evaluator: a prediction on a hand
slot whose GT is not annotated in that frame is neither a true nor a false
positive.
HOT3D has untracked stretches in which the hand is plainly visible (one
test recording leaves 44\% of its hand slots unannotated), and all twelve
such ``false positives'' we inspected were real hands; off-screen hands keep
their GT, so a hand pushed into the image by a wrong depth still counts.
Only FAcc and F1 depend on the rule: Table~\ref{tab:strict} in
Appendix~\ref{sec:app_results} gives HOT3D under the strict rule, which charges every prediction on an unannotated slot, with
Recall and all pose columns unchanged.

\subsection{Baselines}
\label{sec:baselines}

\begin{table*}[t]
    \centering
    \caption{\textbf{Every re-run method} on the HOT3D and ARCTIC tests (Tab.~\ref{tab:main} shows the strongest
    ones). Units as in Tab.~\ref{tab:main}. $^{a}$OmniHands' HOT3D jitter comes from 0.17\% of hand-frames where its PnP
    placement diverges.}
    \label{tab:all}
    \scriptsize
    \setlength{\tabcolsep}{4pt}
    \renewcommand{\arraystretch}{0.92}
    \setlength{\aboverulesep}{0pt}
    \setlength{\belowrulesep}{0pt}
    \begin{tabular}{@{}l*{10}{c}@{}}
        \toprule
        \rowcolor{HandRSIHeader}
        \textbf{Method} & FAcc$\uparrow$ & Recall$\uparrow$ & F1$\uparrow$ & MPJPE-p$\downarrow$ & PA-p$\downarrow$ & MPJPE$^{+\mathrm{OOS}}{\downarrow}$
        & EPE$_{\mathrm{2D}}$-p$\downarrow$ & GO-p$\downarrow$ & CT-p$\downarrow$ & Jitter$\downarrow$ \\
        \midrule
        \multicolumn{11}{l}{\textit{HOT3D (27 test recordings, 1{,}162 segments)}} \\
        HaMeR~\cite{hamer} & 0.735 & 0.890 & 0.895 & 64.51 & 10.07 & 73.03 & 64.86 & 47.78 & 0.095 & 19.47 \\
        Hamba~\cite{hamba} & 0.510 & 0.739 & 0.768 & 72.79 & 9.83 & 79.84 & 169.64 & 64.16 & 0.150 & 14.73 \\
        WiLoR~\cite{wilor} & 0.861 & 0.911 & 0.949 & 26.82 & 7.44 & 41.67 & 61.48 & 22.75 & 0.082 & 16.22 \\
        WildHands~\cite{wildhands} & 0.710 & 0.868 & 0.873 & 51.59 & 12.13 & 56.88 & 102.30 & 49.61 & 0.140 & 23.29 \\
        InterWild~\cite{interwild} & 0.729 & 0.885 & 0.891 & 72.16 & 11.10 & 80.78 & 69.45 & 54.14 & 0.110 & 50.93 \\
        \cmidrule[0.2pt](lr){1-11}
        HaWoR~\cite{hawor} & 0.835 & 0.895 & 0.942 & 24.94 & 6.92 & 40.12 & 55.05 & 22.20 & 0.072 & 11.86 \\
        OmniHands~\cite{omnihands} & 0.583 & 0.874 & 0.819 & 61.35 & 10.35 & 63.74 & 87.27 & 49.12 & 0.215 & 327.39$^{a}$ \\
        EgoForce~\cite{egoforce} & 0.781 & 0.858 & 0.917 & 38.52 & 7.43 & 53.01 & 88.39 & 33.99 & 0.105 & 18.34 \\
        ACE-Ego-Hand, K-free~\cite{ace_ego_hand} & 0.922 & 0.952 & 0.961 & 17.34 & 6.38 & 13.61 & 21.20 & 12.21 & 0.045 & 2.59 \\
        ACE-Ego-Hand, K-given~\cite{ace_ego_hand} & 0.970 & 0.986 & 0.990 & 12.97 & 5.90 & 12.84 & 9.66 & 6.90 & 0.024 & 2.82 \\
        \cmidrule[0.2pt](lr){1-11}
        HandRSI network ($\theta_2$), 42.5 FPS & 0.982 & 0.994 & 0.994 & 9.37 & 4.71 & 10.35 & 5.31 & 5.80 & 0.013 & 2.51 \\
        \rowcolor{HandRSILite}
        HandRSI-Lite, 41.1 FPS & 0.982 & \textbf{0.996} & 0.994 & 9.06 & 4.69 & 10.31 & 4.76 & 5.44 & \textbf{0.012} & 1.47 \\
        \rowcolor{HandRSIMedium}
        HandRSI-Medium, 34.0 FPS & 0.982 & \textbf{0.996} & 0.994 & 9.04 & 4.64 & 10.30 & 4.79 & 5.46 & 0.013 & \textbf{1.33} \\
        \rowcolor{HandRSIHeavy}
        HandRSI-Heavy, 20.7 FPS & \textbf{0.983} & \textbf{0.996} & \textbf{0.995} & 8.41 & \textbf{4.35} & 9.71 & 4.16 & 5.17 & 0.013 & 1.35 \\
        \rowcolor{HandRSIHeavy}
        \textbf{HandRSI} ($\mathcal S_5$), 16.2 FPS & 0.982 & \textbf{0.996} & 0.994 & \textbf{8.37} & 4.37 & \textbf{9.70} & \textbf{4.07} & \textbf{5.16} & 0.013 & 1.34 \\
        \midrule
        \multicolumn{11}{l}{\textit{ARCTIC (held-out subject s05, 291 segments)}} \\
        HaMeR~\cite{hamer} & 0.869 & 0.937 & 0.955 & 29.71 & 9.94 & 37.02 & 55.53 & 25.61 & 0.096 & 17.93 \\
        Hamba~\cite{hamba} & 0.766 & 0.892 & 0.909 & 33.15 & 10.14 & 39.83 & 89.84 & 30.76 & 0.119 & 15.20 \\
        WiLoR~\cite{wilor} & 0.906 & 0.945 & 0.971 & 20.80 & 7.48 & 29.12 & 52.23 & 18.38 & 0.068 & 23.55 \\
        WildHands~\cite{wildhands} & 0.870 & 0.938 & 0.956 & 26.66 & 9.67 & 33.19 & 54.60 & 23.68 & 0.064 & 12.71 \\
        InterWild~\cite{interwild} & 0.865 & 0.934 & 0.953 & 31.31 & 11.34 & 38.17 & 60.18 & 26.55 & 0.092 & 39.55 \\
        \cmidrule[0.2pt](lr){1-11}
        HaWoR~\cite{hawor} & 0.880 & 0.932 & 0.963 & 33.62 & 9.24 & 41.02 & 59.60 & 27.65 & 0.083 & 13.37 \\
        OmniHands~\cite{omnihands} & 0.835 & 0.964 & 0.931 & 30.52 & 10.05 & 34.91 & 40.36 & 24.29 & 0.107 & 50.83 \\
        EgoForce~\cite{egoforce} & 0.860 & 0.920 & 0.956 & 23.14 & 8.82 & 31.57 & 66.62 & 22.19 & 0.076 & 14.87 \\
        ACE-Ego-Hand, K-free~\cite{ace_ego_hand} & 0.969 & 0.984 & 0.987 & 16.61 & 7.71 & 16.43 & 16.77 & 13.99 & 0.030 & 2.45 \\
        ACE-Ego-Hand, K-given~\cite{ace_ego_hand} & 0.986 & 0.995 & 0.996 & 15.58 & 7.43 & 16.54 & 9.36 & 12.45 & 0.024 & 2.58 \\
        \cmidrule[0.2pt](lr){1-11}
        HandRSI network ($\theta_2$), 42.5 FPS & 0.987 & 0.995 & 0.997 & 15.15 & 7.28 & 16.11 & 9.12 & 12.34 & 0.021 & 2.36 \\
        \rowcolor{HandRSILite}
        HandRSI-Lite, 41.1 FPS & 0.990 & 0.999 & 0.997 & 14.53 & 7.25 & 15.84 & 8.80 & 11.76 & \textbf{0.019} & 1.58 \\
        \rowcolor{HandRSIMedium}
        HandRSI-Medium, 34.0 FPS & 0.990 & 0.999 & 0.997 & 14.47 & 7.14 & 15.78 & 8.78 & 11.76 & \textbf{0.019} & \textbf{1.45} \\
        \rowcolor{HandRSIHeavy}
        HandRSI-Heavy, 20.7 FPS & \textbf{0.991} & \textbf{1.000} & \textbf{0.998} & 14.00 & 6.44 & 15.33 & 7.69 & \textbf{11.59} & \textbf{0.019} & 1.46 \\
        \rowcolor{HandRSIHeavy}
        \textbf{HandRSI} ($\mathcal S_5$), 16.2 FPS & 0.990 & \textbf{1.000} & 0.997 & \textbf{13.85} & \textbf{6.34} & \textbf{15.18} & \textbf{7.61} & 11.61 & \textbf{0.019} & \textbf{1.45} \\
 
        \bottomrule
    \end{tabular}
\end{table*}

We re-run nine published single-frame and video methods, including
ACE-Ego-Hand with and without intrinsics (K-given, K-free), with their
official code and released weights under ACE-Ego-Hand's baseline protocol,
so that every row shares segments, intrinsics and evaluator.
WiLoR, HaWoR and EgoForce use their own detectors; HaMeR, Hamba, WildHands,
OmniHands and InterWild share a front end of Detectron2 ViTDet-H person boxes
and ViTPose+-Huge whole-body hand keypoints~\cite{vitdet,vitpose,vitposepp},
with each method's demo rule for turning keypoints into boxes, as in
ACE-Ego-Hand's protocol.
Every method that converts a crop camera to the full image receives the
clip's GT intrinsics; HaWoR runs its camera-space front end with SLAM and
infilling off, EgoForce runs its causal Kalman filter~\cite{kalman1960}, and
InterWild uses its egocentric checkpoint (trained with the egocentric
cameras of ReInterHand~\cite{reinterhand}).
ACE-Ego-Hand's K-free row takes existence from its 2D head, the head its
video deployment uses (with its 3D head, HOT3D FAcc drops from 0.92 to 0.71).
Each method's MANO parameters are re-posed with the evaluator's MANO model
and checked against its own mesh.
PA-p of the baselines is lower than the values ACE-Ego-Hand quotes from
ViDiHand's evaluation~\cite{vidihand}, while MPJPE-p matches them for most
methods, and HaWoR is much stronger here than in the quoted numbers (HOT3D
24.9 vs.\ 71.4\,mm) because it receives GT intrinsics and its detector finds
90\% of the hands.
We did not run Dyn-HaMR~\cite{dyn_hamr}, whose released body-model
parameters we could not verify, or ViDiHand, whose code is not released.
Table~\ref{tab:all} lists all ten re-run configurations on HOT3D and ARCTIC.

\subsection{Implementation Details}
\label{sec:impl}

\begin{figure*}[t]
    \centering
    {\footnotesize
\setlength{\qw}{\dimexpr(\linewidth-1em-14pt-8pt)/8\relax}%
\begin{tabular}{@{}c@{\hspace{2pt}}c@{\hspace{2pt}}c@{\hspace{2pt}}c@{\hspace{2pt}}c@{\hspace{8pt}}c@{\hspace{2pt}}c@{\hspace{2pt}}c@{\hspace{2pt}}c@{}}
 & Ground truth & ACE-Ego-Hand & network & \textbf{\color{HandRSIBlue}HandRSI} & Ground truth & ACE-Ego-Hand & network & \textbf{\color{HandRSIBlue}HandRSI} \\[1pt]
\qrow{frame} & \qtile{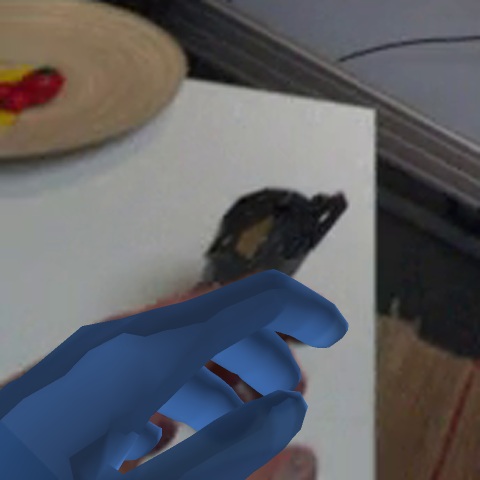}{}{QTag} & \qtile{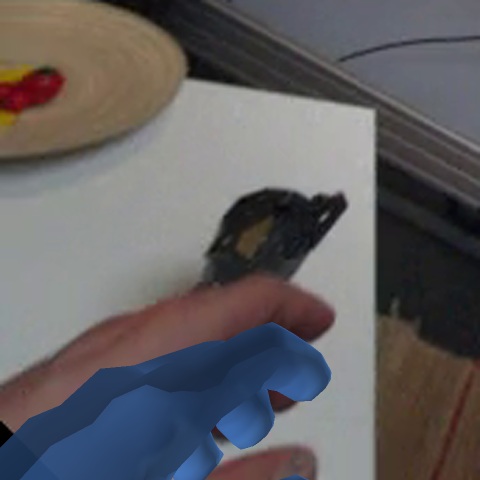}{}{QTag} & \qtile{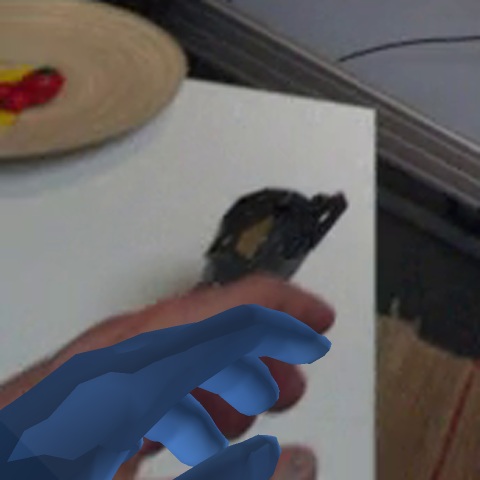}{}{QTag} & \qtile{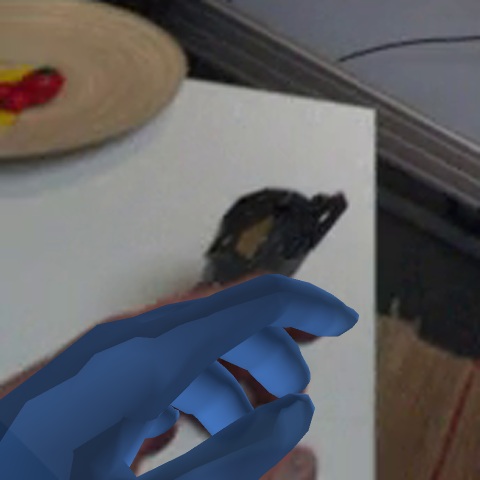}{}{QTag} & \qtile{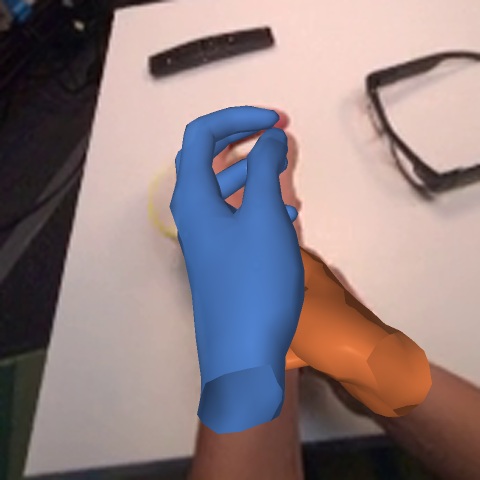}{}{QTag} & \qtile{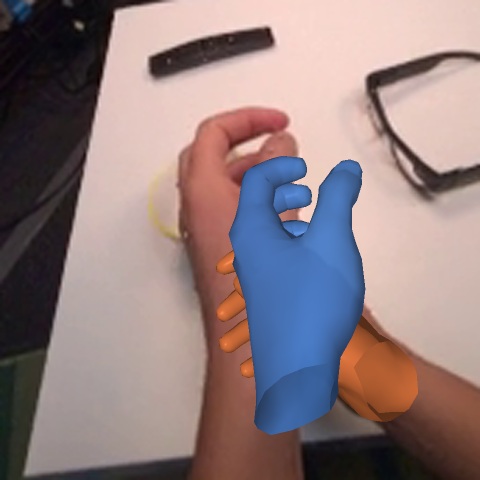}{}{QTag} & \qtile{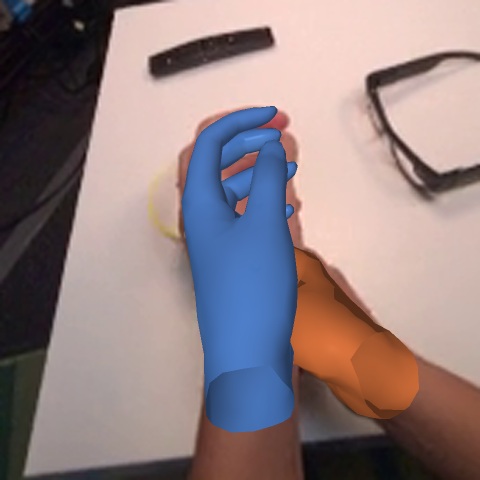}{}{QTag} & \qtile{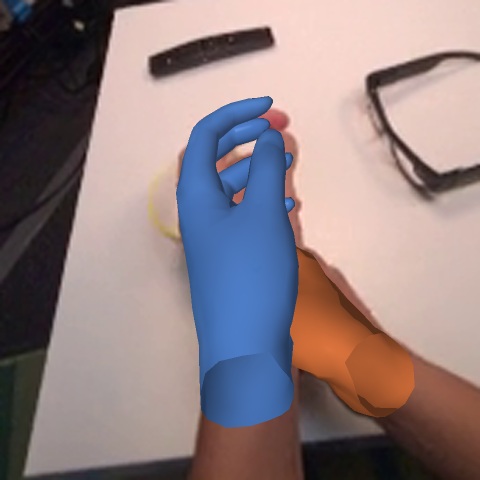}{}{QTag} \\[2pt]
\qrow{frame} & \qtile{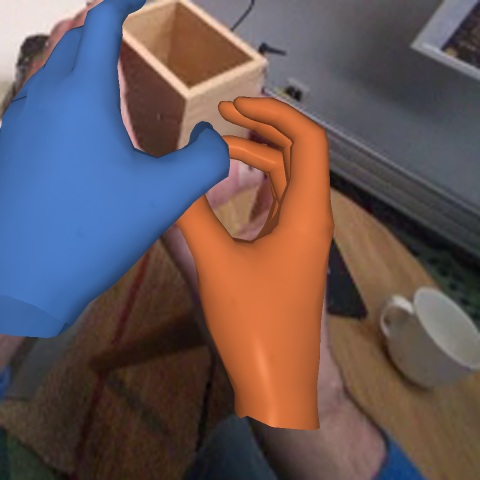}{}{QTag} & \qtile{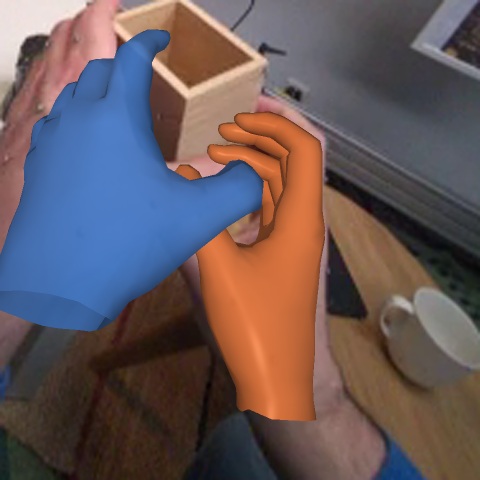}{}{QTag} & \qtile{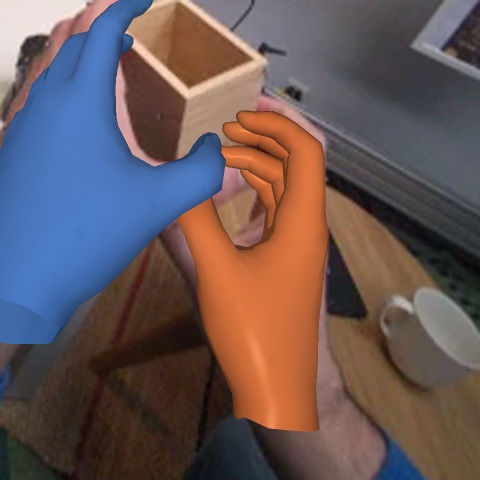}{}{QTag} & \qtile{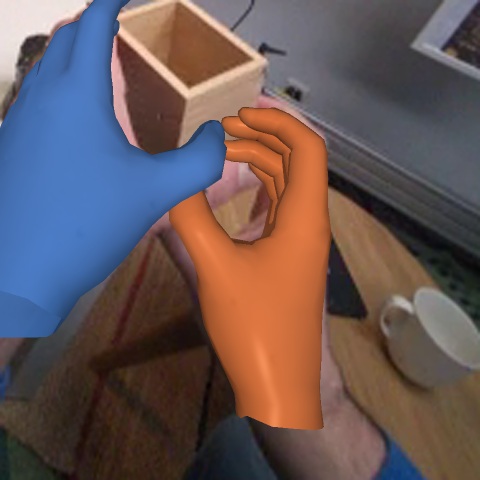}{}{QTag} & \qtile{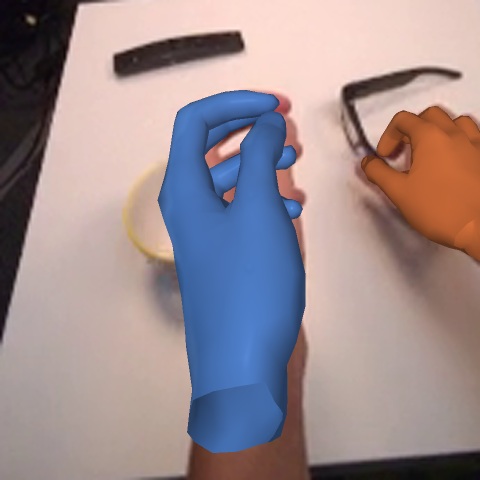}{}{QTag} & \qtile{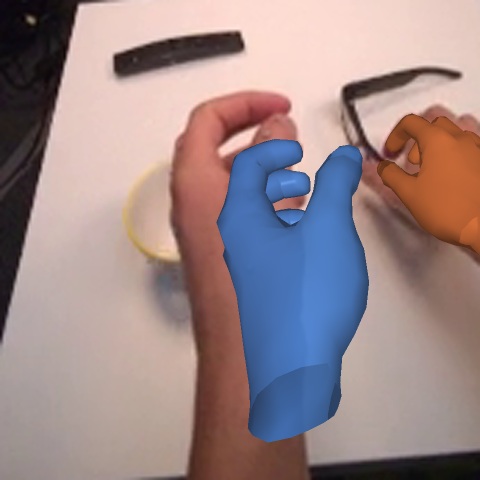}{}{QTag} & \qtile{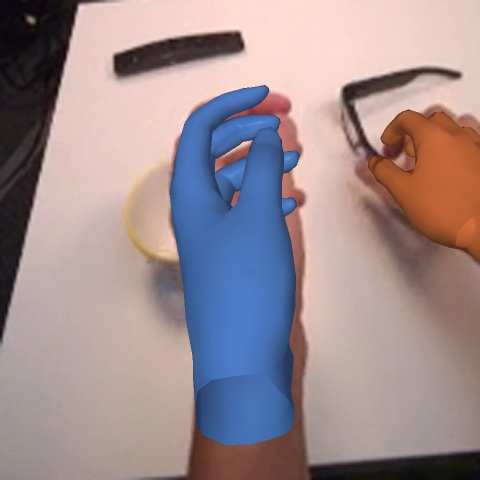}{}{QTag} & \qtile{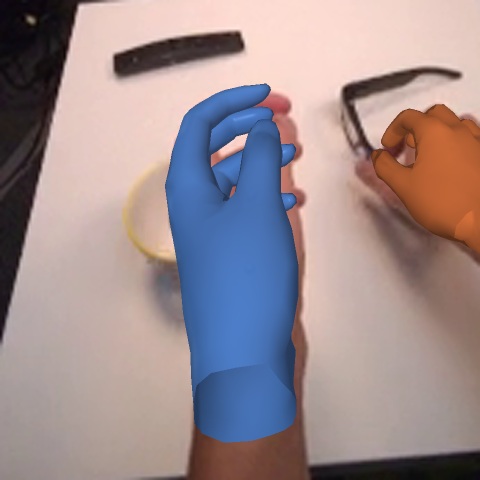}{}{QTag} \\[2pt]
\qrow{3D (camera)} & \qtile{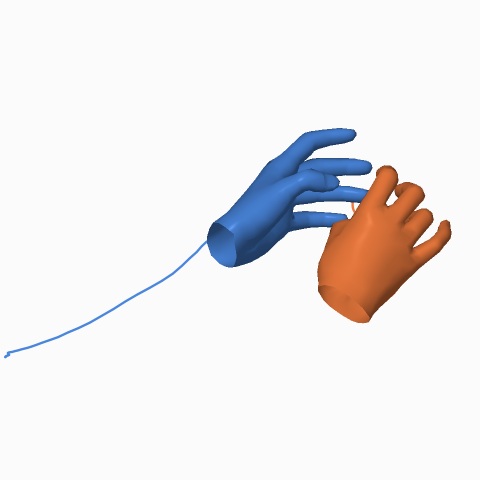}{}{QTag} & \qtile{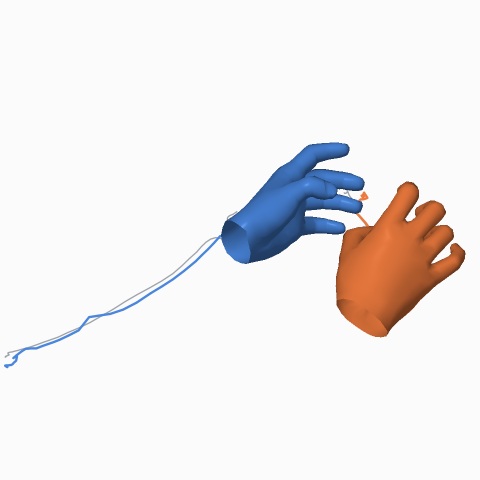}{}{QTag} & \qtile{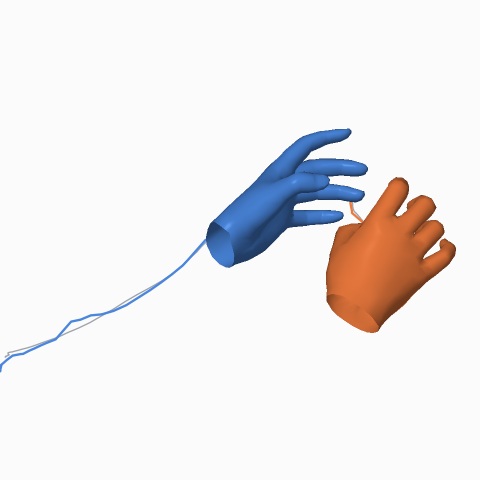}{}{QTag} & \qtile{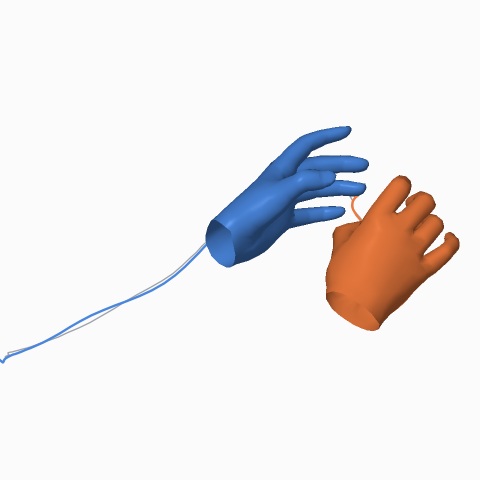}{}{QTag} & \qtile{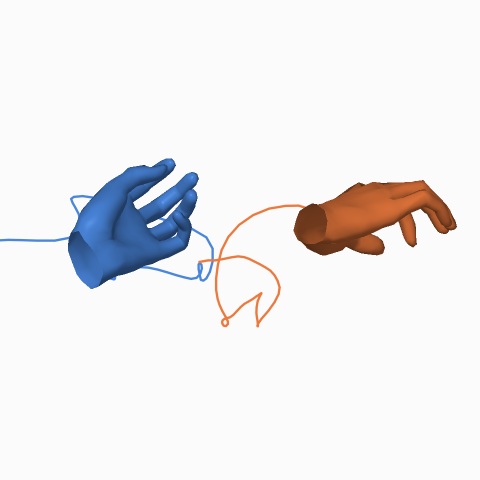}{}{QTag} & \qtile{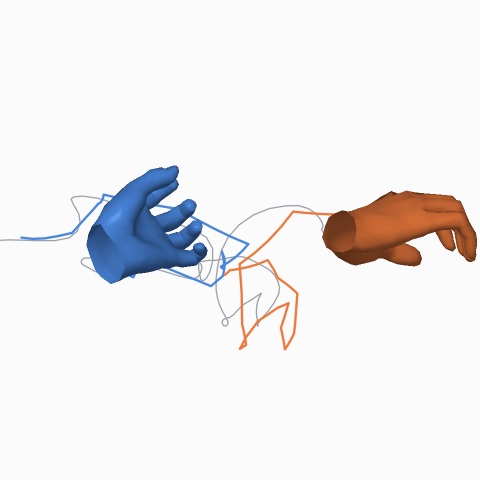}{}{QTag} & \qtile{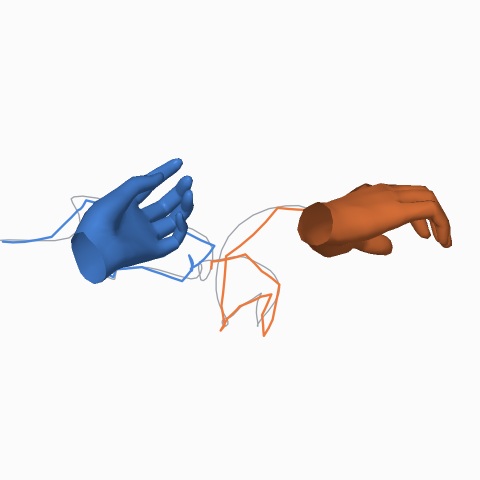}{}{QTag} & \qtile{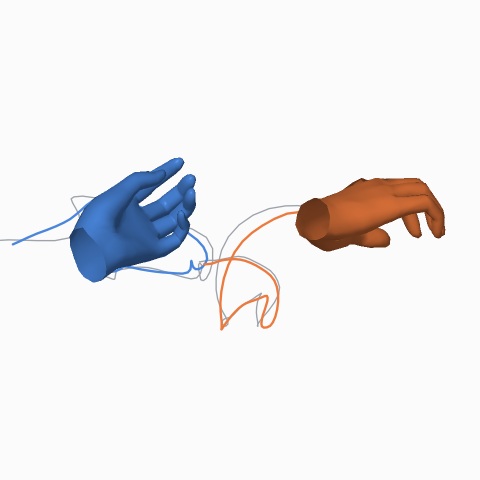}{}{QTag} \\[2pt]
 &  & \scriptsize 36.5\,mm, 24.0\,px & \scriptsize 30.8\,mm, 19.5\,px & \scriptsize\textbf{\color{HandRSIBlue}7.2\,mm, 3.5\,px} &  & \scriptsize 29.7\,mm, 70.2\,px & \scriptsize 11.0\,mm, 5.0\,px & \scriptsize\textbf{\color{HandRSIBlue}10.0\,mm, 3.6\,px} \\
\end{tabular}
}
    \caption{
        \textbf{Two HOT3D test clips} with the network alone: two frames (cropped to the hand that
        ACE-Ego-Hand gets most wrong) and the clip's hands in camera space with wrist trails (ground truth in
        gray). Below: the clip's MPJPE-p and EPE$_{\mathrm{2D}}$-p. ACE-Ego-Hand misses the hand at the image
        border (left) and misplaces the occluded hand (right); the HandRSI network recovers both.
    }
    \label{fig:clips}
\end{figure*}

Loop~A runs eight agents per round, four with Claude Opus~5.5 and four with
GPT-6-sol (Sec.~\ref{sec:loopA}); their modules may call one pretrained
network besides the estimator, WiLoR~\cite{wilor}.
The reflection agent of Loop~C runs Claude Opus~5.5.
Loop~B trains on public data only, HOT3D and ARCTIC with their 3D GT plus
62 unlabeled HOT3D sequences and ARCTIC's subject s03 for the teacher terms
(Tabs.~\ref{tab:loopB} and~\ref{tab:splits}).
A round takes about half a day on eight GPUs, and the Loop~A agents write
about 1.7M output tokens per round (Tab.~\ref{tab:loopa}).
The public release contains the HandRSI pipeline (Loops~A--C, the gates,
Loop~B training, re-validation and the tier search), the final harness and
the three tiers, our re-implementation of the ACE-Ego-Hand evaluator, the
runners of the published baselines and of the classical and ensemble
controls, and the result files behind every table, each mapped to its table
in a README of the public repository (\codeurl); the network weights and the harness of every round are
released separately (\weightsurl).

\subsection{Use of Validation and Test Data}
\label{sec:data_use}

Agents worked on dev only: val was reached only through the gates, and no
agent saw val ground truth or test data.
Rounds~3--5, the tier selection and the classical baselines
(Tab.~\ref{tab:classical}) used dev, val and timing only, and from round~2
on Loop~B checkpoints are selected on val.
In rounds~1--2 the operators consulted test scores for three decisions: the
round-1 Loop~B recipe among the variants of Tab.~\ref{tab:distill}, and two
gate rules (ARCTIC regressions do not block a gate; HOT3D hand slots without
GT are not scored).
Round~1 was accepted by the operators although HOT3D FAcc regressed on val
by $r_m=-5.6\%$, past the 5\% cap but not significantly (CI$_{90}$ $-$16.1 to
$+$6.9\%), while the miss rate fell by 16\%: the new network trades missed
hands for false positives.
The strict rule would have kept $\mathcal H^I_1$ on $\theta_0$, which passes
$G_A$ ($\mathcal C^2_1$, $\Delta=+4.22\%$), and round~2 would have started from
$\theta_0$.
In round~2 the operators removed one module that degraded a single val
recording before the round gate was run again; Loop~C later turned this into
an automatic guard (Sec.~\ref{sec:loopC}).

\section{Results}
\label{sec:results}

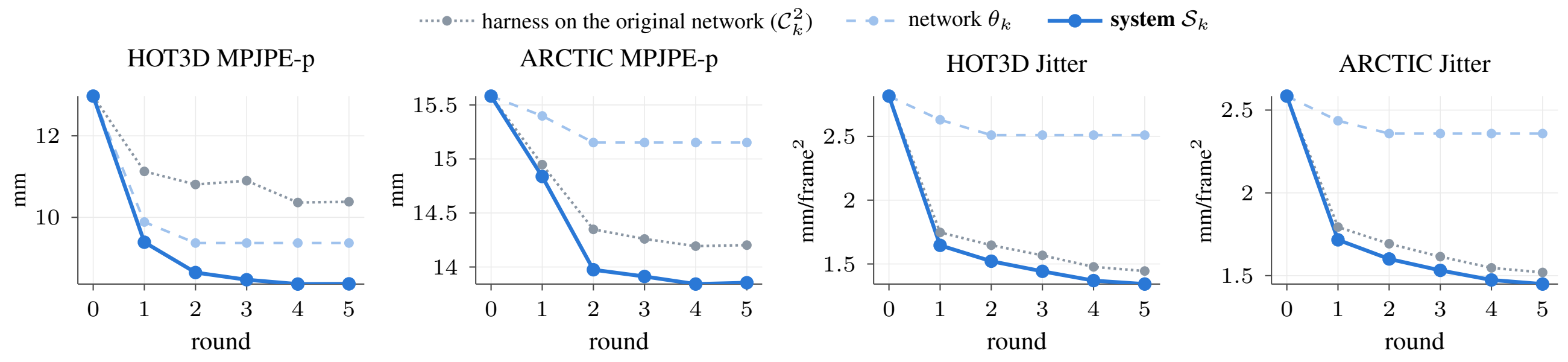
\begin{figure*}[t]
    \centering
\begin{tikzpicture}
\begin{groupplot}[hr, group style={group size=4 by 1, horizontal sep=0.07\linewidth},
    width=0.18\linewidth, height=0.118\linewidth, xmin=-0.3, xmax=5.3, xtick={0,...,5}, xlabel={round},
    ylabel style={font=\scriptsize, yshift=-2pt}, title style={yshift=-2pt},
    legend style={at={(2.583,1.28)}, anchor=south, legend columns=3, /tikz/every even column/.append style={column sep=8pt}}]
\nextgroupplot[title={HOT3D MPJPE-p}, ylabel={mm}, ytick={8,10,12}]
\addplot[CGray, line width=0.8pt, densely dotted, mark=*, mark size=1.1pt, mark options={solid}] table[x=k, y=hot3d_mpjpe_c2]{fig/data/rounds.dat};
\addplot[HandRSIBlue!45, line width=0.8pt, dashed, mark=*, mark size=1.1pt, mark options={solid}] table[x=k, y=hot3d_mpjpe_net]{fig/data/rounds.dat};
\addplot[HandRSIBlue, line width=1.2pt, mark=*, mark size=1.5pt] table[x=k, y=hot3d_mpjpe_sys]{fig/data/rounds.dat};
\addlegendentry{harness on the original network ($\mathcal C^2_k$)}
\addlegendentry{network $\theta_k$}
\addlegendentry{\textbf{system $\mathcal S_k$}}
\nextgroupplot[title={ARCTIC MPJPE-p}, ylabel={mm}, ytick={14,14.5,15,15.5}]
\addplot[CGray, line width=0.8pt, densely dotted, mark=*, mark size=1.1pt, mark options={solid}] table[x=k, y=arctic_mpjpe_c2]{fig/data/rounds.dat};
\addplot[HandRSIBlue!45, line width=0.8pt, dashed, mark=*, mark size=1.1pt, mark options={solid}] table[x=k, y=arctic_mpjpe_net]{fig/data/rounds.dat};
\addplot[HandRSIBlue, line width=1.2pt, mark=*, mark size=1.5pt] table[x=k, y=arctic_mpjpe_sys]{fig/data/rounds.dat};
\nextgroupplot[title={HOT3D Jitter}, ylabel={mm/frame$^2$}, ytick={1.5,2,2.5}]
\addplot[CGray, line width=0.8pt, densely dotted, mark=*, mark size=1.1pt, mark options={solid}] table[x=k, y=hot3d_jitter_c2]{fig/data/rounds.dat};
\addplot[HandRSIBlue!45, line width=0.8pt, dashed, mark=*, mark size=1.1pt, mark options={solid}] table[x=k, y=hot3d_jitter_net]{fig/data/rounds.dat};
\addplot[HandRSIBlue, line width=1.2pt, mark=*, mark size=1.5pt] table[x=k, y=hot3d_jitter_sys]{fig/data/rounds.dat};
\nextgroupplot[title={ARCTIC Jitter}, ylabel={mm/frame$^2$}, ytick={1.5,2,2.5}]
\addplot[CGray, line width=0.8pt, densely dotted, mark=*, mark size=1.1pt, mark options={solid}] table[x=k, y=arctic_jitter_c2]{fig/data/rounds.dat};
\addplot[HandRSIBlue!45, line width=0.8pt, dashed, mark=*, mark size=1.1pt, mark options={solid}] table[x=k, y=arctic_jitter_net]{fig/data/rounds.dat};
\addplot[HandRSIBlue, line width=1.2pt, mark=*, mark size=1.5pt] table[x=k, y=arctic_jitter_sys]{fig/data/rounds.dat};
\end{groupplot}
\end{tikzpicture}
    \caption{
        \textbf{Recursive improvement} on the sealed test sets: the system $\mathcal S_k$, its network $\theta_k$, and
        the round's harness on the original network ($\mathcal C^2_k$); round~0 is ACE-Ego-Hand; val gains in
        Tab.~\ref{tab:loop}.
    }
    \label{fig:rounds}
\end{figure*}

\begin{table}[t]
    \centering
    \caption{
        \textbf{The five rounds}: Loop~A submissions (accepted by $G_A$), modules in the accepted harness, network,
        and the round gate's val gain over the previous system (CI$_{90}$ lower bound).
        From round~3 on, no student passed $G_B$, so the network stays $\theta_2$.
        $^{a}$Accepted with FAcc past the 5\% cap ($r_m=-5.6\%$, not significant), as the network trades missed
        hands for false positives (Sec.~\ref{sec:data_use}); $^{b}$below the round gate, kept by the
        harness-only rule.
    }
    \label{tab:loop}
    \footnotesize
    \setlength{\tabcolsep}{5pt}
    \setlength{\aboverulesep}{0pt}
    \setlength{\belowrulesep}{0pt}
    \begin{tabular}{@{}cccc>{\raggedleft\arraybackslash}p{1.95cm}@{}}
        \toprule
        \rowcolor{HandRSIHeader}
        Round & subm.\ (acc.) & modules & network & val $\Delta$ (\%)\phantom{$^{a}$} \\
        \midrule
        1 & 70 (43) & 18 & $\theta_1$ & $+$3.37 ($+$2.46)$^{a}$ \\
        2 & 36 (6) & 24 & $\theta_2$ & $+$3.54 ($+$2.86)\phantom{$^{a}$} \\
        3 & 26 (12) & 32 & $\theta_2$ & $+$3.17 ($+$2.74)\phantom{$^{a}$} \\
        4 & 17 (9) & 37 & $\theta_2$ & $+$2.06 ($+$1.84)\phantom{$^{a}$} \\
        5 & 7 (3) & 39 & $\theta_2$ & $+$0.77 ($+$0.47)$^{b}$ \\
        \bottomrule
    \end{tabular}
\end{table}

\begin{figure*}[t]
    \centering
    {\footnotesize
\setlength{\qw}{\dimexpr(\linewidth-1em-14.5pt)/7\relax}%
\begin{tabular}{@{}c@{\hspace{2pt}}c@{\hspace{2pt}}c@{\hspace{2pt}}c@{\hspace{2pt}}c@{\hspace{2pt}}c@{\hspace{2pt}}c@{\hspace{2pt}}c@{}}
 & Ground truth & ACE-Ego-Hand & $\mathcal{S}_1$ & $\mathcal{S}_2$ & $\mathcal{S}_3$ & $\mathcal{S}_4$ & \textbf{\color{HandRSIBlue}$\mathcal{S}_5$ (HandRSI)} \\[1pt]
\qrow{frame 24} & \qtile{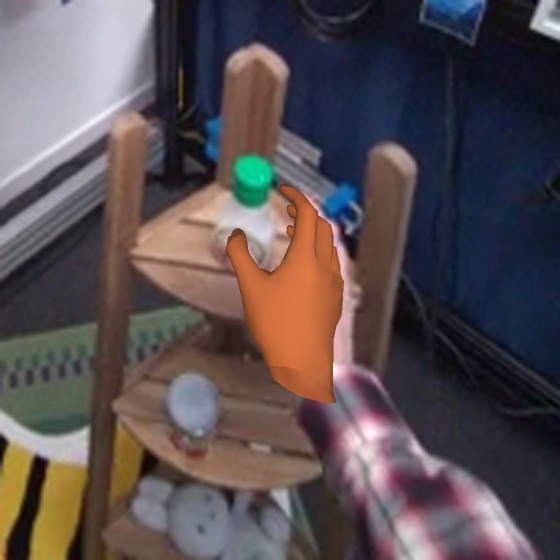}{}{QTag} & \qtile{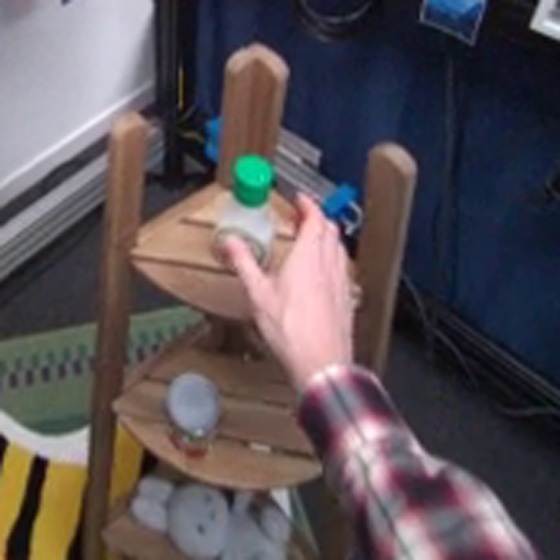}{}{QTag} & \qtile{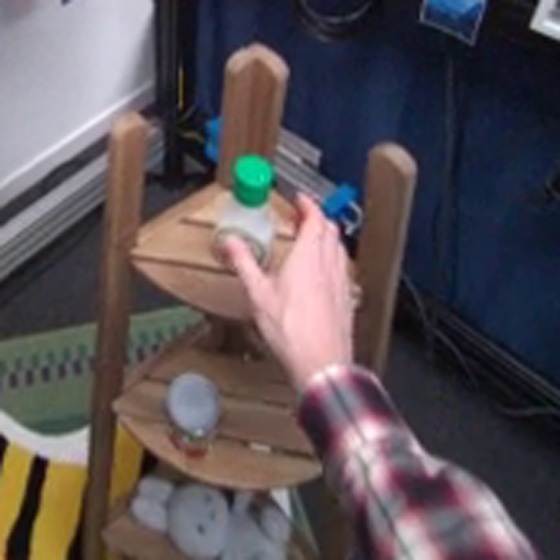}{}{QTag} & \qtile{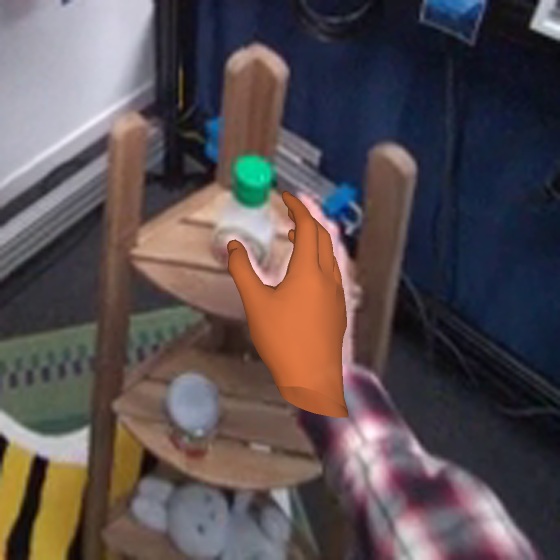}{}{QTag} & \qtile{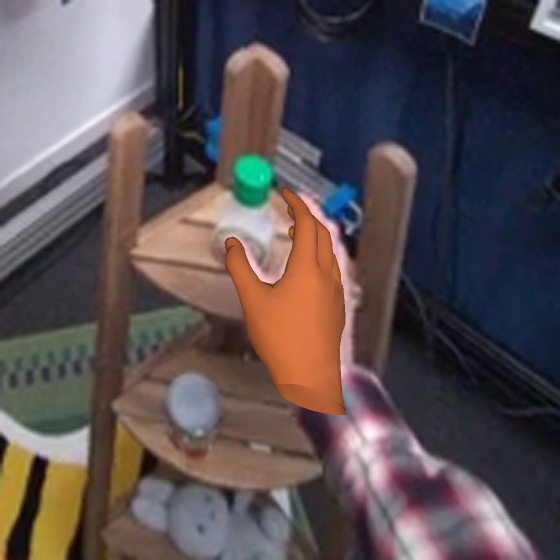}{}{QTag} & \qtile{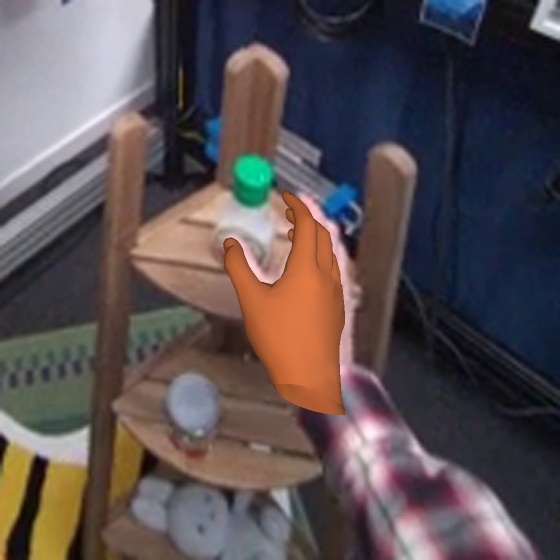}{}{QTag} & \qtile{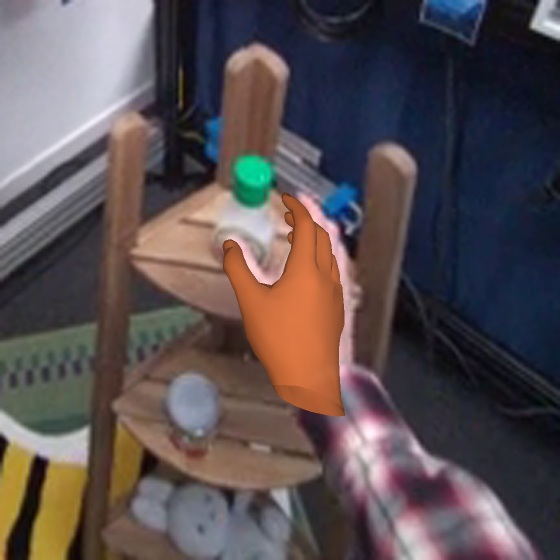}{}{QTag} \\[2pt]
\qrow{frame 51} & \qtile{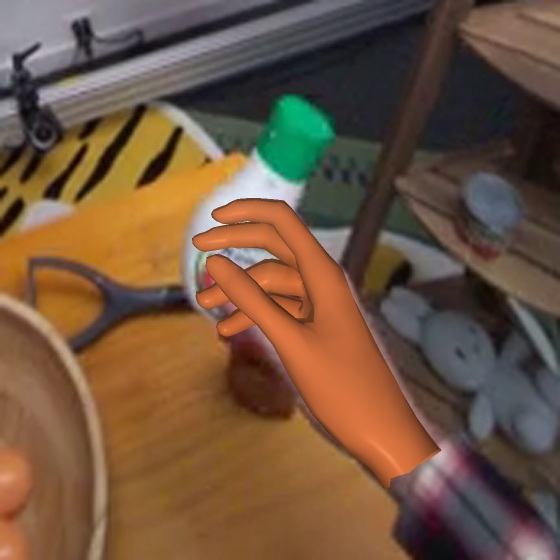}{}{QTag} & \qtile{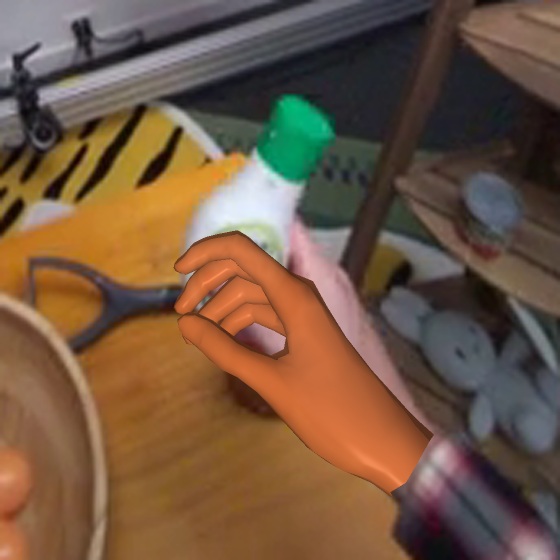}{}{QTag} & \qtile{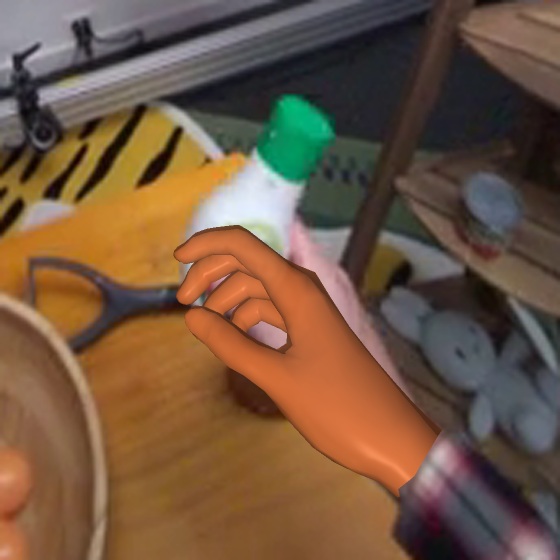}{}{QTag} & \qtile{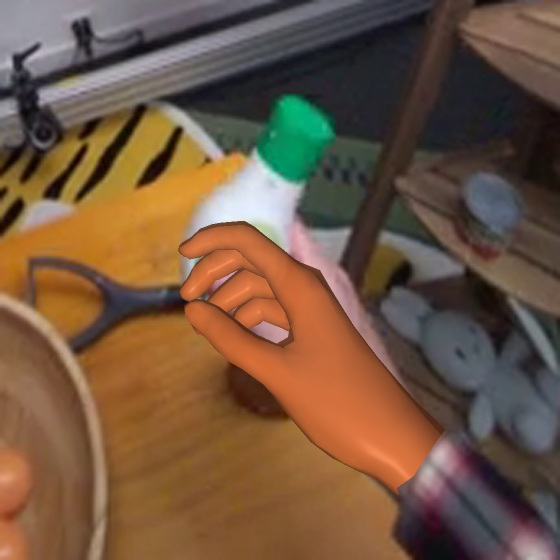}{}{QTag} & \qtile{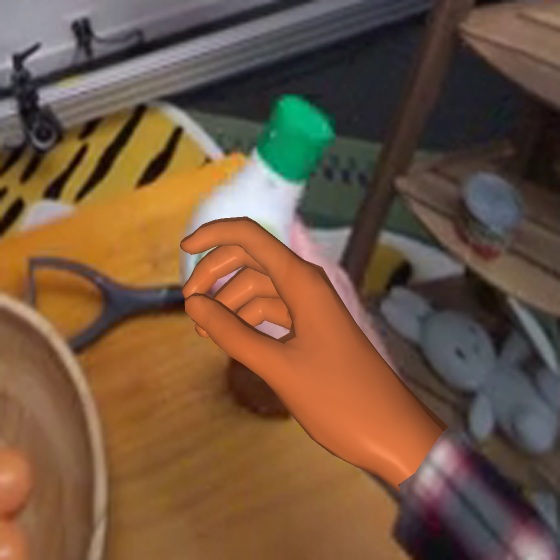}{}{QTag} & \qtile{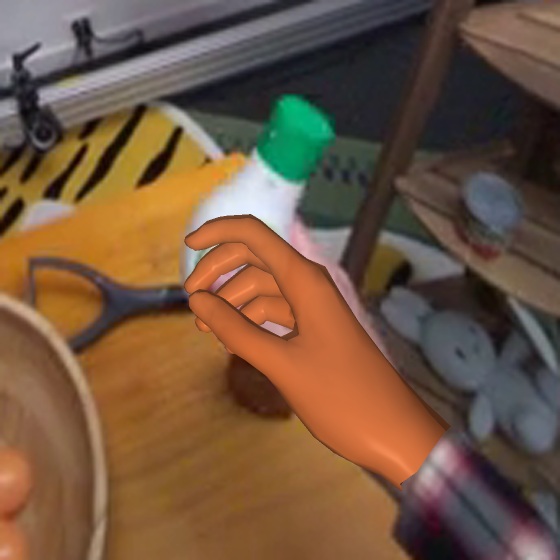}{}{QTag} & \qtile{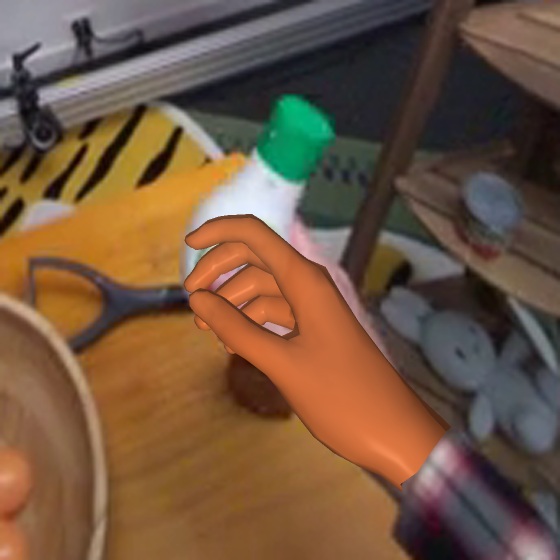}{}{QTag} \\[2pt]
\qrow{error} & \parbox[c]{\qw}{\centering\scriptsize per-frame error\\ (0--40\,mm)\\ \textcolor{QMiss}{\rule{4pt}{4pt}}~missed} & \qtile{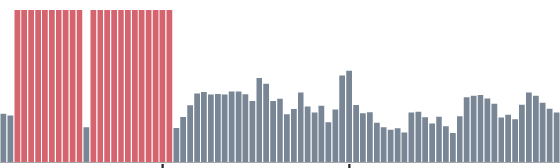}{}{QTag} & \qtile{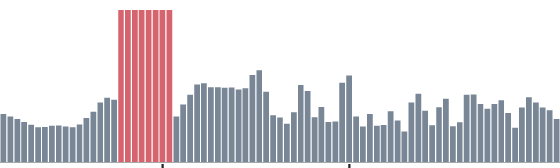}{}{QTag} & \qtile{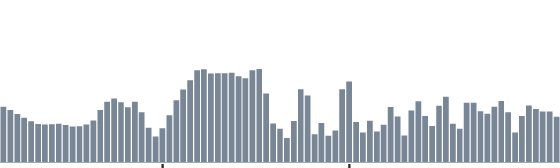}{}{QTag} & \qtile{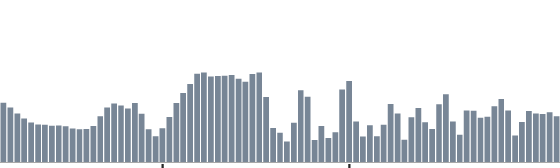}{}{QTag} & \qtile{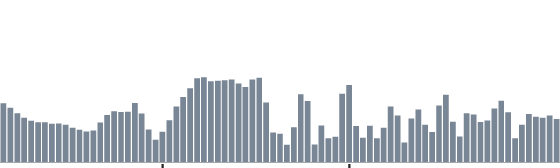}{}{QTag} & \qtile{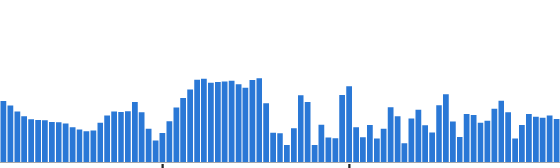}{}{QTag} \\[1pt]
 & \footnotesize MPJPE-p & \footnotesize 47.0\,mm & \footnotesize 28.7\,mm & \footnotesize 16.2\,mm & \footnotesize 15.6\,mm & \footnotesize 15.0\,mm & \footnotesize\textbf{\color{HandRSIBlue}14.9\,mm} \\
\end{tabular}
}
    \caption{
        \textbf{One HOT3D test clip after every round.} Top: two frames of the right hand (same crop per column);
        bottom: its wrist-aligned joint error in each of the 81 frames (red: missed; ticks: the frames above) and the
        clip's MPJPE-p. ACE-Ego-Hand misses the hand in 22 frames and $\mathcal S_1$ in 8; from round~2 on it is
        always found, and later rounds lower its error on the detected frames (14.4$\to$12.5\,mm).
    }
    \label{fig:evolution}
\end{figure*}

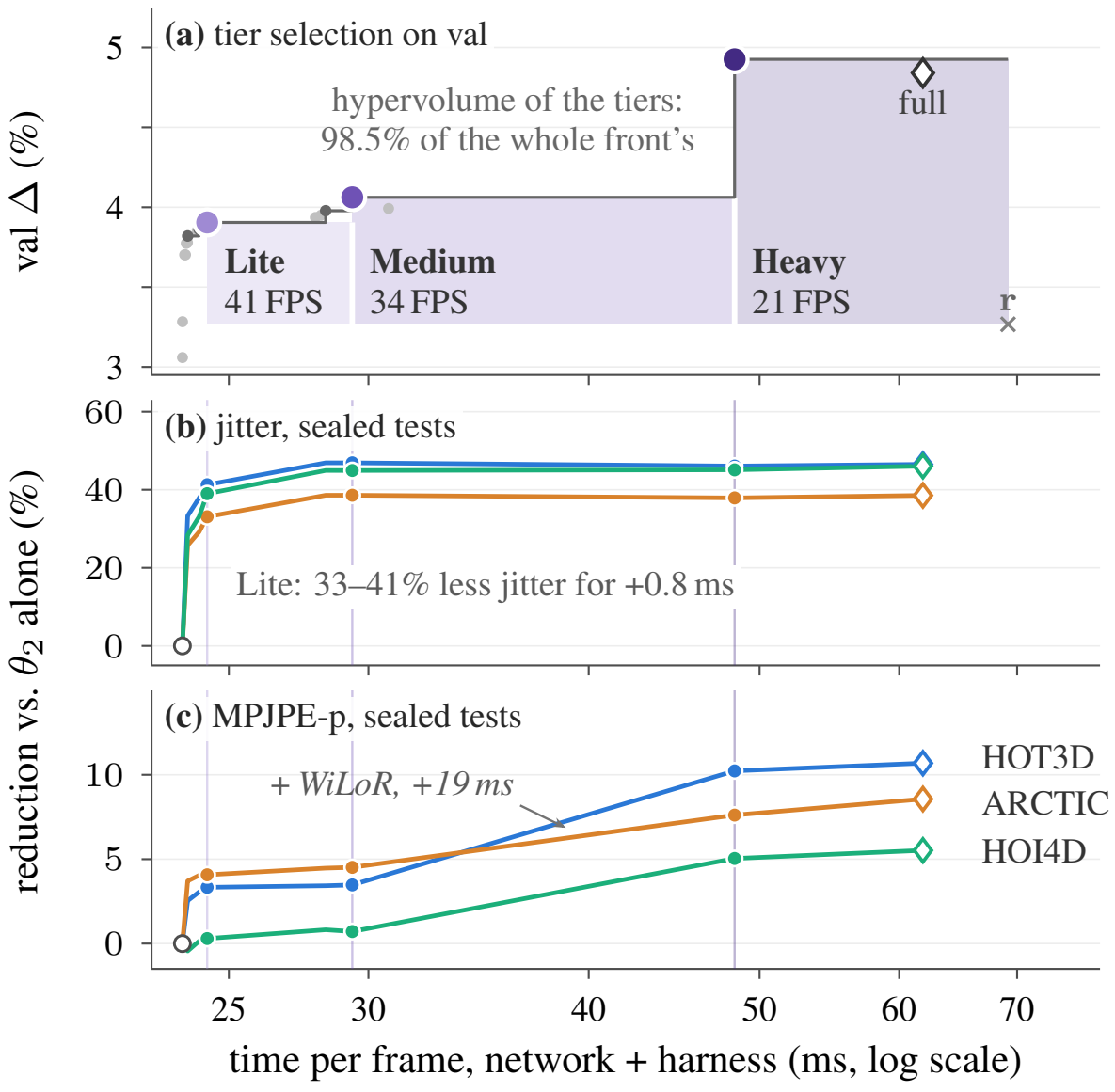
\begin{figure}[t]
    \centering
\definecolor{CHot}{HTML}{2A78D6}\definecolor{CArc}{HTML}{D9822B}\definecolor{CHoi}{HTML}{1BAF7A}
\tikzset{tierguide/.style={line width=0.45pt, opacity=0.35},
    tierlab/.style={font=\scriptsize, align=left, anchor=south west, inner sep=1.5pt, text=black!80, yshift=0.5pt},
    dslab/.style={font=\scriptsize, anchor=west, inner sep=0.5pt, text=black!80},
    paneltag/.style={font=\scriptsize, text=black!85, anchor=north west, inner sep=1.2pt, fill=white, fill opacity=0.9, text opacity=1,
        xshift=1.2pt, yshift=-1.2pt}}
\begin{tikzpicture}
\begin{groupplot}[hr, group style={group size=1 by 3, vertical sep=5pt, x descriptions at=edge bottom},
    width=0.80\linewidth, xmode=log, xmin=22.6, xmax=78, xtick={25,30,40,50,60,70}, xticklabels={25,30,40,50,60,70},
    minor xtick={}, xmajorgrids=false, ymajorgrids=true, grid style={black!7, line width=0.3pt},
    xlabel={time per frame, network + harness (ms, log scale)}, clip mode=individual,
    ylabel style={at={(rel axis cs:-0.105,0.5)}, anchor=south}]
\nextgroupplot[height=0.31\linewidth, ymin=2.95, ymax=5.25, ytick={3,3.5,4,4.5,5}, yticklabels={3,,4,,5}, ylabel={val $\Delta$ (\%)}]
\fill[CLite!20] (axis cs:24.3024,3.2673) rectangle (axis cs:29.3735,3.9049);
\fill[CMedium!20] (axis cs:29.3735,3.2673) rectangle (axis cs:48.3963,4.0622);
\fill[CHeavy!20] (axis cs:48.3963,3.2673) rectangle (axis cs:69.1717,4.9260);
\draw[white, line width=1.1pt] (axis cs:29.3735,3.2673) -- (axis cs:29.3735,3.9049);
\draw[white, line width=1.1pt] (axis cs:48.3963,3.2673) -- (axis cs:48.3963,4.0622);
\addplot[only marks, mark=*, mark size=0.9pt, black!25] table[x=x, y=y]{fig/data/tiers_val.dat};
\addplot[black!60, line width=0.6pt, line join=round] table[x=x, y=y]{fig/data/tiers_val_front.dat};
\addplot[only marks, mark=*, mark size=0.95pt, black!60] table[x=x, y=y]{fig/data/tiers_val_frontpts.dat};
\addplot[only marks, mark=*, mark size=2.6pt, mark options={fill=CLite, draw=white, line width=0.8pt}] coordinates {(24.3024,3.9049)};
\addplot[only marks, mark=*, mark size=2.6pt, mark options={fill=CMedium, draw=white, line width=0.8pt}] coordinates {(29.3735,4.0622)};
\addplot[only marks, mark=*, mark size=2.6pt, mark options={fill=CHeavy, draw=white, line width=0.8pt}] coordinates {(48.3963,4.9260)};
\addplot[only marks, mark=diamond*, mark size=2.8pt, mark options={fill=white, draw=black!80, line width=0.7pt}] coordinates {(61.8917,4.8406)};
\node[font=\scriptsize, text=black!75, anchor=north, inner sep=3.2pt] at (axis cs:61.8917,4.8406) {full};
\addplot[only marks, mark=x, mark size=2pt, black!50, line width=0.6pt] coordinates {(69.1717,3.2673)};
\node[font=\scriptsize, text=black!60, anchor=south, inner sep=2.5pt] at (axis cs:69.1717,3.2673) {$\mathbf r$};
\node[tierlab] at (axis cs:24.5940,3.2673) {\textbf{Lite}\\41\,FPS};
\node[tierlab] at (axis cs:29.7260,3.2673) {\textbf{Medium}\\34\,FPS};
\node[tierlab] at (axis cs:48.9771,3.2673) {\textbf{Heavy}\\21\,FPS};
\node[font=\scriptsize, text=black!60, anchor=center, align=center] at (axis cs:36,4.55) {hypervolume of the tiers:\\98.5\% of the whole front's};
\node[paneltag] at (rel axis cs:0,1) {\textbf{(a)} tier selection on val};
\nextgroupplot[height=0.224\linewidth, ymin=-5, ymax=63, ytick={0,20,40,60}, ylabel={reduction vs.\ $\theta_2$ alone (\%)},
    ylabel style={at={(rel axis cs:-0.105,-0.06)}}]
\draw[CLite, tierguide] ({axis cs:24.3024,0}|-{rel axis cs:0,0}) -- ({axis cs:24.3024,0}|-{rel axis cs:0,1});
\draw[CMedium, tierguide] ({axis cs:29.3735,0}|-{rel axis cs:0,0}) -- ({axis cs:29.3735,0}|-{rel axis cs:0,1});
\draw[CHeavy, tierguide] ({axis cs:48.3963,0}|-{rel axis cs:0,0}) -- ({axis cs:48.3963,0}|-{rel axis cs:0,1});
\addplot[CHot, line width=0.9pt, line join=round, mark=*, mark indices={4,6,7}, mark size=1.55pt, mark options={fill=CHot, draw=white, line width=0.5pt}] table[x=x, y=hot3d_jitter]{fig/data/tiers_test.dat};
\addplot[CArc, line width=0.9pt, line join=round, mark=*, mark indices={4,6,7}, mark size=1.55pt, mark options={fill=CArc, draw=white, line width=0.5pt}] table[x=x, y=arctic_jitter]{fig/data/tiers_test.dat};
\addplot[CHoi, line width=0.9pt, line join=round, mark=*, mark indices={4,6,7}, mark size=1.55pt, mark options={fill=CHoi, draw=white, line width=0.5pt}] table[x=x, y=hoi4d_jitter]{fig/data/tiers_test.dat};
\addplot[only marks, mark=diamond*, mark size=2.3pt, mark options={fill=white, draw=CHot, line width=0.7pt}, forget plot] coordinates {(61.8917,46.5009)};
\addplot[only marks, mark=diamond*, mark size=2.3pt, mark options={fill=white, draw=CArc, line width=0.7pt}, forget plot] coordinates {(61.8917,38.5436)};
\addplot[only marks, mark=diamond*, mark size=2.3pt, mark options={fill=white, draw=CHoi, line width=0.7pt}, forget plot] coordinates {(61.8917,46.0444)};
\addplot[only marks, mark=*, mark size=1.6pt, mark options={fill=white, draw=black!70, line width=0.55pt}] coordinates {(23.5287,0.0000)};
\node[paneltag] at (rel axis cs:0,1) {\textbf{(b)} jitter, sealed tests};
\node[font=\scriptsize, text=black!65, anchor=west, align=left, inner sep=1.2pt, fill=white] at (axis cs:25.0314,15) {Lite: 33--41\% less jitter for +0.8\,ms};
\nextgroupplot[height=0.235\linewidth, ymin=-1.5, ymax=15, ytick={0,5,10}]
\draw[CLite, tierguide] ({axis cs:24.3024,0}|-{rel axis cs:0,0}) -- ({axis cs:24.3024,0}|-{rel axis cs:0,1});
\draw[CMedium, tierguide] ({axis cs:29.3735,0}|-{rel axis cs:0,0}) -- ({axis cs:29.3735,0}|-{rel axis cs:0,1});
\draw[CHeavy, tierguide] ({axis cs:48.3963,0}|-{rel axis cs:0,0}) -- ({axis cs:48.3963,0}|-{rel axis cs:0,1});
\addplot[CHot, line width=0.9pt, line join=round, mark=*, mark indices={4,6,7}, mark size=1.55pt, mark options={fill=CHot, draw=white, line width=0.5pt}] table[x=x, y=hot3d_mpjpe]{fig/data/tiers_test.dat};
\addplot[CArc, line width=0.9pt, line join=round, mark=*, mark indices={4,6,7}, mark size=1.55pt, mark options={fill=CArc, draw=white, line width=0.5pt}] table[x=x, y=arctic_mpjpe]{fig/data/tiers_test.dat};
\addplot[CHoi, line width=0.9pt, line join=round, mark=*, mark indices={4,6,7}, mark size=1.55pt, mark options={fill=CHoi, draw=white, line width=0.5pt}] table[x=x, y=hoi4d_mpjpe]{fig/data/tiers_test.dat};
\addplot[only marks, mark=diamond*, mark size=2.3pt, mark options={fill=white, draw=CHot, line width=0.7pt}, forget plot] coordinates {(61.8917,10.6884)};
\node[dslab, yshift=1.2pt] at (axis cs:66.5336,10.6884) {HOT3D};
\addplot[only marks, mark=diamond*, mark size=2.3pt, mark options={fill=white, draw=CArc, line width=0.7pt}, forget plot] coordinates {(61.8917,8.5622)};
\node[dslab, yshift=-0.8pt] at (axis cs:66.5336,8.5622) {ARCTIC};
\addplot[only marks, mark=diamond*, mark size=2.3pt, mark options={fill=white, draw=CHoi, line width=0.7pt}, forget plot] coordinates {(61.8917,5.5222)};
\node[dslab] at (axis cs:66.5336,5.5222) {HOI4D};
\addplot[only marks, mark=*, mark size=1.6pt, mark options={fill=white, draw=black!70, line width=0.55pt}] coordinates {(23.5287,0.0000)};
\node[paneltag] at (rel axis cs:0,1) {\textbf{(c)} MPJPE-p, sealed tests};
\draw[-{Stealth[length=2.6pt]}, black!55, line width=0.45pt] (axis cs:36.5726,8.1840) -- (axis cs:38.8348,6.8522);
\node[font=\scriptsize\itshape, text=black!65, anchor=south east, inner sep=1pt] at (axis cs:36.5726,8.1840) {+\,WiLoR, +19\,ms};
\end{groupplot}
\end{tikzpicture}
    \caption{
        \textbf{Accuracy--speed tiers.} (a) Selection on val: timed sub-chains
        of the final harness (gray), the front, and the three tiers that maximize
        the hypervolume (Eq.~\ref{eq:tiers}) for $\mathbf r$, shaded by the best
        tier per time budget; sub-chains are scored after the loop, so one can
        exceed the full chain. (b, c) Reduction of jitter and MPJPE-p relative to
        $\theta_2$ alone along the val front on the sealed tests.
    }
    \label{fig:pareto}
\end{figure}

\subsection{Comparison with Existing Methods}
\label{sec:exp_comparison}

\begin{figure}[t]
    \centering
\begin{tikzpicture}
\begin{axis}[hr, width=0.70\linewidth, height=0.40\linewidth, ybar stacked, bar width=0.55, xmin=0.4, xmax=5.6, ymin=0, ymax=45,
    xtick={1,...,5}, ytick={0,10,...,40}, xmajorgrids=false, xlabel={round}, ylabel={modules in $\mathcal H^I_k$}, clip=false,
    legend style={at={(0.5,1.04)}, anchor=south, legend columns=2, /tikz/every even column/.append style={column sep=8pt}},
    legend image code/.code={\fill[#1] (0cm,-0.07cm) rectangle (0.2cm,0.11cm);}, reverse legend]
\addplot[fill=KImg, draw=white, line width=0.3pt] table[x=k, y=img]{fig/data/harness.dat};
\addlegendentry{image evidence (WiLoR)}
\addplot[fill=KDet, draw=white, line width=0.3pt] table[x=k, y=det]{fig/data/harness.dat};
\addlegendentry{detection}
\addplot[fill=KTemp, draw=white, line width=0.3pt] table[x=k, y=temp]{fig/data/harness.dat};
\addlegendentry{temporal}
\addplot[fill=KShape, draw=white, line width=0.3pt] table[x=k, y=shape]{fig/data/harness.dat};
\addlegendentry{shape \& size}
\addplot[fill=KArtic, draw=white, line width=0.3pt] table[x=k, y=artic]{fig/data/harness.dat};
\addlegendentry{articulation \& orientation}
\addplot[fill=KPlace, draw=white, line width=0.3pt] table[x=k, y=place]{fig/data/harness.dat};
\addlegendentry{placement}
\node[font=\tiny, anchor=south, inner sep=1.5pt] at (axis cs:1,18) {$+$18};
\node[font=\tiny, anchor=south, inner sep=1.5pt] at (axis cs:2,24) {$+$8\,/\,$-$2};
\node[font=\tiny, anchor=south, inner sep=1.5pt] at (axis cs:3,32) {$+$9\,/\,$-$1};
\node[font=\tiny, anchor=south, inner sep=1.5pt] at (axis cs:4,37) {$+$10\,/\,$-$5};
\node[font=\tiny, anchor=south, inner sep=1.5pt] at (axis cs:5,39) {$+$3\,/\,$-$1};
\end{axis}
\end{tikzpicture}
    \caption{
        \textbf{The harness over the rounds}: modules of each accepted $\mathcal H^I_k$ by function (from each
        module's own description); above the bars, modules added\,/\,removed relative to the previous round.
    }
    \label{fig:harness}
\end{figure}
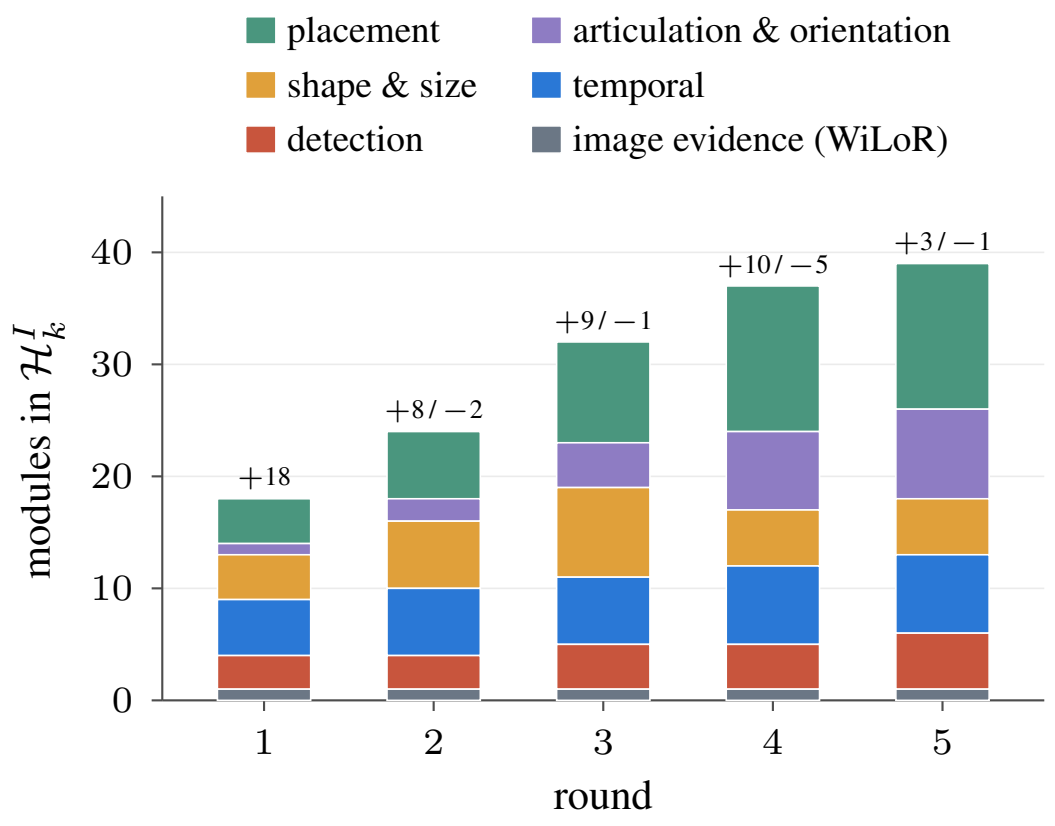

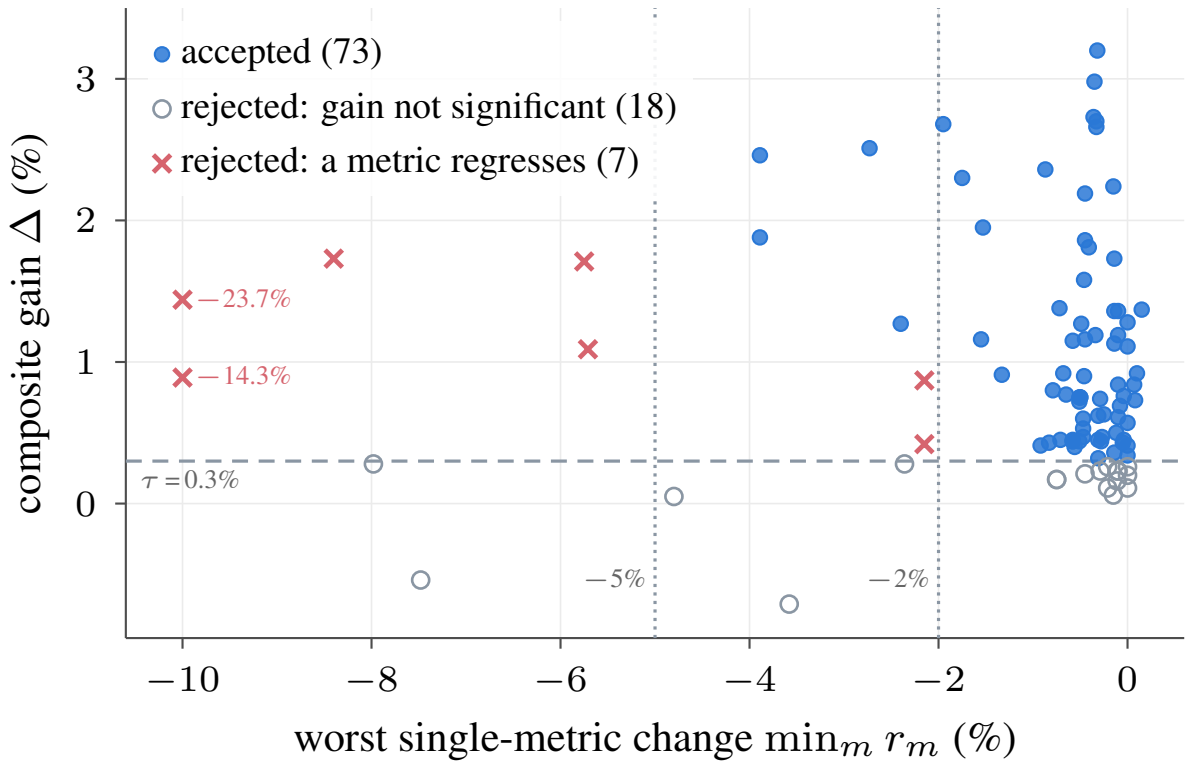
\begin{figure}[t]
    \centering
\begin{tikzpicture}
\begin{axis}[hr, width=0.84\linewidth, height=0.50\linewidth, xmin=-10.6, xmax=0.6, ymin=-0.95, ymax=3.5,
    xtick={-10,-8,...,0}, ytick={0,1,2,3}, clip mode=individual,
    xlabel={worst single-metric change $\min_m r_m$ (\%)}, ylabel={composite gain $\Delta$ (\%)},
    legend style={at={(0.02,0.98)}, anchor=north west, fill=white, fill opacity=0.9, text opacity=1}]
\addplot[CGray, densely dashed, line width=0.6pt, forget plot] coordinates {(-10.6,0.3) (0.6,0.3)};
\addplot[CGray, densely dotted, line width=0.6pt, forget plot] coordinates {(-2,-0.95) (-2,3.5)};
\addplot[CGray, densely dotted, line width=0.6pt, forget plot] coordinates {(-5,-0.95) (-5,3.5)};
\node[font=\tiny, text=black!60, anchor=north west, inner sep=1.5pt] at (axis cs:-10.55,0.3) {$\tau$\,=\,0.3\%};
\node[font=\tiny, text=black!60, anchor=north east, inner sep=1.5pt] at (axis cs:-2,-0.4) {$-$2\%};
\node[font=\tiny, text=black!60, anchor=north east, inner sep=1.5pt] at (axis cs:-5,-0.4) {$-$5\%};
\addplot[only marks, mark=*, mark size=1.35pt, HandRSIBlue, fill opacity=0.85, restrict expr to domain={\thisrow{c}}{0:0}] table[x=x, y=y]{fig/data/gate.dat};
\addlegendentry{accepted (73)}
\addplot[only marks, mark=o, mark size=1.6pt, CGray, line width=0.5pt, restrict expr to domain={\thisrow{c}}{1:1}] table[x=x, y=y]{fig/data/gate.dat};
\addlegendentry{rejected: gain not significant (18)}
\addplot[only marks, mark=x, mark size=2.3pt, CAce, line width=0.9pt, restrict expr to domain={\thisrow{c}}{2:2}] table[x=x, y=y]{fig/data/gate.dat};
\addlegendentry{rejected: a metric regresses (7)}
\node[font=\tiny, text=CAce, anchor=west, inner sep=2pt] at (axis cs:-10.0,1.440) {$-$23.7\%};
\node[font=\tiny, text=CAce, anchor=west, inner sep=2pt] at (axis cs:-10.0,0.890) {$-$14.3\%};
\end{axis}
\end{tikzpicture}
    \caption{
        \textbf{Every Loop~A gate decision} on val (the 98 of 156 submissions that passed the dev pre-check, five
        rounds): composite gain against the
        agent's incumbent and the worst single-metric change. Dashed: $\tau$; dotted: the regression thresholds
        ($r_m<-2\%$ when significant, $r_m<-5\%$ always). Two rejections beyond $-10\%$ are drawn at the border.
    }
    \label{fig:gate}
\end{figure}

\begin{table}[t]
    \centering
    \caption{\textbf{Loop~A per round}: submissions, of which gated on val (after the
    dev pre-check) and accepted by $G_A$; accepted/gated per agent family;
    output tokens of all agents.}
    \label{tab:loopa}
    \footnotesize
    \setlength{\tabcolsep}{3.5pt}
    \setlength{\aboverulesep}{0pt}
    \setlength{\belowrulesep}{0pt}
    \begin{tabular}{@{}lcccccc@{}}
        \toprule
        \rowcolor{HandRSIHeader}
        Round & subm. & gated & acc. & Claude & GPT & tokens (M) \\
        \midrule
        1 & 70 & 53 & 43 & 14/21 & 29/32 & 1.67 \\
        2 & 36 & 13 & 6 & 4/9 & 2/4 & 2.04 \\
        3 & 26 & 14 & 12 & 3/5 & 9/9 & 2.02 \\
        4 & 17 & 12 & 9 & 3/4 & 6/8 & 1.77 \\
        5 & 7 & 6 & 3 & 2/4 & 1/2 & 1.19 \\
        \midrule
        all & 156 & 98 & 73 & 26/43 & 47/55 & 8.68 \\
 
        \bottomrule
    \end{tabular}
\end{table}

\begin{table*}[t]
    \centering
    \caption{
        \textbf{System-level ablation} on HOT3D test, built from the controls
        scored in every round: network only, harness (or a baseline) only, and
        both.
    }
    \label{tab:rounds}
    \footnotesize
    \setlength{\tabcolsep}{3.2pt}
    \setlength{\aboverulesep}{0pt}
    \setlength{\belowrulesep}{0pt}
    \begin{tabular}{@{}l*{10}{c}@{}}
        \toprule
        \rowcolor{HandRSIHeader}
        \textbf{Variant} & FAcc$\uparrow$ & Recall$\uparrow$ & F1$\uparrow$
        & MPJPE-p$\downarrow$ & PA-p$\downarrow$ & MPJPE$^{+\mathrm{OOS}}{\downarrow}$
        & EPE$_{\mathrm{2D}}$-p$\downarrow$ & GO-p$\downarrow$ & CT-p$\downarrow$ & Jitter$\downarrow$ \\
        \midrule
        \rowcolor{HandRSIInit}
        ACE-Ego-Hand $\theta_0$, no harness & 0.970 & 0.986 & 0.990 & 12.97 & 5.90 & 12.84 & 9.66 & 6.90 & 0.024 & 2.82 \\
        \cmidrule[0.2pt](lr){1-11}
        GT-only fine-tuning of $\theta_0$ ($\mathcal C^1_1$) & 0.982 & 0.994 & \textbf{0.994} & 9.91 & 5.00 & 10.76 & 5.56 & 5.71 & 0.014 & 2.77 \\
        GT + distillation, round 1 ($\theta_1$) & 0.981 & 0.994 & \textbf{0.994} & 9.89 & 5.00 & 10.75 & 5.77 & 5.80 & 0.014 & 2.63 \\
        GT-only fine-tuning of $\theta_1$ ($\mathcal C^1_2$) & \textbf{0.983} & 0.994 & \textbf{0.994} & 9.39 & 4.76 & 10.33 & 5.32 & 5.71 & 0.013 & 2.73 \\
        GT + distillation, round 2 ($\theta_2$) & 0.982 & 0.994 & \textbf{0.994} & 9.37 & 4.71 & 10.35 & 5.31 & 5.80 & 0.013 & 2.51 \\
        \cmidrule[0.2pt](lr){1-11}
        $\mathcal H^I_1$ on $\theta_0$ ($\mathcal C^2_1$) & 0.976 & 0.993 & 0.992 & 11.13 & 5.09 & 11.73 & 7.96 & 5.92 & 0.021 & 1.75 \\
        $\mathcal H^I_5$ on $\theta_0$ ($\mathcal C^2_5$) & 0.972 & 0.986 & 0.991 & 10.38 & 5.21 & 10.34 & 7.56 & 6.56 & 0.024 & 1.44 \\
        Savitzky--Golay on $\theta_2$, tuned on val & 0.982 & 0.994 & \textbf{0.994} & 9.34 & 4.70 & 10.33 & 5.24 & 5.76 & 0.013 & 2.18 \\
        Naive WiLoR ensemble on $\theta_2$, tuned on val & 0.982 & 0.995 & \textbf{0.994} & 9.32 & 4.64 & 10.42 & 5.20 & 5.69 & 0.013 & 2.21 \\
        \cmidrule[0.2pt](lr){1-11}
        \rowcolor{HandRSILite}
        One round, $\mathcal S_1$ (18 modules) & 0.977 & \textbf{0.998} & 0.993 & 9.39 & 4.83 & 10.78 & 5.38 & 5.20 & \textbf{0.012} & 1.65 \\
        \rowcolor{HandRSIMedium}
        Two rounds, $\mathcal S_2$ (24 modules) & 0.978 & 0.997 & 0.993 & 8.65 & 4.51 & 10.07 & 4.53 & \textbf{5.14} & \textbf{0.012} & 1.52 \\
        \rowcolor{HandRSIHeavy}
        Five rounds, $\mathcal S_5$ (39 modules) & 0.982 & 0.996 & \textbf{0.994} & \textbf{8.37} & \textbf{4.37} & \textbf{9.70} & \textbf{4.07} & 5.16 & 0.013 & \textbf{1.34} \\
 
        \bottomrule
    \end{tabular}
\end{table*}

Table~\ref{tab:main} compares HandRSI with the strongest baselines and
Tab.~\ref{tab:all} with all ten re-run configurations of the nine published
methods.
On both benchmarks the full system outperforms every published method in
all ten metrics, and the best value of every column belongs to a HandRSI
system: relative to ACE-Ego-Hand, MPJPE-p drops by \hotRedMPJPE\% on HOT3D
and \arcRedMPJPE\% on ARCTIC, the 2D error on HOT3D more than halves, and
jitter roughly halves on both.
Against ACE-Ego-Hand fine-tuned on our GT ($\mathcal C^1_1$), which the
released baselines lack, HandRSI lowers MPJPE-p, 2D error and jitter by
\hotRedCOneMPJPE/\hotRedCOneEPE/\hotRedCOneJitter\% on HOT3D and
\arcRedCOneMPJPE/\arcRedCOneEPE/\arcRedCOneJitter\% on ARCTIC.
Even the Lite tier, at nearly the network's speed and without WiLoR, beats
every published method in every column.
On the unseen HOI4D camera (Tab.~\ref{tab:main}, all methods in Tab.~\ref{tab:hoi4d}),
zero-shot for every method, HandRSI has the lowest 3D pose errors of all published methods (MPJPE-p, PA-p
and MPJPE$^{+\mathrm{OOS}}$) and less than half of the next-lowest jitter,
while its 2D error, placement and detection stay behind ACE-Ego-Hand's:
neither the harness nor Loop~B saw HOI4D's camera.
This comes mainly from fine-tuning on HOT3D and ARCTIC cameras: GT
fine-tuning alone moves CT-p from \hoiAceCT{} to \hoiCOneCT\,m, while the
final harness on the original network ($\mathcal C^2_5$) keeps
\hoiCTwoFiveCT\,m, lowers MPJPE-p to \hoiCTwoFiveMPJPE\,mm and halves
jitter at a small 2D cost, so it can serve new cameras without retraining.

\paragraph{Where HOI4D's 2D cost comes from.}
On HOI4D, Lite raises $\theta_2$'s EPE$_{\mathrm{2D}}$-p from 17.39 to 19.86\,px.
About two thirds of the difference are 67 more missed hand-frames (358 vs.\
291 of 38{,}131, 0.18\%), each charged the image diagonal, and the rest is
the error of detected hands (11.16 vs.\ 10.41\,px).
Leaving out any one of Lite's 20 modules recovers at most 0.69\,px (a
temporal-consistency module, which also accounts for 22 of the extra
misses), followed by an anchor-fusion module (0.57\,px) and the
root-trajectory prior (0.45\,px); no single module causes the cost, and 3D
pose and jitter still improve (MPJPE-p 24.11 to 24.03\,mm, jitter 1.87 to
1.14).

The PCK curves (Fig.~\ref{fig:pck}, top) show that the gain is not driven by a
few outliers: HandRSI leads at every threshold above 4\,mm on both
benchmarks.
Figure~\ref{fig:qualitative} shows typical baseline failures that HandRSI
corrects, and Fig.~\ref{fig:temporal} (bottom) shows that the harness removes
the acceleration peaks that \mbox{ACE-Ego-Hand's} video latent causes every four
frames.
The network alone already accounts for part of this: Fig.~\ref{fig:clips} shows two
clips in which $\theta_2$, without any harness, finds a hand at the image border
that ACE-Ego-Hand misses and places an occluded hand that ACE-Ego-Hand misplaces.
Appendix~\ref{sec:app_gallery} extends Fig.~\ref{fig:qualitative} to four more
frames per test set, each with the hands in 3D from a third-person view; there
too, the baselines fail mostly by missing a hand or by placing it at a wrong
depth.

\subsection{Recursive Improvement}
\label{sec:exp_recursion}

\begin{table}[t]
    \centering
    \caption{
        \textbf{What to distill} (round~1, HOT3D dev): Loop~B recipes from the
        same GT-fine-tuned network and budget; $\Delta$ against GT only.
        $^\ddagger$Indented rows build on this row.
    }
    \label{tab:distill}
    \footnotesize
    \setlength{\tabcolsep}{3pt}
    \setlength{\aboverulesep}{0pt}
    \setlength{\belowrulesep}{0pt}
    \begin{tabular}{@{}l*{5}{c}@{}}
        \toprule
        \rowcolor{HandRSIHeader}
        \textbf{Loop B recipe} & FAcc & MPJPE-p
        & EPE$_{\mathrm{2D}}$-p & Jitter & $\Delta$ (\%) \\
        \midrule
        GT only & 0.954 & 10.24 & 5.37 & 2.73 & -- \\
        + teacher articulation & 0.965 & 10.60 & 5.75 & 2.76 & $-$1.62 \\
        + articulation, temp.\ 0.3 & 0.942 & 10.38 & 5.53 & 2.33 & $-$0.70 \\
        + presence, 2D, temp.\ 0.3 & 0.944 & 10.40 & 5.58 & 2.32 & $-$1.01 \\
        + presence, temp.\ 0.3$^\ddagger$ & 0.954 & 10.19 & 5.39 & 2.32 & $+$1.39 \\
        \quad temp.\ 0.1 & 0.961 & 10.24 & 5.33 & 2.54 & $+$1.95 \\
        \quad + border mask & 0.962 & 10.58 & 5.62 & 2.65 & $-$1.14 \\
        \rowcolor{HandRSIHeavy}
        \quad + visibility (final) & 0.965 & 10.42 & 5.50 & 2.61 & $+$0.22 \\
 
        \bottomrule
    \end{tabular}
\end{table}

\begin{table}[t]
    \centering
    \caption{\textbf{Harness modules kept by re-validation} for each Loop~B recipe.}
    \label{tab:reval}
    \small
    \setlength{\tabcolsep}{4pt}
    \resizebox{\linewidth}{!}{%
    \setlength{\aboverulesep}{0pt}
    \setlength{\belowrulesep}{0pt}
    \begin{tabular}{@{}llc@{}}
        \toprule
        \rowcolor{HandRSIHeader}
        Round & Loop B recipe & kept / total \\
        \midrule
        1 (v1) & teacher pose and placement (frames without GT) & 6 / 18 \\
        1 (v2) & teacher articulation only & 15 / 18 \\
        1 (v3) & teacher presence + temporal 0.3 & 6 / 18 \\
        1 (final) & teacher presence + temporal 0.1, border mask, visibility & 18 / 18 \\
        2 & same as final & 25 / 25 \\
        \bottomrule
    \end{tabular}}
\end{table}

\begin{table}[t]
    \centering
    \caption{
        \textbf{Deployment tiers} chosen on val (Fig.~\ref{fig:pareto}); test
        accuracy in Tab.~\ref{tab:main}.
        Harness time per frame on one RTX PRO~5000, after the network.
    }
    \label{tab:tiers}
    \footnotesize
    \setlength{\tabcolsep}{2.6pt}
    \setlength{\aboverulesep}{0pt}
    \setlength{\belowrulesep}{0pt}
    \begin{tabular}{@{}lcccc@{}}
        \toprule
        \rowcolor{HandRSIHeader}
        \textbf{System} & Modules & FPS$\uparrow$ & harness ms/frame$\downarrow$ & val $\Delta$ (\%)$\uparrow$ \\
        \midrule
        ACE-Ego-Hand ($\theta_0$) & -- & 42.7 & 0 & -- \\
        $\theta_2$ alone & 0 & 42.5 & 0 & $+$0.00 \\
        \rowcolor{HandRSILite}
        HandRSI-Lite & 20 & 41.1 & 0.9 & $+$3.90 \\
        \rowcolor{HandRSIMedium}
        HandRSI-Medium & 28 & 34.0 & 6.0 & $+$4.06 \\
        \rowcolor{HandRSIHeavy}
        HandRSI-Heavy & 29 & 20.7 & 25.0 & $+$4.93 \\
        $\mathcal S_5$ (loop output) & 39 & 16.2 & 38.5 & $+$4.84 \\
 
        \bottomrule
    \end{tabular}
\end{table}

\paragraph{Rounds.}
Pose and jitter improve on both benchmarks in each of the first four
rounds; in round~5 pose levels off while jitter still improves
(Figs.~\ref{fig:rounds} and~\ref{fig:evolution}, Tab.~\ref{tab:loop}).
Distillation saturates after round~2: under three Loop~C recipes, every
student trades 2D, orientation and recall for jitter on val (Tab.~\ref{tab:supp_loopb_sat}), so $\theta_2$ stays while the harness
gains until the round gate stops the loop.

\paragraph{Why distillation saturates.}
From round~3 on, GT fine-tuning of $\theta_2$ is no longer ahead on dev (a
$\Delta$ of $+$0.3 to $+$0.9\% whose interval includes zero, with 5--6\% more
jitter), so the teacher is $\theta_2$ with the round's harness and the student
starts from $\theta_2$.
Loop~C changed the recipe for each of these rounds
(Tab.~\ref{tab:supp_loopb_sat}), and all twelve student checkpoints have the
same profile on val: jitter improves by 1--3\%, but 2D alignment,
orientation and recall lose 1.7--9.3\% each, for a $\Delta$ of $-$0.8 to
$-$2.7\%, whereas round~2's student gained 6\% MPJPE-p and 11\%
EPE$_{\mathrm{2D}}$-p.
At this LoRA budget, more training on the same labeled clips and teacher
terms no longer helps the network alone, and what the harness still adds,
placement and articulation fitted to image evidence, is what
Tab.~\ref{tab:distill} shows a student cannot take from its teacher without
losing pose.

\begin{table}[t]
    \centering
    \caption{Loop~B's best student checkpoint per round on val (network alone vs.\ $\theta_{k-1}$): change set
    by Loop~C, $\Delta$ with its CI$_{90}$ lower bound, and $r_m$ (\%) of five columns. All four checkpoints of
    rounds 3--5 have the same signs.}
    \label{tab:supp_loopb_sat}
    \small
    \setlength{\tabcolsep}{3pt}
    \resizebox{\linewidth}{!}{%
    \setlength{\aboverulesep}{0pt}
    \setlength{\belowrulesep}{0pt}
    \begin{tabular}{@{}clcccccc@{}}
        \toprule
        \rowcolor{HandRSIHeader}
        R & Loop~C change & $\Delta$ ($L_\Delta$) & MPJPE & EPE$_{\mathrm{2D}}$ & GO & Jitter & Recall \\
        \midrule
        2 & -- & $+$1.66 ($+$0.80) & $+$6.4 & $+$10.6 & $-$0.8 & $+$4.0 & $+$2.2 \\
        3 & $\lambda_R$ 0.5 & $-$0.84 ($-$2.12) & $-$1.5 & $-$2.6 & $-$3.3 & $+$2.1 & $-$1.7 \\
        4 & articulation on HOT3D & $-$1.09 ($-$2.55) & $-$1.0 & $-$3.3 & $-$4.8 & $+$1.9 & $-$2.2 \\
        5 & $\lambda_{\mathrm{KD}}$ 1, lr $5{\cdot}10^{-5}$ & $-$2.35 ($-$3.21) & $-$2.1 & $-$6.8 & $-$8.0 & $+$1.1 & $-$6.5 \\
        \bottomrule
    \end{tabular}}
\end{table}

\paragraph{Every stage of every round.}
Appendix~\ref{sec:app_results} gives every stage of every round on the three
test sets, with the GT-only control, the teacher and the harness on the
original network (Tabs.~\ref{tab:stages_hot3d}--\ref{tab:stages_hoi4d}).
In every round the system lowers its network's MPJPE-p on all three test
sets, and the gains compound: on HOT3D, MPJPE-p falls from 12.97\,mm
($\theta_0$) to 8.36\,mm by round~4, and the harness of a later round is
stronger even on the original network ($\mathcal C^2_5$ 10.38 vs.\
$\mathcal C^2_1$ 11.13\,mm, from 8.7M vs.\ 1.7M Loop~A output tokens,
Tab.~\ref{tab:loopa}).
From round~3 on the network stays $\theta_2$, and the harness alone lowers
\mbox{MPJPE-p} from 8.65 to 8.36\,mm by round~4.
Figure~\ref{fig:app_rounds_gallery} in the appendix follows three more HOT3D
test clips through the rounds.

\subsection{What the Agents Wrote}
\label{sec:exp_agents}

\paragraph{Harness content.}
Early rounds write temporal, shape and placement modules; later rounds
refine the output against image evidence and rescue hands at the image
border (Fig.~\ref{fig:harness}).
The 39 modules of $\mathcal H^I_5$ form six families: \emph{temporal} (7),
minimum-jerk and local-polynomial priors on root and MANO rotations, one
aware of the network's four-frame keyframes; \emph{placement} (13), bounded
root corrections fitted to a fusion of the network's 2D keypoints and its
projected MANO joints; \emph{articulation} (8), regularized finger fits to
reliable 2D evidence; \emph{shape} (5), one size per hand track, consistent
across both hands; \emph{presence} (5), border rescue and image checks; and
one \emph{image} module that runs WiLoR~\cite{wilor} (its released weights and
hand detector) on a crop around each hand and blends WiLoR's finger rotations
into the network's where both agree in 2D; four later modules also use
WiLoR's 2D and 3D keypoints as image evidence.
WiLoR is the only pretrained network besides the estimator that a module may
call; the Lite and Medium tiers do not contain it, and without it the
modules that read its output use the network's predictions alone.
Later rounds add mostly placement and articulation modules that refine the
network's output against image evidence (4 and 1 modules in round~1, 13 and
8 in round~5).

\paragraph{The gate at work.}
The multi-metric gate shapes this code.
Replaying the final harness without one module on 278 HOT3D test segments,
removing the WiLoR-based image-evidence module costs 0.7\,mm MPJPE-p, whereas
removing a detection module lowers frame accuracy but slightly improves
pose, so a gate on any single metric would have rejected one of them.
Of 156 submissions, 98 passed the dev pre-check and were gated on val
(Tab.~\ref{tab:loopa}, Fig.~\ref{fig:gate}).
The GPT-6-sol agents passed $G_A$ more often (47 of 55 gated submissions)
than the Claude Opus~5.5 agents (26 of 43); both agent families contributed
modules to every round's harness.
Seven submissions with a significant composite gain were rejected because
one metric regressed by 2.2--23.7\%, which a composite-only gate
would have accepted, and accepted submissions regress no metric by more than
3.9\%, and only where the regression is not significant.

\subsection{Ablation Studies}
\label{sec:exp_ablation}

\begin{table*}[t]
    \centering
    \caption{
        \textbf{Classical post-processing} against the agent-written harness on the sealed tests, on the
        untouched network $\theta_0$ and on $\theta_2$. Savitzky--Golay (window 7, order 2) is the best of 30
        filter settings on HOT3D val; 2D alignment refits the hand's image-plane translation to the network's own
        2D keypoints by least squares before smoothing (the best alignment variant on val). WiLoR ensemble:
        WiLoR's finger rotations blended into the network's with one global weight (0.25, selected on val), then
        Savitzky--Golay. Units as in Tab.~\ref{tab:main}; best in bold.
    }
    \label{tab:classical}
    \footnotesize
    \setlength{\tabcolsep}{3.2pt}
    \setlength{\aboverulesep}{0pt}
    \setlength{\belowrulesep}{0pt}
    \begin{tabular}{@{}l*{10}{c}@{}}
        \toprule
        \rowcolor{HandRSIHeader}
        \textbf{Method} & FAcc$\uparrow$ & Recall$\uparrow$ & F1$\uparrow$ & MPJPE-p$\downarrow$ & PA-p$\downarrow$ & MPJPE$^{+\mathrm{OOS}}{\downarrow}$
        & EPE$_{\mathrm{2D}}$-p$\downarrow$ & GO-p$\downarrow$ & CT-p$\downarrow$ & Jitter$\downarrow$ \\
        \midrule
        \multicolumn{11}{l}{\textit{HOT3D test}} \\
        \rowcolor{HandRSIInit}
        ACE-Ego-Hand $\theta_0$, no harness & 0.970 & 0.986 & 0.990 & 12.97 & 5.90 & 12.84 & 9.66 & 6.90 & 0.024 & 2.82 \\
        $\theta_0$ + Savitzky--Golay & 0.970 & 0.986 & 0.990 & 12.94 & 5.89 & 12.82 & 9.60 & 6.86 & 0.024 & 2.44 \\
        $\theta_0$ + 2D alignment + Savitzky--Golay & 0.863 & 0.997 & 0.958 & 11.63 & 5.86 & 12.82 & 8.70 & 5.26 & 0.022 & 2.41 \\
        $\mathcal H^I_5$ on $\theta_0$ ($\mathcal C^2_5$) & 0.972 & 0.986 & 0.991 & 10.38 & 5.21 & 10.34 & 7.56 & 6.56 & 0.024 & 1.44 \\
        \cmidrule[0.2pt](lr){1-11}
        HandRSI network $\theta_2$, no harness & 0.982 & 0.994 & 0.994 & 9.37 & 4.71 & 10.35 & 5.31 & 5.80 & 0.013 & 2.51 \\
        $\theta_2$ + Savitzky--Golay & 0.982 & 0.994 & 0.994 & 9.34 & 4.70 & 10.33 & 5.24 & 5.76 & 0.013 & 2.18 \\
        $\theta_2$ + 2D alignment + Savitzky--Golay & 0.862 & \textbf{0.999} & 0.958 & 8.68 & 4.68 & 10.33 & 4.87 & \textbf{5.00} & 0.013 & 2.14 \\
        $\theta_2$ + WiLoR ensemble + Savitzky--Golay & 0.982 & 0.995 & 0.994 & 9.32 & 4.64 & 10.42 & 5.20 & 5.69 & 0.013 & 2.21 \\
        \rowcolor{HandRSILite}
        HandRSI-Lite & 0.982 & 0.996 & 0.994 & 9.06 & 4.69 & 10.31 & 4.76 & 5.44 & \textbf{0.012} & 1.47 \\
        \rowcolor{HandRSIMedium}
        HandRSI-Medium & 0.982 & 0.996 & 0.994 & 9.04 & 4.64 & 10.30 & 4.79 & 5.46 & 0.013 & \textbf{1.33} \\
        \rowcolor{HandRSIHeavy}
        HandRSI-Heavy (with WiLoR) & \textbf{0.983} & 0.996 & \textbf{0.995} & 8.41 & \textbf{4.35} & 9.71 & 4.16 & 5.17 & 0.013 & 1.35 \\
        \rowcolor{HandRSIHeavy}
        HandRSI ($\mathcal S_5$) & 0.982 & 0.996 & 0.994 & \textbf{8.37} & 4.37 & \textbf{9.70} & \textbf{4.07} & 5.16 & 0.013 & 1.34 \\
        \midrule
        \multicolumn{11}{l}{\textit{ARCTIC test}} \\
        \rowcolor{HandRSIInit}
        ACE-Ego-Hand $\theta_0$, no harness & 0.986 & 0.995 & 0.996 & 15.58 & 7.43 & 16.54 & 9.36 & 12.45 & 0.024 & 2.58 \\
        $\theta_0$ + Savitzky--Golay & 0.988 & 0.996 & 0.997 & 15.52 & 7.42 & 16.53 & 9.20 & 12.38 & 0.023 & 2.40 \\
        $\theta_0$ + 2D alignment + Savitzky--Golay & 0.932 & \textbf{1.000} & 0.981 & 15.17 & 7.41 & 16.53 & 8.84 & 11.79 & 0.021 & 2.33 \\
        $\mathcal H^I_5$ on $\theta_0$ ($\mathcal C^2_5$) & \textbf{0.992} & 0.999 & \textbf{0.998} & 14.20 & 6.35 & 15.50 & 7.65 & 11.77 & 0.021 & 1.52 \\
        \cmidrule[0.2pt](lr){1-11}
        HandRSI network $\theta_2$, no harness & 0.987 & 0.995 & 0.997 & 15.15 & 7.28 & 16.11 & 9.12 & 12.34 & 0.021 & 2.36 \\
        $\theta_2$ + Savitzky--Golay & 0.988 & 0.996 & 0.997 & 15.10 & 7.27 & 16.10 & 8.99 & 12.28 & 0.021 & 2.21 \\
        $\theta_2$ + 2D alignment + Savitzky--Golay & 0.930 & \textbf{1.000} & 0.981 & 14.76 & 7.25 & 16.10 & 8.61 & 11.70 & \textbf{0.019} & 2.15 \\
        $\theta_2$ + WiLoR ensemble + Savitzky--Golay & 0.989 & 0.996 & 0.997 & 15.00 & 6.80 & 16.06 & 8.88 & 12.19 & 0.021 & 2.23 \\
        \rowcolor{HandRSILite}
        HandRSI-Lite & 0.990 & 0.999 & 0.997 & 14.53 & 7.25 & 15.84 & 8.80 & 11.76 & \textbf{0.019} & 1.58 \\
        \rowcolor{HandRSIMedium}
        HandRSI-Medium & 0.990 & 0.999 & 0.997 & 14.47 & 7.14 & 15.78 & 8.78 & 11.76 & \textbf{0.019} & \textbf{1.45} \\
        \rowcolor{HandRSIHeavy}
        HandRSI-Heavy (with WiLoR) & 0.991 & \textbf{1.000} & \textbf{0.998} & 14.00 & 6.44 & 15.33 & 7.69 & \textbf{11.59} & \textbf{0.019} & 1.46 \\
        \rowcolor{HandRSIHeavy}
        HandRSI ($\mathcal S_5$) & 0.990 & \textbf{1.000} & 0.997 & \textbf{13.85} & \textbf{6.34} & \textbf{15.18} & \textbf{7.61} & 11.61 & \textbf{0.019} & \textbf{1.45} \\
 
        \bottomrule
    \end{tabular}
\end{table*}

\paragraph{System-level contributions.}
Table~\ref{tab:rounds} isolates the factors with one-component
controls.
\textbf{GT fine-tuning} ($\mathcal C^1$) is a strong baseline on HOT3D
($\mathcal C^1_1$: \hotCOneMPJPE\,mm, Tab.~\ref{tab:main}) but gives only
\arcFtShareMPJPE\% of the MPJPE-p gain on ARCTIC and at most \ftShareJitterMax\% of
the jitter gain; the rest is the loop's.
\textbf{Distillation} on top of the same GT leaves 3D pose unchanged but
transfers temporal stability ($\theta_k$ vs.\ $\mathcal C^1_k$).
\textbf{Harness evolution alone} ($\mathcal C^2$) substantially improves even
the untouched base network, which neither classical post-processing nor a
naive WiLoR ensemble does (below).
\textbf{Recursion} compounds the two: harnesses from later rounds are
stronger even on the original network
($\mathcal H^I_5$ vs.\ $\mathcal H^I_1$ on $\theta_0$: 10.38 vs.\ 11.13\,mm), and
only their combination with the retrained network reaches 8.37\,mm.

\paragraph{What to distill.}
Distilling the teacher's articulation hurts pose and placement, and strong
temporal targets keep hands alive out of view
(Tab.~\ref{tab:distill}); the two refinements against border false
positives lower $\Delta$ on dev.
The last three variants (temporal 0.1, border mask, final) were also scored
on val, where only the final recipe, which takes presence and in-image
temporal behavior from the teacher, has no blocking regression, though
temporal 0.1 is ahead on dev; the other two regress HOT3D FAcc past the
gate's limits.
It also keeps every harness module useful under re-validation
(Tab.~\ref{tab:reval}), whereas recipes that distill the teacher's placement
or strong temporal smoothing make two thirds of the harness redundant: with
the final recipe, the network and the harness complement each other.

\paragraph{Classical post-processing.}
To see how much of the harness a standard filter recovers, we apply
classical post-processing to the network's outputs (MANO rotations,
projected back to $SO(3)$, translation and 2D keypoints), with the same
interface as a harness module and no GT: zero-phase Gaussian (6 settings)
and Savitzky--Golay~\cite{savitzky1964smoothing} smoothing (8) and the
causal One-Euro filter~\cite{casiez2012one} (16), each also after a
least-squares alignment of the translation to the network's 2D keypoints.
Settings are chosen by $\Delta$ on HOT3D val against $\theta_2$, as a
Loop~A submission would be.
The best filter, Savitzky--Golay, passes $G_A$ ($\Delta=+0.90\%$) but on
test cuts jitter by 13\% with almost no change in pose; One-Euro lags behind
fast motion and regresses pose in every setting.
The best alignment variant reaches $\Delta=+2.03\%$ and lowers MPJPE-p to
8.68\,mm on HOT3D test, but it moves hands whose 2D keypoints are unreliable
and drops frame accuracy by 12\%, which the gate rejects
(Tab.~\ref{tab:classical}).
The harness keeps detection intact while lowering pose error and jitter
further, on $\theta_0$ as on $\theta_2$.

\paragraph{Naive WiLoR ensemble.}
Heavy owes its last pose gain to modules that read WiLoR~\cite{wilor}, so we
also test whether a plain two-model ensemble explains it.
WiLoR runs on its demo crop around each hand the network detects, and its
finger rotations are blended into the network's by spherical interpolation
with one global weight $\alpha$, with no outlier gate and no temporal
smoothing of WiLoR's output, optionally followed by the Savitzky--Golay
filter above.
Selected on HOT3D val like the filters ($\alpha\in\{0.25,0.5,0.75,1\}$, with
or without smoothing), the best setting, $\alpha=0.25$ with smoothing, passes
$G_A$ ($\Delta=+2.49\%$, below Lite's $+3.90\%$); larger weights raise the
pose error.
On test it lowers the MPJPE-p of $\theta_2$ by 0.05\,mm on HOT3D and
0.15\,mm on ARCTIC, against 0.96 and 1.15\,mm for Heavy
(Tab.~\ref{tab:classical}): the image evidence pays off as the agents
integrated it, with perspective crops, agreement gates and band-limited
fusion (Sec.~\ref{sec:loopA}, Fig.~\ref{fig:harness}).

\subsection{Accuracy--Speed Tiers}
\label{sec:exp_tiers}

Most of the gain comes from cheap modules (Fig.~\ref{fig:pareto},
Tabs.~\ref{tab:main} and~\ref{tab:tiers}): Lite keeps half of the modules at
almost the network's speed and already cuts the network's jitter by a third
or more, Medium adds shape, articulation and smoothing modules for the
lowest HOT3D jitter, and Heavy adds WiLoR's image evidence for the
remaining pose gain, within 0.15\,mm MPJPE-p of the full system at 28\%
higher speed; together the three tiers cover \hvRatio\% of the whole
front's hypervolume.
Fig.~\ref{fig:pareto}b,c shows the same split on all three tests.
Appendix~\ref{sec:app_results} shows all \nSearched{} searched configurations
(Fig.~\ref{fig:supp_search}).

\subsection{Reproduction}
\label{sec:exp_repro}

The released runner and evaluator, with the final weights and harnesses,
reproduce all eight HOT3D test rows of rounds 0--5 on a different GPU type
(NVIDIA H20) within 0.01\,mm MPJPE-p, and its video interface matches the
evaluation predictions to about 0.3\,mm on average.
Decoded with whole-sequence context instead of 81-frame segments, the
network alone improves further (Appendix~\ref{sec:app_results},
Fig.~\ref{fig:seqlen}).

\section{Conclusion}
\label{sec:conclu}

HandRSI lets coding agents improve a 3D hand pose system, its harness and,
by distillation, its network, under a gate on every benchmark
metric.
Five rounds give the best values in all ten metrics on ARCTIC and HOT3D
among nine re-run methods, and three tiers trade accuracy for
speed.
Three lessons emerge: gate on all
metrics at once; take from the teacher only what GT lacks; and re-check the harness after each model update and for
deployment cost.

\FloatBarrier
\section*{LLM Usage Statement}

The agents of the method (Sec.~\ref{sec:method}) run Claude Opus~5.5 (Claude Code) and GPT-6-sol (Codex).
The authors directed the study; Claude Opus~5.5 (Claude Code) assisted them in implementing and running
experiment code, producing figures and tables, checking references, discussing design options, and drafting
and editing text.
The authors reviewed all of it and take full responsibility for the report.

{
    \small
    \bibliographystyle{ieeenat_fullname}
    \bibliography{main}
}

\appendix
\section{Complete Results}
\label{sec:app_results}
\raggedbottom

\renewcommand{\dblfloatpagefraction}{0.75}
\begin{table*}[p]
    \centering
    \caption{\textbf{HOT3D test, every stage of every round.} Per round~$k$: the GT-only control $\mathcal C^1_k$, the
    teacher $\mathcal T_{k-1}$, the network $\theta_k$, the system $\mathcal S_k$ and the round's harness on the original
    network ($\mathcal C^2_k$). Units as in Tab.~\ref{tab:main}.}
    \label{tab:stages_hot3d}
    \scriptsize
    \setlength{\tabcolsep}{4pt}
    \renewcommand{\arraystretch}{0.92}
    \setlength{\aboverulesep}{0pt}
    \setlength{\belowrulesep}{0pt}
    \begin{tabular}{@{}l*{10}{c}@{}}
        \toprule
        \rowcolor{HandRSIHeader}
        \textbf{System} & FAcc$\uparrow$ & Recall$\uparrow$ & F1$\uparrow$ & MPJPE-p$\downarrow$ & PA-p$\downarrow$ & MPJPE$^{+\mathrm{OOS}}{\downarrow}$
        & EPE$_{\mathrm{2D}}$-p$\downarrow$ & GO-p$\downarrow$ & CT-p$\downarrow$ & Jitter$\downarrow$ \\
        \midrule
        \rowcolor{HandRSIInit}
        $\theta_0$ & 0.970 & 0.986 & 0.990 & 12.97 & 5.90 & 12.84 & 9.66 & 6.90 & 0.024 & 2.82 \\
        \multicolumn{11}{l}{\textit{Round 1}} \\
        \quad $\mathcal C^1_{k}$ & 0.982 & 0.994 & 0.994 & 9.91 & 5.00 & 10.76 & 5.56 & 5.71 & 0.014 & 2.77 \\
        \quad teacher $\mathcal T_{k-1}$ & \textbf{0.984} & 0.995 & \textbf{0.995} & 9.82 & 4.96 & 10.75 & 5.01 & 5.58 & 0.013 & 1.74 \\
        \quad $\theta_1$ & 0.981 & 0.994 & 0.994 & 9.89 & 5.00 & 10.75 & 5.77 & 5.80 & 0.014 & 2.63 \\
        \rowcolor{HandRSILite}
        \quad $\mathcal S_1$ & 0.977 & \textbf{0.998} & 0.993 & 9.39 & 4.83 & 10.78 & 5.38 & 5.20 & \textbf{0.012} & 1.65 \\
        \quad $\mathcal C^2_{k}$ & 0.976 & 0.993 & 0.992 & 11.13 & 5.09 & 11.73 & 7.96 & 5.92 & 0.021 & 1.75 \\
        \multicolumn{11}{l}{\textit{Round 2}} \\
        \quad $\mathcal C^1_{k}$ & 0.983 & 0.994 & 0.994 & 9.39 & 4.76 & 10.33 & 5.32 & 5.71 & 0.013 & 2.73 \\
        \quad teacher $\mathcal T_{k-1}$ & 0.980 & \textbf{0.998} & 0.994 & 8.47 & 4.54 & 9.96 & 4.40 & \textbf{4.93} & \textbf{0.012} & 1.61 \\
        \quad $\theta_2$ & 0.982 & 0.994 & 0.994 & 9.37 & 4.71 & 10.35 & 5.31 & 5.80 & 0.013 & 2.51 \\
        \rowcolor{HandRSILite}
        \quad $\mathcal S_2$ & 0.978 & 0.997 & 0.993 & 8.65 & 4.51 & 10.07 & 4.53 & 5.14 & \textbf{0.012} & 1.52 \\
        \quad $\mathcal C^2_{k}$ & 0.974 & 0.990 & 0.992 & 10.81 & 5.22 & 11.12 & 7.90 & 6.07 & 0.022 & 1.65 \\
        \multicolumn{11}{l}{\textit{Round 3}} \\
        \quad $\mathcal C^1_{k}$ & 0.983 & 0.994 & 0.994 & 9.21 & 4.62 & 10.27 & 5.22 & 5.74 & 0.013 & 2.66 \\
        \quad Loop~B best ckpt (rejected) & 0.982 & 0.994 & 0.994 & 9.47 & 4.71 & 10.48 & 5.34 & 5.90 & 0.014 & 2.47 \\
        \quad $\theta_3$ & 0.982 & 0.994 & 0.994 & 9.37 & 4.71 & 10.35 & 5.31 & 5.80 & 0.013 & 2.51 \\
        \rowcolor{HandRSILite}
        \quad $\mathcal S_3$ & 0.980 & 0.996 & 0.994 & 8.47 & 4.45 & 9.83 & 4.21 & 5.16 & \textbf{0.012} & 1.44 \\
        \quad $\mathcal C^2_{k}$ & 0.971 & 0.986 & 0.991 & 10.90 & 5.37 & 10.87 & 7.97 & 6.55 & 0.024 & 1.57 \\
        \multicolumn{11}{l}{\textit{Round 4}} \\
        \quad $\mathcal C^1_{k}$ & 0.983 & 0.994 & 0.994 & 9.19 & 4.62 & 10.25 & 5.21 & 5.74 & 0.013 & 2.67 \\
        \quad Loop~B best ckpt (rejected) & 0.981 & 0.994 & 0.994 & 9.55 & 4.72 & 10.53 & 5.38 & 6.00 & 0.014 & 2.46 \\
        \quad $\theta_4$ & 0.982 & 0.994 & 0.994 & 9.37 & 4.71 & 10.35 & 5.31 & 5.80 & 0.013 & 2.51 \\
        \rowcolor{HandRSILite}
        \quad $\mathcal S_4$ & 0.980 & 0.997 & 0.994 & \textbf{8.36} & 4.39 & 9.72 & \textbf{4.06} & 5.13 & \textbf{0.012} & 1.37 \\
        \quad $\mathcal C^2_{k}$ & 0.971 & 0.986 & 0.991 & 10.36 & 5.22 & 10.35 & 7.55 & 6.53 & 0.024 & 1.48 \\
        \multicolumn{11}{l}{\textit{Round 5}} \\
        \quad $\mathcal C^1_{k}$ & 0.983 & 0.994 & 0.994 & 9.23 & 4.64 & 10.23 & 5.24 & 5.75 & 0.013 & 2.66 \\
        \quad Loop~B best ckpt (rejected) & 0.980 & 0.993 & 0.993 & 9.60 & 4.74 & 10.43 & 5.44 & 6.05 & 0.014 & 2.49 \\
        \quad $\theta_5$ & 0.982 & 0.994 & 0.994 & 9.37 & 4.71 & 10.35 & 5.31 & 5.80 & 0.013 & 2.51 \\
        \rowcolor{HandRSILite}
        \quad $\mathcal S_5$ & 0.982 & 0.996 & 0.994 & 8.37 & \textbf{4.37} & \textbf{9.70} & 4.07 & 5.16 & 0.013 & \textbf{1.34} \\
        \quad $\mathcal C^2_{k}$ & 0.972 & 0.986 & 0.991 & 10.38 & 5.21 & 10.34 & 7.56 & 6.56 & 0.024 & 1.44 \\
 
        \bottomrule
    \end{tabular}
\end{table*}
\begin{table*}[p]
    \centering
    \caption{\textbf{ARCTIC s05 (test), every stage of every round} (as Tab.~\ref{tab:stages_hot3d}).}
    \label{tab:stages_arctic}
    \scriptsize
    \setlength{\tabcolsep}{4pt}
    \renewcommand{\arraystretch}{0.92}
    \setlength{\aboverulesep}{0pt}
    \setlength{\belowrulesep}{0pt}
    \begin{tabular}{@{}l*{10}{c}@{}}
        \toprule
        \rowcolor{HandRSIHeader}
        \textbf{System} & FAcc$\uparrow$ & Recall$\uparrow$ & F1$\uparrow$ & MPJPE-p$\downarrow$ & PA-p$\downarrow$ & MPJPE$^{+\mathrm{OOS}}{\downarrow}$
        & EPE$_{\mathrm{2D}}$-p$\downarrow$ & GO-p$\downarrow$ & CT-p$\downarrow$ & Jitter$\downarrow$ \\
        \midrule
        \rowcolor{HandRSIInit}
        $\theta_0$ & 0.986 & 0.995 & 0.996 & 15.58 & 7.43 & 16.54 & 9.36 & 12.45 & 0.024 & 2.58 \\
        \multicolumn{11}{l}{\textit{Round 1}} \\
        \quad $\mathcal C^1_{k}$ & 0.988 & 0.996 & 0.997 & 15.51 & 7.36 & 16.49 & 9.28 & 12.40 & 0.023 & 2.57 \\
        \quad teacher $\mathcal T_{k-1}$ & 0.990 & 0.997 & 0.997 & 15.31 & 6.84 & 16.41 & 8.82 & 12.19 & 0.022 & 1.80 \\
        \quad $\theta_1$ & 0.987 & 0.996 & 0.996 & 15.40 & 7.36 & 16.38 & 9.20 & 12.50 & 0.022 & 2.44 \\
        \rowcolor{HandRSILite}
        \quad $\mathcal S_1$ & 0.987 & \textbf{1.000} & 0.997 & 14.84 & 6.69 & 16.18 & 8.31 & 11.92 & 0.019 & 1.72 \\
        \quad $\mathcal C^2_{k}$ & 0.989 & \textbf{1.000} & 0.997 & 14.95 & 6.67 & 16.29 & 8.26 & 11.84 & 0.020 & 1.79 \\
        \multicolumn{11}{l}{\textit{Round 2}} \\
        \quad $\mathcal C^1_{k}$ & 0.988 & 0.996 & 0.997 & 15.26 & 7.32 & 16.25 & 9.10 & 12.43 & 0.022 & 2.51 \\
        \quad teacher $\mathcal T_{k-1}$ & 0.989 & 0.999 & 0.997 & 14.10 & 6.60 & 15.40 & 7.97 & 11.76 & 0.019 & 1.68 \\
        \quad $\theta_2$ & 0.987 & 0.995 & 0.997 & 15.15 & 7.28 & 16.11 & 9.12 & 12.34 & 0.021 & 2.36 \\
        \rowcolor{HandRSILite}
        \quad $\mathcal S_2$ & 0.989 & 0.999 & 0.997 & 13.97 & 6.59 & 15.26 & 7.89 & 11.65 & \textbf{0.018} & 1.60 \\
        \quad $\mathcal C^2_{k}$ & 0.991 & 0.999 & \textbf{0.998} & 14.35 & 6.61 & 15.62 & 8.02 & 11.79 & 0.021 & 1.69 \\
        \multicolumn{11}{l}{\textit{Round 3}} \\
        \quad $\mathcal C^1_{k}$ & 0.989 & 0.996 & 0.997 & 15.23 & 7.27 & 16.26 & 9.00 & 12.38 & 0.021 & 2.48 \\
        \quad Loop~B best ckpt (rejected) & 0.988 & 0.995 & 0.997 & 15.07 & 7.29 & 16.02 & 9.19 & 12.42 & 0.021 & 2.32 \\
        \quad $\theta_3$ & 0.987 & 0.995 & 0.997 & 15.15 & 7.28 & 16.11 & 9.12 & 12.34 & 0.021 & 2.36 \\
        \rowcolor{HandRSILite}
        \quad $\mathcal S_3$ & 0.990 & \textbf{1.000} & 0.997 & 13.91 & 6.48 & 15.20 & 7.72 & \textbf{11.59} & 0.019 & 1.53 \\
        \quad $\mathcal C^2_{k}$ & 0.991 & 0.999 & \textbf{0.998} & 14.26 & 6.50 & 15.54 & 7.84 & 11.76 & 0.021 & 1.61 \\
        \multicolumn{11}{l}{\textit{Round 4}} \\
        \quad $\mathcal C^1_{k}$ & 0.989 & 0.996 & 0.997 & 15.17 & 7.26 & 16.18 & 9.01 & 12.42 & 0.021 & 2.47 \\
        \quad Loop~B best ckpt (rejected) & 0.989 & 0.996 & 0.997 & 15.14 & 7.28 & 16.18 & 9.10 & 12.37 & 0.021 & 2.32 \\
        \quad $\theta_4$ & 0.987 & 0.995 & 0.997 & 15.15 & 7.28 & 16.11 & 9.12 & 12.34 & 0.021 & 2.36 \\
        \rowcolor{HandRSILite}
        \quad $\mathcal S_4$ & 0.989 & \textbf{1.000} & 0.997 & \textbf{13.84} & 6.37 & \textbf{15.17} & \textbf{7.61} & 11.61 & 0.019 & 1.47 \\
        \quad $\mathcal C^2_{k}$ & 0.991 & 0.999 & \textbf{0.998} & 14.19 & 6.37 & 15.50 & 7.66 & 11.77 & 0.021 & 1.55 \\
        \multicolumn{11}{l}{\textit{Round 5}} \\
        \quad $\mathcal C^1_{k}$ & 0.988 & 0.996 & 0.997 & 15.25 & 7.27 & 16.23 & 9.05 & 12.40 & 0.021 & 2.48 \\
        \quad Loop~B best ckpt (rejected) & 0.987 & 0.995 & 0.996 & 15.23 & 7.29 & 16.17 & 9.74 & 12.43 & 0.021 & 2.33 \\
        \quad $\theta_5$ & 0.987 & 0.995 & 0.997 & 15.15 & 7.28 & 16.11 & 9.12 & 12.34 & 0.021 & 2.36 \\
        \rowcolor{HandRSILite}
        \quad $\mathcal S_5$ & 0.990 & \textbf{1.000} & 0.997 & 13.85 & \textbf{6.34} & 15.18 & \textbf{7.61} & 11.61 & 0.019 & \textbf{1.45} \\
        \quad $\mathcal C^2_{k}$ & \textbf{0.992} & 0.999 & \textbf{0.998} & 14.20 & 6.35 & 15.50 & 7.65 & 11.77 & 0.021 & 1.52 \\
 
        \bottomrule
    \end{tabular}
\end{table*}
\begin{table*}[p]
    \centering
    \caption{\textbf{HOI4D (zero-shot), every stage of every round} (as Tab.~\ref{tab:stages_hot3d}).}
    \label{tab:stages_hoi4d}
    \scriptsize
    \setlength{\tabcolsep}{4pt}
    \renewcommand{\arraystretch}{0.92}
    \setlength{\aboverulesep}{0pt}
    \setlength{\belowrulesep}{0pt}
    \begin{tabular}{@{}l*{10}{c}@{}}
        \toprule
        \rowcolor{HandRSIHeader}
        \textbf{System} & FAcc$\uparrow$ & Recall$\uparrow$ & F1$\uparrow$ & MPJPE-p$\downarrow$ & PA-p$\downarrow$ & MPJPE$^{+\mathrm{OOS}}{\downarrow}$
        & EPE$_{\mathrm{2D}}$-p$\downarrow$ & GO-p$\downarrow$ & CT-p$\downarrow$ & Jitter$\downarrow$ \\
        \midrule
        \rowcolor{HandRSIInit}
        $\theta_0$ & \textbf{0.994} & \textbf{0.994} & \textbf{0.994} & 23.45 & 11.02 & 22.93 & \textbf{15.60} & 19.39 & \textbf{0.058} & 2.10 \\
        \multicolumn{11}{l}{\textit{Round 1}} \\
        \quad $\mathcal C^1_{k}$ & \textbf{0.994} & \textbf{0.994} & \textbf{0.994} & 23.83 & 11.01 & 23.29 & 15.71 & 19.35 & 0.062 & 2.07 \\
        \quad $\theta_1$ & 0.993 & \textbf{0.994} & \textbf{0.994} & 23.71 & 11.08 & 23.13 & 16.18 & 19.43 & 0.062 & 1.98 \\
        \rowcolor{HandRSILite}
        \quad $\mathcal S_1$ & 0.990 & 0.990 & 0.990 & 23.51 & 10.36 & 22.54 & 20.02 & 19.98 & 0.064 & 1.26 \\
        \quad $\mathcal C^2_{k}$ & 0.991 & 0.991 & 0.992 & 23.22 & 10.25 & 22.39 & 18.28 & 19.80 & 0.060 & 1.31 \\
        \multicolumn{11}{l}{\textit{Round 2}} \\
        \quad $\mathcal C^1_{k}$ & 0.993 & 0.993 & \textbf{0.994} & 23.90 & 11.12 & 23.29 & 16.48 & 19.44 & 0.062 & 2.02 \\
        \quad $\theta_2$ & 0.992 & 0.992 & 0.993 & 24.11 & 11.13 & 23.42 & 17.39 & 19.57 & 0.064 & 1.87 \\
        \rowcolor{HandRSILite}
        \quad $\mathcal S_2$ & 0.990 & 0.990 & 0.991 & 23.19 & 10.43 & 22.27 & 18.87 & 19.65 & 0.065 & 1.14 \\
        \quad $\mathcal C^2_{k}$ & 0.992 & 0.992 & 0.993 & 22.57 & 10.23 & 21.85 & 16.64 & 19.41 & 0.059 & 1.20 \\
        \multicolumn{11}{l}{\textit{Round 3}} \\
        \quad $\mathcal C^1_{k}$ & 0.993 & \textbf{0.994} & \textbf{0.994} & 23.94 & 11.20 & 23.37 & 16.11 & 19.43 & 0.063 & 1.97 \\
        \quad Loop~B best ckpt (rejected) & 0.993 & 0.993 & 0.993 & 23.98 & 11.19 & 23.36 & 16.82 & 19.47 & 0.064 & 1.84 \\
        \quad $\theta_3$ & 0.992 & 0.992 & 0.993 & 24.11 & 11.13 & 23.42 & 17.39 & 19.57 & 0.064 & 1.87 \\
        \rowcolor{HandRSILite}
        \quad $\mathcal S_3$ & 0.991 & 0.991 & 0.992 & 23.07 & 10.34 & 22.23 & 18.60 & 19.44 & 0.064 & 1.06 \\
        \quad $\mathcal C^2_{k}$ & 0.993 & 0.993 & \textbf{0.994} & 22.44 & 10.14 & 21.80 & 16.19 & \textbf{19.21} & \textbf{0.058} & 1.11 \\
        \multicolumn{11}{l}{\textit{Round 4}} \\
        \quad $\mathcal C^1_{k}$ & 0.993 & 0.993 & 0.993 & 24.07 & 11.19 & 23.40 & 17.08 & 19.45 & 0.063 & 1.97 \\
        \quad Loop~B best ckpt (rejected) & 0.993 & 0.993 & \textbf{0.994} & 23.92 & 11.22 & 23.31 & 16.73 & 19.72 & 0.064 & 1.83 \\
        \quad $\theta_4$ & 0.992 & 0.992 & 0.993 & 24.11 & 11.13 & 23.42 & 17.39 & 19.57 & 0.064 & 1.87 \\
        \rowcolor{HandRSILite}
        \quad $\mathcal S_4$ & 0.991 & 0.991 & 0.992 & 22.76 & 10.21 & 21.92 & 18.53 & 19.42 & 0.064 & 1.02 \\
        \quad $\mathcal C^2_{k}$ & 0.993 & 0.993 & \textbf{0.994} & \textbf{22.23} & 10.04 & \textbf{21.59} & 16.11 & \textbf{19.21} & \textbf{0.058} & 1.06 \\
        \multicolumn{11}{l}{\textit{Round 5}} \\
        \quad $\mathcal C^1_{k}$ & 0.993 & 0.993 & 0.993 & 24.05 & 11.17 & 23.39 & 16.87 & 19.45 & 0.063 & 1.97 \\
        \quad Loop~B best ckpt (rejected) & 0.990 & 0.990 & 0.991 & 24.18 & 11.28 & 23.30 & 19.51 & 19.93 & 0.067 & 1.83 \\
        \quad $\theta_5$ & 0.992 & 0.992 & 0.993 & 24.11 & 11.13 & 23.42 & 17.39 & 19.57 & 0.064 & 1.87 \\
        \rowcolor{HandRSILite}
        \quad $\mathcal S_5$ & 0.991 & 0.991 & 0.992 & 22.77 & 10.21 & 21.93 & 18.61 & 19.44 & 0.064 & \textbf{1.01} \\
        \quad $\mathcal C^2_{k}$ & 0.993 & 0.993 & \textbf{0.994} & 22.24 & \textbf{10.03} & \textbf{21.59} & 16.13 & 19.22 & \textbf{0.058} & 1.05 \\
 
        \bottomrule
    \end{tabular}
\end{table*}
\begin{table*}[p]
    \centering
    \caption{\textbf{Zero-shot HOI4D test}, every re-run method (units as in Tab.~\ref{tab:main}).}
    \label{tab:hoi4d}
    \footnotesize
    \setlength{\tabcolsep}{3.2pt}
    \setlength{\aboverulesep}{0pt}
    \setlength{\belowrulesep}{0pt}
    \begin{tabular}{@{}l*{10}{c}@{}}
        \toprule
        \rowcolor{HandRSIHeader}
        \textbf{Method} & FAcc$\uparrow$ & Recall$\uparrow$ & F1$\uparrow$
        & MPJPE-p$\downarrow$ & PA-p$\downarrow$ & MPJPE$^{+\mathrm{OOS}}{\downarrow}$
        & EPE$_{\mathrm{2D}}$-p$\downarrow$ & GO-p$\downarrow$ & CT-p$\downarrow$ & Jitter$\downarrow$ \\
        \midrule
        \multicolumn{11}{l}{\textit{HOI4D (zero-shot, 498 segments)}} \\
        HaMeR~\cite{hamer} & 0.930 & 0.926 & 0.961 & 45.10 & 11.66 & 45.12 & 75.83 & 34.14 & 0.170 & 20.95 \\
        Hamba~\cite{hamba} & 0.920 & 0.913 & 0.954 & 46.25 & 11.24 & 46.22 & 90.45 & 35.73 & 0.167 & 18.77 \\
        WiLoR~\cite{wilor} & 0.978 & 0.976 & 0.988 & 34.10 & 10.37 & 34.11 & 30.59 & 25.36 & 0.105 & 16.44 \\
        WildHands~\cite{wildhands} & 0.930 & 0.925 & 0.961 & 44.16 & 13.96 & 44.17 & 81.98 & 46.01 & 0.152 & 18.57 \\
        InterWild~\cite{interwild} & 0.929 & 0.925 & 0.960 & 53.83 & 12.98 & 53.84 & 77.67 & 41.82 & 0.173 & 48.22 \\
        \cmidrule[0.2pt](lr){1-11}
        HaWoR~\cite{hawor} & 0.985 & 0.984 & 0.991 & 33.04 & 11.43 & 32.93 & 22.73 & 25.66 & 0.063 & 16.76 \\
        OmniHands~\cite{omnihands} & 0.991 & 0.991 & 0.992 & 41.41 & 10.69 & 40.88 & 19.77 & 29.44 & 0.086 & 32.36 \\
        EgoForce~\cite{egoforce} & 0.939 & 0.935 & 0.964 & 45.64 & 11.97 & 45.86 & 106.88 & 39.50 & 0.123 & 42.03 \\
        ACE-Ego-Hand, K-free~\cite{ace_ego_hand} & \textbf{0.996} & \textbf{0.996} & \textbf{0.997} & 23.14 & 11.14 & 22.82 & 17.42 & \textbf{18.33} & \textbf{0.050} & 2.22 \\
        ACE-Ego-Hand, K-given~\cite{ace_ego_hand} & 0.994 & 0.994 & 0.994 & 23.45 & 11.02 & 22.93 & \textbf{15.60} & 19.39 & 0.058 & 2.10 \\
        \cmidrule[0.2pt](lr){1-11}
        HandRSI network ($\theta_2$), 42.5 FPS & 0.992 & 0.992 & 0.993 & 24.11 & 11.13 & 23.42 & 17.39 & 19.57 & 0.064 & 1.87 \\
        \rowcolor{HandRSILite}
        HandRSI-Lite, 41.1 FPS & 0.991 & 0.991 & 0.992 & 24.03 & 11.13 & 23.18 & 19.86 & 19.68 & 0.065 & 1.14 \\
        \rowcolor{HandRSIMedium}
        HandRSI-Medium, 34.0 FPS & 0.991 & 0.991 & 0.992 & 23.93 & 10.98 & 23.07 & 19.83 & 19.68 & 0.065 & 1.03 \\
        \rowcolor{HandRSIHeavy}
        HandRSI-Heavy, 20.7 FPS & 0.991 & 0.991 & 0.992 & 22.89 & 10.32 & 22.02 & 18.75 & 19.48 & 0.065 & 1.02 \\
        \rowcolor{HandRSIHeavy}
        \textbf{HandRSI} ($\mathcal S_5$), 16.2 FPS & 0.991 & 0.991 & 0.992 & \textbf{22.77} & \textbf{10.21} & \textbf{21.93} & 18.61 & 19.44 & 0.064 & \textbf{1.01} \\
 
        \bottomrule
    \end{tabular}
\end{table*}
\begin{table*}[p]
    \centering
    \caption{\textbf{HOT3D test under the strict detection rule} (predictions on unannotated hand slots count as
    false positives).}
    \label{tab:strict}
    \footnotesize
    \setlength{\tabcolsep}{3.5pt}
    \setlength{\aboverulesep}{0pt}
    \setlength{\belowrulesep}{0pt}
    \begin{tabular}{@{}l*{10}{c}@{}}
        \toprule
        \rowcolor{HandRSIHeader}
        \textbf{System} & FAcc$\uparrow$ & Recall$\uparrow$ & F1$\uparrow$ & MPJPE-p$\downarrow$ & PA-p$\downarrow$ & MPJPE$^{+\mathrm{OOS}}{\downarrow}$
        & EPE$_{\mathrm{2D}}$-p$\downarrow$ & GO-p$\downarrow$ & CT-p$\downarrow$ & Jitter$\downarrow$ \\
        \midrule
        ACE-Ego-Hand ($\theta_0$) & \textbf{0.959} & 0.986 & \textbf{0.987} & 12.97 & 5.90 & 12.84 & 9.66 & 6.90 & 0.024 & 2.82 \\
        $\theta_2$ alone & 0.930 & 0.994 & 0.974 & 9.37 & 4.71 & 10.35 & 5.31 & 5.80 & 0.013 & 2.51 \\
        HandRSI-Lite & 0.930 & \textbf{0.996} & 0.974 & 9.06 & 4.69 & 10.31 & 4.76 & 5.44 & \textbf{0.012} & 1.47 \\
        HandRSI-Medium & 0.930 & \textbf{0.996} & 0.974 & 9.04 & 4.64 & 10.30 & 4.79 & 5.46 & 0.013 & \textbf{1.33} \\
        HandRSI-Heavy & 0.932 & \textbf{0.996} & 0.974 & 8.41 & \textbf{4.35} & 9.71 & 4.16 & 5.17 & 0.013 & 1.35 \\
        $\mathcal S_5$ & 0.932 & \textbf{0.996} & 0.975 & \textbf{8.37} & 4.37 & \textbf{9.70} & \textbf{4.07} & \textbf{5.16} & 0.013 & 1.34 \\
 
        \bottomrule
    \end{tabular}
\end{table*}
\begin{figure}[t]
    \centering
\definecolor{CMotion}{HTML}{1BAF7A}
\begin{tikzpicture}
\begin{axis}[hr, width=0.84\linewidth, height=0.52\linewidth, xtick={2.9957,3.9120,4.6052,5.2983,6.2146,6.9078,7.6009,8.5172}, xticklabels={20,50,100,200,500,1k,2k,5k},
    xmin=2.7081, xmax=9.1050, ymin=-0.2, ymax=5.3, xlabel={summed module cost per 81-frame clip (ms, estimate)},
    ylabel={val $\Delta$ vs.\ network alone (\%)}, clip=false,
    legend style={at={(0.98,0.03)}, anchor=south east, fill=white, fill opacity=0.85, text opacity=1}]
\addplot[CGray!80, line width=0.7pt, forget plot] table[x=x, y=y]{fig/data/search_front.dat};
\addplot[only marks, mark=*, mark size=1.15pt, CGray, restrict expr to domain={\thisrow{g}}{0:0}] table[x=x, y=y]{fig/data/search_all.dat};
\addlegendentry{cheap modules only (104)}
\addplot[only marks, mark=*, mark size=1.15pt, CMotion, restrict expr to domain={\thisrow{g}}{1:1}] table[x=x, y=y]{fig/data/search_all.dat};
\addlegendentry{+ image-motion refinement (11)}
\addplot[only marks, mark=*, mark size=1.15pt, CHeavy, restrict expr to domain={\thisrow{g}}{2:2}] table[x=x, y=y]{fig/data/search_all.dat};
\addlegendentry{+ WiLoR module (9)}
\addplot[only marks, mark=o, mark size=2.1pt, black!80, line width=0.45pt, restrict expr to domain={\thisrow{timed}}{1:1}] table[x=x, y=y]{fig/data/search_all.dat};
\addlegendentry{timed (20)}
\addplot[only marks, mark=*, mark size=2.3pt, mark options={fill=CLite, draw=white, line width=0.5pt}, forget plot] coordinates {(4.4710,3.9049)};
\node[font=\scriptsize, text=black!75, anchor=south, inner sep=1pt, yshift=7pt] at (axis cs:4.4710,3.9049) {Lite};
\addplot[only marks, mark=*, mark size=2.3pt, mark options={fill=CMedium, draw=white, line width=0.5pt}, forget plot] coordinates {(6.1748,4.0622)};
\node[font=\scriptsize, text=black!75, anchor=north, inner sep=1pt, yshift=-7pt] at (axis cs:6.1748,4.0622) {Medium};
\addplot[only marks, mark=*, mark size=2.3pt, mark options={fill=CHeavy, draw=white, line width=0.5pt}, forget plot] coordinates {(8.9003,4.9260)};
\node[font=\scriptsize, text=black!75, anchor=south east, inner sep=1pt, xshift=-3pt, yshift=3pt] at (axis cs:8.9003,4.9260) {Heavy};
\addplot[only marks, mark=diamond*, mark size=2.6pt, mark options={fill=white, draw=black!85, line width=0.6pt}, forget plot] coordinates {(9.0328,4.8406)};
\node[font=\scriptsize, anchor=north, inner sep=1pt, yshift=-6pt] at (axis cs:9.0328,4.8406) {$\mathcal S_5$};
\end{axis}
\end{tikzpicture}
    \caption{\textbf{All \nSearched{} searched configurations} on val against the
    summed per-module cost (an estimate that overstates WiLoR about 2.5$\times$;
    Fig.~\ref{fig:pareto} uses measured time).}
    \label{fig:supp_search}
\end{figure}
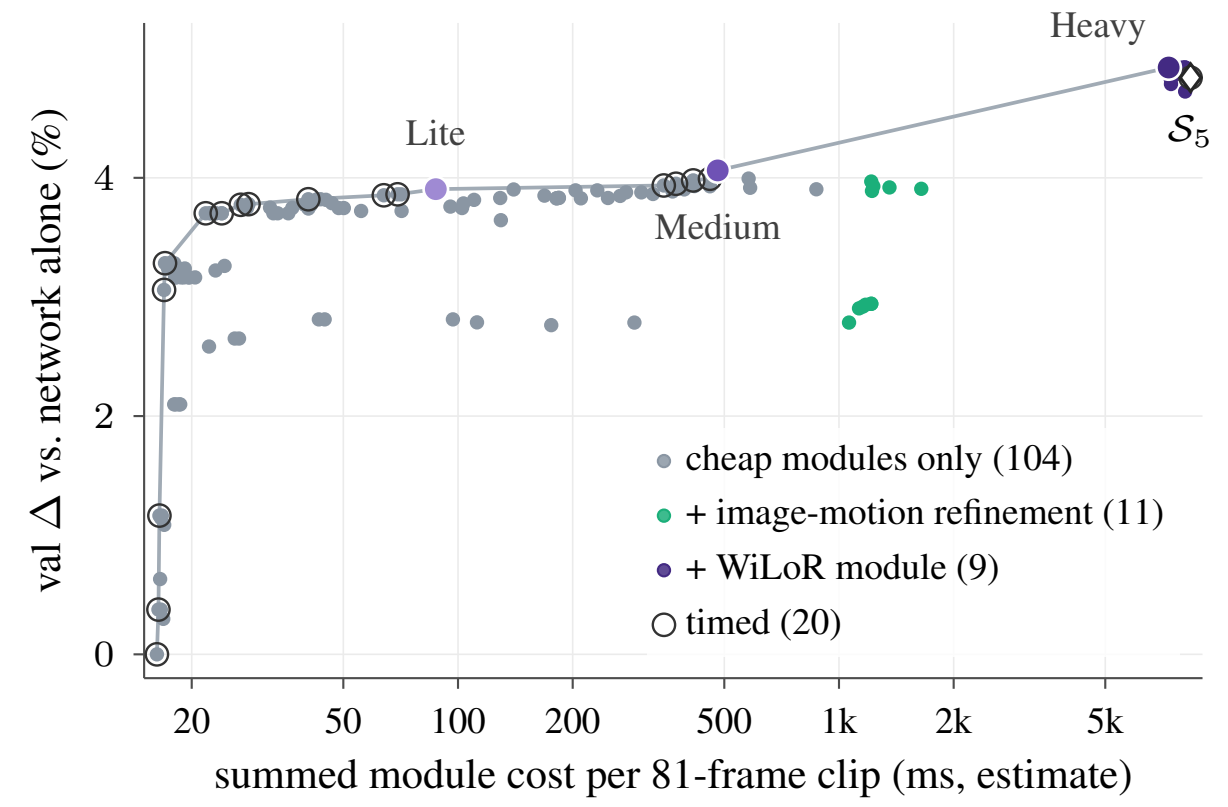

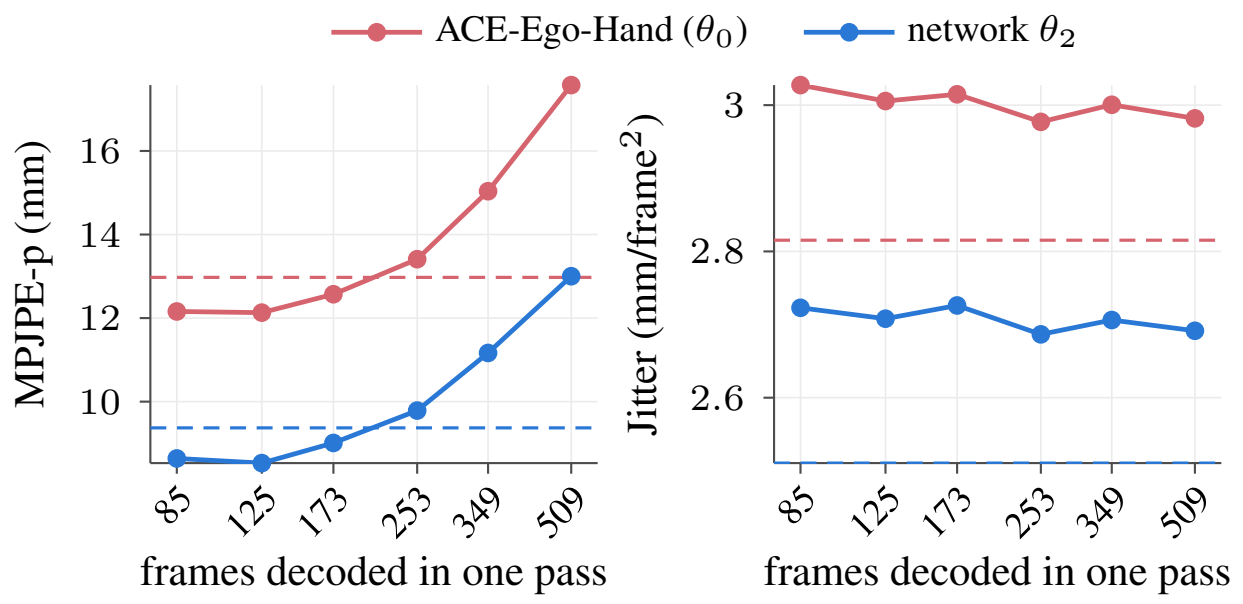
\begin{figure}[t]
    \centering
\begin{tikzpicture}
\begin{groupplot}[hr, group style={group size=2 by 1, horizontal sep=0.14\linewidth}, width=0.355\linewidth, height=0.30\linewidth,
    xmin=4.3227, xmax=6.3524, xtick={4.4427,4.8283,5.1533,5.5334,5.8551,6.2324}, xticklabels={85,125,173,253,349,509}, xticklabel style={rotate=45, anchor=north east, inner sep=1pt},
    xlabel={frames decoded in one pass},
    legend style={at={(1.25,1.05)}, anchor=south, legend columns=2, /tikz/every even column/.append style={column sep=8pt}}]
\nextgroupplot[ylabel={MPJPE-p (mm)}]
\addplot[CAce, line width=0.9pt, mark=*, mark size=1.3pt] table[x=x, y=m0]{fig/data/seqlen.dat};
\addplot[HandRSIBlue, line width=0.9pt, mark=*, mark size=1.3pt] table[x=x, y=m2]{fig/data/seqlen.dat};
\addplot[CAce, densely dashed, line width=0.6pt, forget plot] coordinates {(4.3227,12.9743) (6.3524,12.9743)};
\addplot[HandRSIBlue, densely dashed, line width=0.6pt, forget plot] coordinates {(4.3227,9.3721) (6.3524,9.3721)};
\addlegendentry{ACE-Ego-Hand ($\theta_0$)}
\addlegendentry{network $\theta_2$}
\nextgroupplot[ylabel={Jitter (mm/frame$^2$)}]
\addplot[CAce, line width=0.9pt, mark=*, mark size=1.3pt] table[x=x, y=j0]{fig/data/seqlen.dat};
\addplot[HandRSIBlue, line width=0.9pt, mark=*, mark size=1.3pt] table[x=x, y=j2]{fig/data/seqlen.dat};
\addplot[CAce, densely dashed, line width=0.6pt, forget plot] coordinates {(4.3227,2.8153) (6.3524,2.8153)};
\addplot[HandRSIBlue, densely dashed, line width=0.6pt, forget plot] coordinates {(4.3227,2.5106) (6.3524,2.5106)};
\end{groupplot}
\end{tikzpicture}
    \caption{\textbf{Decoded sequence length}: network alone on HOT3D test against the length decoded in one
    pass (dashed: 81-frame segments).}
    \label{fig:seqlen}
\end{figure}

\paragraph{Every stage of every round.}
Tables~\ref{tab:stages_hot3d}--\ref{tab:stages_hoi4d} give every stage of every round on the three test sets:
the GT-only control $\mathcal C^1_k$, the teacher, the network $\theta_k$, the system $\mathcal S_k$, and the
round's harness on the original network ($\mathcal C^2_k$).
In the block of round~$k$, the teacher is $\mathcal T_{k-1}=(\mathcal C^1_k,\mathcal H^{I,\star}_{k-1})$ in the
notation of Alg.~\ref{alg:round}; in rounds 3--5 it equals $\mathcal S_k$ and is not repeated, and the
teachers of rounds 1--2 on HOI4D were not kept.
Table~\ref{tab:hoi4d} lists every re-run method on the zero-shot HOI4D test (Sec.~\ref{sec:exp_comparison}).

\paragraph{Decoded sequence length.}
Following ACE-Ego-Hand's Fig.~S2, Fig.~\ref{fig:seqlen} decodes HOT3D test with
whole-sequence context in tiles of 85 to 509 frames.
$\theta_2$ leads $\theta_0$ by 3.5--4.6\,mm at every length, and tiles of
85--125 frames lower $\theta_2$'s MPJPE-p from 9.37\,mm (81-frame segments) to
8.53\,mm.

\paragraph{Tier search.}
Figure~\ref{fig:supp_search} shows all \nSearched{} configurations searched for the deployment tiers
(Secs.~\ref{sec:tiers} and~\ref{sec:exp_tiers}) against their summed per-module cost: chains of cheap modules
alone reach $+$4.06\% of the full system's $+$4.84\% val gain.

\paragraph{Strict detection rule.}
Table~\ref{tab:strict} scores HOT3D under the strict detection rule (Sec.~\ref{sec:eval}), which counts every
prediction on a hand slot without GT as a false positive; only FAcc and F1 change.

\section{Extended Qualitative Results}
\label{sec:app_gallery}

\begin{figure*}[p]
    \centering
    {\footnotesize
\setlength{\qw}{\dimexpr(\linewidth-1em-12pt)/6\relax}%
\begin{tabular}{@{}c@{\hspace{2pt}}c@{\hspace{2pt}}c@{\hspace{2pt}}c@{\hspace{2pt}}c@{\hspace{2pt}}c@{\hspace{2pt}}c@{}}
 & Ground truth & \textbf{\color{HandRSIBlue}HandRSI (ours)} & ACE-Ego-Hand~\cite{ace_ego_hand} & EgoForce~\cite{egoforce} & HaWoR~\cite{hawor} & WiLoR~\cite{wilor} \\[1pt]
\qrow{HOT3D \#1} & \qtile{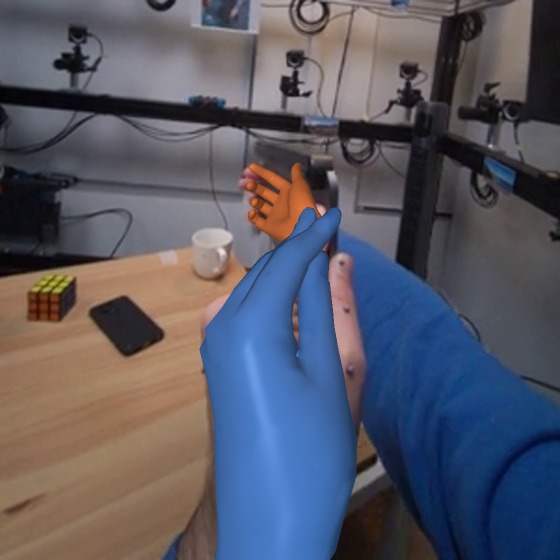}{}{QTag} & \qtile{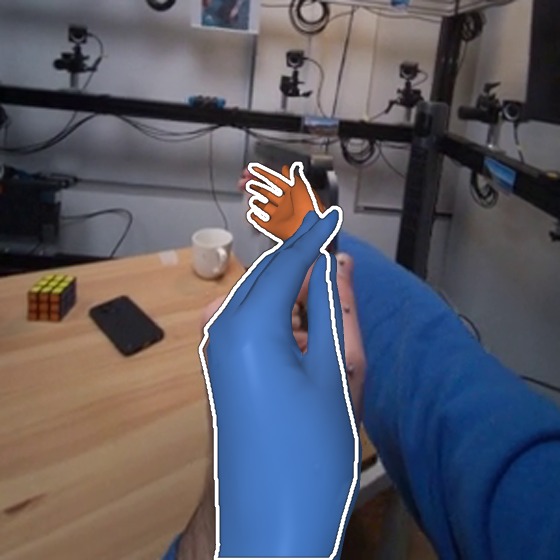}{3\,px}{HandRSIBlue} & \qtile{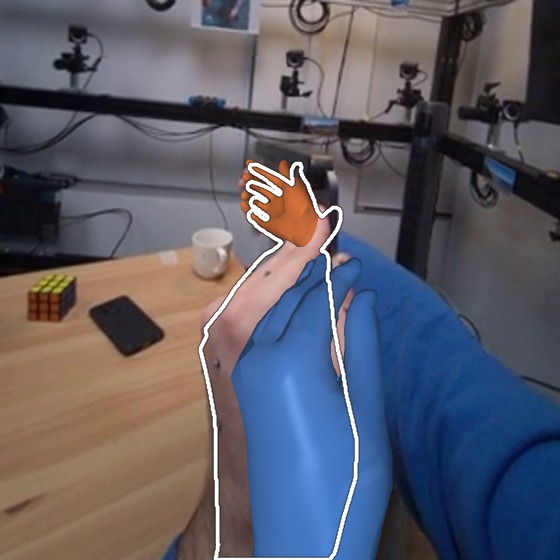}{13\,px}{QTag} & \qtile{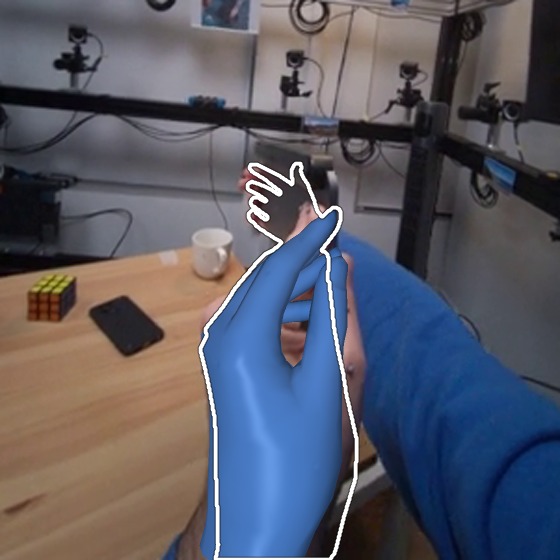}{missed}{QTag} & \qtile{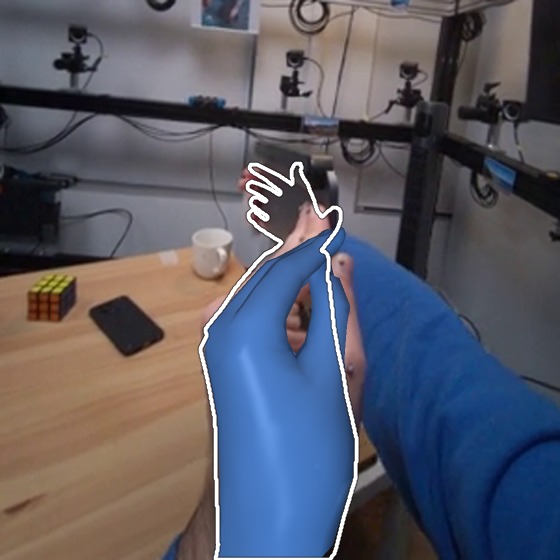}{missed}{QTag} & \qtile{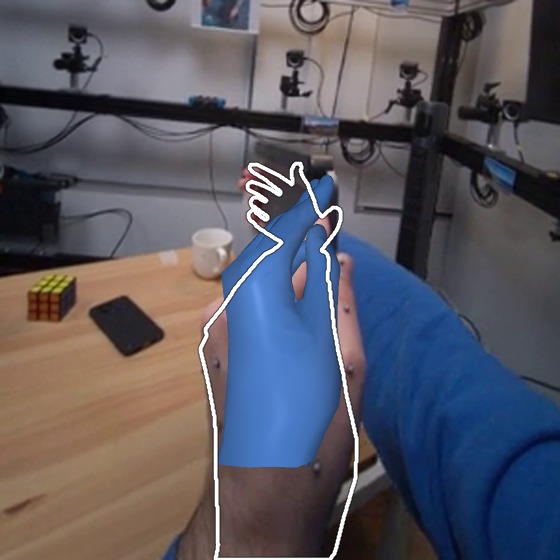}{missed}{QTag} \\[1pt]
\qrow{3D view} & \qtile{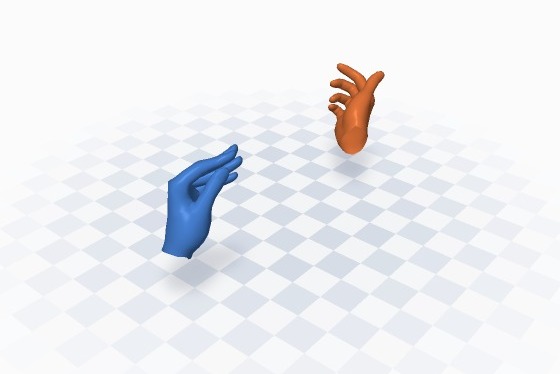}{}{QTag} & \qtile{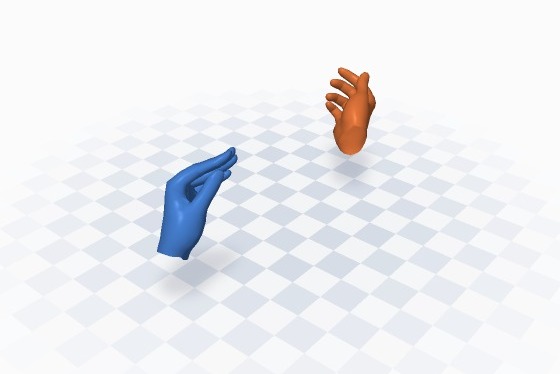}{}{QTag} & \qtile{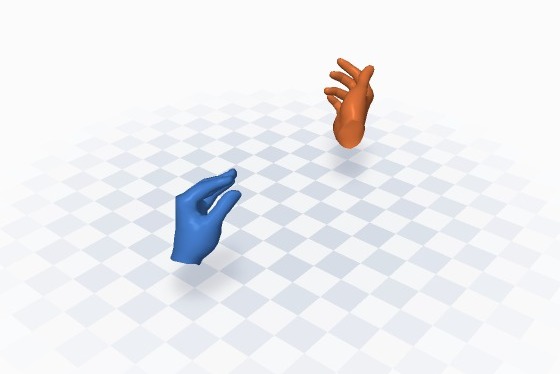}{}{QTag} & \qtile{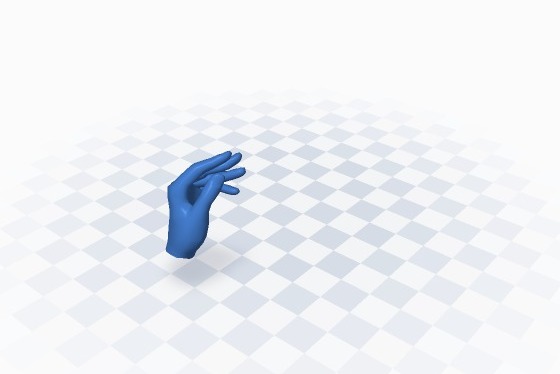}{}{QTag} & \qtile{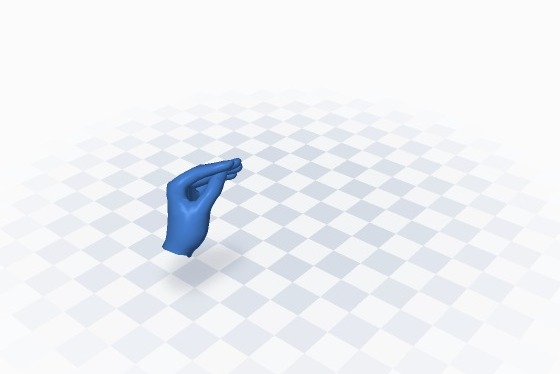}{}{QTag} & \qtile{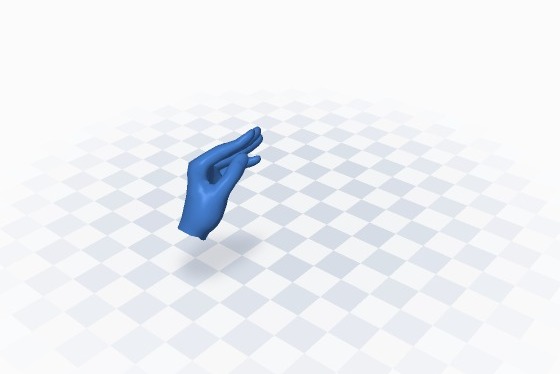}{}{QTag} \\[5pt]
\qrow{HOT3D \#2} & \qtile{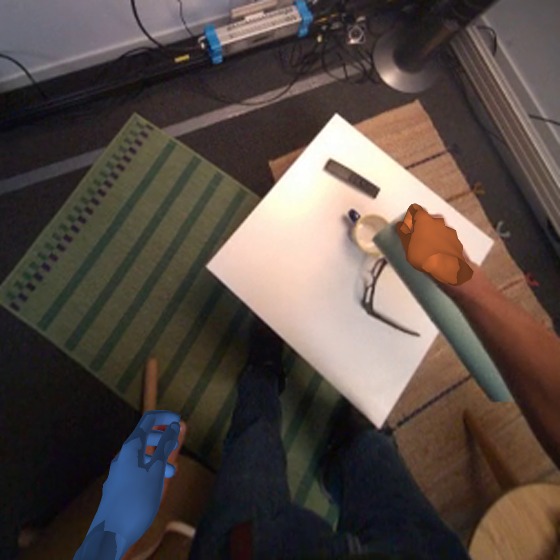}{}{QTag} & \qtile{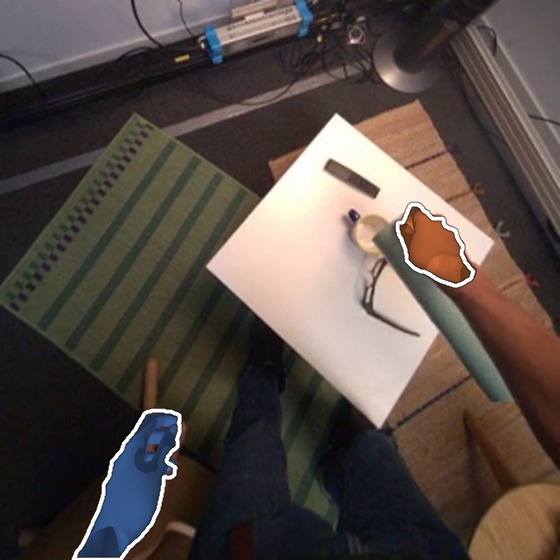}{2\,px}{HandRSIBlue} & \qtile{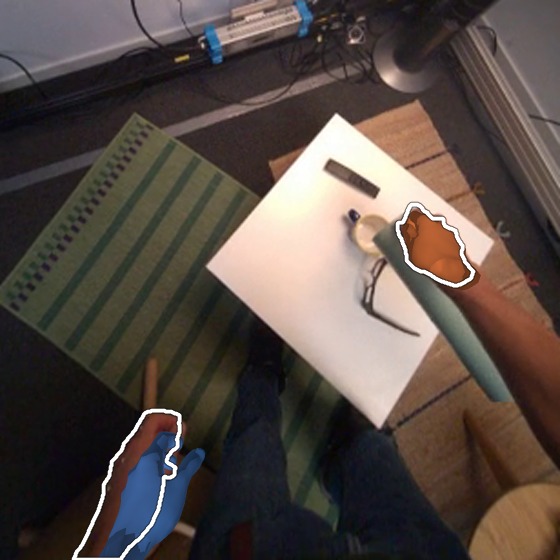}{8\,px}{QTag} & \qtile{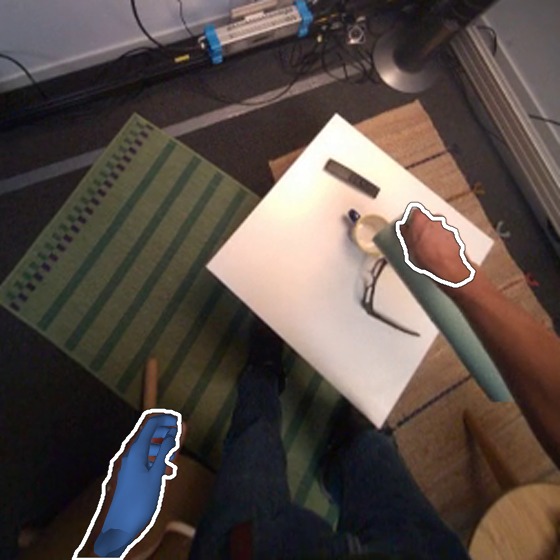}{missed}{QTag} & \qtile{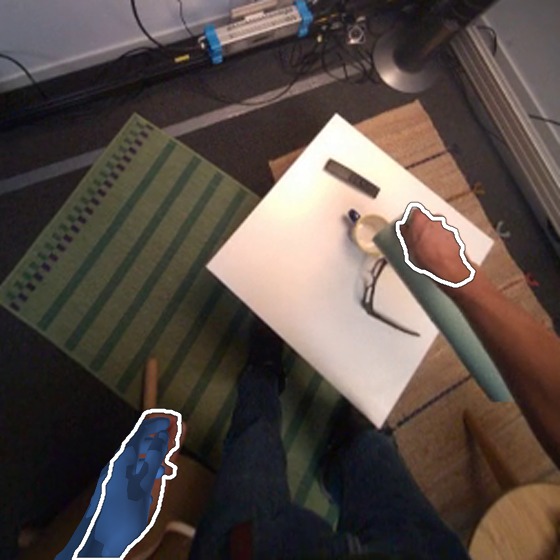}{missed}{QTag} & \qtile{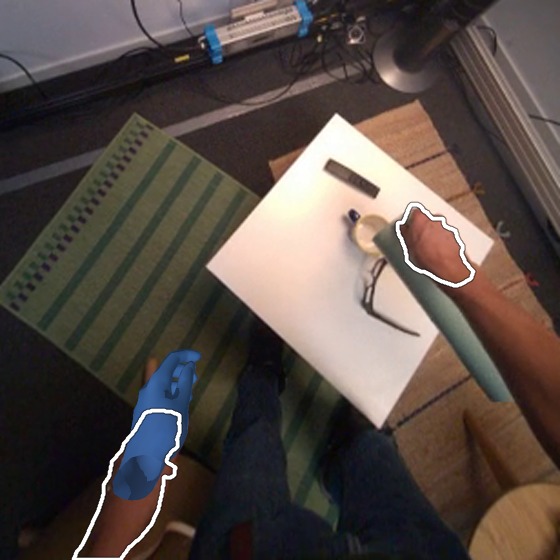}{missed}{QTag} \\[1pt]
\qrow{3D view} & \qtile{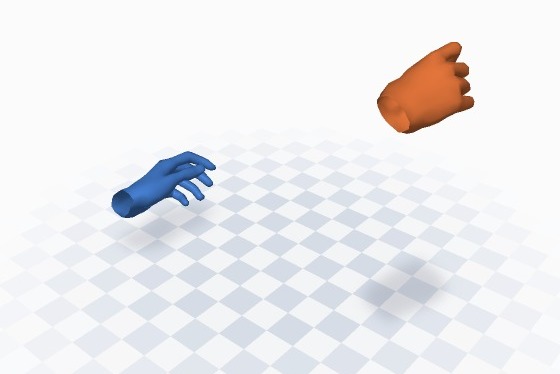}{}{QTag} & \qtile{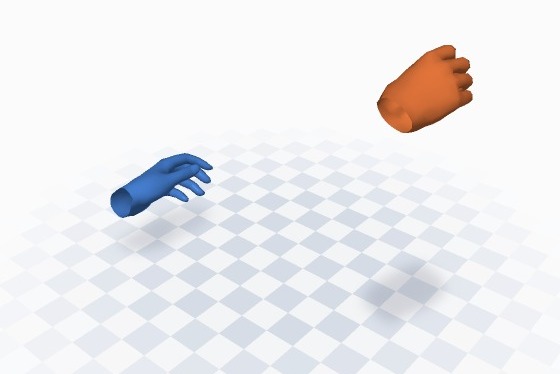}{}{QTag} & \qtile{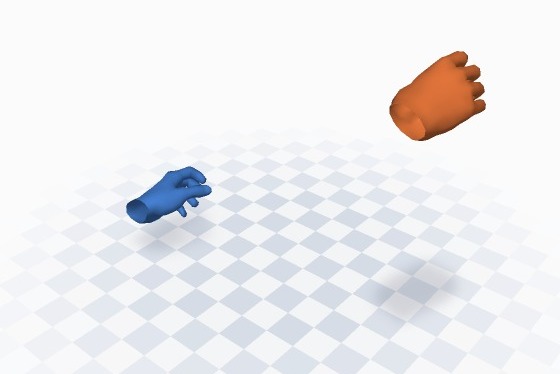}{}{QTag} & \qtile{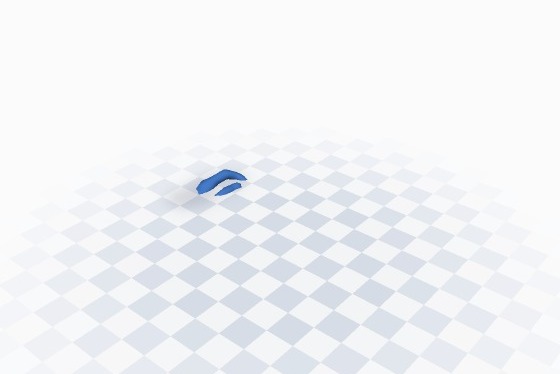}{}{QTag} & \qtile{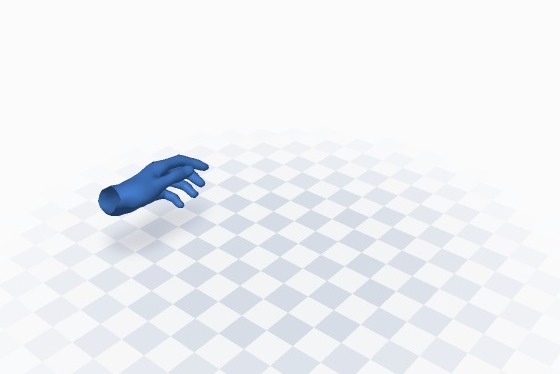}{}{QTag} & \qtile{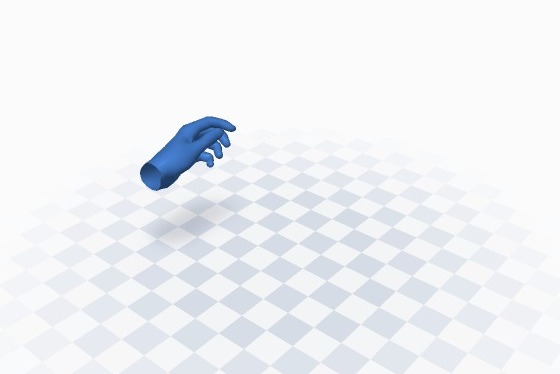}{}{QTag} \\[5pt]
\qrow{HOT3D \#3} & \qtile{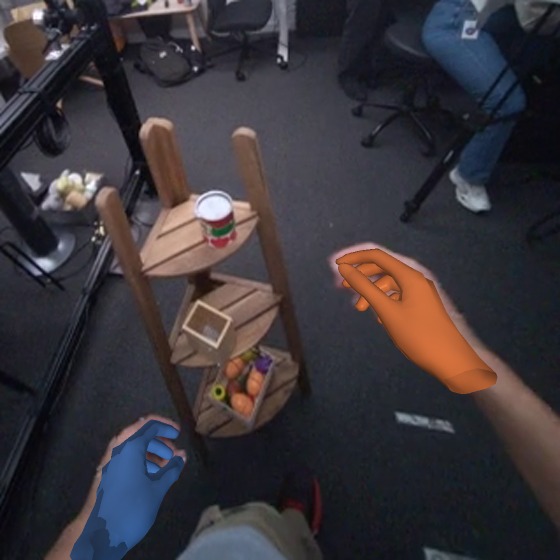}{}{QTag} & \qtile{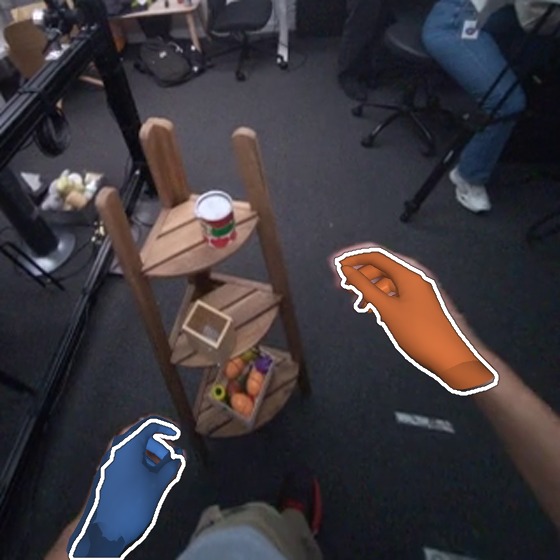}{3\,px}{HandRSIBlue} & \qtile{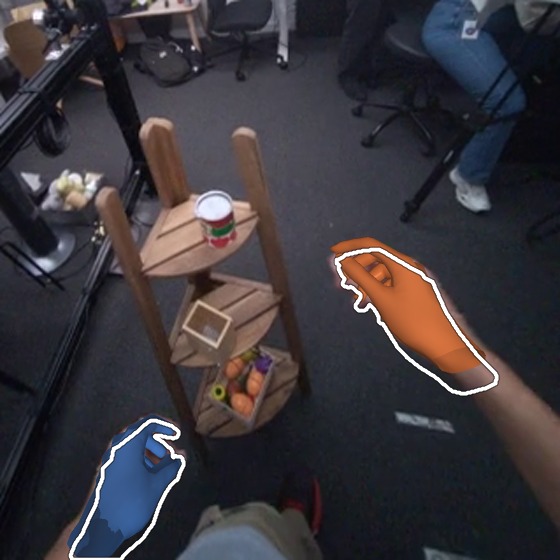}{6\,px}{QTag} & \qtile{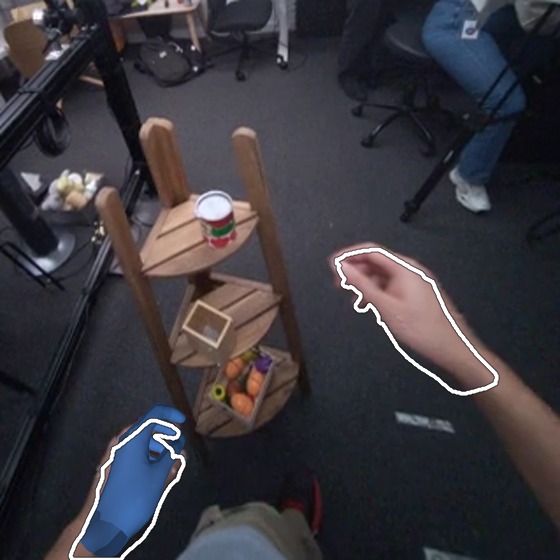}{$>$99\,px}{QTag} & \qtile{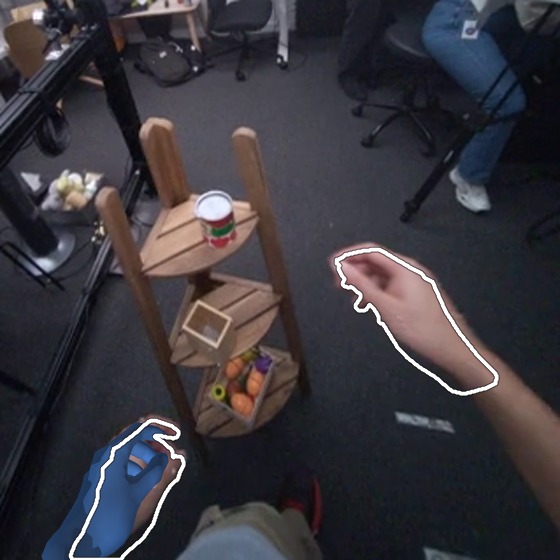}{$>$99\,px}{QTag} & \qtile{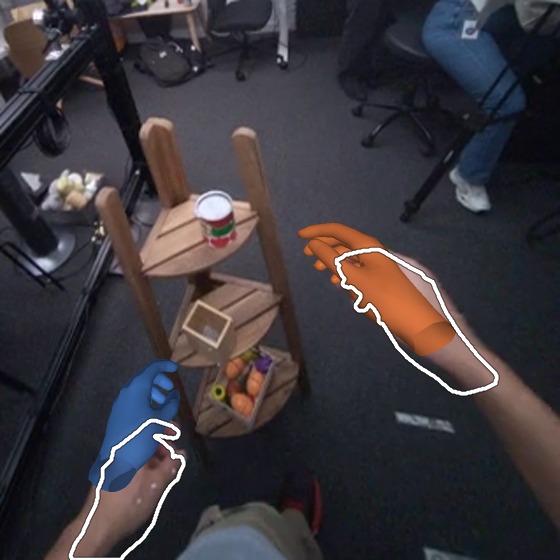}{34\,px}{QTag} \\[1pt]
\qrow{3D view} & \qtile{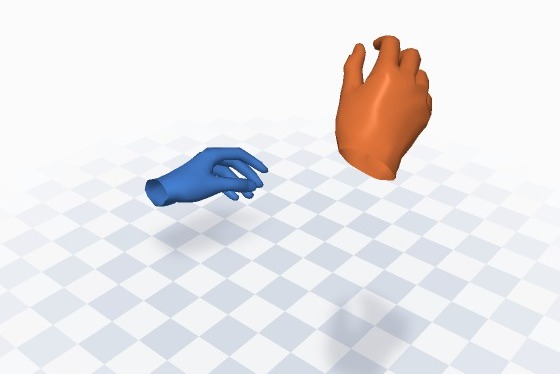}{}{QTag} & \qtile{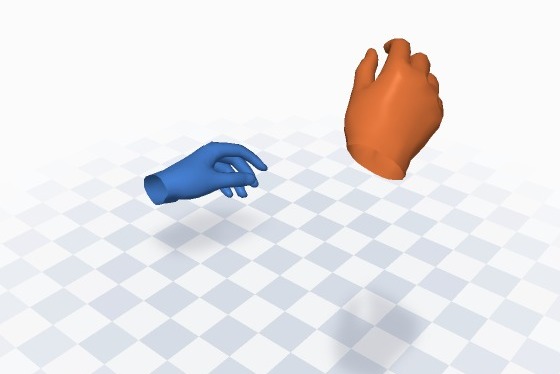}{}{QTag} & \qtile{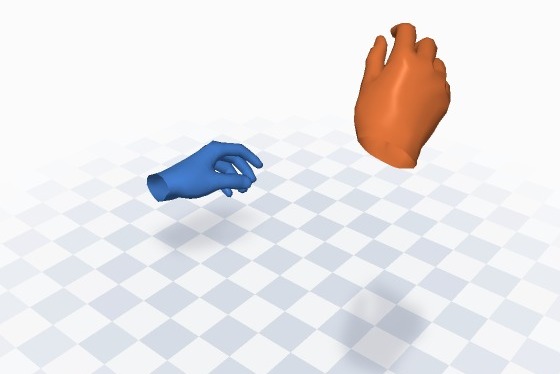}{}{QTag} & \qtile{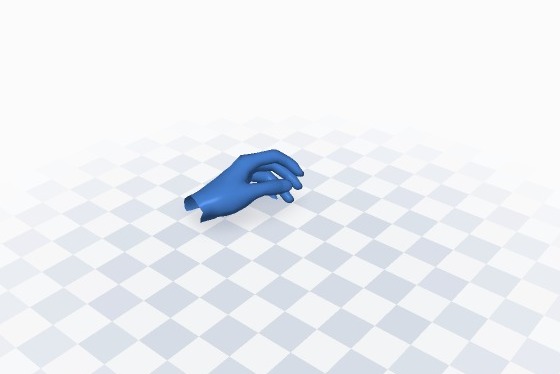}{}{QTag} & \qtile{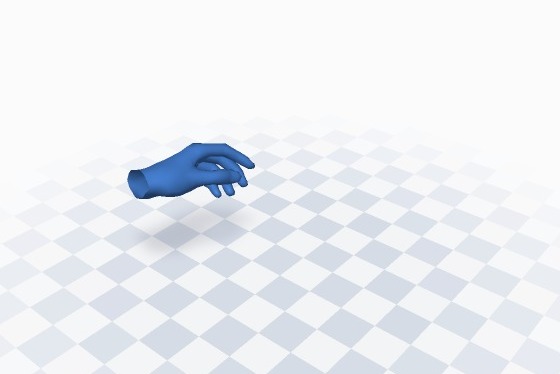}{}{QTag} & \qtile{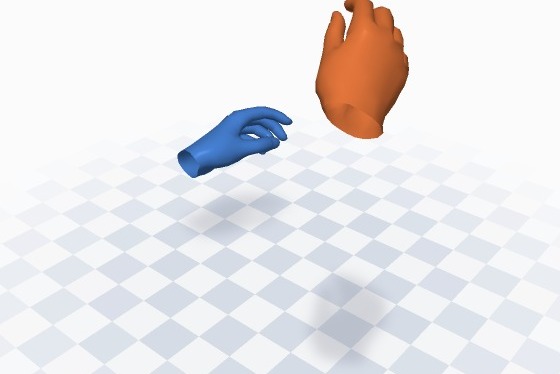}{}{QTag} \\[5pt]
\qrow{HOT3D \#4} & \qtile{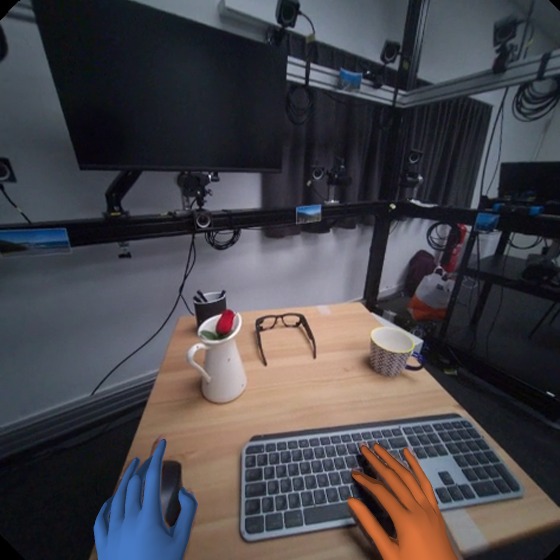}{}{QTag} & \qtile{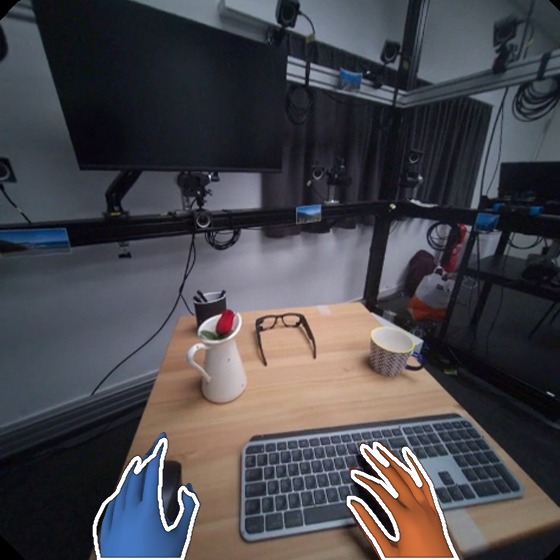}{4\,px}{HandRSIBlue} & \qtile{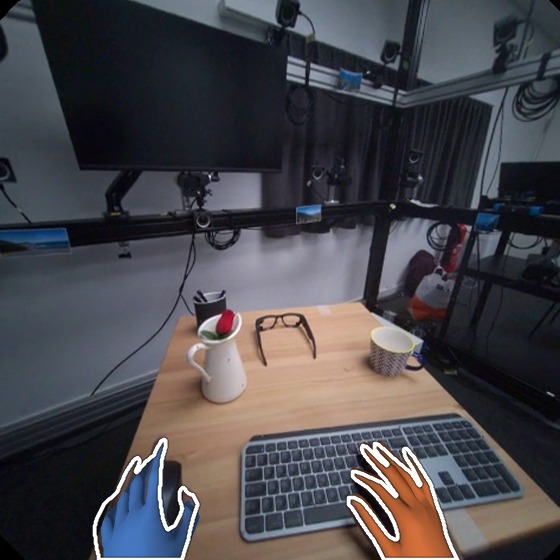}{4\,px}{QTag} & \qtile{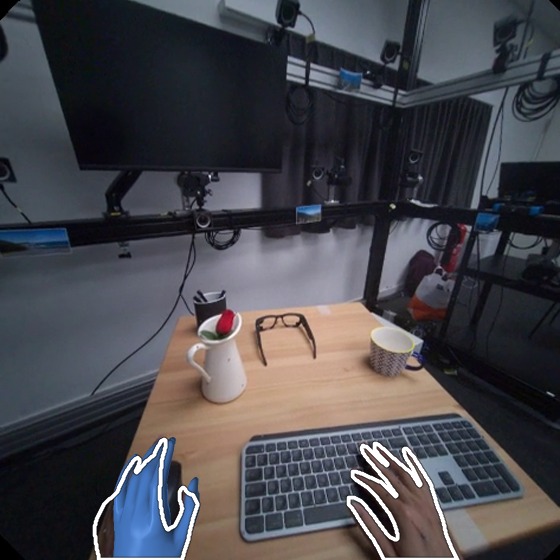}{$>$99\,px}{QTag} & \qtile{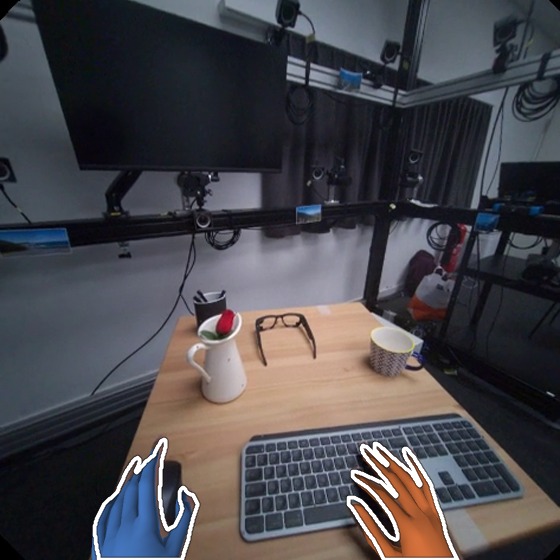}{7\,px}{QTag} & \qtile{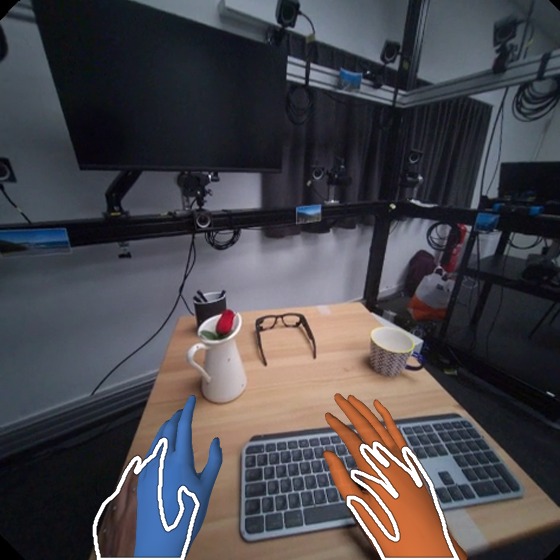}{51\,px}{QTag} \\[1pt]
\qrow{3D view} & \qtile{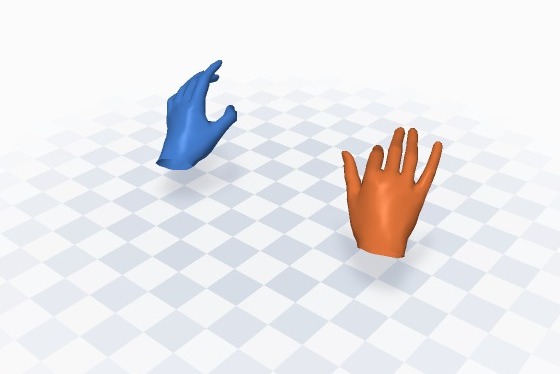}{}{QTag} & \qtile{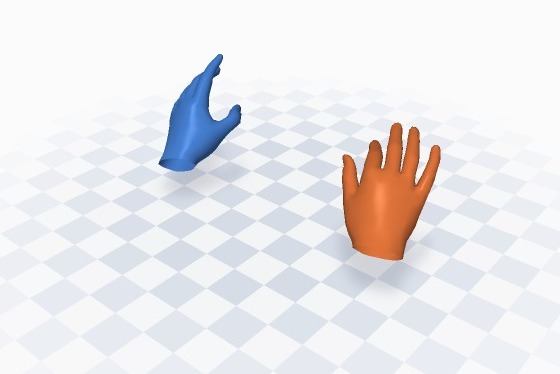}{}{QTag} & \qtile{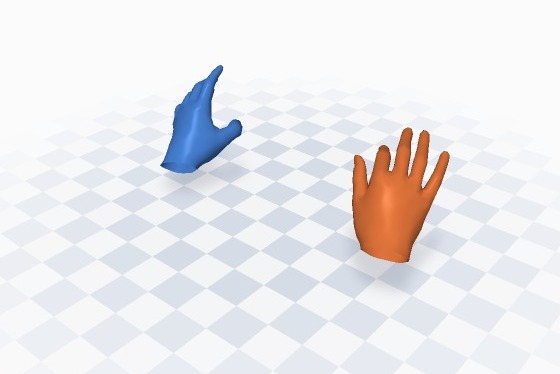}{}{QTag} & \qtile{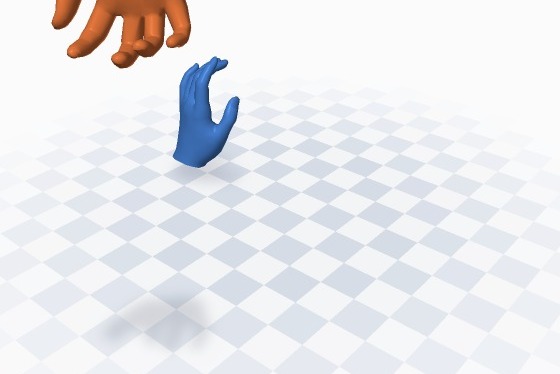}{}{QTag} & \qtile{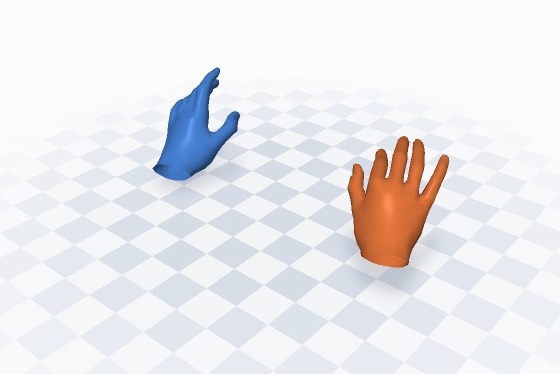}{}{QTag} & \qtile{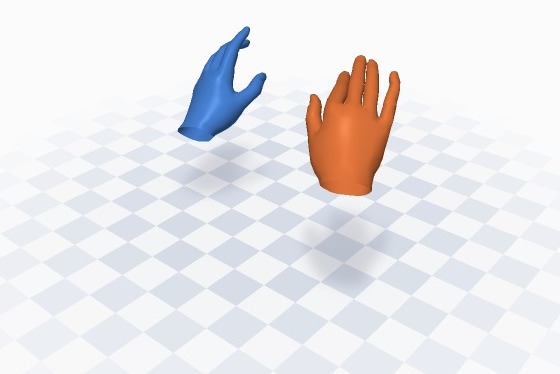}{}{QTag} \\
\end{tabular}
}
    \caption{
        \textbf{Extended comparison on the sealed HOT3D test.} Per example: the frame with each method's
        meshes (white outline: ground truth; tag: mean 2D joint error of the visible hands, ``missed'' if one
        of them is not output) and the same hands in 3D from one third-person view above a floor, identical
        in every column. Frames: the two with the largest gap between the baselines' and HandRSI's 2D error,
        then the two largest among frames where every method outputs every visible hand; one per recording.
    }
    \label{fig:app_gal_hot3d}
\end{figure*}
\begin{figure*}[p]
    \centering
    {\footnotesize
\setlength{\qw}{\dimexpr(\linewidth-1em-12pt)/6\relax}%
\begin{tabular}{@{}c@{\hspace{2pt}}c@{\hspace{2pt}}c@{\hspace{2pt}}c@{\hspace{2pt}}c@{\hspace{2pt}}c@{\hspace{2pt}}c@{}}
 & Ground truth & \textbf{\color{HandRSIBlue}HandRSI (ours)} & ACE-Ego-Hand~\cite{ace_ego_hand} & EgoForce~\cite{egoforce} & HaWoR~\cite{hawor} & WiLoR~\cite{wilor} \\[1pt]
\qrow{ARCTIC \#1} & \qtile{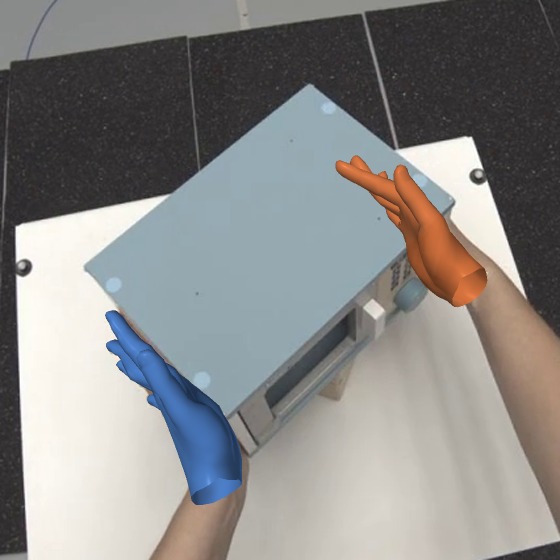}{}{QTag} & \qtile{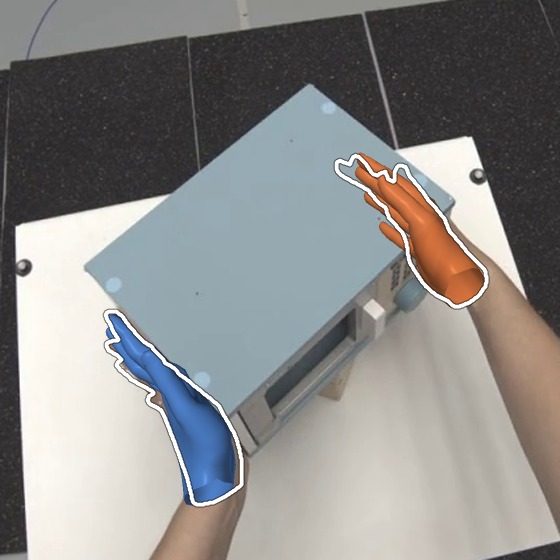}{8\,px}{HandRSIBlue} & \qtile{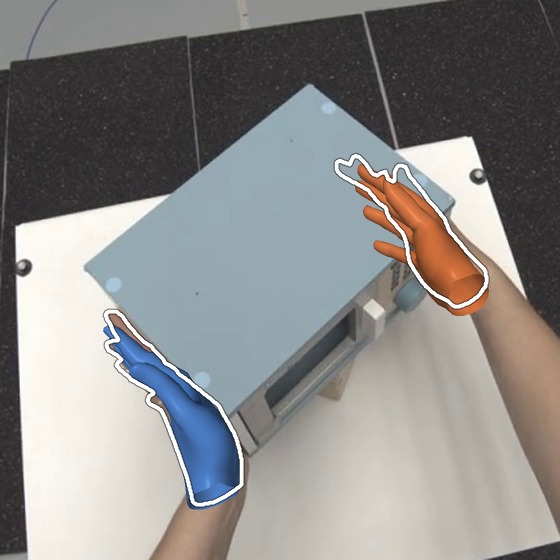}{11\,px}{QTag} & \qtile{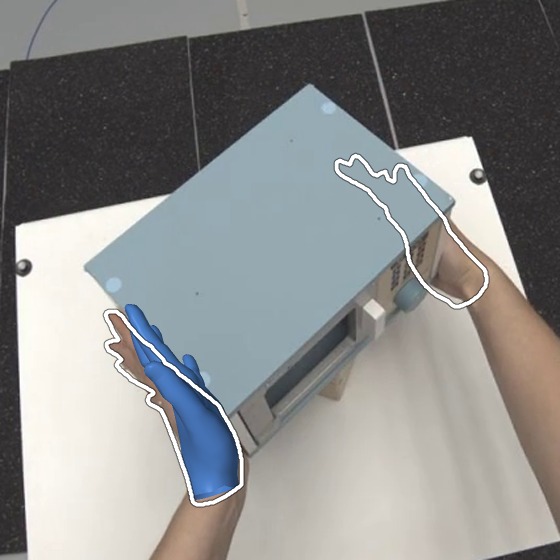}{missed}{QTag} & \qtile{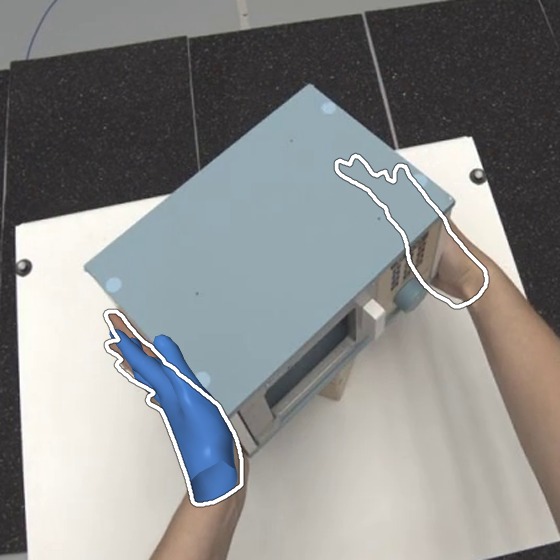}{missed}{QTag} & \qtile{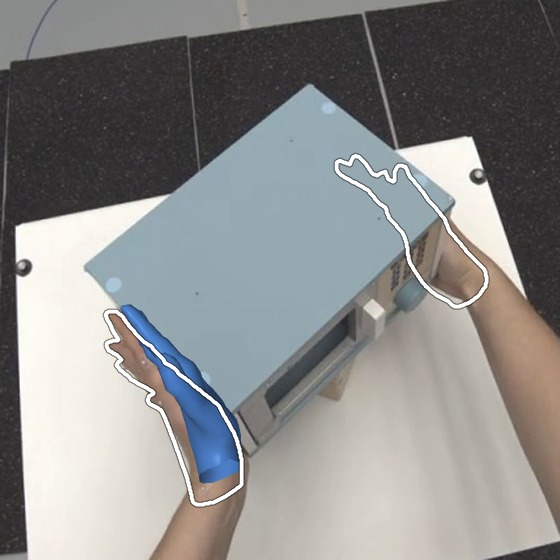}{missed}{QTag} \\[1pt]
\qrow{3D view} & \qtile{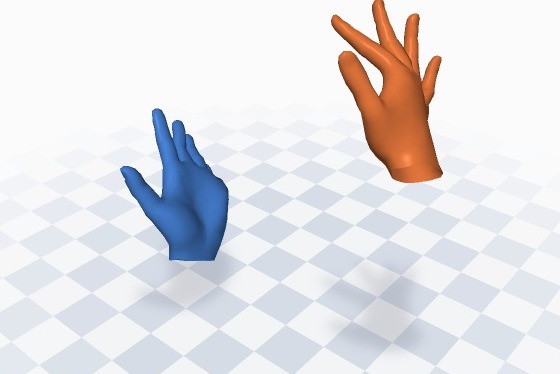}{}{QTag} & \qtile{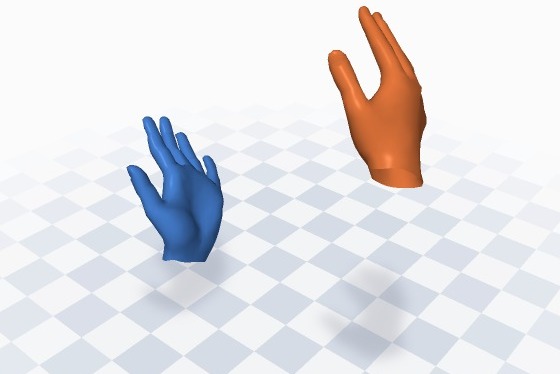}{}{QTag} & \qtile{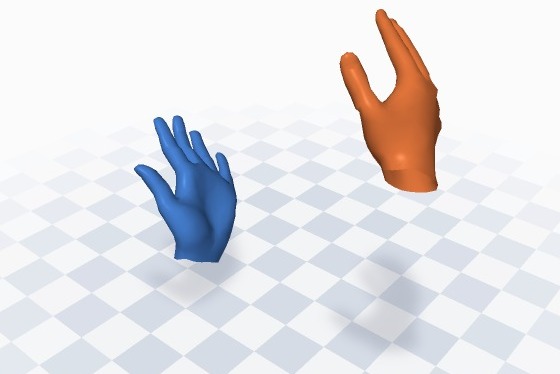}{}{QTag} & \qtile{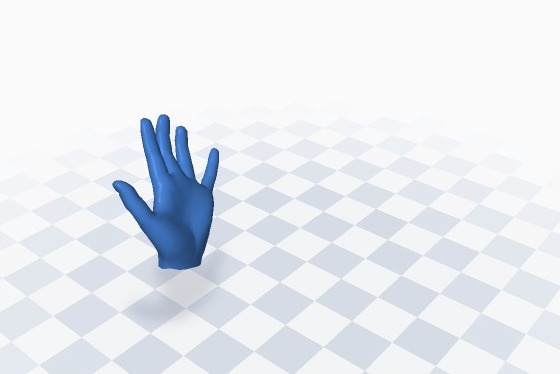}{}{QTag} & \qtile{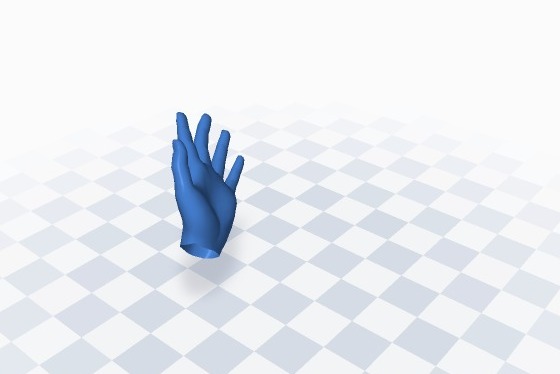}{}{QTag} & \qtile{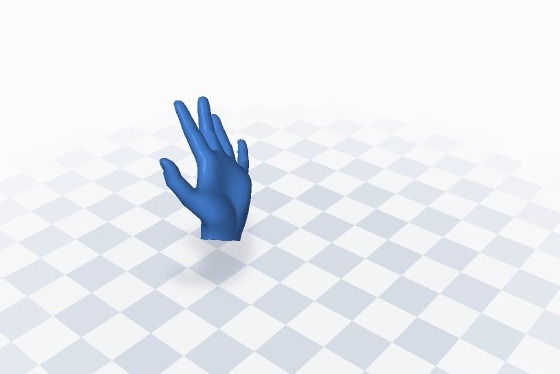}{}{QTag} \\[5pt]
\qrow{ARCTIC \#2} & \qtile{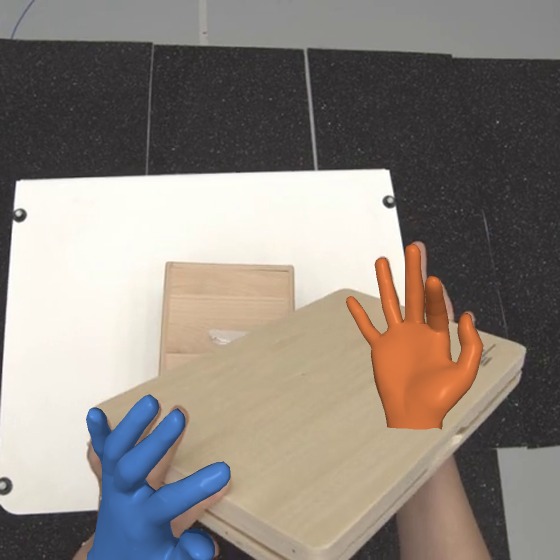}{}{QTag} & \qtile{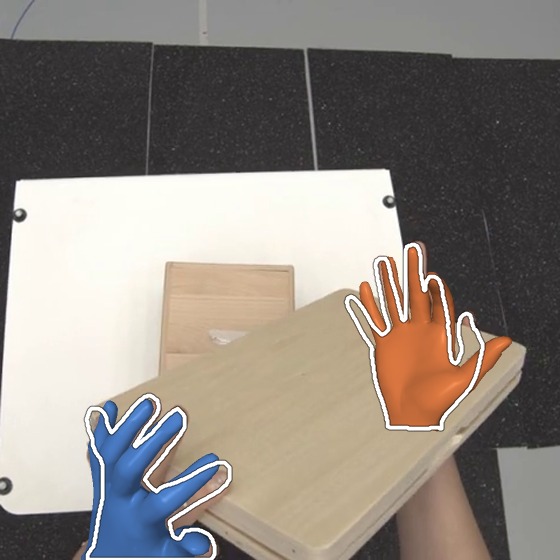}{8\,px}{HandRSIBlue} & \qtile{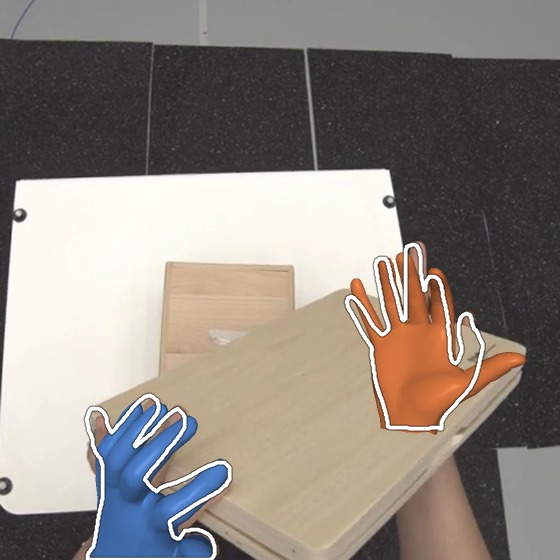}{10\,px}{QTag} & \qtile{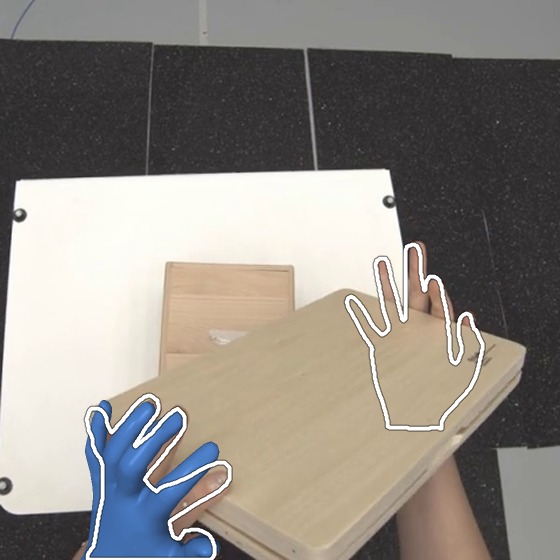}{missed}{QTag} & \qtile{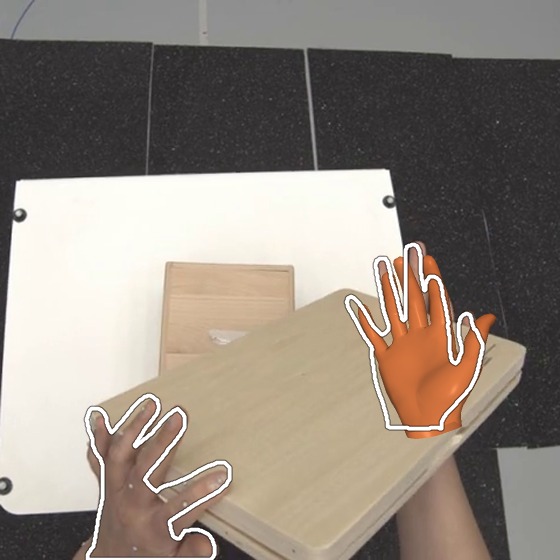}{missed}{QTag} & \qtile{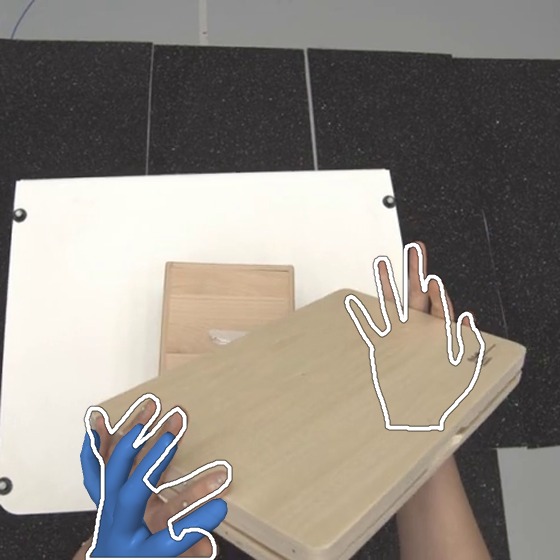}{missed}{QTag} \\[1pt]
\qrow{3D view} & \qtile{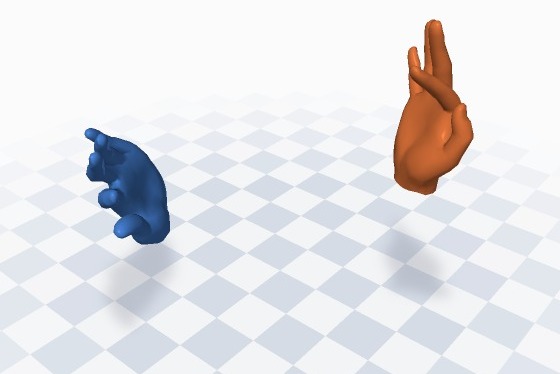}{}{QTag} & \qtile{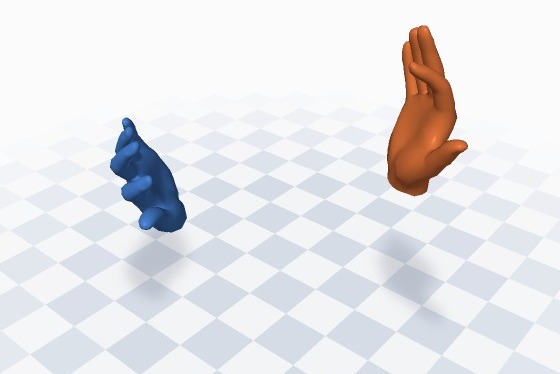}{}{QTag} & \qtile{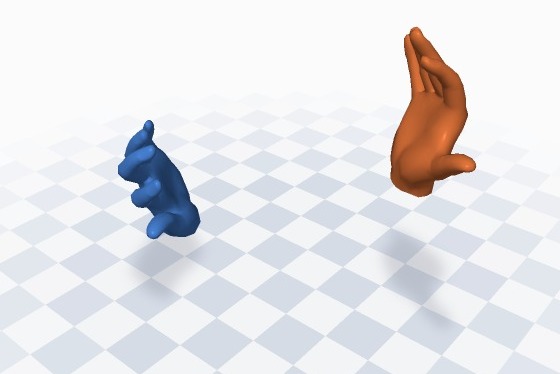}{}{QTag} & \qtile{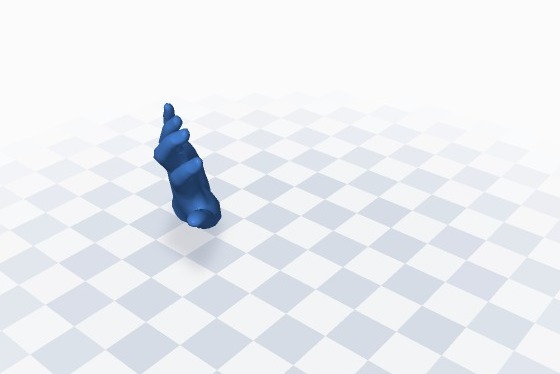}{}{QTag} & \qtile{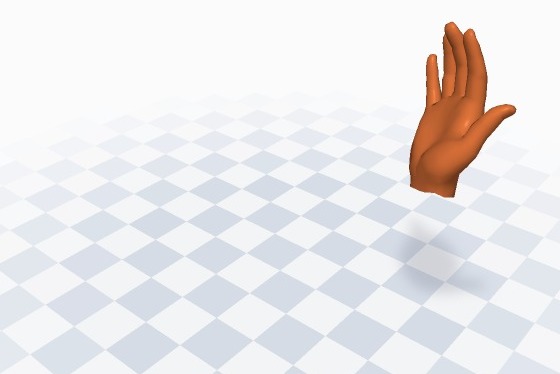}{}{QTag} & \qtile{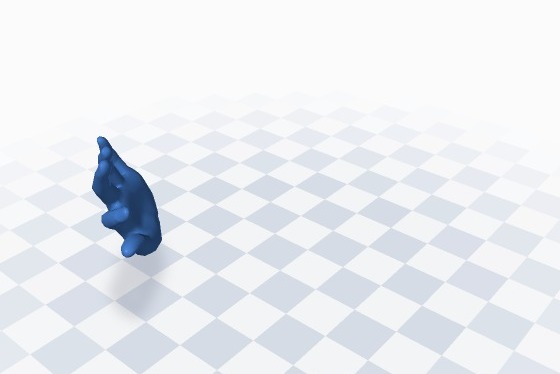}{}{QTag} \\[5pt]
\qrow{ARCTIC \#3} & \qtile{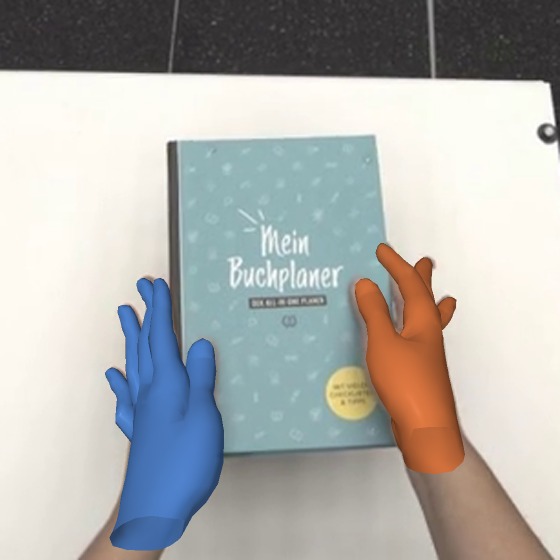}{}{QTag} & \qtile{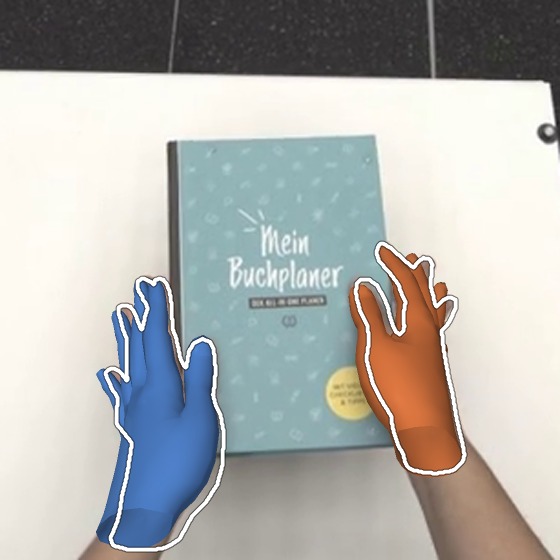}{5\,px}{HandRSIBlue} & \qtile{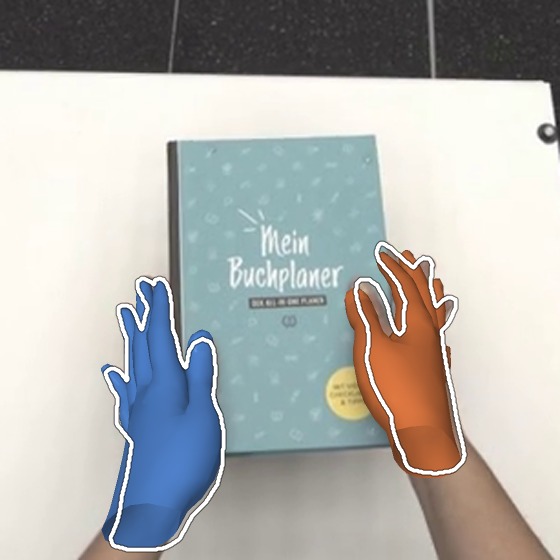}{5\,px}{QTag} & \qtile{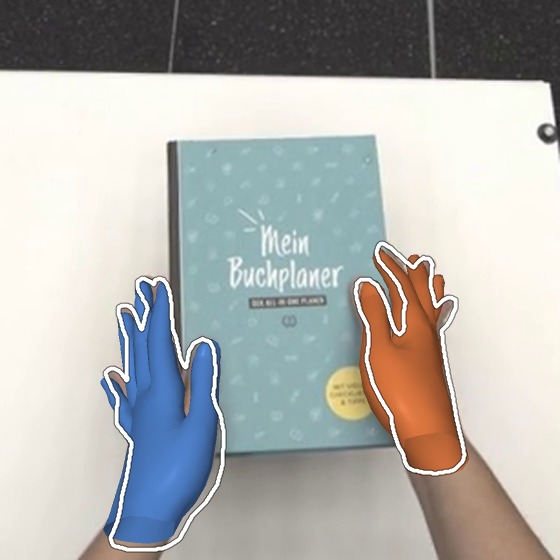}{5\,px}{QTag} & \qtile{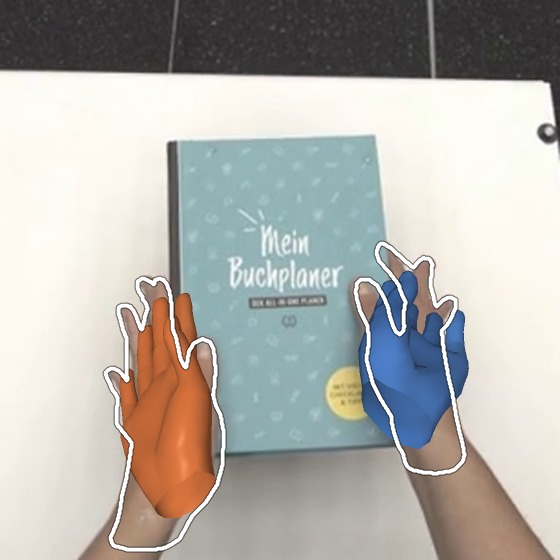}{$>$99\,px}{QTag} & \qtile{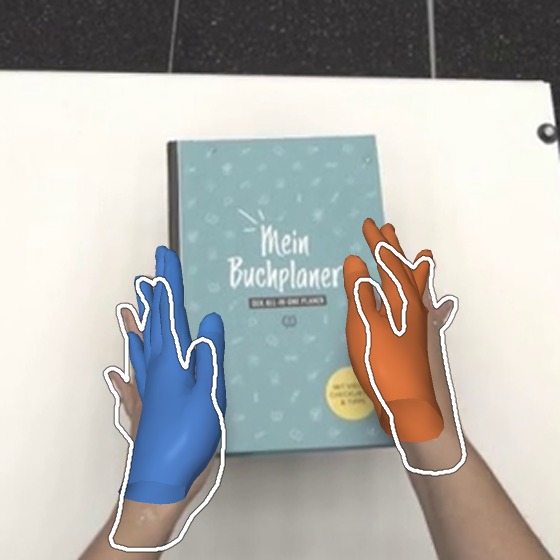}{20\,px}{QTag} \\[1pt]
\qrow{3D view} & \qtile{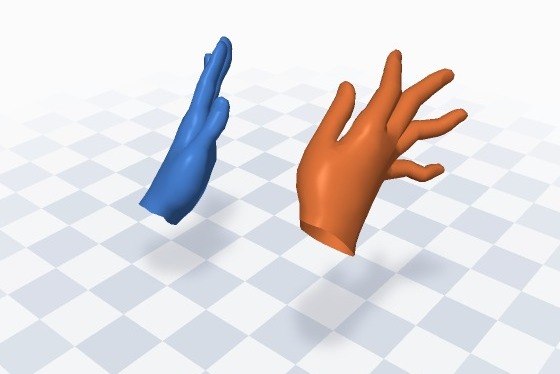}{}{QTag} & \qtile{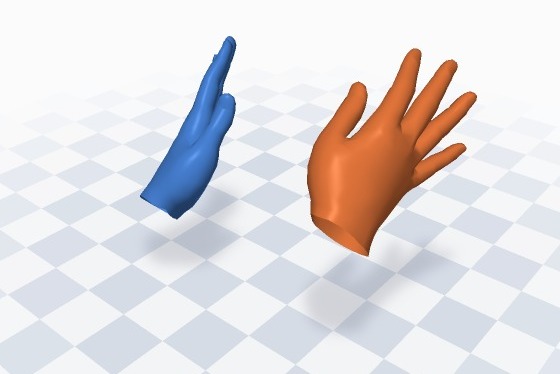}{}{QTag} & \qtile{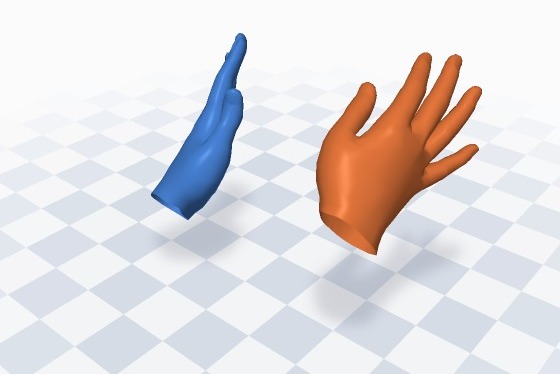}{}{QTag} & \qtile{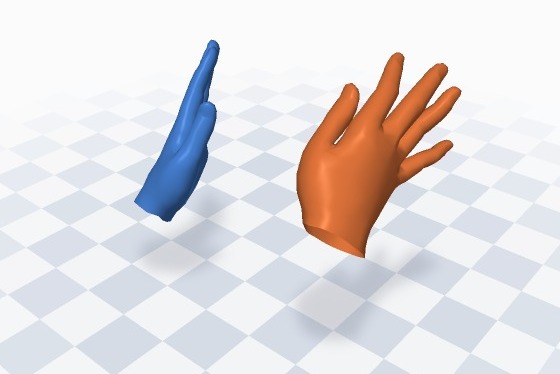}{}{QTag} & \qtile{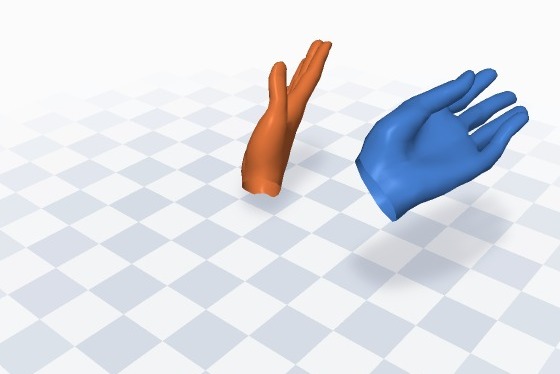}{}{QTag} & \qtile{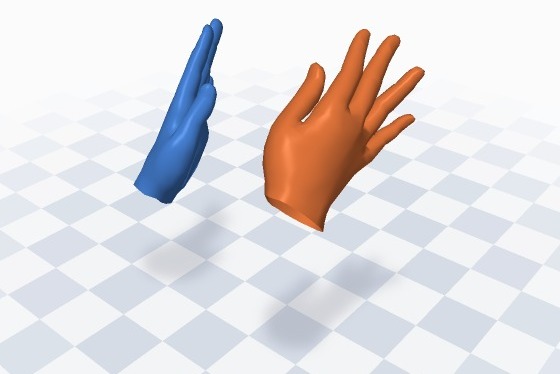}{}{QTag} \\[5pt]
\qrow{ARCTIC \#4} & \qtile{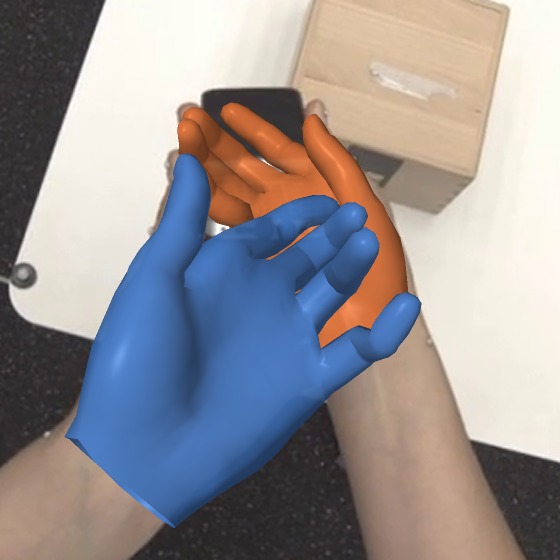}{}{QTag} & \qtile{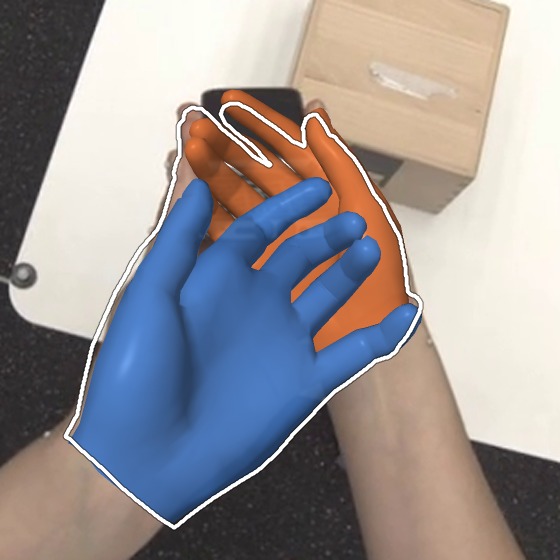}{7\,px}{HandRSIBlue} & \qtile{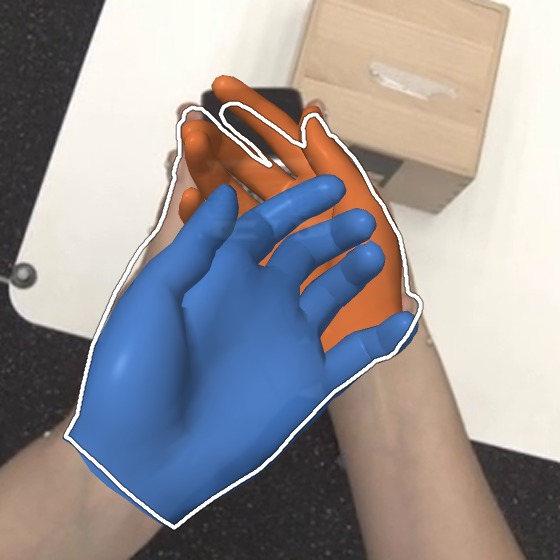}{8\,px}{QTag} & \qtile{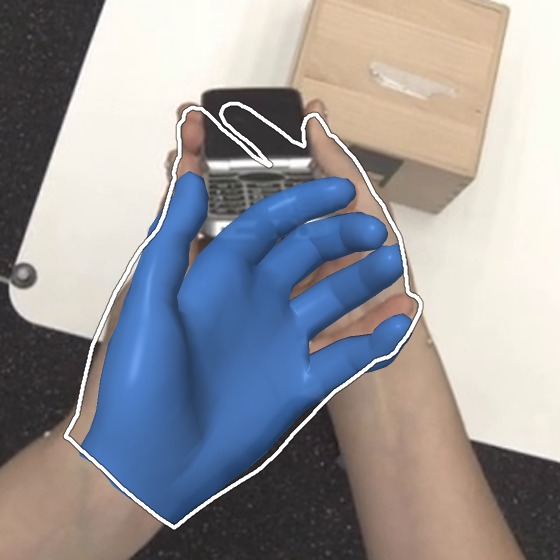}{$>$99\,px}{QTag} & \qtile{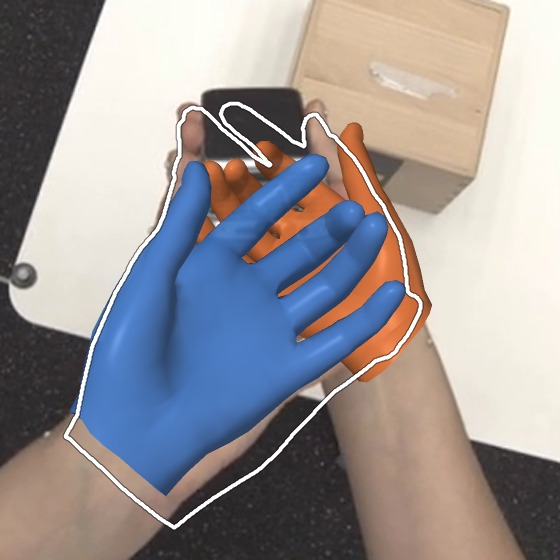}{16\,px}{QTag} & \qtile{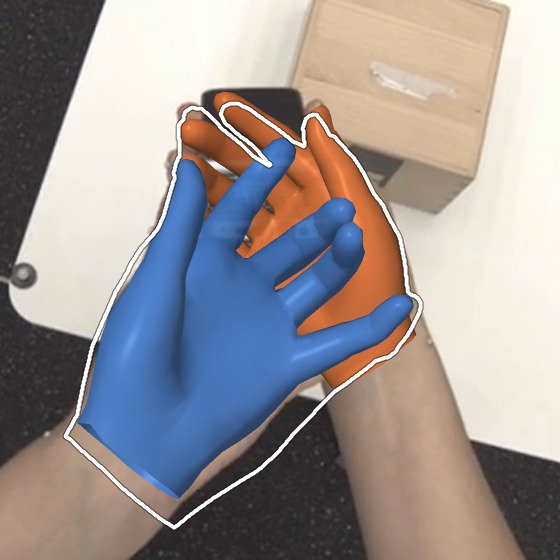}{11\,px}{QTag} \\[1pt]
\qrow{3D view} & \qtile{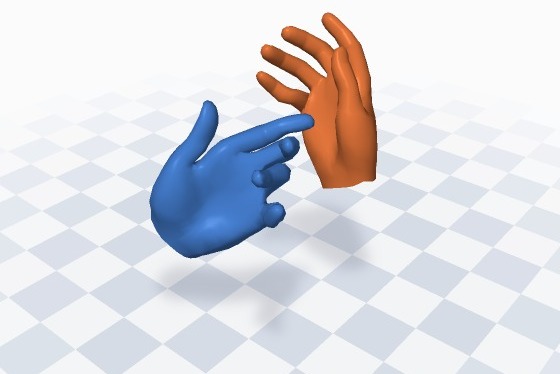}{}{QTag} & \qtile{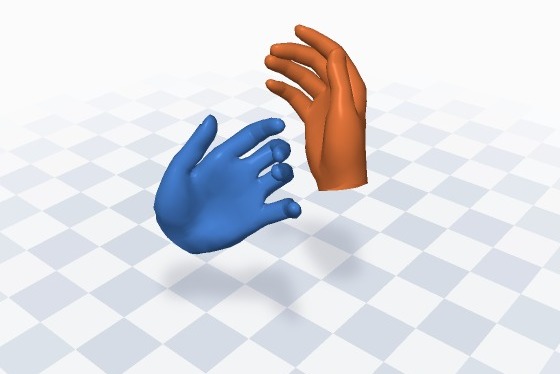}{}{QTag} & \qtile{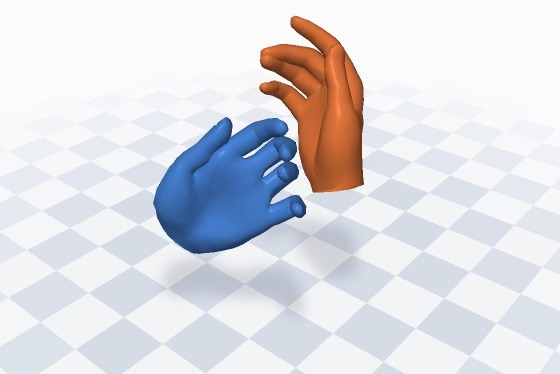}{}{QTag} & \qtile{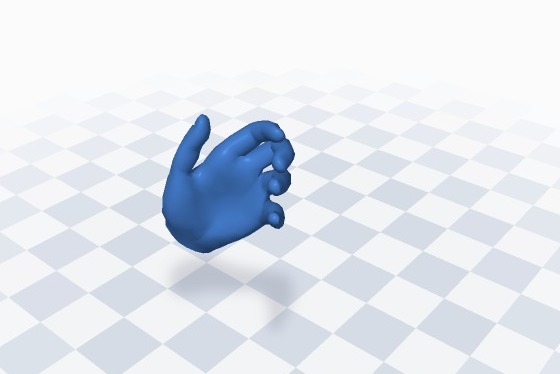}{}{QTag} & \qtile{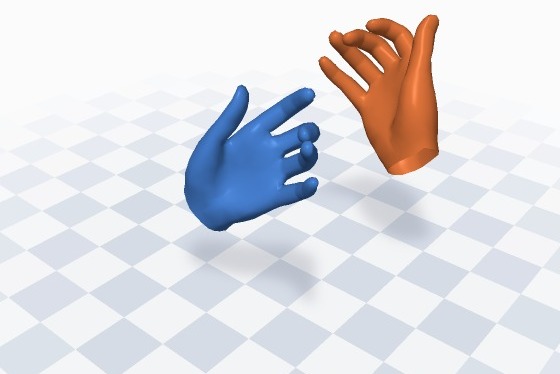}{}{QTag} & \qtile{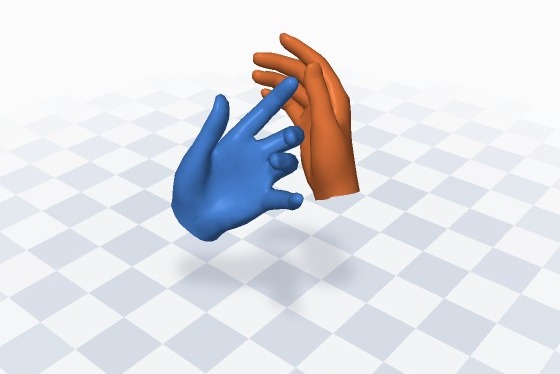}{}{QTag} \\
\end{tabular}
}
    \caption{
        \textbf{Extended comparison on the sealed ARCTIC test} (held-out subject s05), as in
        Fig.~\ref{fig:app_gal_hot3d}.
    }
    \label{fig:app_gal_arctic}
\end{figure*}
\begin{figure*}[p]
    \centering
    {\footnotesize
\setlength{\qw}{\dimexpr(\linewidth-1em-12pt)/6\relax}%
\begin{tabular}{@{}c@{\hspace{2pt}}c@{\hspace{2pt}}c@{\hspace{2pt}}c@{\hspace{2pt}}c@{\hspace{2pt}}c@{\hspace{2pt}}c@{}}
 & Ground truth & \textbf{\color{HandRSIBlue}HandRSI (ours)} & ACE-Ego-Hand~\cite{ace_ego_hand} & EgoForce~\cite{egoforce} & HaWoR~\cite{hawor} & WiLoR~\cite{wilor} \\[1pt]
\qrow{HOI4D \#1} & \qtile{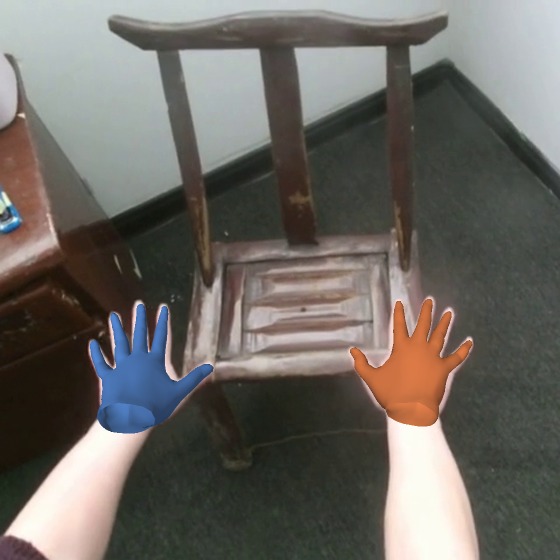}{}{QTag} & \qtile{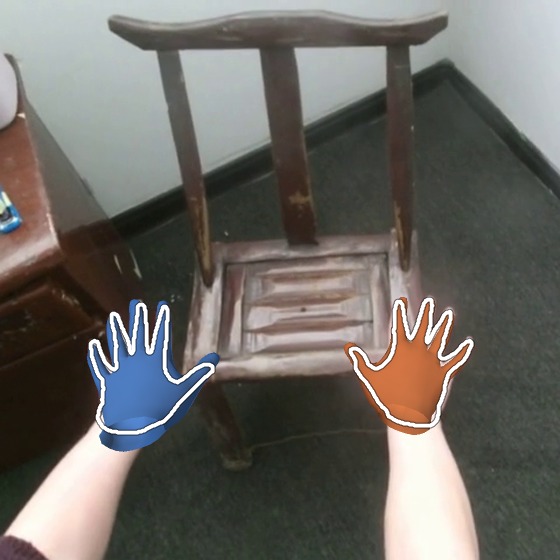}{5\,px}{HandRSIBlue} & \qtile{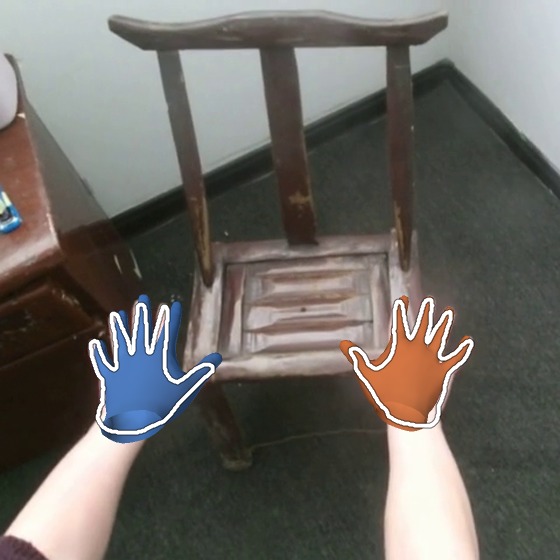}{7\,px}{QTag} & \qtile{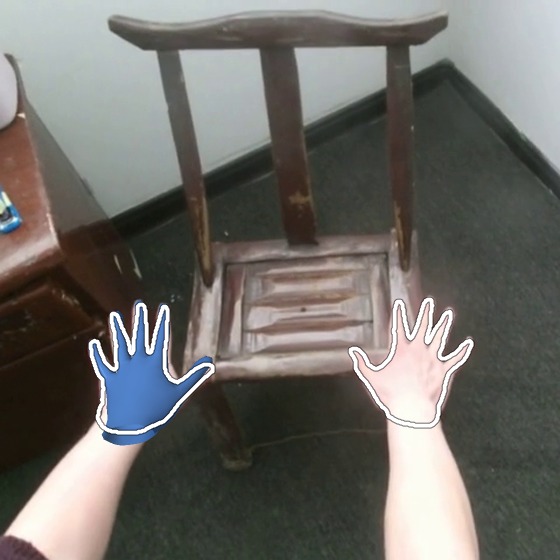}{missed}{QTag} & \qtile{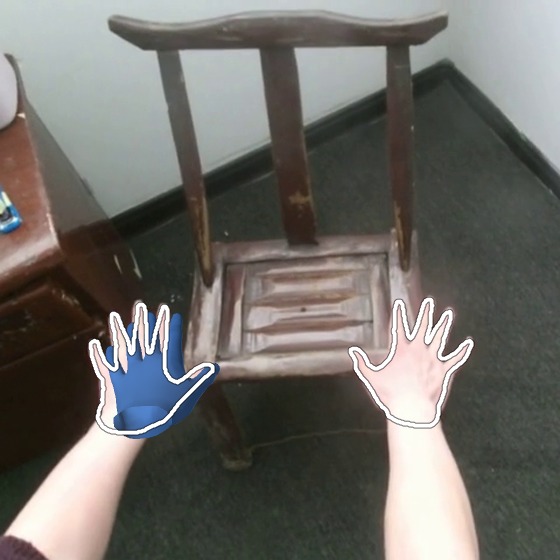}{missed}{QTag} & \qtile{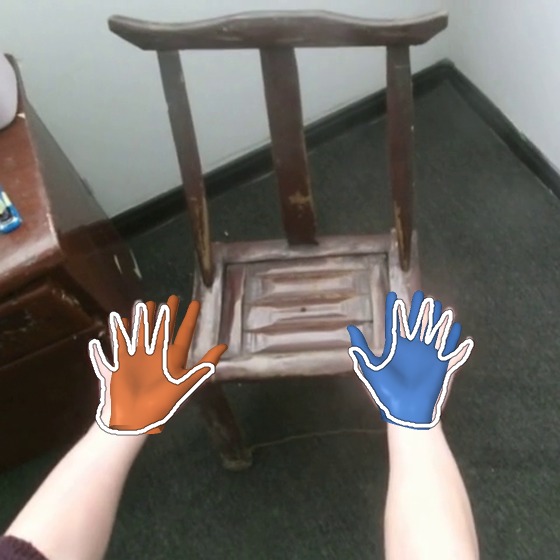}{$>$99\,px}{QTag} \\[1pt]
\qrow{3D view} & \qtile{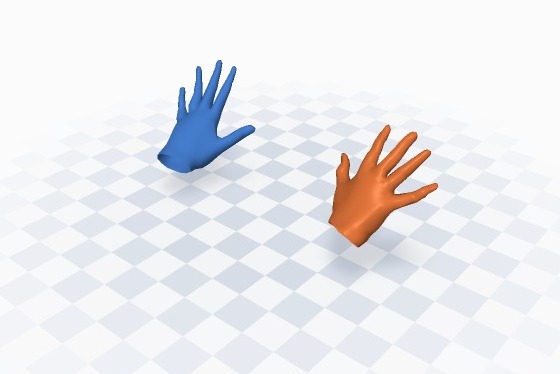}{}{QTag} & \qtile{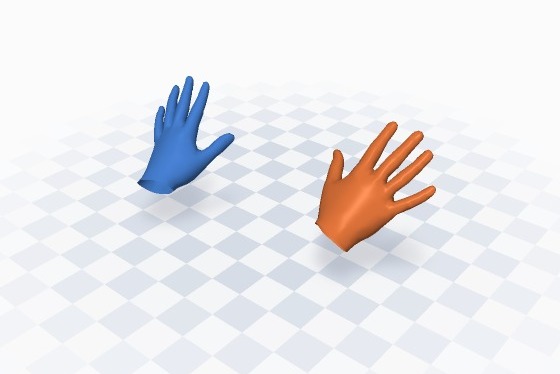}{}{QTag} & \qtile{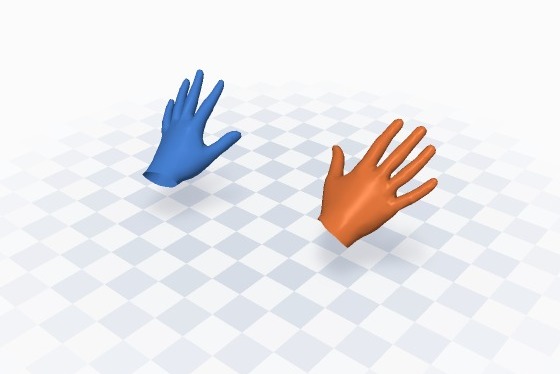}{}{QTag} & \qtile{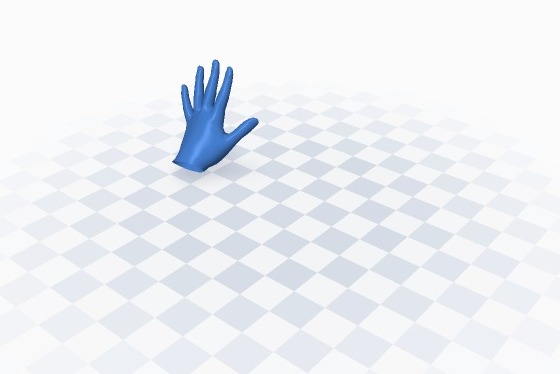}{}{QTag} & \qtile{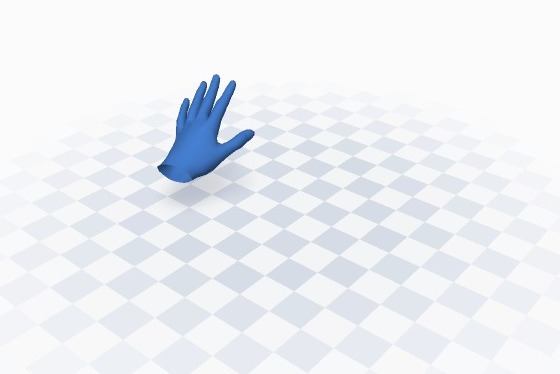}{}{QTag} & \qtile{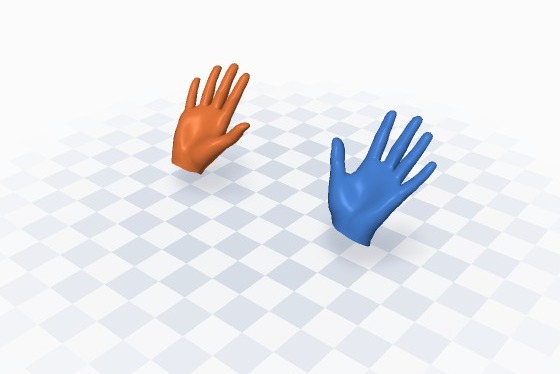}{}{QTag} \\[5pt]
\qrow{HOI4D \#2} & \qtile{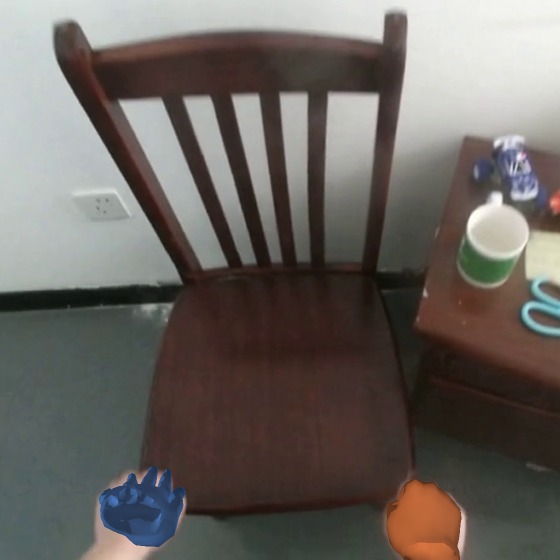}{}{QTag} & \qtile{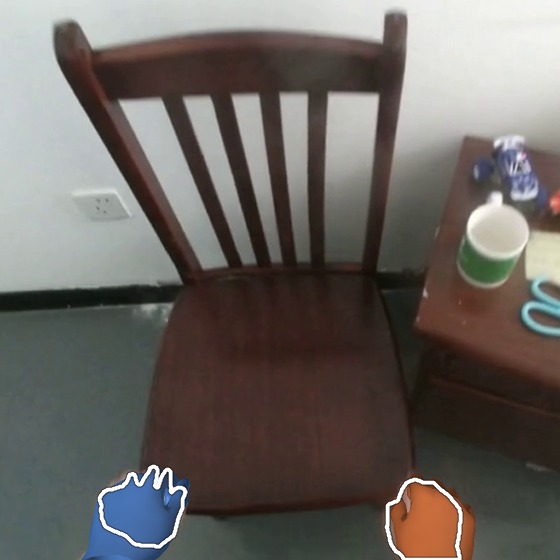}{17\,px}{HandRSIBlue} & \qtile{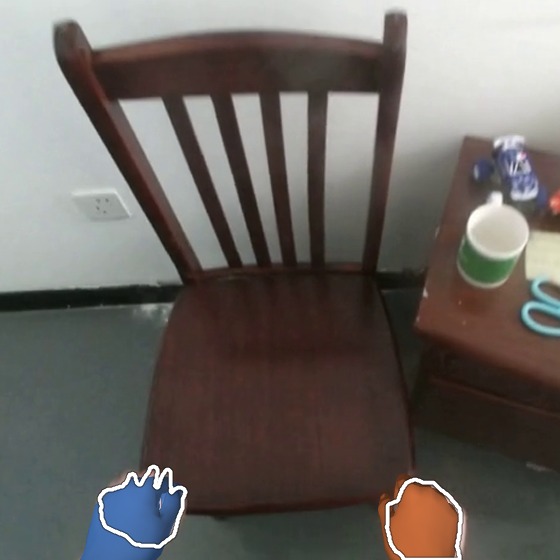}{14\,px}{QTag} & \qtile{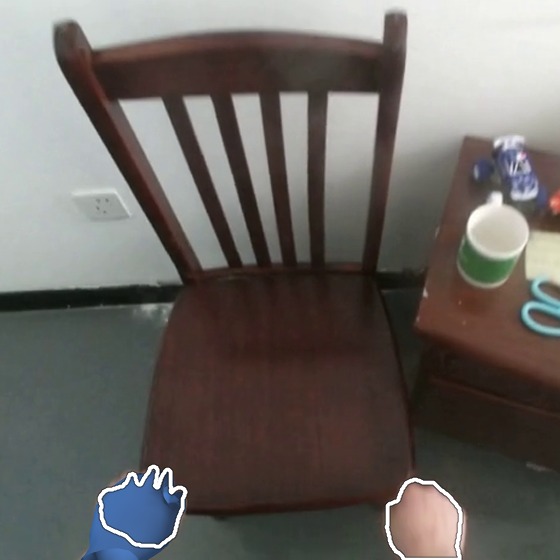}{missed}{QTag} & \qtile{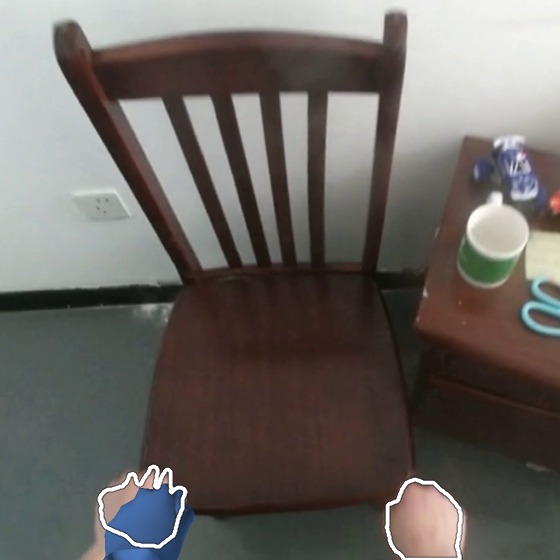}{missed}{QTag} & \qtile{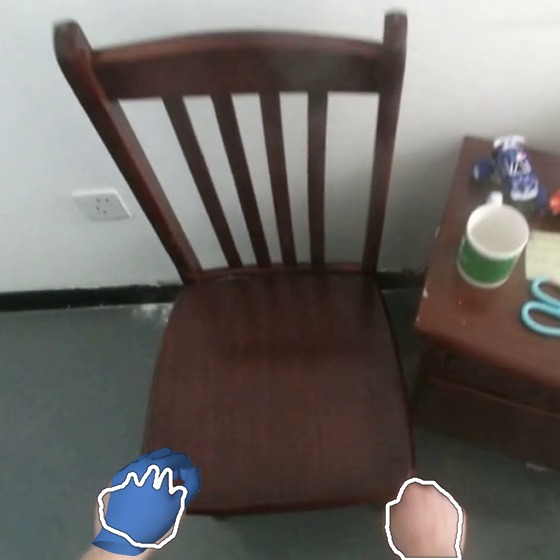}{missed}{QTag} \\[1pt]
\qrow{3D view} & \qtile{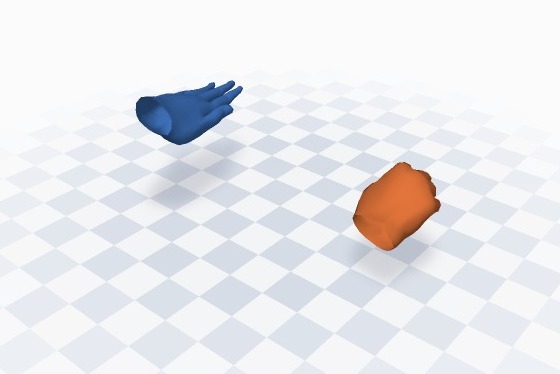}{}{QTag} & \qtile{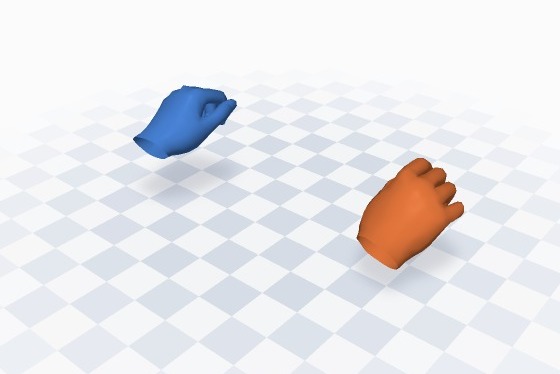}{}{QTag} & \qtile{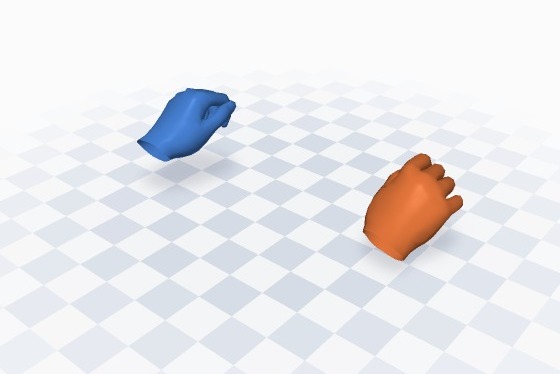}{}{QTag} & \qtile{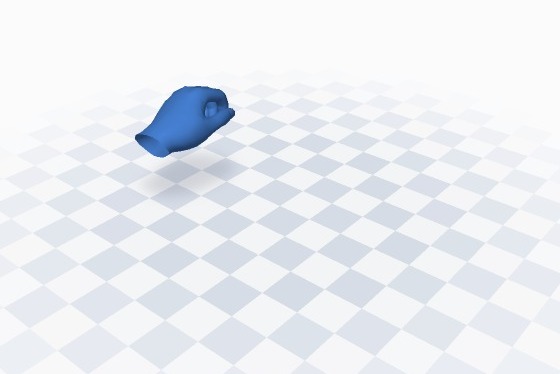}{}{QTag} & \qtile{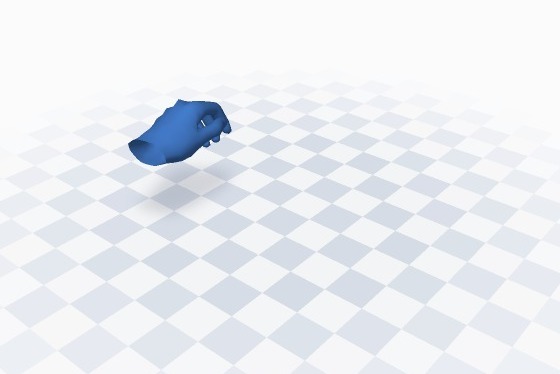}{}{QTag} & \qtile{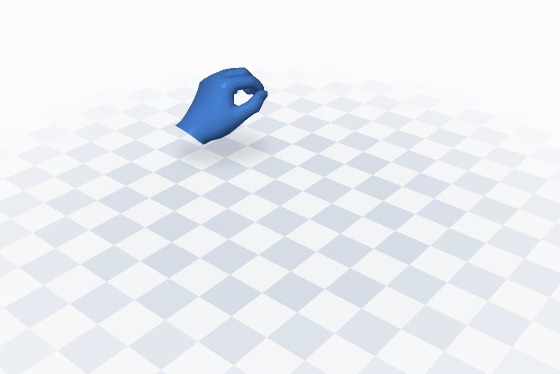}{}{QTag} \\[5pt]
\qrow{HOI4D \#3} & \qtile{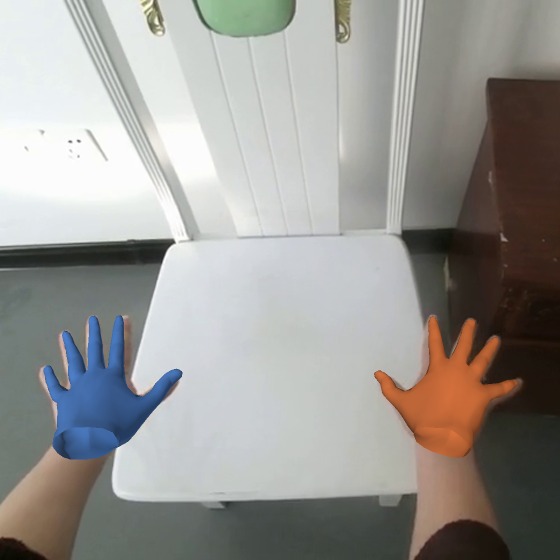}{}{QTag} & \qtile{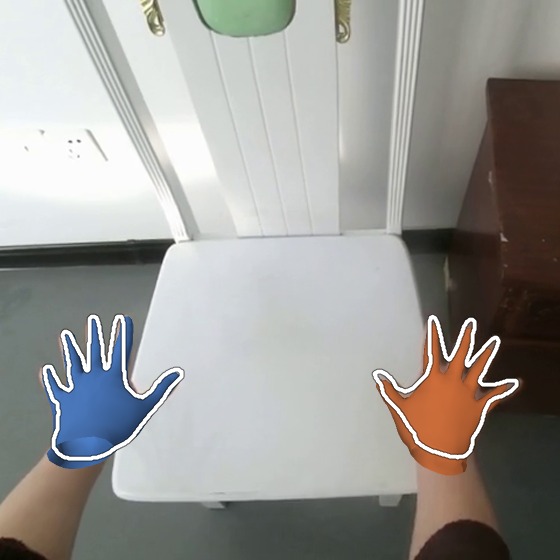}{7\,px}{HandRSIBlue} & \qtile{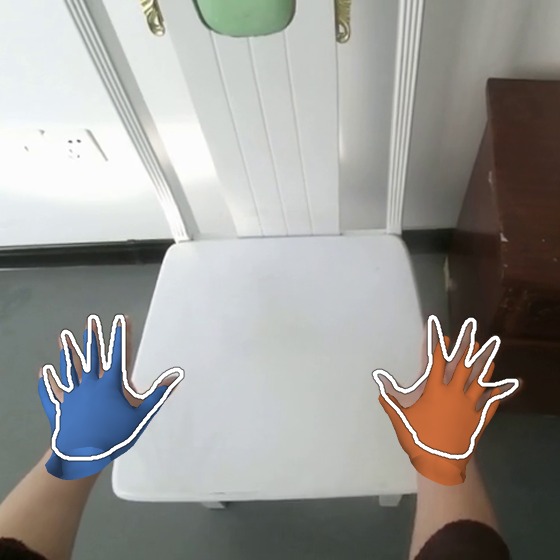}{16\,px}{QTag} & \qtile{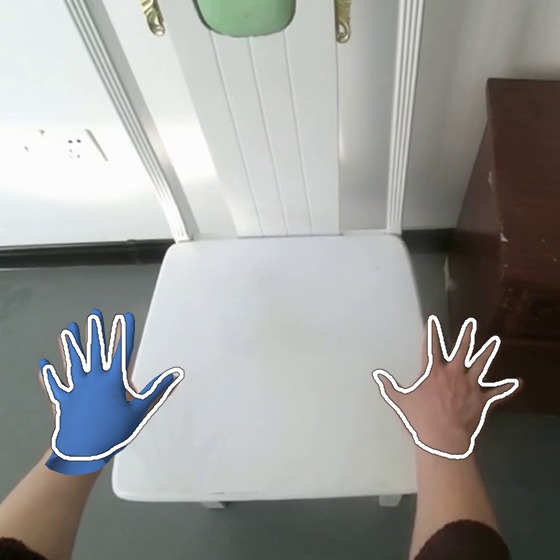}{$>$99\,px}{QTag} & \qtile{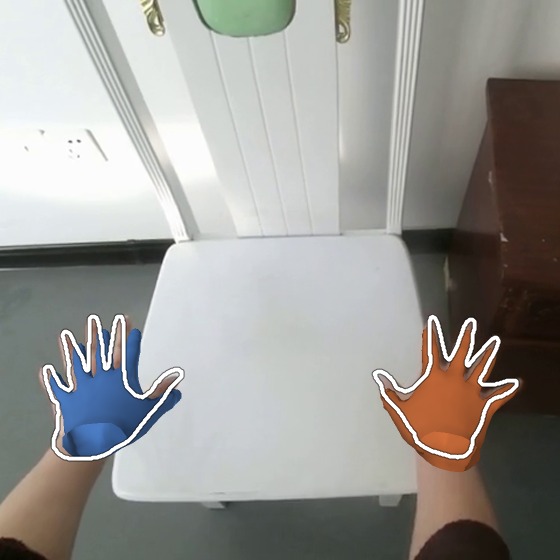}{11\,px}{QTag} & \qtile{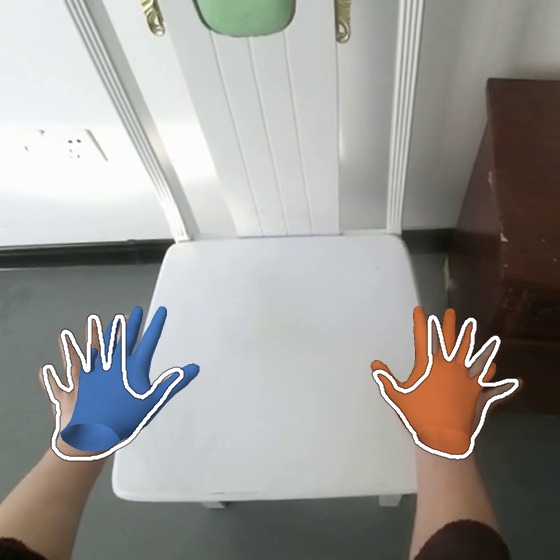}{22\,px}{QTag} \\[1pt]
\qrow{3D view} & \qtile{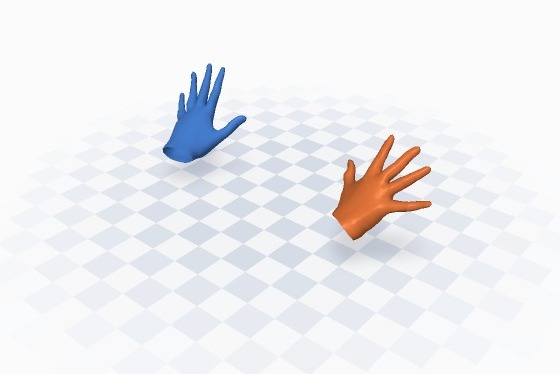}{}{QTag} & \qtile{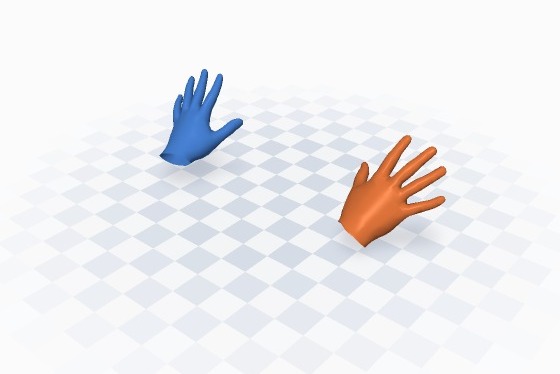}{}{QTag} & \qtile{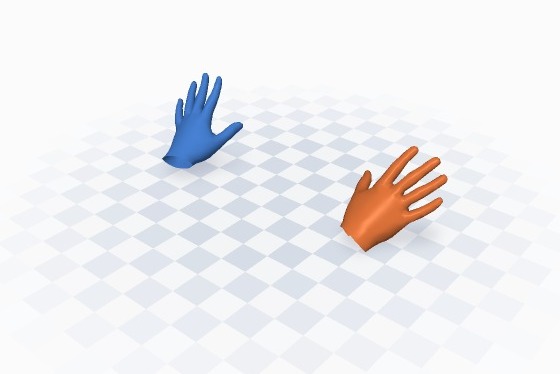}{}{QTag} & \qtile{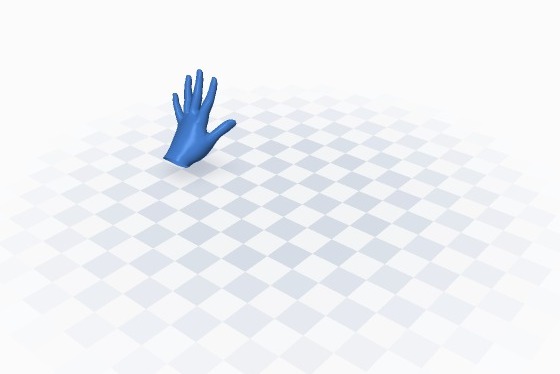}{}{QTag} & \qtile{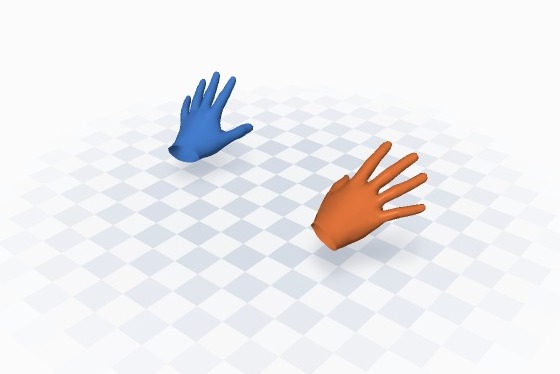}{}{QTag} & \qtile{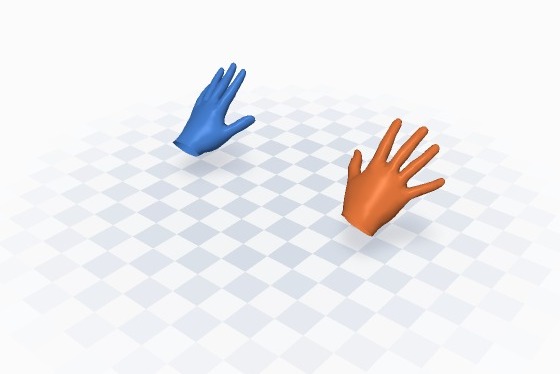}{}{QTag} \\[5pt]
\qrow{HOI4D \#4} & \qtile{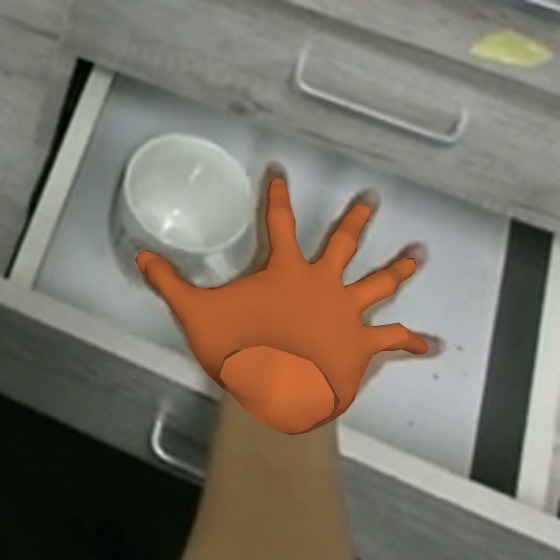}{}{QTag} & \qtile{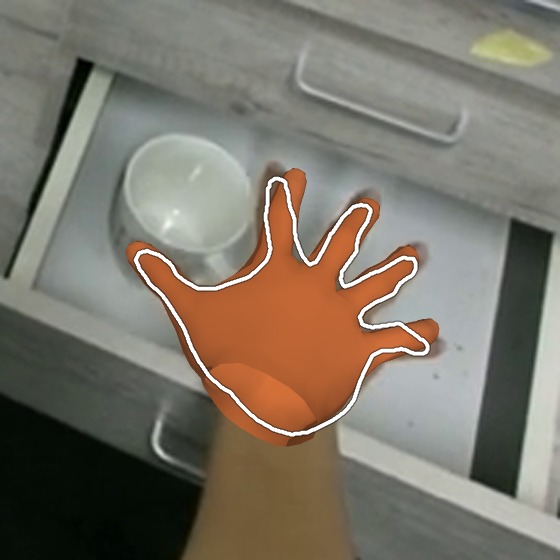}{4\,px}{HandRSIBlue} & \qtile{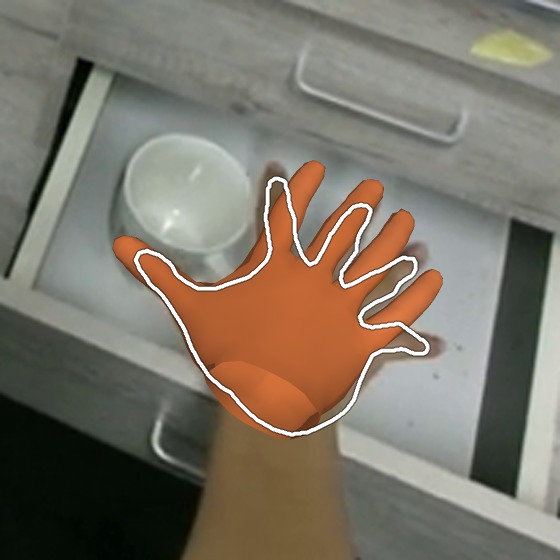}{9\,px}{QTag} & \qtile{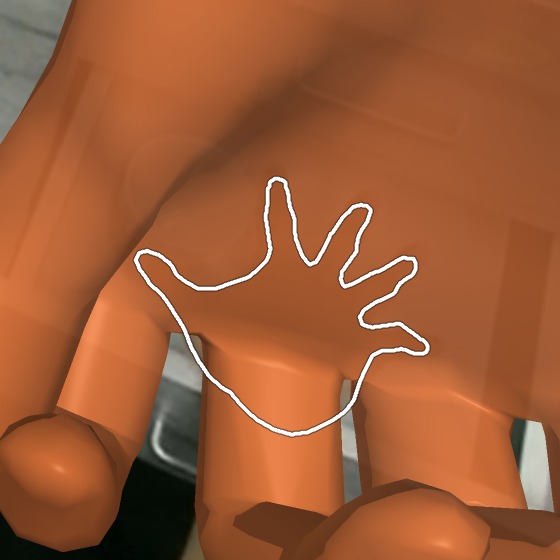}{$>$99\,px}{QTag} & \qtile{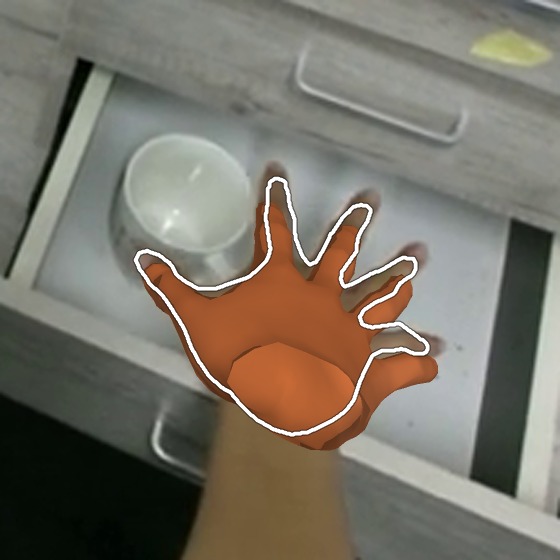}{9\,px}{QTag} & \qtile{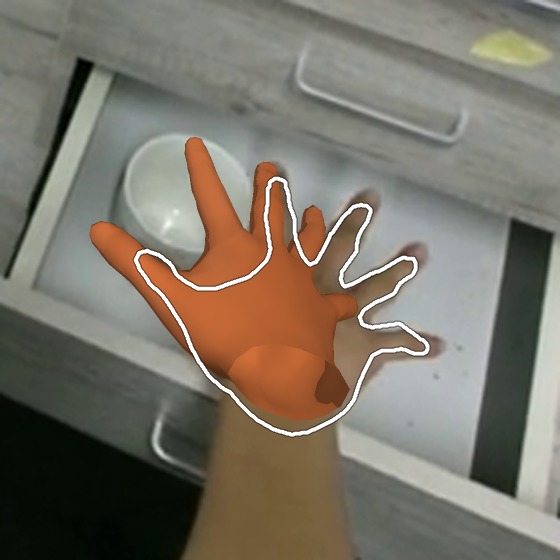}{27\,px}{QTag} \\[1pt]
\qrow{3D view} & \qtile{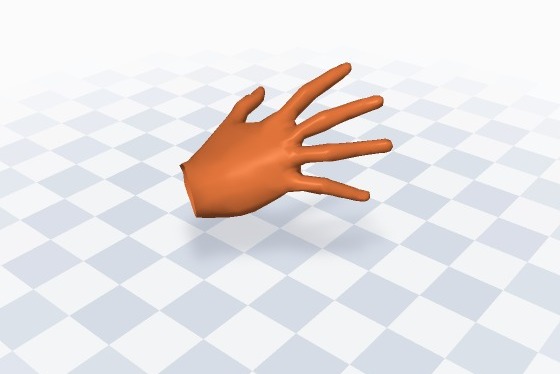}{}{QTag} & \qtile{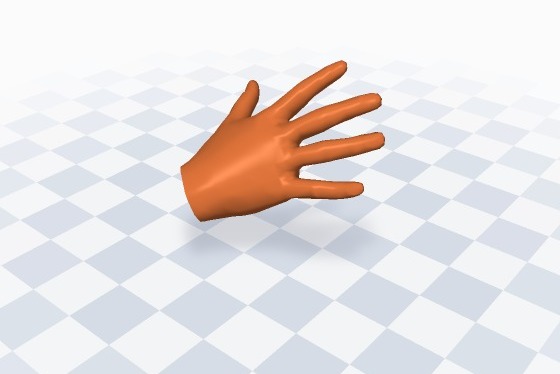}{}{QTag} & \qtile{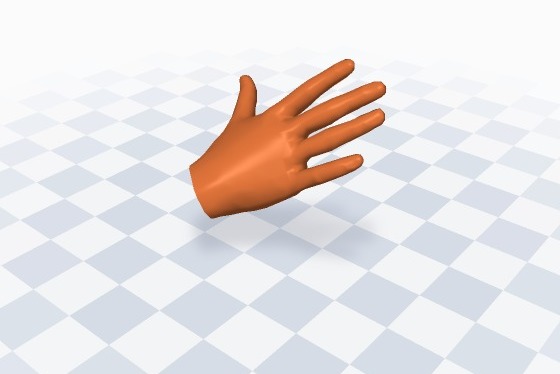}{}{QTag} & \qtile{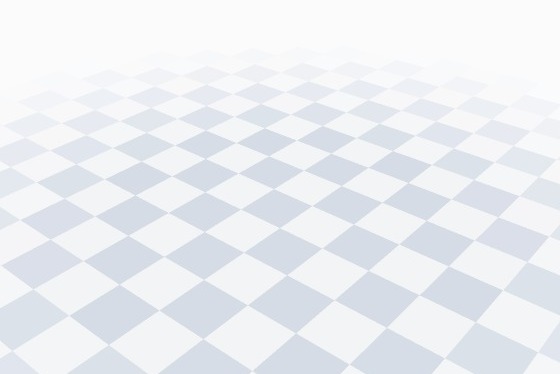}{}{QTag} & \qtile{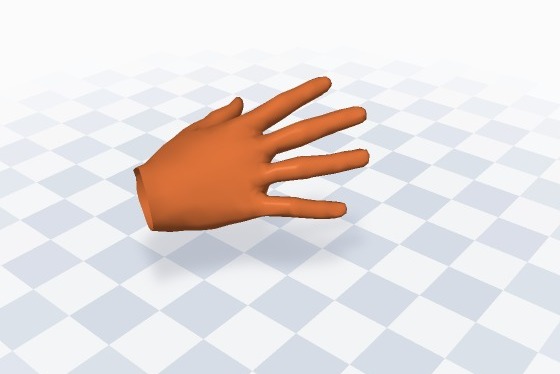}{}{QTag} & \qtile{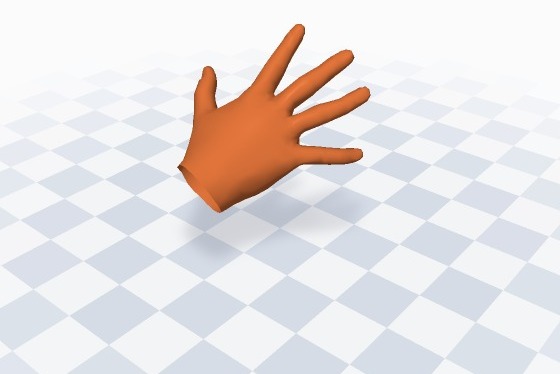}{}{QTag} \\
\end{tabular}
}
    \caption{
        \textbf{Extended comparison on HOI4D}, zero-shot for every method, as in Fig.~\ref{fig:app_gal_hot3d}.
    }
    \label{fig:app_gal_hoi4d}
\end{figure*}
\begin{figure*}[p]
    \centering
    {\footnotesize
\setlength{\qw}{\dimexpr(\linewidth-1em-14.5pt)/7\relax}%
\begin{tabular}{@{}c@{\hspace{2pt}}c@{\hspace{2pt}}c@{\hspace{2pt}}c@{\hspace{2pt}}c@{\hspace{2pt}}c@{\hspace{2pt}}c@{\hspace{2pt}}c@{}}
 & Ground truth & ACE-Ego-Hand & $\mathcal{S}_1$ & $\mathcal{S}_2$ & $\mathcal{S}_3$ & $\mathcal{S}_4$ & \textbf{\color{HandRSIBlue}$\mathcal{S}_5$ (HandRSI)} \\[1pt]
\qrow{frame 36} & \qtile{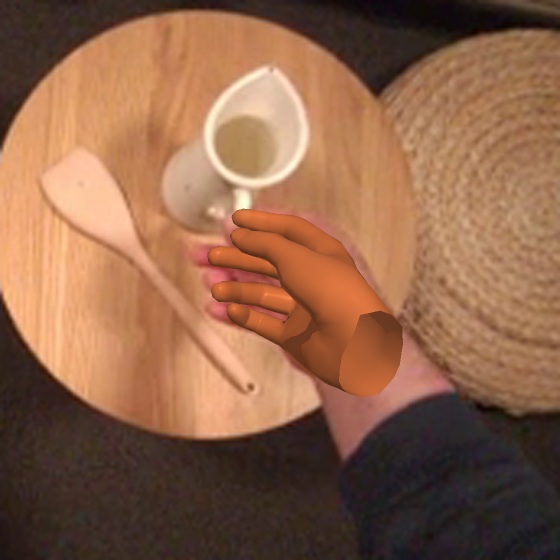}{}{QTag} & \qtile{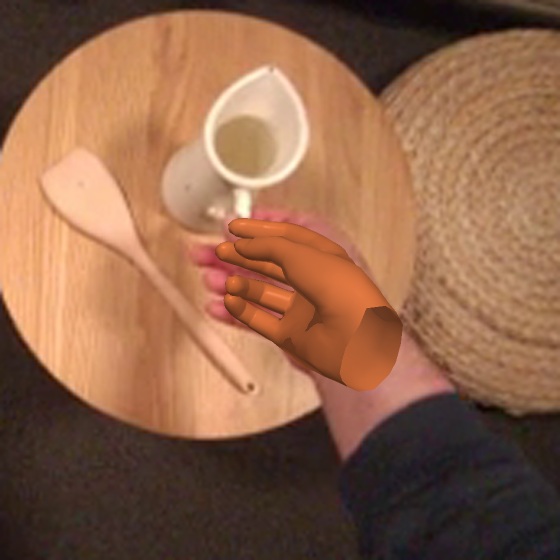}{}{QTag} & \qtile{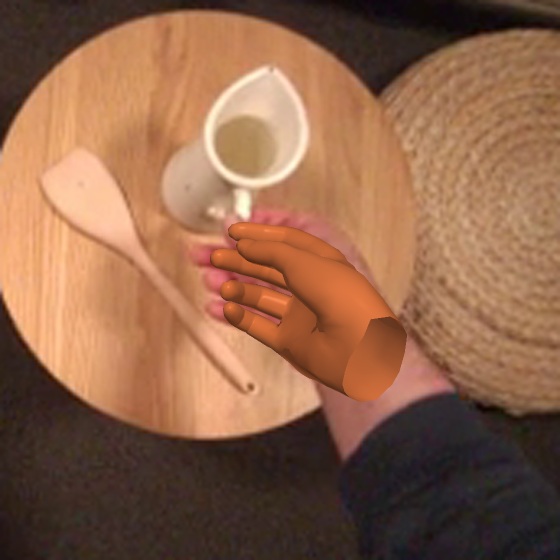}{}{QTag} & \qtile{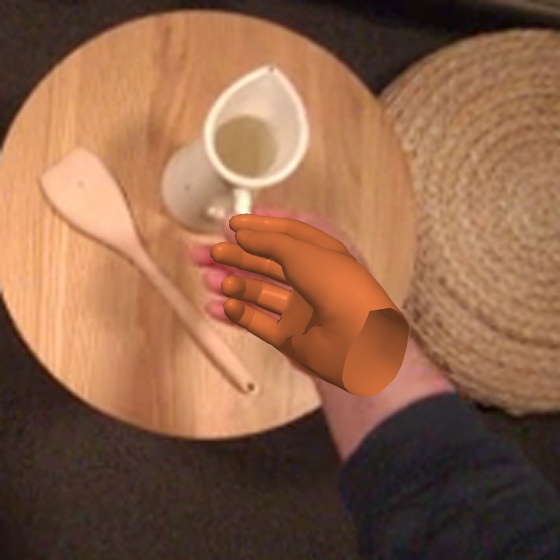}{}{QTag} & \qtile{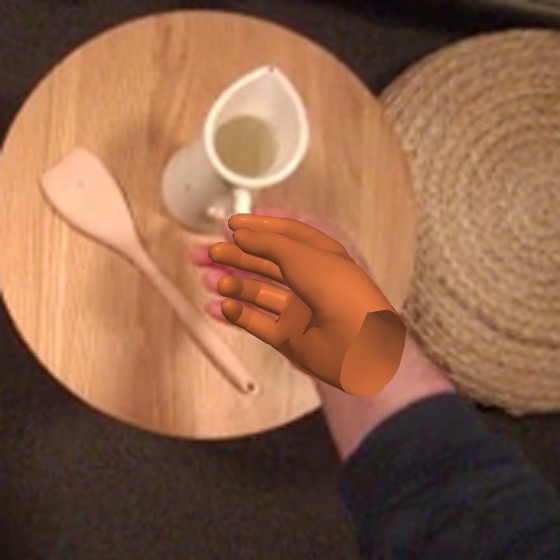}{}{QTag} & \qtile{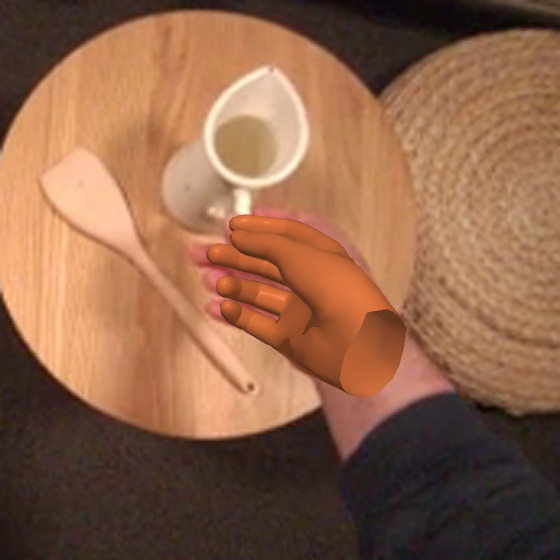}{}{QTag} & \qtile{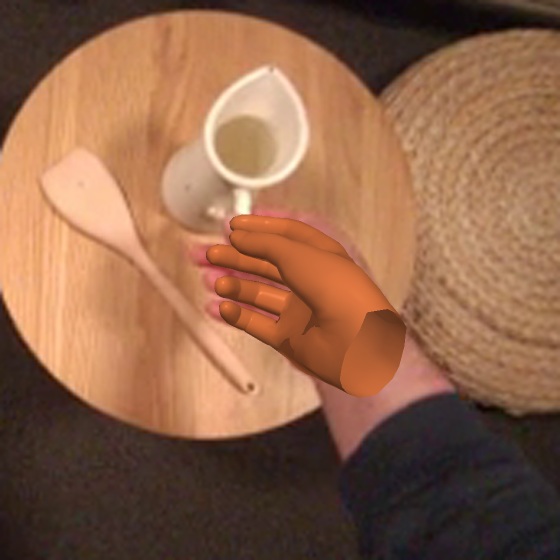}{}{QTag} \\[2pt]
\qrow{frame 62} & \qtile{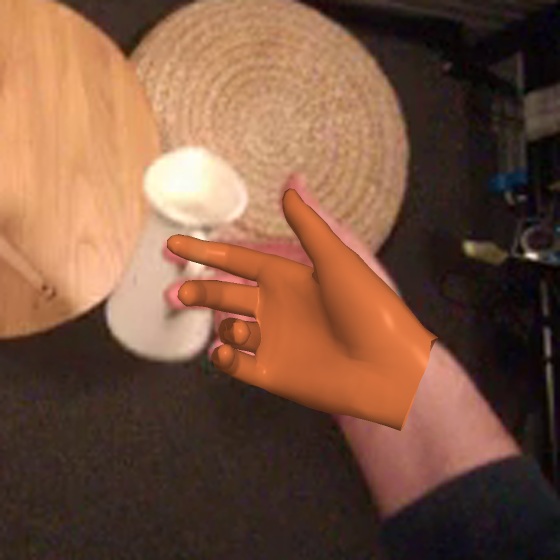}{}{QTag} & \qtile{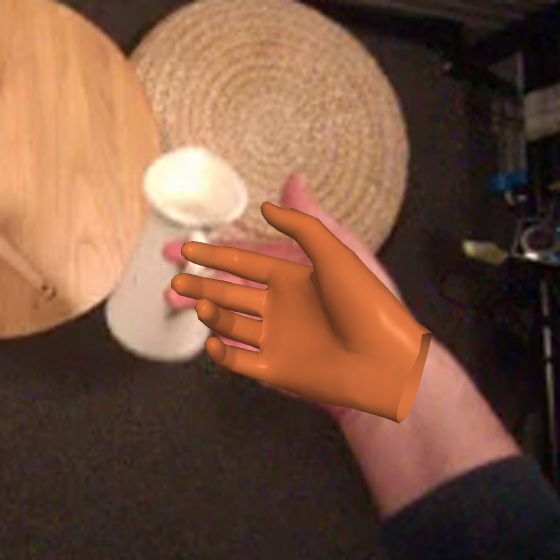}{}{QTag} & \qtile{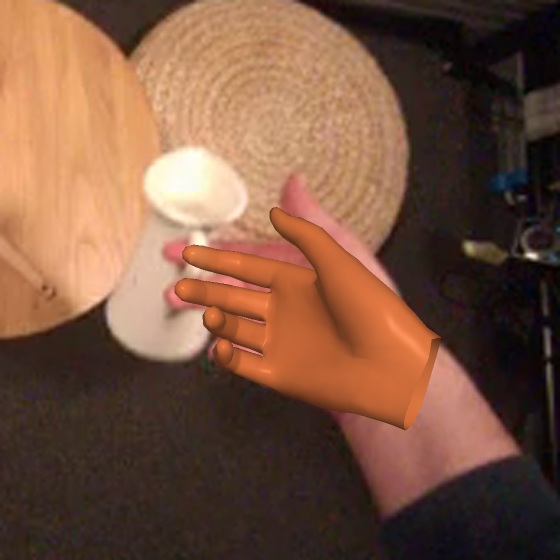}{}{QTag} & \qtile{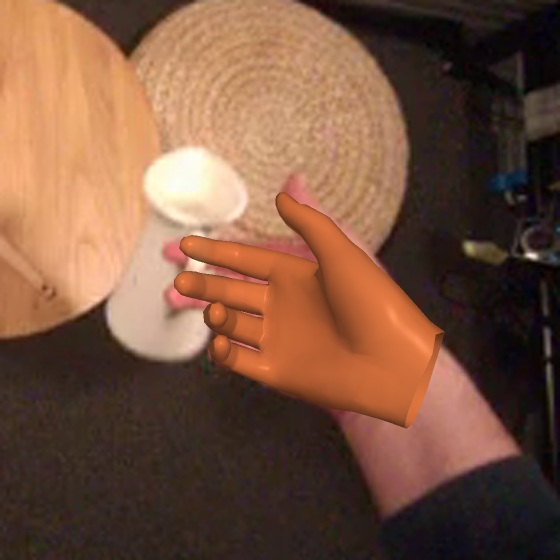}{}{QTag} & \qtile{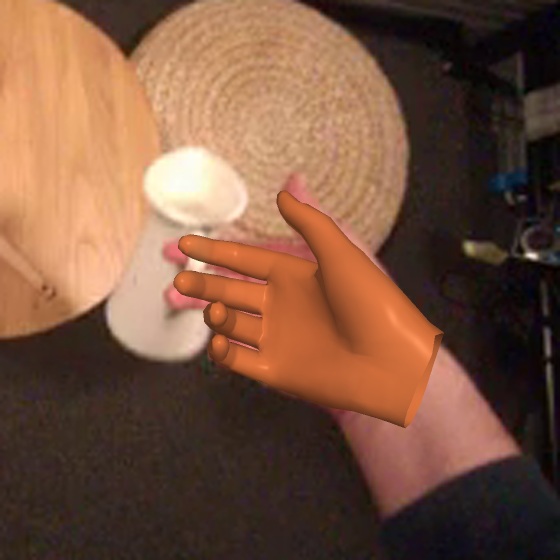}{}{QTag} & \qtile{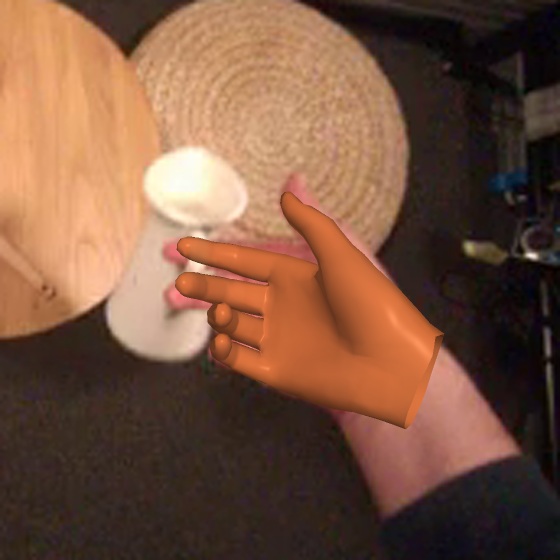}{}{QTag} & \qtile{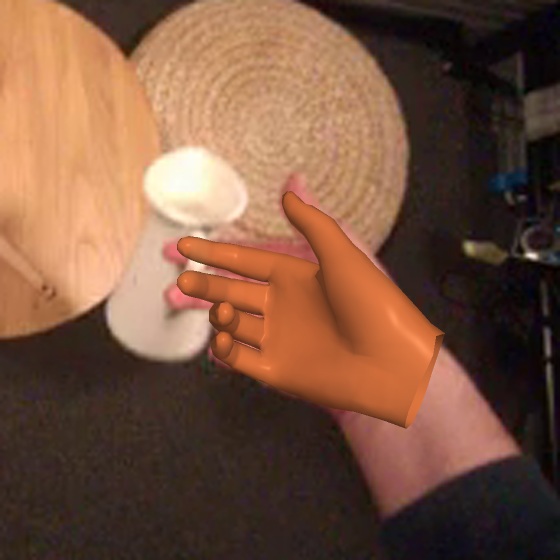}{}{QTag} \\[2pt]
\qrow{error} & \parbox[c]{\qw}{\centering\scriptsize per-frame error\\ (0--40\,mm)\\ \textcolor{QMiss}{\rule{4pt}{4pt}}~missed} & \qtile{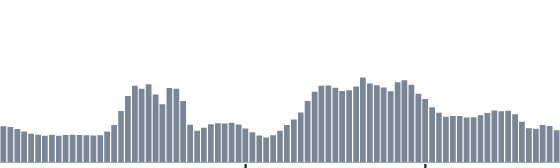}{}{QTag} & \qtile{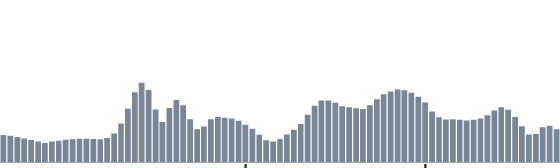}{}{QTag} & \qtile{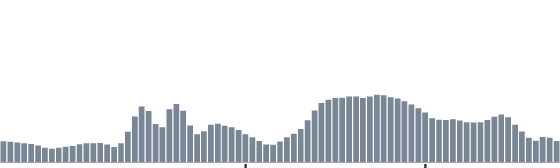}{}{QTag} & \qtile{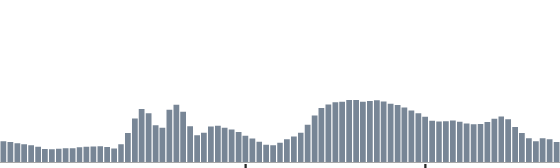}{}{QTag} & \qtile{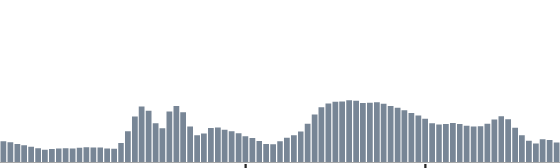}{}{QTag} & \qtile{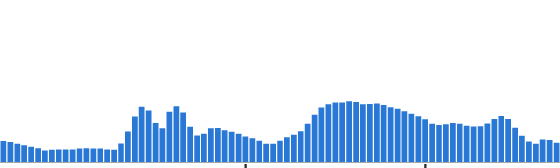}{}{QTag} \\[1pt]
 & \footnotesize MPJPE-p & \footnotesize 45.9\,mm & \footnotesize 10.2\,mm & \footnotesize 8.7\,mm & \footnotesize 8.2\,mm & \footnotesize 8.0\,mm & \footnotesize\textbf{\color{HandRSIBlue}7.9\,mm} \\
\end{tabular}
}\\[7pt]
    {\footnotesize
\setlength{\qw}{\dimexpr(\linewidth-1em-14.5pt)/7\relax}%
\begin{tabular}{@{}c@{\hspace{2pt}}c@{\hspace{2pt}}c@{\hspace{2pt}}c@{\hspace{2pt}}c@{\hspace{2pt}}c@{\hspace{2pt}}c@{\hspace{2pt}}c@{}}
 & Ground truth & ACE-Ego-Hand & $\mathcal{S}_1$ & $\mathcal{S}_2$ & $\mathcal{S}_3$ & $\mathcal{S}_4$ & \textbf{\color{HandRSIBlue}$\mathcal{S}_5$ (HandRSI)} \\[1pt]
\qrow{frame 8} & \qtile{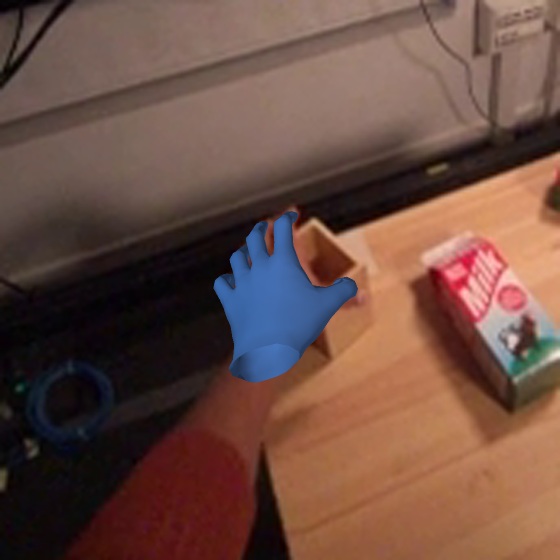}{}{QTag} & \qtile{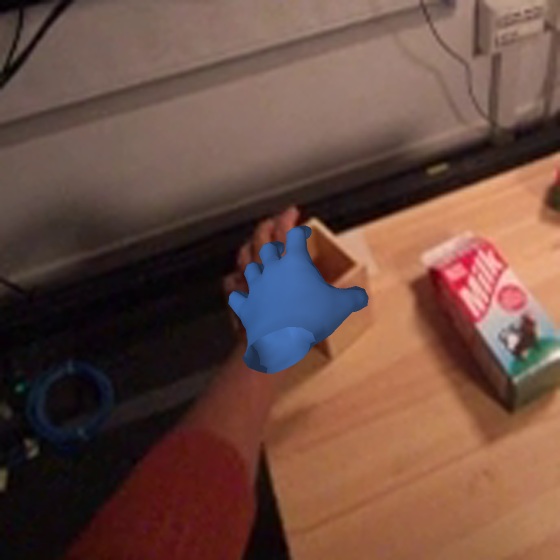}{}{QTag} & \qtile{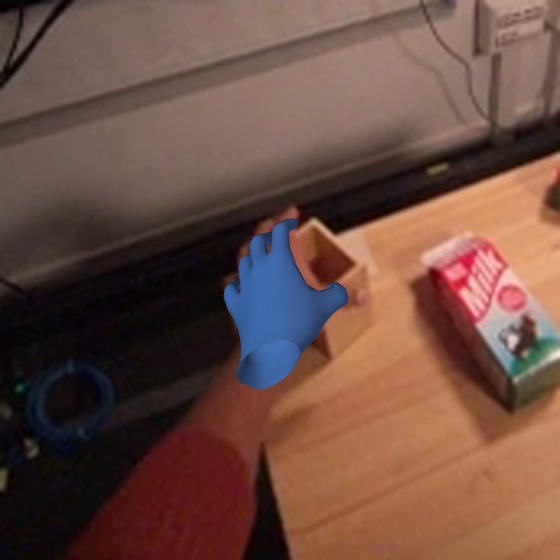}{}{QTag} & \qtile{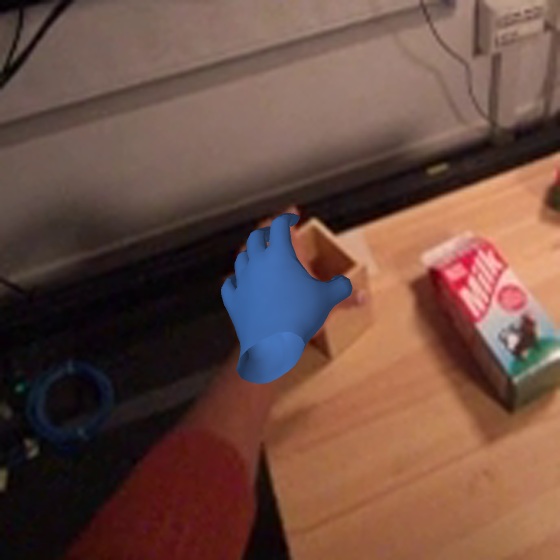}{}{QTag} & \qtile{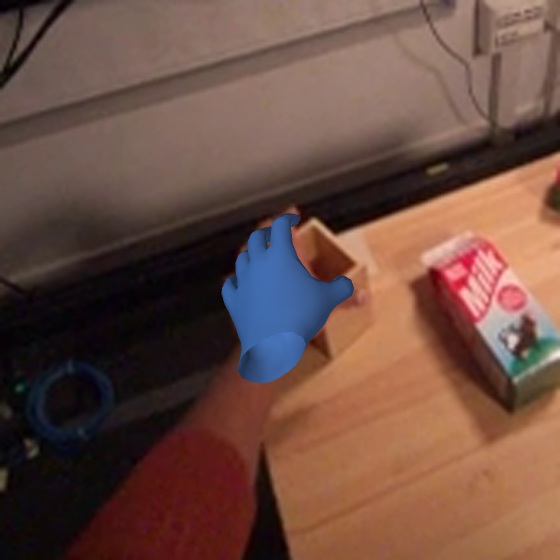}{}{QTag} & \qtile{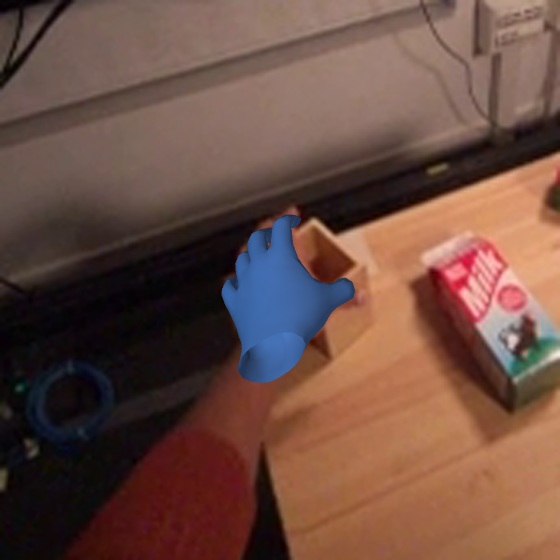}{}{QTag} & \qtile{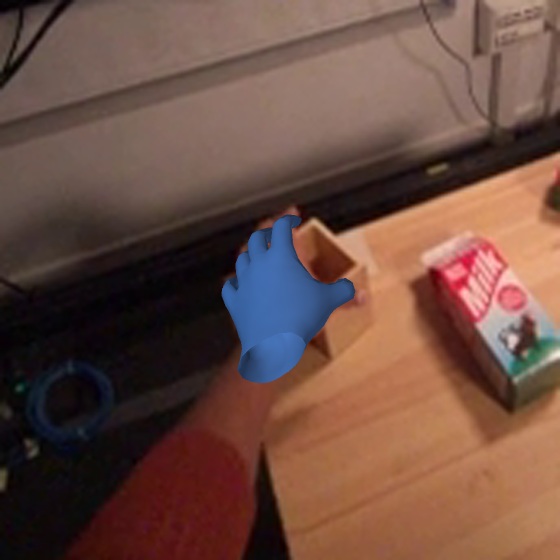}{}{QTag} \\[2pt]
\qrow{frame 69} & \qtile{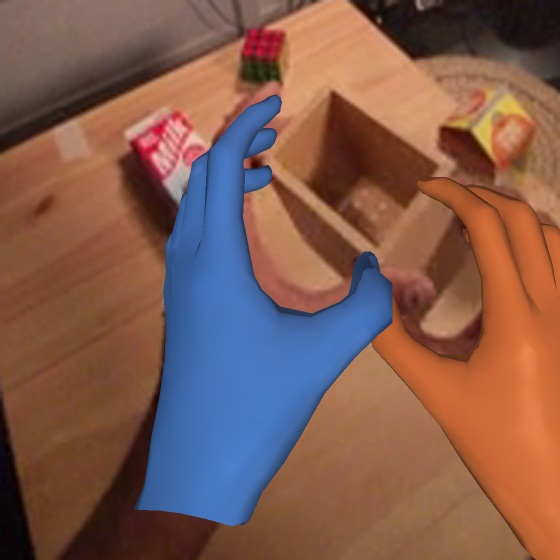}{}{QTag} & \qtile{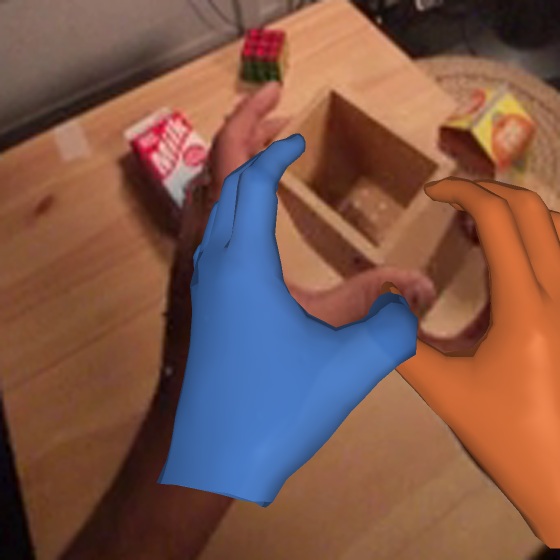}{}{QTag} & \qtile{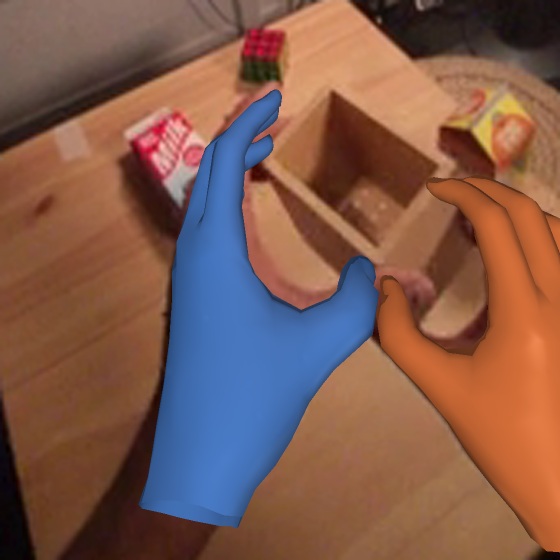}{}{QTag} & \qtile{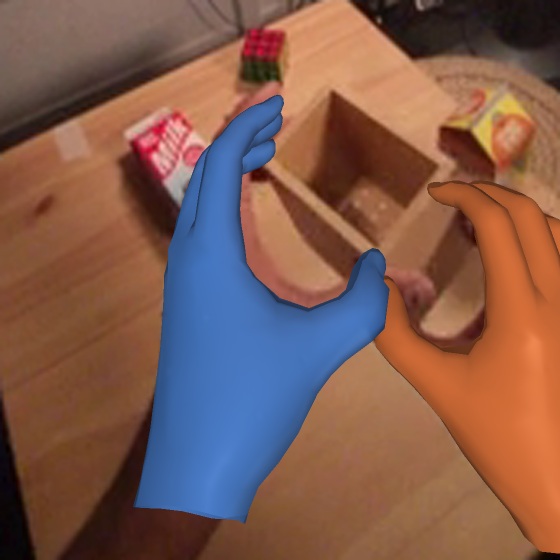}{}{QTag} & \qtile{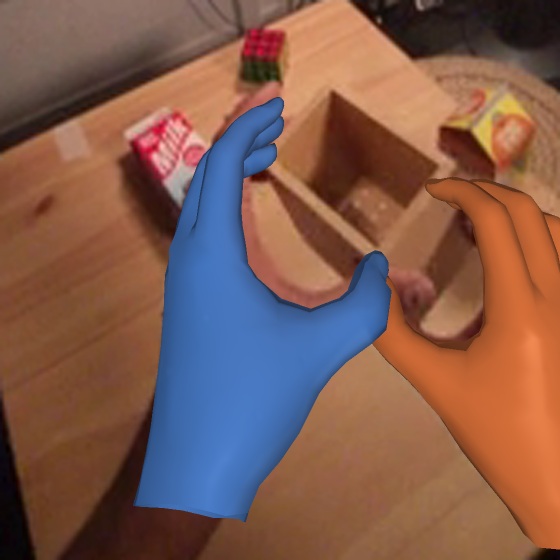}{}{QTag} & \qtile{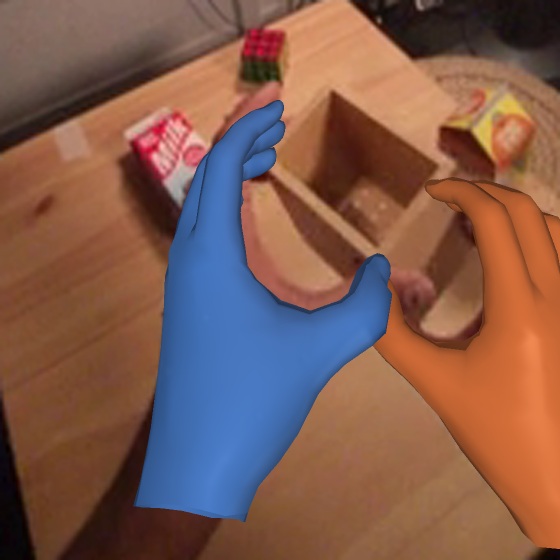}{}{QTag} & \qtile{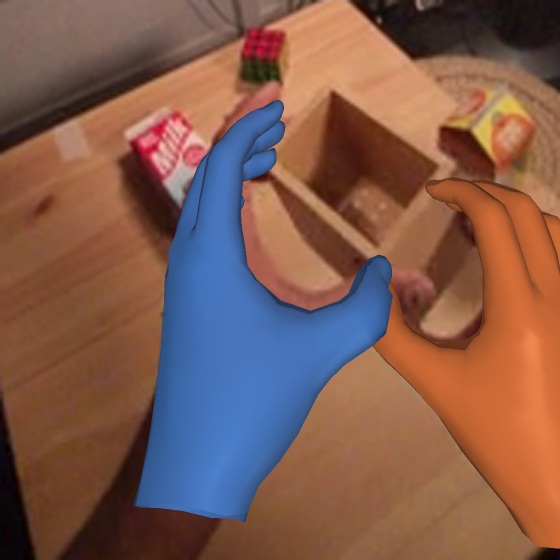}{}{QTag} \\[2pt]
\qrow{error} & \parbox[c]{\qw}{\centering\scriptsize per-frame error\\ (0--40\,mm)\\ \textcolor{QMiss}{\rule{4pt}{4pt}}~missed} & \qtile{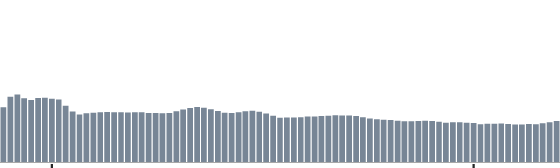}{}{QTag} & \qtile{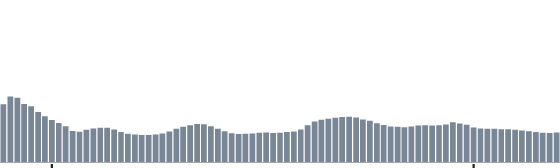}{}{QTag} & \qtile{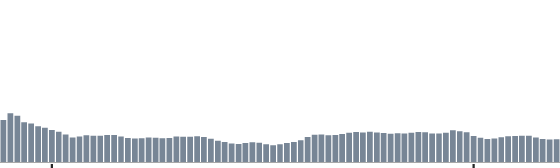}{}{QTag} & \qtile{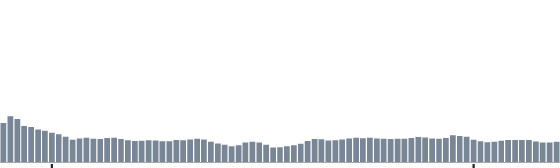}{}{QTag} & \qtile{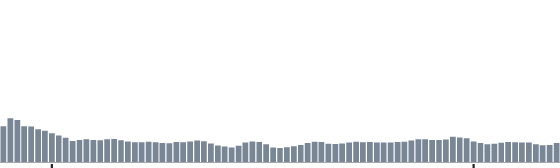}{}{QTag} & \qtile{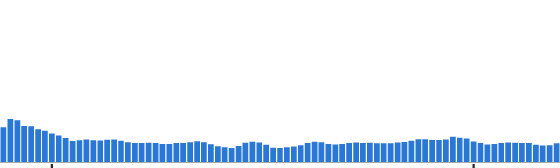}{}{QTag} \\[1pt]
 & \footnotesize MPJPE-p & \footnotesize 30.2\,mm & \footnotesize 9.0\,mm & \footnotesize 7.6\,mm & \footnotesize 6.9\,mm & \footnotesize 6.6\,mm & \footnotesize\textbf{\color{HandRSIBlue}6.5\,mm} \\
\end{tabular}
}\\[7pt]
    {\footnotesize
\setlength{\qw}{\dimexpr(\linewidth-1em-14.5pt)/7\relax}%
\begin{tabular}{@{}c@{\hspace{2pt}}c@{\hspace{2pt}}c@{\hspace{2pt}}c@{\hspace{2pt}}c@{\hspace{2pt}}c@{\hspace{2pt}}c@{\hspace{2pt}}c@{}}
 & Ground truth & ACE-Ego-Hand & $\mathcal{S}_1$ & $\mathcal{S}_2$ & $\mathcal{S}_3$ & $\mathcal{S}_4$ & \textbf{\color{HandRSIBlue}$\mathcal{S}_5$ (HandRSI)} \\[1pt]
\qrow{frame 2} & \qtile{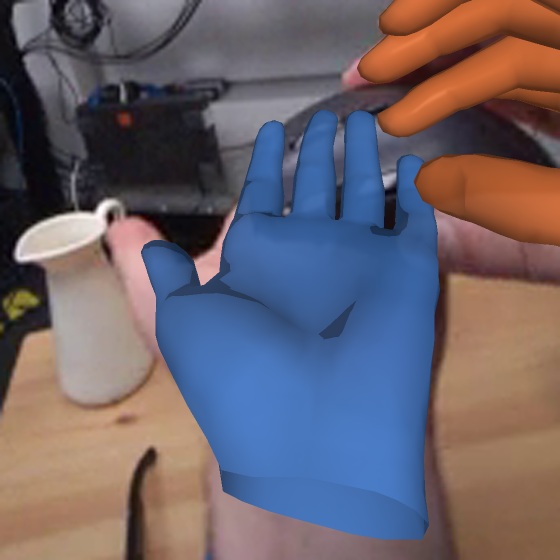}{}{QTag} & \qtile{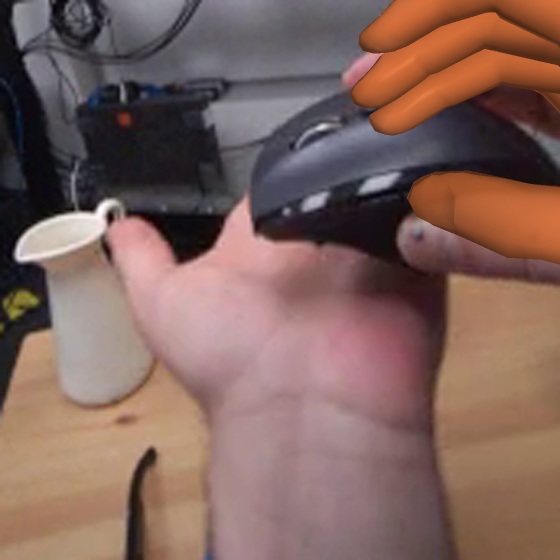}{}{QTag} & \qtile{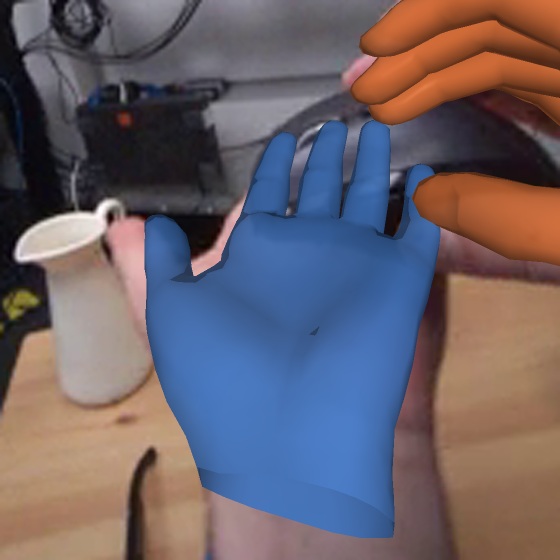}{}{QTag} & \qtile{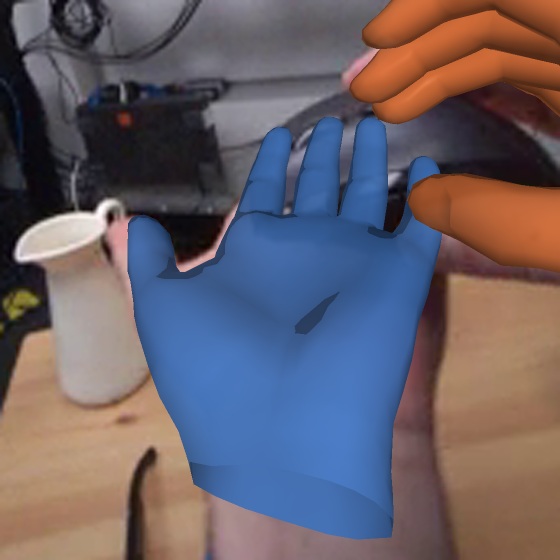}{}{QTag} & \qtile{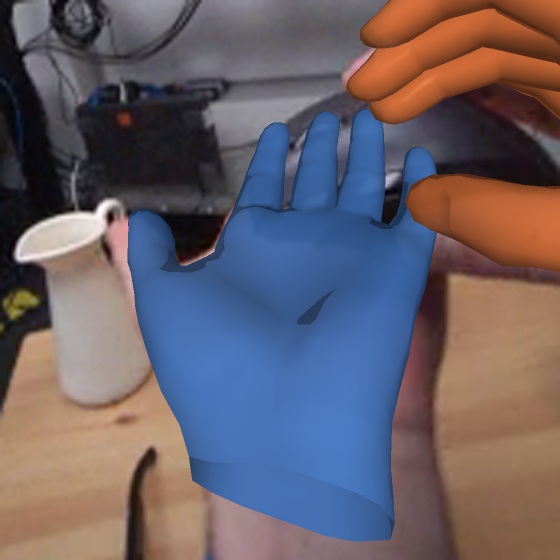}{}{QTag} & \qtile{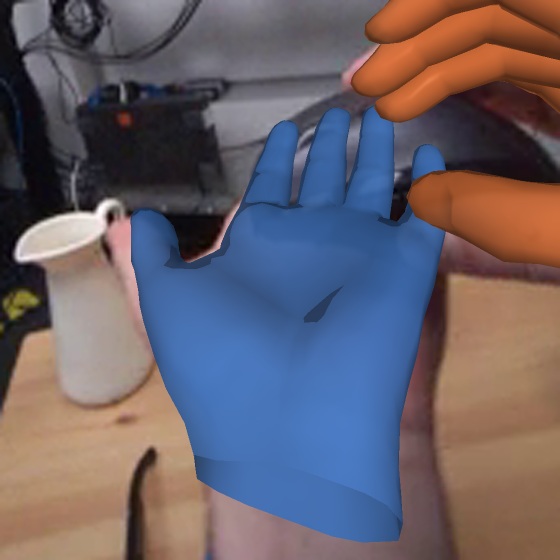}{}{QTag} & \qtile{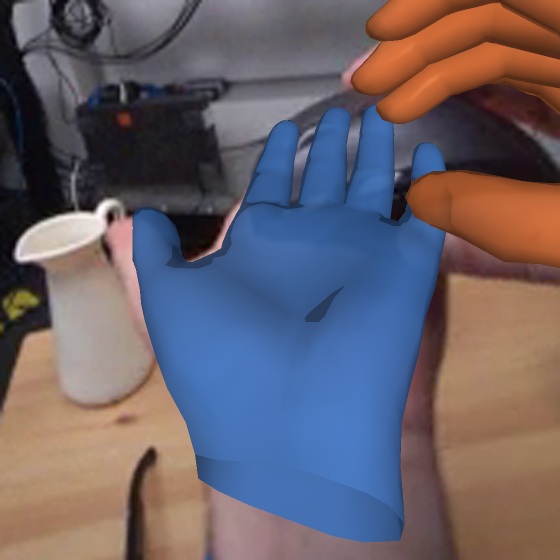}{}{QTag} \\[2pt]
\qrow{frame 12} & \qtile{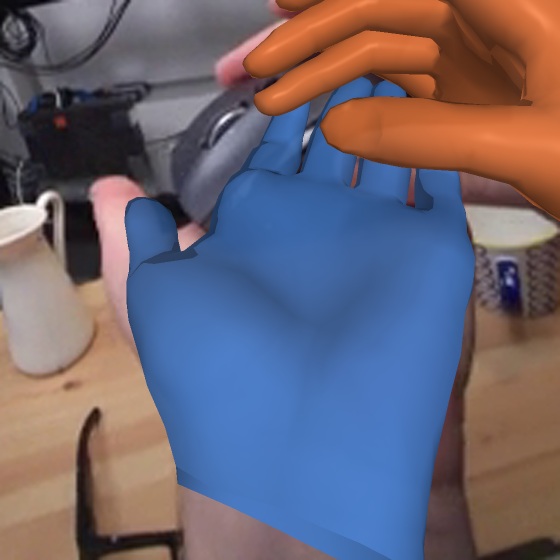}{}{QTag} & \qtile{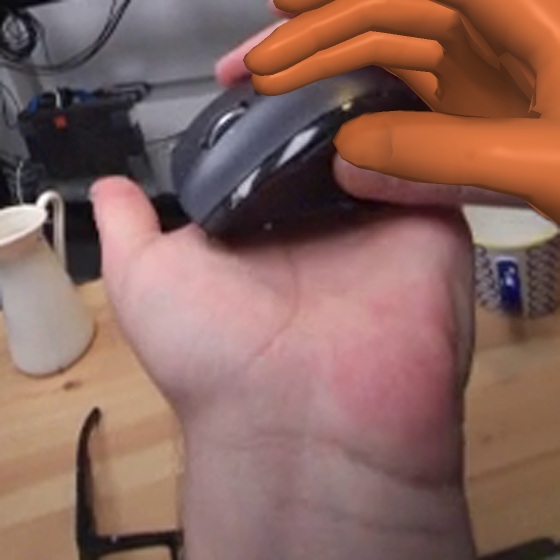}{}{QTag} & \qtile{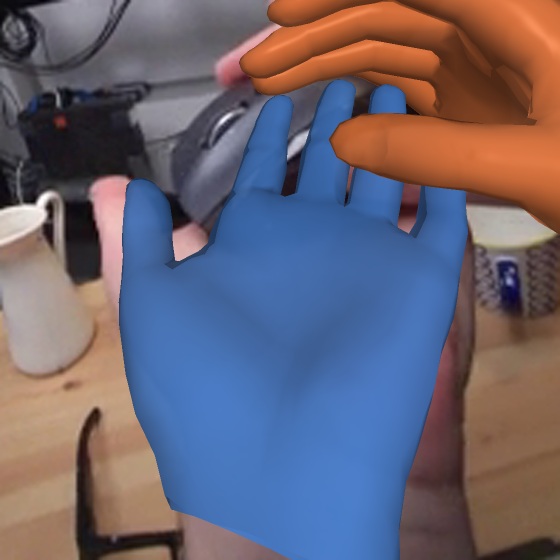}{}{QTag} & \qtile{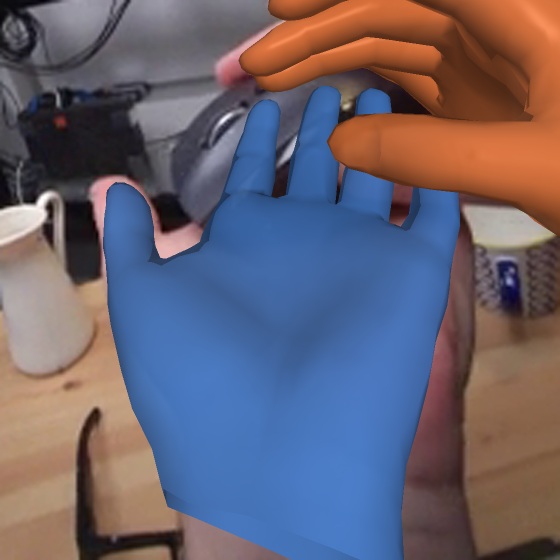}{}{QTag} & \qtile{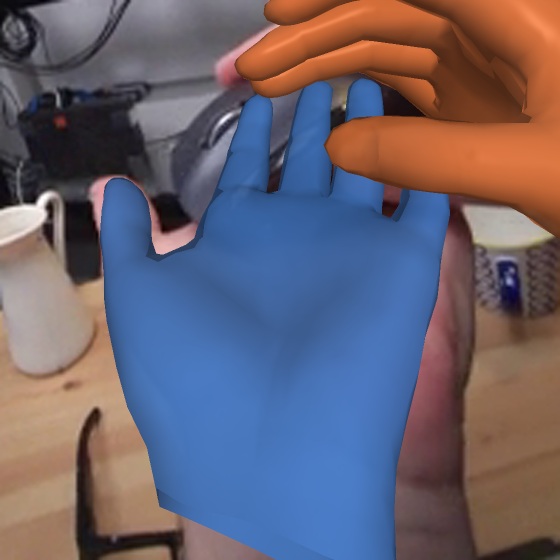}{}{QTag} & \qtile{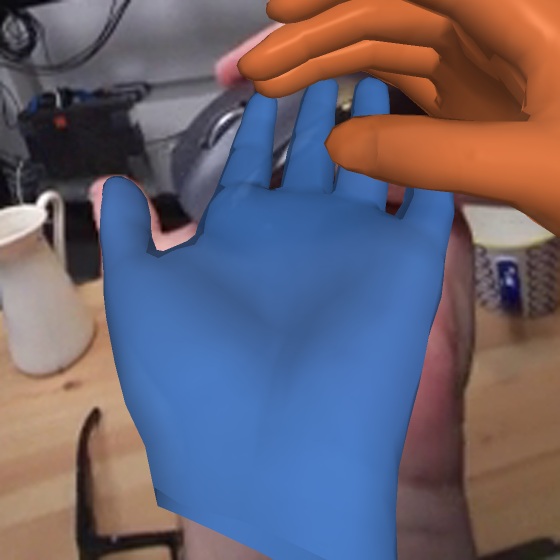}{}{QTag} & \qtile{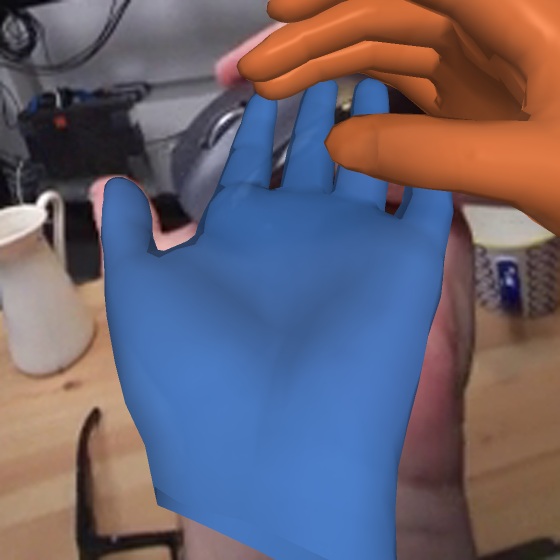}{}{QTag} \\[2pt]
\qrow{error} & \parbox[c]{\qw}{\centering\scriptsize per-frame error\\ (0--40\,mm)\\ \textcolor{QMiss}{\rule{4pt}{4pt}}~missed} & \qtile{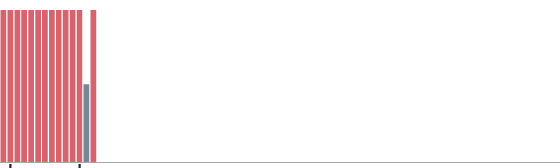}{}{QTag} & \qtile{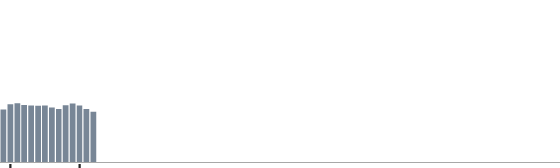}{}{QTag} & \qtile{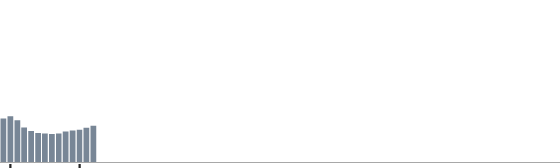}{}{QTag} & \qtile{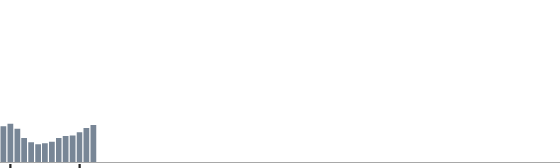}{}{QTag} & \qtile{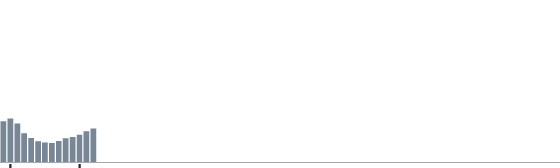}{}{QTag} & \qtile{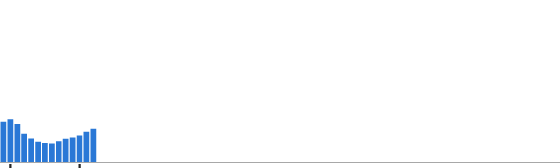}{}{QTag} \\[1pt]
 & \footnotesize MPJPE-p & \footnotesize 22.6\,mm & \footnotesize 7.3\,mm & \footnotesize 6.0\,mm & \footnotesize 5.8\,mm & \footnotesize 5.7\,mm & \footnotesize\textbf{\color{HandRSIBlue}5.7\,mm} \\
\end{tabular}
}
    \caption{
        \textbf{Three more HOT3D test clips after every round} (as Fig.~\ref{fig:evolution}),
        the three with the largest total drop among the clips whose MPJPE-p falls in every round. Per clip: two
        frames of the hand that improves most over the clip (same crop in every column), that hand's wrist-aligned
        joint error in each of the 81 frames (red: missed; ticks: the two frames shown), and the clip's MPJPE-p,
        which also counts the other hand.
    }
    \label{fig:app_rounds_gallery}
\end{figure*}

Figures~\ref{fig:app_gal_hot3d}--\ref{fig:app_gal_hoi4d} extend Fig.~\ref{fig:qualitative}
to four more frames per test set: each method's meshes on the frame and the same hands in 3D, seen from one
third-person view that is identical in every column.
Per set we take, one per recording, the two frames with the largest gap between the baselines' mean 2D error
and HandRSI's, and the two largest among frames in which every method outputs every visible hand; segments of
Fig.~\ref{fig:qualitative} are excluded.
The baselines fail mostly by missing a hand or by placing it at a wrong depth, which the 3D views show as a
hand far from its ground-truth position.
On ARCTIC, HandRSI's meshes follow the outline of hands that grasp large objects more closely than
ACE-Ego-Hand's; on HOI4D, zero-shot for every method, WiLoR swaps the two hands in the first example and
EgoForce places a hand at the camera in the fourth.

Figure~\ref{fig:app_rounds_gallery} follows three more HOT3D test clips through the rounds, as
Fig.~\ref{fig:evolution}.
Of the 1{,}162 test clips, 319 improve in every round; we show the three with the largest total drop, one per
recording, from recordings not shown in Figs.~\ref{fig:qualitative}, \ref{fig:evolution}
or~\ref{fig:clips}.
Most of the gain comes in rounds~1 and~2, when the network is retrained and missed hands are found (third
clip), and rounds~3--5 lower the clips' MPJPE-p by a further 0.3--1.1\,mm with the harness alone.

\end{document}